%% file: Tese.tex
\documentclass[12pt,    
    openright,	         
	twoside,		    
	a4paper,			
    main=english,				
    sumario=tradicional 
	]{TemplateABNT/abntex2}
	
\input{PacotesEConfigs} 

\titulo{Current Challenges of Symbolic Regression: Optimization, Selection, Model Simplification, and Benchmarking}
\autor{Guilherme Seidyo Imai Aldeia} \autorcite{Imai Aldeia, Guilherme Seidyo}
\local{Santo André} \uf{SP}
\data{17 de novembro de 2025} \ano{2025} 
\orientador{Fabrício Olivetti de Fran{\c{c}}a}
\coorientador{William G. La Cava}
\ttorientador{Federal University of ABC} 

\instituicao{Federal University of ABC} \sigla{UFABC}
\instituto{Center of Mathematics, Computing and Cognition}
\departamento{Graduate Program in Computer Science}
\curso{Graduate Program in Computer Science}
\tipotrabalho{Thesis}
\grau{Doctor of Science}

\hypersetup{pdfauthor   = \theauthor,
            pdftitle    = \thetitle,
            pdfsubject  = Doctorate Thesis in Computer Science,
            pdfkeywords = { symbolic regression; simplification; $\epsilon$-lexicse selection; non-linear optimization; multi-objective optimization},
            pdfproducer = pdfLaTex 2019,
            pdfcreator  = pdfLaTex 2019}

\makeindex   

\begin{document} 

\frenchspacing  

\pagenumbering{roman}

\imprimircapa
\imprimirfolhaderosto

\blankpage
\include{PreTextuais/ApoioCAPES}
\include{PreTextuais/Dedicatoria}
\include{PreTextuais/Agradecimento}
\include{PreTextuais/Epigrafe}
\include{PreTextuais/Resumo}


\pdfbookmark[0]{\listfigurename}{lof}
\listoffigures*   
\cleardoublepage

\pdfbookmark[0]{\listtablename}{lot}
\listoftables*  
\cleardoublepage

\pdfbookmark[0]{\listalgorithmname}{lof}
\listofalgorithms   
\cleardoublepage

\include{PreTextuais/ListaSiglas}

\pdfbookmark[0]{\contentsname}{toc}
\tableofcontents*
\cleardoublepage

\pagenumbering{arabic} \setcounter{page}{1}

\textual


\part{Research Summary}
\include{Capitulos/Introduction/introduction}
\include{Capitulos/Background/background}

\part{Current Challenges of Symbolic Regression}
\include{Capitulos/Optimization/optimization}
\include{Capitulos/Selection/selection}
\include{Capitulos/Simplification/simplification}
\include{Capitulos/Benchmarking/benchmarking}

\part{Conclusions}
\include{Capitulos/Conclusions/conclusions}

\postextual 


\bibliographystyle{TemplateABNT/abntex2-alf}
\bibliography{Bibliografia} 

\phantompart  \printindex  

\end{document}

%% file: PacotesEConfigs.tex
\usepackage[T1]{fontenc}		
\usepackage[utf8]{inputenc}		
\usepackage{lastpage}		    
\usepackage{indentfirst}		
\usepackage{color, colortbl}	
\usepackage{graphicx}			
\usepackage{tabularx}
\usepackage{microtype} 		  	
\usepackage{zref-totpages}   

\usepackage{longtable} 

\usepackage{rotating} 
\usepackage{pdflscape} 

\usepackage{ragged2e} 

\usepackage{subfig}             

\usepackage{pdfpages} 

\usepackage{draftwatermark} 
\SetWatermarkScale{0.75}
\SetWatermarkText{} 

\usepackage{float}              
\usepackage{booktabs}           
\usepackage{adjustbox}          
\usepackage{tikz}               
\usepackage{pgfgantt}           
\usepackage{multirow, array}    
\usepackage{enumerate}          
\usepackage{icomma}             
\usepackage{microtype}          
\usepackage{ae, aecompl}        
\usepackage{latexsym}           
\usepackage{lscape}             
\usepackage{adjustbox}          
\usepackage{fancyvrb}           

\usepackage{pdflscape}
\usepackage{afterpage} 

\usepackage{hyperref}      

\PassOptionsToPackage{unicode}{hyperref}
\PassOptionsToPackage{naturalnames}{hyperref}

\usepackage{caption}

\usepackage{amsfonts, amssymb, amsmath, amsthm}     

\usepackage[noend]{algorithmic}
\usepackage[chapter]{algorithm}

\newcommand{\CALL}[3][]{\texttt{#2}(#3)}%

\usepackage[alf,abnt-etal-list=0,abnt-etal-cite=2,abnt-emphasize=bf,abnt-and-type=&]{TemplateABNT/abntex2cite}

\makeatletter
\newcommand*{\citelinktext}[2]{%
  \hyper@@link[cite]{}{cite.#1}{#2}}
\makeatother

\usepackage{etoolbox}

\newcount\ccellA
\newcommand*\ccell[2]
  {%
    \def\tmpa{}%
    \ccellA=1
    \loop
      \ifnum#1=\ccellA
        \edef\tmpa{\unexpanded\expandafter{\tmpa\cellcolor{gray}}}%
      \fi
    \ifnum#2>\ccellA
      \advance\ccellA1
      \edef\tmpa{\unexpanded\expandafter{\tmpa&}}%
    \repeat
    \tmpa
  }

\usepackage{appendix}

\AtBeginEnvironment{subappendices}{%
\chapter*{Appendices}
\addcontentsline{toc}{chapter}{Appendices}
\counterwithin{figure}{section}
\counterwithin{table}{section}
}

\AtEndEnvironment{subappendices}{%
\counterwithout{figure}{section}
\counterwithout{table}{section}
}

\counterwithin{figure}{chapter}
\counterwithin{table}{chapter}

\definecolor{blue}{RGB}{25,25,112}
\makeatletter 
\hypersetup{
	pdftitle={\@title}, 
	pdfauthor={\@author},
	colorlinks=true,    
    linkcolor=black,     
    citecolor=black,     
    filecolor=magenta, 	
	urlcolor=blue,
	bookmarksdepth=4
}
\makeatother

\newenvironment{laysummary} {
  \section*{\centering \large Summary}
  \thispagestyle{empty}
}

\usepackage{pifont}
\newcommand{\xmark}{\ding{55}}

\usepackage{makecell} 

\def\CPP{C\raisebox{0.5ex}{\tiny\textbf{++}}}

\makeatletter

\let\LaTeXStandardPart\part%
\newcommand{\unstarredpart@@noopt}[1]{%
  \unstarredpart@@opt[#1]{#1}%
}%

\newcommand{\unstarredpart@@opt}[2][]{%
  \cleardoublepage
  \begingroup%
  \let\newpage\relax%
  \LaTeXStandardPart[#1]{#2}%
  \endgroup%
}%

\newcommand{\starredpart}[1]{%
  \LaTeXStandardPart*{#1}%
}%

\newcommand{\unstarredpart}{%
  \@ifnextchar[{\unstarredpart@@opt}{\unstarredpart@@noopt}%
}%

\renewcommand{\part}{%
  \@ifstar{\starredpart}{\unstarredpart}%
}%

\usepackage{etoolbox}

\cftsetindents{figure}{0em}{3.5em}
\cftsetindents{table}{0em}{3.5em}

\makeatletter
\pretocmd{\chapter}{\addtocontents{toc}{\protect\addvspace{0.5cm}}}{}{}
\makeatother

 \def\blankpage{%
      \clearpage%
      \thispagestyle{empty}%
      \addtocounter{page}{-1}%
      \null%
      \clearpage}
      
\usepackage{tgtermes}
\renewcommand{\ABNTEXchapterfont}{\rmfamily\bfseries}

\providecommand{\imprimirautorcite}{}
\newcommand{\autorcite}[1]{\renewcommand{\imprimirautorcite}{#1}} 
\providecommand{\imprimirsubtitulo}{}
 
\providecommand{\imprimirsigla}{}
\newcommand{\sigla}[1]{\renewcommand{\imprimirsigla}{#1}}
\providecommand{\imprimiruf}{}
\newcommand{\uf}[1]{\renewcommand{\imprimiruf}{#1}}
\providecommand{\imprimircurso}{}
\newcommand{\curso}[1]{\renewcommand{\imprimircurso}{#1}}
\providecommand{\imprimirinstituto}{}
\newcommand{\instituto}[1]{\renewcommand{\imprimirinstituto}{#1}}
\providecommand{\imprimirdepartamento}{}
\newcommand{\departamento}[1]{\renewcommand{\imprimirdepartamento}{#1}}
\providecommand{\imprimirano}{}
\newcommand{\ano}[1]{\renewcommand{\imprimirano}{#1}}
\providecommand{\imprimirgrau}{}
\newcommand{\grau}[1]{\renewcommand{\imprimirgrau}{#1}}

\providecommand{\imprimirttorientador}{}
\newcommand{\ttorientador}[1]{
    \renewcommand{\imprimirttorientador}{#1}}

\renewcommand{\imprimircapa}{ 
\begin{capa}
        \begin{center}
                \begin{DoubleSpace}
                \MakeUppercase{\imprimirinstituicao } \\
                \MakeUppercase{\imprimirinstituto } \\
                \MakeUppercase{\imprimirdepartamento} \\
                \end{DoubleSpace}
                \vspace{5cm}
				\MakeUppercase{\imprimirautor}  \\
                        				
				\vspace{5cm}
             \textbf{{\large\MakeUppercase{\imprimirtitulo}}} \\
			 \textbf{{\large \MakeUppercase{\imprimirsubtitulo}}} \\
				\vfill
        {\large{\imprimirlocal,~\imprimiruf \\ \imprimirano  }}
        \end{center}
\end{capa}   
} 

\renewcommand{\imprimirfolhaderosto}{
       \begin{center}
				\MakeUppercase{\imprimirautor}  \\
				
                \vspace{6cm}
			    \begin{DoubleSpace}
                \MakeUppercase{\textbf{\imprimirtitulo} } \\
                \MakeUppercase{\textbf{\imprimirsubtitulo}} \\
                \end{DoubleSpace} 
      \end{center}
    \vfill 
    \begin{flushright} 
    \parbox{0.6\linewidth}{
		\imprimirtipotrabalho~submitted to the \imprimircurso~of the \imprimirinstituicao~in partial fulfillment of the requirements for the degree of \imprimirgrau. \\ \\
		\textbf{\imprimirorientadorRotulo}~\imprimirorientador \\
		\textbf{\imprimircoorientadorRotulo}~\imprimircoorientador}
   \end{flushright} 
		\vfill
   \begin{center}
   {\large{\imprimirlocal,~\imprimiruf \\
   \imprimirano} }
   \end{center} }  

\usepackage{abstract}

\newcommand\Chapter[2]{\chapter
  [#1:\texorpdfstring{\hfil\hbox{}}{}\protect\linebreak{\itshape#2}]%
  {#1\\[2ex]\Large\itshape#2}%
}

\usepackage{hyphenat}

\definecolor{blue}{RGB}{41,5,195}

\definecolor{barblue}{RGB}{153,204,254}
\definecolor{groupblue}{RGB}{51,102,254}

\makeatletter
\ganttset{
    prog default/.initial=100,
    prog/.code={
        \pgfutil@in@{:}{#1}
        \ifpgfutil@in@
            \pgfqkeysalso{/pgfgantt}{@prog={#1}}
        \else
            \pgfqkeysalso{/pgfgantt}{@prog={\pgfkeysvalueof{/pgfgantt/prog default}:#1}}
        \fi
    },
    @prog/.code args={#1:#2}{
        \edef\pgf@tempa{#1}%
        \ifx\pgf@tempa\tikz@nonetext
            \pgfqkeysalso{/pgfgantt}{progress={100},progress label text={#2}}
        \else
            \pgfqkeysalso{/pgfgantt}{progress={#1},progress label text={#2 (##1\,\%)}}
        \fi
    }
}
\makeatother

\fvset{frame=single,framesep=1mm,framerule=.3mm,commandchars=\\\{\}}

\makeatletter
\let\old@l@part\l@part
\renewcommand{\l@part}[2]{%
  \old@l@part{#1}{#2}
  \vspace{2.0em}}
\makeatother

%% file: PreTextuais/ApoioCAPES.tex
\noindent This study was financed in part by the Coordenação de Aperfeiçoamento de Pessoal de Nível Superior – Brasil (CAPES) – Finance Code 001, and grant 88887.802848/2023-00.

\blankpage

%% file: PreTextuais/Dedicatoria.tex
\begin{dedicatoria}
   \vspace*{\fill}
   \centering
   \noindent
   \textit{In dedication to my Family.} 
   \vspace*{\fill}
\end{dedicatoria}


%% file: PreTextuais/Agradecimento.tex
\begin{agradecimentos}

As I could easily discuss for hours, writing a thesis is not merely the production of a document, but also a process of personal growth and coming of age in many respects.
At no other time in my life have I felt such a profound transformation between the person I was and the person I have become.
I wholehearthfully want to thank those who were part of this journey, some without even realizing it.

First and foremost, I thank my family, Gabrielle, Sueli, and Wagner, for their unwavering support, patience, encouragement, and faith in me.

I am deeply grateful to my advisors, Fabrício de França and William La Cava, for their invaluable teaching, guidance, preparation, opportunities, and countless meetings throughout the years. 

I also thank the committee members and all the researchers I had the privilege to collaborate and learn from. Your feedback, critiques, suggestions, and insights were essential in shaping this thesis. 

I extend my gratitude to my colleagues and professors with whom I shared lunches, coffees, and conversations --- sometimes about science, sometimes about everything but science.

My thanks also go to everyone at the Computational Health Informatics Program, Boston Children’s Hospital, Harvard Medical School, and the Graduate Program in Computer Science at the Center for Mathematics, Computing, and Cognition, Federal University of ABC.  

Last but not least, I thank my close friends for their genuine support, and for making this journey joyful. Despite the sobriety and technicality of this thesis as a whole ---except for this single page--- a part of each one of you is here. Some of you are getting married, some are planning to have kids. Well, I wrote a thesis.

It is both daunting and rewarding to look back at all the progress I have made, and exciting to look ahead and imagine all that is yet to come. To my future self, I hope you never lose your ambition. In moments of doubt, may this thesis remind you that you will always be capable of going a little further.  

\end{agradecimentos}

%% file: PreTextuais/Epigrafe.tex
\begin{epigrafe}
\vspace*{\fill}

\begin{flushright}
    \textit{Completing a thesis represents a coming of age not just scientifically, but also educationally and personally. It signals the passing of an intellectual milestone — from a student under the care of a supervisor to an individual who asks questions of their own.}
    
    \citelinktext{past_present_future_phd}{The past, present and future of the PhD thesis. \textit{Nature} (2016, vol. 535, nº 7610, pp. 7-7)}
\end{flushright}

\end{epigrafe}

%% file: PreTextuais/Resumo.tex

\setlength{\absparsep}{18pt} 
\begin{resumo}[Resumo]
    
    A Regressão Simbólica (SR) é um método de regressão que busca descobrir expressões matemáticas que descrevem a relação entre variáveis, e costuma ser implementado por meio da programação genética, uma metáfora para o processo de evolução biológico.
    Seu apelo está em combinar alto desempenho preditivo com modelos interpretáveis, mas essa promessa é limitada por alguns desafios existentes há tempos na literatura da área: os parâmetros livres são difíceis de otimizar, a seleção de soluções candidatas pode afetar a direção da busca, os modelos frequentemente crescem de forma desnecessariamente complexa.
    Além das limitações dos métodos, há também limitações metodológicas na realização e apresentação de resultados empíricos quando comparando diversos métodos para compreender o panorama geral da área.
    Esta tese aborda esses diferentes desafios por meio de uma sequência de estudos realizados ao longo do doutorado, cada um focado em um aspecto importante do processo de busca em SR.
    Primeiro, a otimização de parâmetros é investigada, buscando entender seu papel na diminuição do erro de predição, mostrando compromissos em tempo de execução e tamanho das expressões em troca. Em seguida, a seleção de soluções pais é estudada, explorando o $\epsilon$-lexicase para escolher candidatos mais propensos a gerar novas soluções com bom desempenho. Então, o foco é direcionado para a simplificação, onde é apresentado um método novo baseado em memoização e hashing sensível à localidade, que reduz a redundância e facilita obtenção de expressões mais simples e com mesmo desempenho preditivo.
    Todas essas contribuições são implementadas em uma biblioteca de SR evolutiva multi-objetivo, que alcança um desempenho de Pareto ótimo em termos de erro de predição e simplicidade em problemas reais e sintéticos, superando diversas abordagens contemporâneas alternativas.
    A tese conclui propondo alterações a um famoso benchmark de regressão simbólica em larga escala, a fim de avaliar o panorama mais recente da área e incluir o trabalho desenvolvido, demonstrando que o um método incorporando as propostas da tese atinge desempenho de Pareto ótimo.

    \vspace{\onelineskip}
    \noindent
    \textbf{Palavras-chave}: regressão simbólica; simplificação; seleção $\epsilon$-lexicase; otimização não-linear; otimização multi-objetivo.
    
\end{resumo}

\begin{resumo}[Abstract]
\begin{otherlanguage*}{english}
   
    Symbolic Regression (SR) is a regression method that aims to discover mathematical expressions that describe the relationship between variables, and it is often implemented through Genetic Programming, a metaphor for the process of biological evolution.
    Its appeal lies in combining predictive accuracy with interpretable models, but its promise is limited by several long-standing challenges: parameters are difficult to optimize, the selection of solutions can affect the search, and models often grow unnecessarily complex. In addition, current methods must be constantly re-evaluated to understand the SR landscape.
    This thesis addresses these challenges through a sequence of studies conducted throughout the doctorate, each focusing on an important aspect of the SR search process.  
    First, I investigate parameter optimization, obtaining insights into its role in improving predictive accuracy, albeit with trade-offs in runtime and expression size. Next, I study parent selection, exploring $\epsilon$-lexicase to select parents more likely to generate good performing offspring. The focus then turns to simplification, where I introduce a novel method based on memoization and locality-sensitive hashing that reduces redundancy and yields simpler, more accurate models.  
    All of these contributions are implemented into a multi-objective evolutionary SR library, which achieves Pareto-optimal performance in terms of accuracy and simplicity on benchmarks of real-world and synthetic problems, outperforming several contemporary SR approaches.  
    The thesis concludes by proposing changes to a famous large-scale symbolic regression benchmark suite, then running the experiments to assess the symbolic regression landscape, demonstrating that a SR method with the contributions presented in this thesis achieves Pareto-optimal performance.

   \vspace{\onelineskip}
   \noindent 
   \textbf{Keywords}: symbolic regression; simplification; $\epsilon$-lexicase selection; non-linear optimization; multi-objective optimization.
   
\end{otherlanguage*}
\end{resumo}

%% file: PreTextuais/ListaSiglas.tex
\begin{siglas}
    \item[ AUC ] \textit{Area Under the Curve}
    \item[ CI ] \textit{Confidence Interval}
    \item[ CD ] \textit{Critical Differences}
    \item[ DL ] \textit{Description Length}
    \item[ ERC ] \textit{Ephemeral Random Constant}
    \item[ FEAT ] \textit{Feature Engineering Automation Tool}
    \item[ GA ] \textit{Genetic Algorithm}
    \item[ GP ] \textit{Genetic Programming}
    \item[ ITEA ] \textit{Interaction-Transformation Evolutionary Algorithm}
    \item[ IT ] \textit{Interaction-Transformation}
    \item[ LLM ] \textit{Large Language Model}
    \item[ LM ] \textit{Levenberg-Marquardt}
    \item[ LRU ] \textit{Least Recently Used}
    \item[ LSH ] \textit{Locality-Sensitive Hashing}
    \item[ MAD ] \textit{Median Absolute Deviation}
    \item[ ML ] \textit{Machine Learning}
    \item[ MO ] \textit{Multi-Objective}
    \item[ MSE ] \textit{Mean Squared Error}
    \item[ MVT ] \textit{Minimum Variance Threshold}
    \item[ NMSE ] \textit{Normalized Mean Squared Error}
    \item[ NSGA2 ] \textit{Non-dominated Sorting Genetic Algorithm II}
    \item[ OLS ] \textit{Ordinary Least Squares}
    \item[PMLB] \textit{Penn Machine Learning Benchmarks}
    \item[ PTC2 ] \textit{Probabilistic Tree Creation 2}
    \item[ RSS ] \textit{Residual Sum of Squares}
    \item[ RL ] \textit{Reinforcement Learning}
    \item[ RMSE ] \textit{Root Mean Squared Error}
    \item[ SotA ] \textit{state-of-the-art}
    \item[ SRBench ] \textit{Symbolic Regression Benchmark}
    \item[ SR ] \textit{Symbolic Regression}
\end{siglas}

%% file: Capitulos/Introduction/introduction.tex
\chapter{Introduction}~\label{chap:introduction}








Symbolic Regression (SR), popularized by \citeonline{GPOnTheProgrammingOfComputersByMeansOfNaturalSelection}, refers to methods that search for mathematical expressions describing the relationship between variables, balancing predictive accuracy and interpretability\footnote{Definitions of \textit{interpretability} and \textit{explainability} have been the subject of long-running discussions in the literature. A comprehensive review of interpretability in the context of symbolic regression was published by \citeonline{Aldeia2022}. See it for further details.}.
This dual optimization enables SR to discover concise and interpretable models that balance predictive accuracy with human-understandable representations \cite{srbench,de2024srbenchplusplus}.

Among its many applications, SR has been successfully used in physics \cite{Angelis2023,Reinbold2021, KUBALIK2021115210,HRUSKA2025118821} and astrophysics \cite{multiviewsr}, aerospace engineering \cite{LACAVA2016892,windturbinedynamics,LIU2024108986}, dynamical systems~\cite{schmidt_distilling_2009}, and chemical catalysis \cite{Weng2020}.
It has also been applied to recovering physical laws from observational data \cite{cranmer2023interpretablemachinelearningscience}, and to medical informatics \cite{la_cava_flexible_2023, Wilstrup2022,10.1145/3205455.3205604}.

One of its main interest lies in applications to physics, where SR is viewed as a means of ``automating Kepler'' ---discovering analytical expressions through trial and error---, and interpretability plays a major role. This is particularly evident in recent works that use SR to recover laws directly from experimental data~\cite{Makke2024}, underscoring its potential as a tool for automated scientific discovery.
It has demonstrated the ability to uncover physical laws purely from data, sometimes using small sample sizes \cite{cranmer2023interpretablemachinelearningscience} --- the same observational data from which known laws, such as Kepler's laws of planetary motion, were originally derived.
It has also been shown to find equations that fit data from different sources simultaneously in astrophysics problems where no known correct equation exists yet \cite{multiviewsr}, yielding promising results for scientific discovery.

Interpretability is essential not only in scientific discovery but also in contexts where models must justify their recommendations, such as healthcare decision support, \textit{e.g.}, the FDA guidelines for clinical decision support software \cite{FDA}. 
In medical informatics, for example, SR applied to electronic health records of patients with different subtypes of hypertension produced an interpretable scoring equation to predict treatment-resistant hypertension \cite{la_cava_flexible_2023}.

SR has repeatedly proven to yield human-readable and competitive results, as demonstrated in interpretability competitions \cite{defranca2023interpretable} and the annual \emph{Human-Competitive Results Produced By Genetic And Evolutionary Computation} (HUMIES) awards\footnote{\url{https://www.human-competitive.org/awards}}.
To this day, many algorithms have been published and are readily available for use \cite{srbench}.

In the broader context of machine learning (ML), SR is a supervised learning approach for the regression task.
The main appeal of SR compared to other ML methods is that the returned mathematical expression can serve as both a prediction function and can be dissected for analytical study, providing \textit{insights} into the underlying data or problem --- transforming data into knowledge.
SR has the potential to find intrinsically interpretable solutions~\cite{GainingDeeperInsightsInSymbolicRegression,SymTree, ExplainingSRPredictions,ApplyingGeneticProgrammingToImproveInterpretability, Aldeia2022,radwan_comparison_2024}, in contrast to many ML methods such as random forests or gradient boosting machines.
Accordingly, many SR algorithms employ multi-objective optimization techniques to balance predictive performance and expression complexity \cite{cranmer2023interpretablemachinelearningscience,Makke2024,10.1162/evco_a_00278}.

However, the search for such expressions is computationally difficult. Symbolic regression is NP-hard \cite{SymbolicRegressionIsNPHard,song2024prove}, meaning that searching for the best regression model is computationally intractable (unless P=NP is proved). The search space of SR can be vast, as it can represent any algebraic function.

Many SR algorithms are based on Genetic Programming (GP), following similar concepts to how it was conceived by Koza. GP is a population-based heuristic, meaning it works with several different candidate solutions in parallel, and this allows it to search in multiple directions at once. GP evolves candidate programs through variation and selection: a randomly initialized population is refined over generations, with solutions recombined, perturbed, and selected based on fitness until termination criteria are met. When a termination criterion is met, the algorithm stops and returns a final solution, selected from the final population in different ways --- such as the most accurate, the simplest, or a compromise between the two. 

More recently, deep learning approaches have emerged, training generative models to produce expressions~\cite{petersen2021deep,pmlr-v139-biggio21a,kamienny_end--end_nodate,shojaee_transformer-based_nodate,UnifiedFrameworkForDeepSR,holt2023deep}. These methods differ in architecture (recurrent network, transformer), training strategy (reinforcement learning, input-output-expression tuples), and treatment of floating-point constants. However, transformer-based models remain constrained by the number of input variables they can handle and the cost of the pre-training.  
Hybrid methods that combine GP with deep generative models have shown promise~\cite{UnifiedFrameworkForDeepSR}, though deep learning approaches still generally underperform when compared to state-of-the-art GP methods~\cite{operon, srbench}.

\section{Current challenges}~\label{sec:introduction:current-challenges}

Symbolic regression has a long history as a promising approach for interpretable ML. Yet many of its celebrated applications ---such as rediscovering physical laws--- remain proofs of concept rather than tools routinely adopted by scientists or practitioners.
The gap lies not in SR's potential but in technical barriers: the interaction between parameter optimization methods used by different algorithms remains underexplored, with limited knowledge about how they influence each other; the parent selection mechanism, intended to mix different parts of promising parents, depends on the selection criteria and can indirectly affect performance, requiring it to promote diversity while also ensuring that good-performing individuals are selected; models often grow unnecessarily complex during evolution, incorporating redundant or non-expressed operations that may affect performance on unseen data, thus benefiting from simplifications to achieve more interpretable models; and robust, comprehensive benchmarking is needed to assess the current SR landscape and identify trends and opportunities in the field, though this remains challenging due to the variety of methods and the need for standardized hardware and postprocessing procedures for in-depth analyses.
This thesis contributes to this effort by tackling four core challenges of SR: parameter optimization, parent selection, model simplification, and benchmarking.

Identifying optimal parameter values for the SR expression is a challenging task \cite{Kommenda2019}, also belonging to the NP-hard category \cite{kotary2021endtoend}. Leaving this task solely to the evolutionary process ---by generating random constants in the expectation that survival pressure will keep promising ones--- makes it harder to find optimal values and drastically slows down convergence speed \cite{8393325}, since these operators lack the capability to obtain the global optimum \cite{WANG20042453}.
In this context, real-valued parameter optimization in SR has been partially addressed with local optimization methods, where authors proposed to use linear combination of terms with linear regression \cite{10.1162/evco_a_00285}, linear regression with a regularization term \cite{McConaghy2011,cava2018learning}, or non-linear regression iterative methods \cite{PrioritizedGrammarEnumeration,operon}, although the interaction between linear and non-linear optimization remains underexplored.
In this sense, understanding the advantages of using linear or non-linear optimization could help shed light on this matter, \textit{e.g.}, whether if they should be used separately or in combination.

An essential aspect of any evolutionary algorithm is selecting parent solutions to recombine and generate new \emph{offspring}. The goal of recombination is to merge parts of highly fit parents, with the expectation that the offspring improves upon them. Effective parent selection therefore requires parents that perform well across different examples in the dataset.
Traditional methods such as tournament selection \cite{miller1995genetic, brindle1980genetic} use aggregate fitness scores to decide which parents will reproduce. However, evidence shows that selection based on aggregate fitness ignores relevant cases \cite{10.1145/2576768.2598288}. An alternative is to reward individuals excelling in difficult subsets of cases. This is the idea behind lexicase selection \cite{10.1145/2330784.2330846}, later adapted for regression problems \cite{la_cava_epsilon-lexicase_2016}, which has shown state-of-the-art performance \cite{ding_probabilistic_2023}. The canonical version of $\epsilon$-lexicase selection uses the median absolute deviation to select an $\epsilon$ tolerance threshold, but was benchmarked with different variants by \citeonline{hernandez_exploration_2021,la_cava_epsilon-lexicase_2016,10.1145/3583131.3590375} and the best way of defining $\epsilon$ is still an open question.

Another challenge investigated is that SR search space includes equivalent mathematical expressions \cite{burlacu2019hash}, introns (parts of the model that do not influence predictions) \cite{GainingDeeperInsightsInSymbolicRegression}, bloat (growth in model size without justified improvements in loss) \cite{luke_comparison_2006}, and redundant parameters \cite{10.1145/3583131.3590346}. These issues inflate model size, reducing simplicity and interpretability.
Several approaches exist to address this: automatic simplification \cite{helmuth_improving_2017}, parametric rewriting rules \cite{kulunchakov_creation_2017}, Bayesian loss metrics for model selection \cite{bomarito_bayesian_2022}, restricting model structure \cite{10.1162/evco_a_00285}, or exact simplification methods such as equality saturation~\cite{10.1145/3583131.3590346}. However, existing techniques do not always yield simpler or more interpretable models, and may require several rules to be implemented in order to effectively work, increasing the implementation burden.

The rapid pace of development in the field ---with new methods published every year--- raises the need for constant evaluation under standardized frameworks. Such benchmarking is often expensive, since GP-based methods evolve hundreds or thousands of candidates simultaneously and still lack strong GPU or parallel computing support.
In the SR field, previous surveys identified benchmarking as a weakness \cite{10.1145/3205455.3205539}. A lack of consensus on standardizing datasets and methods has already been observed \cite{srbench}, with some works emphasizing the importance of open-source and ease-of-use principles \cite{egklitz2020}. Still, several authors have performed benchmarks using only a small set of problems or algorithms \cite{esg1352,egklitz2020,matsubara2024rethinking}, typically considering up to a dozen methods.
Benchmarking remains essential to measure progress, redefine the state-of-the-art, and gather insights into promising research directions, ensuring symbolic regression research remains meaningful. However, benchmarking practices are still fragmented.

\section{Research gaps}~\label{sec:introduction:limitations}

There are dozens of SR algorithms published to date. Rather than proposing a new method, this thesis focuses on deriving insights into the critical components of SR algorithms. 
These opportunities can contribute to the symbolic regression community by improving SR methods from multiple facets.

First, combining parameter optimization heuristics can help balance predictive accuracy and model size. Closed-form methods provide a fast solution for linear parameters, but they require that parameters appear linearly in the model structure. In contrast, iterative non-linear optimizers can tune arbitrary parameters and improve predictive performance, yet they converge to a local minimum and are sensitive to the choice of initialization. Despite their complementary strengths, the interaction between linear and non-linear optimization remains underexplored, and a better understanding of how they synergize could yield more accurate expressions.

Second, parent selection mechanisms can be improved by incorporating more informative selection criteria. $\epsilon$-lexicase relies on the Median Absolute Deviation to compute an $\epsilon$ threshold for filtering candidates. However, this threshold is determined relative to the best individual, which may focus on near-elite selection. Introducing an information-theoretic threshold offers an alternative to modify the parent selection behavior.

Third, model simplification remains essential for enhancing interpretability and improving generalization. Existing simplification strategies ---such as random deletions or replacements, hashing of commutative operations, or algebraic rewriting rules--- are often complex to implement. Since simplification should be both effective and practical, there is a need for data-driven approaches that reduce unnecessary expression complexity without relying on handcrafted rules.

Lastly, benchmarking is itself also an area with room for improvement. Many new SR algorithms have emerged, including transformer-based generative approaches, and the community has suggested several improvements to existing benchmarking practices. Proposing a benchmark is therefore crucial to reliably assess progress and support the development of future SR methods.

\section{Objectives}~\label{sec:introduction:research-questions}

The main objective of this thesis is to improve symbolic regression algorithms by addressing the challenges outlined above and validating them through robust benchmarking against alternative approaches.
The specific objectives are as follows:

\begin{description}
    \item[Parameter optimization.] To investigate whether separating the optimization of linear and non-linear parameters improves predictive accuracy and model size, and to assess the trade-off in runtime when applying iterative non-linear optimization and closed-form linear optimization.
    \item[Parent selection.] 
    To use a different criterion based on information theory for $\epsilon$-lexicase and evaluate its effects on parent selection with a focus on error convergence, as well as to determine in which problem domains the new criterion enhances overall performance.
    \item[Simplification.] To develop and assess a new data-driven method for automatic simplification of symbolic regression expressions, reducing unnecessary complexity while preserving predictive accuracy, and to explore whether fast and inexact simplification can reduce implementation burden without harming performance.
    \item[Benchmarking.] To perform a comprehensive benchmarking study of contemporary symbolic regression algorithms, establishing standardized evaluation criteria and clarifying the broader landscape of SR research. 
\end{description}

\section{Implementations on a single framework}~\label{sec:introduction:brush}

This thesis resulted from multiple research questions investigated during the doctorate. For roughly three out of the four years of the doctorate development, the student worked on \textit{Brush}, a framework initially proposed by William La Cava and Joseph D. Romano for working with electronic health records. Many functionalities implemented on Brush were derived from, or are related to, the independent investigations into SR problems presented in this thesis.

Brush ---although not the main focus of this thesis--- is described in Chapter \ref{chap:Background}. Thus, even if experiments were not directly run with Brush, the final implementations were validated within it, unifying the contributions into a single system and contrasting it to existing literature.

The Chapter \ref{chap:benchmarking} presents a large-scale benchmark that includes Brush, among other SR algorithms, evaluated on different real-world, and observational data. In comparison to many recently published SR algorithms, Brush was capable of achieving state-of-the-art performance, balancing accuracy and simplicity in the real-world problems, and returned highly accurate models on synthetic problems.

\section{Contributions}~\label{sec:introduction:contributions}

The contributions of this thesis are threefold.
First, it provides an empirical analysis on current challenges in symbolic regression, focusing on components that influence both predictive performance and model complexity.
Second, it introduces and evaluates a novel simplification method based on fast, inexact approximations, which improves predictive accuracy while reducing model complexity, demonstrating that simplification can simultaneously serve performance and interpretability.
Third, it conducts a comprehensive benchmarking of symbolic regression algorithms across diverse datasets and multiple state-of-the-art methods, resulting in a clearer view of the current landscape, providing practical guidance for future research.

\section{Text structure}~\label{sec:introduction:text-structure}

This thesis is organized as follows. 
The Chapter \ref{chap:Background} provides an extensive background for each problem investigated in this thesis, covering the related literature.
After that, each chapter is based on independent research addressing one of the symbolic regression challenges outlined in this introduction. This reflects a modern thesis style \cite{Gould2016,past_present_future_phd}, where chapters correspond to accepted scientific publications.

This modern thesis style, built upon scientific publications, provides valuable feedback through peer review, and helps the student establish a presence in the international research community \cite{BurroughBoenisch2016}. A thesis is more than a final product on a subject --- it is part of the process of developing expertise, with a central role in the student's education \cite{https://doi.org/10.1111/area.12779}. Collaborating with other researchers also strengthens teamwork, communication, management, and research ethics skills \cite{Gould2016}.

Authorship for this thesis is disclosed clearly: the student was first author and made significant contributions to all included papers. 
Parts of the chapters were published in peer review venues, and in this thesis they are expanded with new discussions and aligned to the broader SR algorithms. Major adjustments were also made to improve readability, such as consolidating the background and related work in a dedicated chapter.

Each chapter study was conducted with the most appropriate SR framework for that specific challenge. For example, parameter optimization was investigated in an algorithm highly dependent on parameter tuning; parent selection was studied in a framework canonically supporting $\epsilon$-lexicase; and simplification was developed in a clean experimental environment with few confounding factors.
To overcome this isolation, contributions were implemented in Brush, a \CPP{} symbolic regression library with a Python wrapper.

The contributions are organized as follows. Chapter \ref{chap:Background} presents the necessary background and introduces the unified framework.
The Chapters \ref{chap:parameteropt} to \ref{chap:inexactsimp} reports the contributions to the symbolic regression field, organized in order of publication, with the most recent listed last. Chapter \ref{chap:parameteropt} examines linear and non-linear parameter optimization. Chapter \ref{chap:parentsel} studies parent selection in evolutionary algorithms, focusing on $\epsilon$-lexicase and proposing an information-theoretic variation. Chapter \ref{chap:inexactsimp} investigates simplification, proposing an implementable algorithm that relaxes strict equivalence to achieve simpler forms. The contributions were implemented on a symbolic regression algorithm, that was benchmarked in Chapter \ref{chap:benchmarking}, including $25$ other algorithms evaluated under the same environment. Finally, Chapter \ref{chap:Conclusions} summarizes the contributions and outlines future perspectives.


%% file: Capitulos/Background/background.tex
\chapter{Background}~\label{chap:Background}

Symbolic Regression (SR) is a supervised machine learning (ML) approach that searches for mathematical expressions, rather than optimizing a black-box structure. 
Its appeal lies in balancing predictive accuracy with interpretability, offering insights into the underlying mechanisms of a system.

This chapter provides the necessary background on SR, covering its history, taxonomy, and open challenges, while situating this thesis within the field. The chapter also describe how SR is commonly implemented through Genetic Programming (GP), and review the relevant literature related to the challenges explored in this work.

This chapter is organized as follows.
Section~\ref{sec:background:history} provides a historical overview of the main milestones in SR.
Section~\ref{sec:background:problem_formulation} presents the mathematical and computational formulation of SR, also discussing multi-objective optimization and GP in detail, highlighting its continued relevance for SR.
Section~\ref{sec:background:search} reviews the challenges addressed in this thesis, focusing on search and optimization of expressions.
Section~\ref{sec:background:benchmarking} covers benchmarking practices in SR and their importance for progress in the field.
Section~\ref{sec:background:recent_trends} introduces the taxonomy of SR methods, the role of SR in the ML ecosystem, and recent research trends.
Section~\ref{sec:background:brush} presents a unified framework that integrates several improvements discussed here, which is evaluated in the benchmarking chapter.
Finally, Section~\ref{sec:background:summary} summarizes the chapter and outlines the key concepts needed for the remainder of the thesis.

\section{Historical Background}~\label{sec:background:history}

The roots of SR can be traced back to early work on evolutionary computation, from initial attempts at automatically discovering systems \cite{lenat11984automated,langley1977bacon,dvzeroski1993discovering} to the foundational works of Genetic Algorithms by \citeonline{adaptationinnaturalartificial} and Genetic Programming by \citeonline{GPOnTheProgrammingOfComputersByMeansOfNaturalSelection}.
Deductive problem-solving capabilities and full theory formation were sought by \citeonline{lenat11984automated} a long time ago, proposing a rule-based system to represent and manipulate mathematical knowledge.
BACON was introduced by \citeonline{langley1977bacon} to identify relationships between variables from data, with the goal of discovering patterns.
The idea of automatically discovering mathematics through machines was explored by \citeonline{dvzeroski1993discovering}, who developed the algorithm LAGRANGE to discover dynamic system equations.
In parallel, genetic algorithms were pioneered by Holland, while Koza established much of the basis that later shaped the symbolic regression field.

By the 1960s, Holland formalized the idea of Genetic Algorithms (GA), encoding candidate solutions as strings that could be manipulated by genetic operators and later decoded into classifier systems. Many of his contributions helped establish the foundations of GAs.
Holland viewed living organisms as problem solvers, emphasizing that natural selection avoids the need to specify in advance all features of a problem and the actions required to solve it~\cite{adaptationinnaturalartificial}. In later work, \citeonline{holland_genetic_1992} argued that GAs make it possible to explore a wider range of solutions than conventional programs, the same way organisms evolve based on natural selection and reproduction. 
In his paper, individuals are encoded as strings and manipulated accordingly: crossover and mutation are performed by swapping values between two strings or flipping individual bits. This encoded representation requires the string to be decoded into a program before it becomes useful, analogous to how DNA encodes an organism but requires expression to yield function.


Meanwhile, Koza proposed a different way of representing and manipulating individuals: instead of evolving string encodings, the representation of programs is done by a hierarchy of functions applied to each other. Genetic Programming (GP) evolves programs directly, using parse trees as a data structure to store functions and terminals~\cite{GPOnTheProgrammingOfComputersByMeansOfNaturalSelection}. 
The first author to propose such structure was \citeonline{cramer1985representation}, described as a simple way to reproduce algorithmic language and making easy for the offspring to remain correct after manipulation by the genetic operators \cite{back1997handbook}.
By representing individuals as trees, subtrees can be swapped through crossover, producing offspring that inherit parts of both parents. Koza also demonstrated that GP could be used not only for evolving computer programs but also for discovering mathematical expressions --- laying the foundation for SR. One of the key limitations of early GP, however, was that the constants (\textit{i.e.}, free numerical parameters) were chosen randomly using a process called Ephemeral Random Constant (ERC), which is limited given that randomly sampling from an infinite space of possible values is unlikely to work optimally.

It is important to notice that GP and GA are distinct\footnote{It is arguable that most of the theoretical work was done under the idea of GA, whereas GP has fewer theoretical basis and often borrow concepts proven to work well in GA to incorporate within their algorithms, although without direct equivalence due to non-linear way of encoding programs.}\footnote{One famous term to refer to optimization and search methods inspired by natural evolution is \textit{evolutionary algorithms} (EAs), often used as an umbrella term for GA and GP.}. GP is focused on evolving the structure, while GA evolves a linear (\textit{e.g.}, a string) encoding used to build solutions by following construction rules based on each position's value. GA is also often used for black box parameter optimization, where the goal is to optimize a set of free parameters (\textit{i.e.}, real values) for a black-box function when gradient descent is not applicable, since it scales for several variables and requires minimum interaction with the inner black-box system.

Extensions of these ideas followed. Memetic algorithms~\cite{moscato1989evolution} incorporated local learning into the representation, arguing that even though our DNA dictates our characteristics, the environment also influences their expression in ways that can alter the structure. 
Nowadays, in the context of SR, this can be interpreted as having parameter tuning within individuals. The interplay of global and local search yields more successful results \cite{Katoch2020,moscato2003gentle}.

Parent selection is also important, and prior work studied its role in maintaining a selective pressure \cite{selective_pressure} in GAs, where parent selection is seen as independent of the search space, but it impacts the exploration-exploitation trade-off. B\"{a}ck concluded that only a few parent selection proposed back at that time, such as the tournament selection, are actually useful.

The genetic operators, \textit{i.e.}, crossover and mutation, were widely accepted at early stages, but started to be questioned as the field progressed. 
The effectiveness of crossover is a long-running discussion in the field, as stated by \citeonline{Ye_2020} in a contemporary paper, when they analyzed mutation-based algorithms.
This is particularly true because swapping two subtrees at random may produce an individual that performs worse than both parents, especially because there is no one-to-one correspondence between parts exchanged, as observed by \citeonline{SymTree}.
Some methods were proposed by relying solely on mutation \cite{Beyer2002} such as evolutionary strategies, or local search, such as simulated annealing~\cite{simulatedannealing}, and this discussion continues to this day, with theoretical works attempting to give mathematical proofs of the benefits of using crossover \cite{dang_proof_2023}.

The ``no free lunch'' theorem~\cite{wolpert_no_1997} later highlighted that no optimization algorithm is universally superior, emphasizing the need for problem-specific insights and benchmarks. The theorem states that when performance is averaged uniformly over all possible functions, no algorithm can outperform any other, including a blind search (\textit{i.e.}, randomly selecting the algorithm's next steps), when it comes to black-box optimization algorithms, such as GAs. In other words, all algorithms have identical performance when evaluated over the entire space of possible problems. This also suggests that it is not advisable to compare algorithms by their performance on few problems, and indicates that it is important to have problem-specific knowledge into the behavior of the algorithm.

Over time, researchers recognized more limitations of early SR. For example, \citeonline{Keijzer2004} noted the importance of scaling data, and proposed a modification by scaling the target data to improve the evaluation of expressions, which alleviated the need of finding the exact values for the free parameters, since parameter tuning was not a common approach yet.
A benchmark of different strategies related to linear scaling was published by \citeonline{chen_relieving_2023}, comparing four strategies for relieving genetic programming from the burden of parameter learning by modifying the fitness measurement. However, their methods only shift the difficulty away from genetic programming without necessarily addressing the parameter optimization problem.

Meanwhile, specialized systems such as Eureqa~\cite{schmidt_distilling_2009} popularized SR by successfully discovering governing equations from experimental data, such as invariants in the chaotic double pendulum system. Schmidt work helped making SR known outside its field by originally releasing Eureqa as a free (though closed-source) software, and his work also received praise in the press. Their SR method was based on GP, and was designed to work with noisy experimental data.
Schmidt argued that without any additional information, system models, or theoretical knowledge, the search produced several analytical expressions that respects known physical laws directly from the data, achieving meaningful and non-trivial invariantes with connection to natural physics. 

The rapid growth of evolutionary computation also gave rise to a proliferation of algorithms inspired by various biological processes.
This diversity has helped push the field forward, but also has made the landscape difficult to assess, with a plethora of new bioinspired algorithms that somehow resemble the classical ideas discussed above. Some authors argue that most of these methods reduce to a common set of components, raising the question of whether new biologically inspired variants contribute real progress or simply repackage existing ideas.
\citeonline{molina_paradox_2025} argue that the field suffers from inadequate benchmarking, problem-specific overfitting, and redundant proposals motivated primarily by metaphors. They call for methodological guidelines, more heterogeneous benchmarks, and higher-quality experimental comparisons, especially against well-established baselines like GA/GP.

\citeonline{10.1145/3205455.3205539} and the follow-up work~\citeonline{srbench} also helped bring SR to broader attention by proposing a robust benchmark called SRBench, and by publishing extensive results in high-profile venues. \citeonline{cranmer2023interpretablemachinelearningscience} further contributed to the accessibility of SR by releasing and promoting a widely used SR library called PySR. 

Despite these challenges, GP remains central to many SR methods. Recent systems such as Operon~\cite{operon} integrate non-linear parameter optimization into GP, and has been widely recognized as the state-of-the-art algorithm by many researchers of the field since it's publication. Different heuristics emerged, such as deterministic alternatives like FFX~\cite{McConaghy2011} that attempt to avoid the complexity of GP heuristics. Neural network-based methods, such as EQL~\cite{martius_extrapolation_2016}, were also introduced to provide differentiable architectures for equation discovery.

Beyond the algorithms, the SR community has grown into an active but relatively niche research area, with conferences such as GECCO, EuroGP, EvoStar, and GPTP providing dedicated workshops. 

\section{Problem formulation}~\label{sec:background:problem_formulation}

Formally, let $D$ be a dataset consisting of $d$ observed points $\left\{(\mathbf{x}_i, y_i) \right\}_{i=1}^{d}$, where $\mathbf{x}$ is an $n$-dimensional feature vector, and $y_i$ is the target value for each observation $\mathbf{x}_i$. 
Let $\mathcal{X} = \left [ \mathbf{x}_{1}, \mathbf{x}_{2}, \ldots, \mathbf{x}_{d} \right ] \in \mathbb{R}^{d \times n}$ denote the feature matrix, and $\mathbf{y} = \left [ y_{1}, y_{2}, \ldots, y_{d} \right ] \in \mathbb{R}^d$ the target values vector. 

Let $\mathcal{H}$ denote the search space (or the hypothesis space) of symbolic expressions, and let each candidate model be denoted by $(f,\theta)$ where $f\in\mathcal{H}$, and $\theta\in\mathbb{R}^p$ are its tunable parameters. 

The goal of symbolic regression is to search through $\mathcal{H}$ for a function $\hat{f}$ and a set of parameter values $\hat{\theta}$ such that $\hat{f}(\mathcal{X}, \hat{\theta}) : \mathbb{R}^n \rightarrow \mathbb{R} \approx \mathbf{y}$, with $\hat{\theta} \in \mathbb{R}^p$ being a $p$-dimensional vector of real values for each free parameter in $\hat{f}$, that minimizes one or more objective functions.

Unlike traditional regression techniques that assume a fixed model form, SR simultaneously searches for both the optimal model structure and its best parameters. The search for $\hat{f}$ adds an additional degree of freedom compared to traditional parametric models --- which begin with a fixed function form $f$ and initial parameters $\theta$, that is, $f(x, \mathbf{\theta})$, and focus only on finding the optimal $\mathbf{\hat{\theta}}$ that minimizes a loss function.

\subsection{Model representation}

SR models are commonly represented as trees. Each internal node encodes a mathematical operator (\textit{e.g.}, $+$, $\times$, $\sin$), while the leaves contain either input \textit{variables} or numeric \textit{constants} (often referred to as \textit{parameters}). 

\begin{figure}[htb]
    \centering
    \caption[Evaluation of a parse tree (left) and a single node (right).]{
    Evaluation of a parse tree (left) and a single node (right). The dashed arrows traverse the tree top-down. Each time a node with children is encountered, its children are evaluated first (\textit{i.e.}, through recursive calls). Once their values are returned (continuous arrows), the node's operator is applied to the children's returned values.}
    \label{fig:background:evaluating}
    \includegraphics[width=0.75\linewidth]{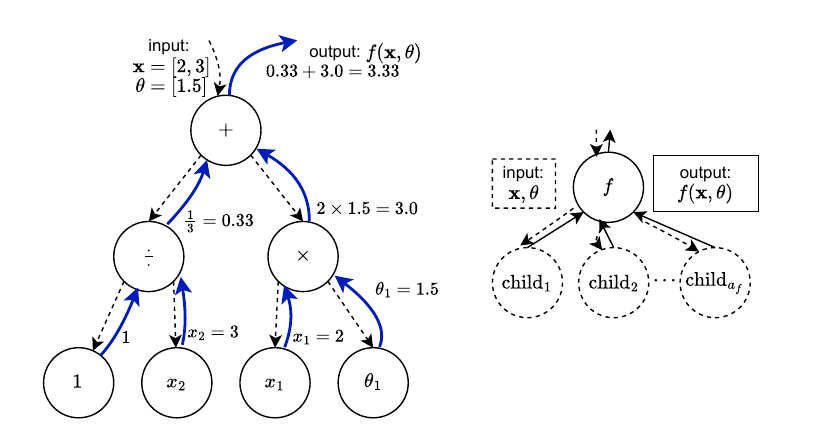}
    \legend{Source: Elaborated by the author.}
\end{figure}

Expressions are evaluated recursively from the leaves to the root. This representation is highly flexible, as virtually any algebraic structure can be encoded, but it also creates a vast search space. Figure \ref{fig:background:evaluating} illustrates the evaluation of a tree representing the function $\frac{1}{x_2} + x_1 \times \theta_1$.

The choice of representation has a strong impact on the efficiency of SR. 
Alternatives include linear representations (\textit{e.g.}, weighted sums of interaction terms \cite{10.1162/evco_a_00285} or smaller expression trees \cite{cava2018learning}), graph-based encodings \cite{randall2022bingo}, and grammar-based rules \cite{espada2022data} that restrict the search space. Representation therefore determines both the expressive power and the tractability of SR methods.

\subsection{Objective functions}\label{subsec:background:objective_functions}

Different metrics can be used in SR as objective functions.
Prediction error is commonly measured with the Mean Squared Error (MSE) between predictions $f(\mathcal{X}, \mathbf{\theta}) = \mathbf{\widehat{y}}$ and expected values $\mathbf{y}$ (Eq. \ref{eq:background:mse}):

\begin{equation}~\label{eq:background:mse}
    \text{MSE}(\mathbf{\widehat{y}}, \mathbf{y}) = \frac{1}{d} \sum_{i=1}^{d} \left ( \widehat{y}_i - y_i \right )^2
\end{equation}

Besides MSE and its variants ---namely Root Mean Squared Error (RMSE), or Mean Absolute Error (MAE)---, the coefficient of determination $R^2$ (Eq. \ref{eq:background:r2}), which measures the amount of variance explained by the model, is also commonly used.
$R^2$ is often prefereable due to its values having a maximum value of 1, and because it allows a natural interpretation that values close to zero means the model performs no better than prediting the mean.

\begin{equation}~\label{eq:background:r2}
    R^2 = 1 - \frac{\sum_{i=1}^{d} \left(\widehat{y}_i - y_i\right)^2}{\sum_{i=1}^{d} \left(y_i - \bar{y}\right)^2},
\end{equation}

\noindent where $\bar{y}$ is the mean of the observed target values.

\citeonline{haut_correlation_2022} tested correlation-based objectives, first optimizing $R^2$ and then applying linear scaling. However, correlation rarely outperformed RMSE, and they concluded RMSE should remain the preferred metric. Correlation is not popular since it captures only associations and ignores scale, being not directly related to predictive error, often failing in selecting high performing models.

Another reason to prefer MSE is that parameter optimization requires differentiable loss functions, and MSE fulfills this requirement, enabling gradient descent or black-box optimization of parameters. Thus, while optimization is generally performed with respect to MSE, $R^2$ remains convenient when comparing models across datasets with different target values scales, as it is invariant to rescaling of $y$, where MSE is not directly comparable.

When it comes to model size (often used as a proxy for interpretability), it is often the secondary objective aiming to improve expression interpretability. \citeonline{10.1145/3236009} states the interpretability is often measured as size of expressions, a vague notion that may not yield good results for some domain-specific problems.

The model size is usually measured as the number of terms in the expression tree, counting every symbol --- operator, constant, features, and multiplicative weights.
It is arguable that there is considerable lack of consensus in the field, as \textit{size}, \textit{complexity}, and \textit{simplicity} are often used interchangeably, but also often have different calculations.
Alternatives to count number of terms are the concept of complexity used by \citeonline{cava2018learning}, or parsimony used by \citeonline{schmidt_distilling_2009}.

The complexity definition provided by \citeonline{cava2018learning} calculates the complexity of a node $n$ with $k$ child nodes as the recursive combination of the complexity of its children with the complexity of the root operator, as shown in Eq. \ref{eq:background:complexity}:

\begin{equation}~\label{eq:background:complexity}
    C(n) = c_n \cdot (\sum_{a=1}^k C(a)),
\end{equation}

\noindent where $c_n$ is an arbitrary positive value ($c_n \in \mathbb{Z}^+$) defined for each possible operator, representing the difficulty of understanding its behavior by humans.

The complexity of each feature increases exponentially with depth and tends to be higher when deeper nodes involve more complex operators. This design discourages the use of complicated operators, promoting more interpretable models.
This penalizes nested, hard-to-interpret operators more strongly. Note that this approach can also affect simple operators such as $+$, which may be nested in linear combinations. To prevent linear models from being unnecessarily penalized, commutative operators (\textit{i.e.}, $+$ and $\times$) may be allowed higher arity, reducing the complexity of linear combinations compared to strictly binary trees.

Parsimony (measured as the inverse of number of terms in the expression) was used by \citeonline{schmidt_distilling_2009} to indicate equations with high complexity (small parsimony value) or simplicity (high parsimony value).

Another approach to balance accuracy and complexity in symbolic regression is the Description Length (DL) principle, grounded in information theory and widely discussed in \citeonline{ExhaustiveSR}. The authors point out that ``the best functional representation of a dataset is the one that compresses the most, requiring the fewest units of information to communicate the data with the help of the function''.
It is calculated as (Eq. \ref{eq:background:mdl}):

\begin{equation}~\label{eq:background:mdl}
    L(D) = L(f) + L(D|f),
\end{equation}

\noindent where $L$ denotes codelength, $f$ is the functional hypothesis, and $D$ the dataset. $L(D|f) = -\log{\mathcal{L}(D|\hat{\theta})}$ denotes the optimum encoding of the residuals that best fits the data, and $L(f)$ denotes the units of information needed to represent it.
DL penalizes overly complex hypotheses while rewarding accurate fits, striking a trade-off between complexity and predictive power. As the amount of data increase, DL tends to favor more accurate solutions even at the cost of higher complexity, since the simplicity penalty becomes relatively less influential. This makes DL a conservative criterion that avoids overfitting on small datasets but leverages richer datasets to capture more complex structures.

Some other attempts of measuring interpretability of models was proposed by \citeonline{vladislavleva_order_2009}, with a polynomial-based complexity measure, though it has not been widely adopted. They count the minimum degree of an approximation polynomial function to the expressions.

Due to the large variety of metrics that emphasize different facets of mathematical expressions generated by SR algorithms, the community does not have a unified consensus on which metric should be used for predictive accuracy or model interpretability. As a result, researchers often rely on whichever metric is most suitable for their specific application or empirical benchmark.

\subsection{Multi-Objective optimization}

SR can optimze both prediction error and model size by being implemented as a multi-objective optimization problem, or by minimizing prediction error while indirectly controlling size through hyperparameters or the evolutionary process.

Multi-objective optimization has been widely studied in SR by explicitly minimizing two or more objectives simultaneously. For instance, \citeonline{vladislavleva_order_2009} explored SR in a multi-objective scenario with different complexity measures as secondary objective.
Beyond error and complexity, alternative objectives have also been proposed, such as \citeonline{Schmidt2010} considered the \emph{age} of solutions as an additional objective to encourage diversity.

In an (explicit) multi-objective setting, the fitness of a candidate solution $f$ is defined as a vector, typically combining one loss function with one penalty for model size or complexity.

Consider the hypothesis space $\mathcal{H}$ and $m$ objective functions. Let $\mathbf{F}$ be the fitness vector whose values are each of the scalar-valued objective functions $\mathcal{O}_i$, with $i \in \{1, 2, \ldots, m\}$, where the first value is called the dominant objective, and the remainder are secondary objectives (Eq. \ref{eq:background:moo}):

\begin{equation}~\label{eq:background:moo}
    \mathbf{F}(f,\theta) = \bigl[\mathcal{O}_1(f,\theta), \mathcal{O}_2(f,\theta), \dots, \mathcal{O}_m(f,\theta)\bigr]^T,
\end{equation}

\noindent for example, MSE and complexity (Eq. \ref{eq:background:moo_examples}):

\begin{equation}~\label{eq:background:moo_examples}
    \begin{split}
        \mathcal{O}_1(f,\theta) &:= \text{MSE}\bigl(f(\mathcal{X},\theta),\mathbf{y}\bigr), \\
        \mathcal{O}_2(f,\theta) &:= \text{Complexity}(f).
    \end{split}
\end{equation}

The corresponding multi-objective optimization problem is (Eq. \ref{eq:background:mo-problem}):

\begin{equation}\label{eq:background:mo-problem}
    \min_{f\in\mathcal{H},\ \theta\in\mathbb{R}^p} \ \mathbf{F}(f,\theta).
\end{equation}

One common way of including multiple objectives in the GP framework is by using the non-dominated-sorting approach introduced in the NSGA-II algorithm by \citeonline{NSGA-II}, where individuals are ranked according to dominance relationships, and non-dominated individuals are preserved in the population.
The NSGA-II uses the definition of Pareto dominance to sort the entire population based on their fitness vector $\mathbf{F}$, selecting the non-dominated solutions to survive to the next generation.

The pareto dominance is defined for two objective vectors $\mathbf{u},\mathbf{v}\in\mathbb{R}^m$. Then, $\mathbf{u}$ \emph{dominates} $\mathbf{v}$ (write $\mathbf{u}\prec\mathbf{v}$) if and only if:

\begin{equation}
    \forall i\in\{1,\dots,m\}:\ u_i \le v_i \quad\text{and}\quad \exists j:\ u_j < v_j.
\end{equation}

In other words, $\mathbf{u}$ is at least as good as $\mathbf{v}$ in all objectives, and strictly better in at least one. This relation is a preorder and can be read as ``preferred to''. 
The Table \ref{tab:background:pareto_relationships} shows the different Pareto dominance relationships two fitness can have, based on \citeonline{ehrgott_multicriteria_2005,emmerich_tutorial_2018}.

\begin{table}[htb]
    \centering
    \caption{Conditions of Pareto dominance in multi-objective optimization.}
    \label{tab:background:pareto_relationships}
    \footnotesize
    \begin{tabular}{lcp{8cm}}
        \toprule\midrule
        \textbf{Pareto Dominance} & \textbf{Condition} & \textbf{Description} \\
        \midrule
        
        \makecell{Strict Dominance \\ ($\mathbf{u} \prec \mathbf{v}$)}
        & \makecell{$\forall i,\; u_i \leq v_i \;\wedge\;$ \\ $\exists j,\; u_j < v_j$}
        & Solution $\mathbf{u}$ is strictly better than $\mathbf{v}$ if it is no worse in \emph{all} objectives and strictly better in at least one objective. \\

        \midrule
        \makecell{Weak Dominance \\ ($\mathbf{u} \preceq \mathbf{v}$)} 
        & $\forall i,\; u_i \leq v_i$ 
        & Solution $\mathbf{u}$ is weakly dominant if it is no worse than $\mathbf{v}$ in all objectives (it may be equal in all of them). \\

        \midrule
        \makecell{Equivalence \\ ($\mathbf{u} \equiv \mathbf{v}$)} 
        & $\forall i,\; u_i = v_i$ 
        & Both solutions achieve exactly the same performance on all objectives, meaning neither dominates the other. \\

        \midrule
        \makecell{Non-dominance \\
        ($\nexists\; \mathbf{u} \prec \mathbf{v} \;\wedge\; \nexists\; \mathbf{v} \prec \mathbf{u}$)} 
        & --- 
        & Neither solution dominates the other: one is better in some objectives, but worse in others. Both are part of the Pareto front. \\

        \midrule\bottomrule
    \end{tabular}
    \legend{Source: Elaborated by the author.}
\end{table}

The Pareto front is the set of all non-dominated individuals in the population, and represent individuals with different trade-offs between the objectives, without being dominated --- they are superior in at least one objective. These solutions represent the best trade-offs among the considered objectives. Figure~\ref{fig:background:pareto_front} illustrates an example of a Pareto front, where the optimal trade-off curve separates feasible solutions from dominated ones.

\begin{figure}[htb]
    \centering
    \caption[Illustration of a Pareto front (dotted line).]{Illustration of a Pareto front (dotted line). The Pareto front consists of the set of non-dominated individuals in a multi-objective setting. Final solutions are usually chosen from this set. Dominated individuals are worse on all objectives when compared with any solution from the Pareto front.}
    \label{fig:background:pareto_front}
    \includegraphics[width=0.5\linewidth,clip=true,trim={0.5cm 0.5cm 0.5cm 0.5cm}]{Capitulos/Background/floats/pareto_front.pdf}
    \legend{Source: Elaborated by the author.}
\end{figure}

The NSGA-II will select the entire Pareto front to compose the next generation, and the remaining slots ---so the next generation has the same size as the previous one--- is filled with the dominated individuals based on their crowding distance to the Pareto front.
There is usually a decision criteria to pick from the final population an individual with a good trade-off between the objectives, for example the pseudo-weight vector approach \cite{mcdm}.

\subsection{Genetic programming}~\label{sec:background:gp}

\begin{figure}[htb]
    \centering
    \caption[Flowchart of common steps of evolutionary algorithms.]{Flowchart of common steps of evolutionary algorithms. Each iteration corresponds to a \textit{generation}. Individuals are randomly generated and then iterativally refined through a process that assemble a random search, but with selection and survival steps helping guiding the search towards promissing regions of the hypothesis space.}
    \label{fig:background:EAs}
    \includegraphics[width=0.6\linewidth,clip=true,trim={0 0.5cm 7.5cm 1cm}]{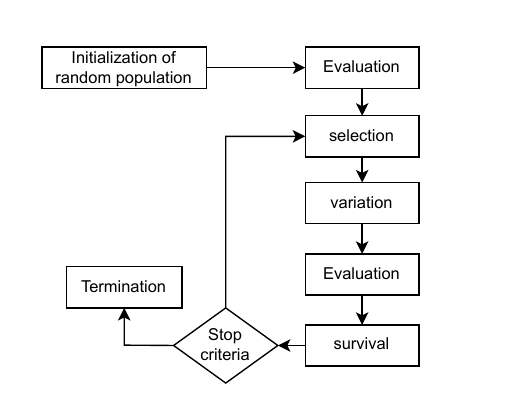}
    \legend{Source: Elaborated by the author.}
\end{figure}

Genetic Programming is the primary heuristic for implementing symbolic regression. 
Figure~\ref{fig:background:EAs} summarizes these steps in a flowchart, illustrating the iterative loop of evolutionary algorithms.

It maintains a population of candidate models and iteratively improves them through selection, variation, and survival steps, aiming to produce high-quality solutions.

\textbf{Initialization.}  
A randomly initialized population is first created in GP. Common strategies are the \textit{grow} and \textit{ramped half-and-half} methods \cite{GPOnTheProgrammingOfComputersByMeansOfNaturalSelection}. In the grow method, child nodes are randomly determined as either functions or terminals. Control is limited to maximum depth, and the total number of nodes has no fine-grained control. In the ramped half-and-half method, grow is alternated with the generation of full trees.
Unlike \textit{grow} and \textit{ramped half-and-half}, the probabilistic tree creator 2 (PTC2) published by \citeonline{PTC2} is designed to explicitly control tree size and depth to manage expression complexity, and is still used to generate individuals within a given maximum size and depth.
The initialization of parameters in expressions was originally proposed through ephemeral random constants \cite{GPOnTheProgrammingOfComputersByMeansOfNaturalSelection} and remains widely used. However, parameter optimization methods are often applied afterward to improve accuracy, serving as a form of local learning in modern algorithms.

\textbf{Evaluation.} 
The evaluation step asserts that every individual have its fitness calculated using the dataset before proceeding to the next step. This step preceeds selection and survival steps, and can also check for Pareto dominance in multi-objective algorithms. 

\begin{figure}[htb]
    \centering
    \caption[Roulette selection, a parent selection mechanism proposed by John Holland, where probabilities are proportional to fitness.]{Roulette selection, a parent selection mechanism proposed by John Holland, where probabilities are proportional to fitness. Each parent is selected by spinning the wheel.}
    \label{fig:background:roulette}
    \includegraphics[width=0.3\linewidth,clip=true,trim={0.5cm 0.5cm 0.5cm 0.5cm}]{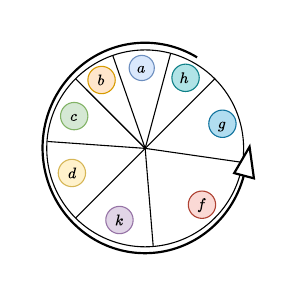}
    \legend{Source: Elaborated by the author.}
\end{figure}

\textbf{Selection.}  
The selection step chooses parents to generate offspring through the \textit{variation} operators. Selection mechanisms can significantly influence exploration-exploitation balance and convergence speed.
Early approaches, such as \textit{roulette selection} \cite{adaptationinnaturalartificial} ---proposed by Holland in the context og GA---, select individuals with probabilities proportional to their fitness, ensuring even poor solutions have a small chance of selection (Figure~\ref{fig:background:roulette}).  

\begin{figure}[htb]
    \centering
    \caption[Tournament selection, where each parent is selected by winning a tournament of $k$ individuals.]{Tournament selection, where each parent is selected by winning a tournament of $k$ individuals. This is the most commonly found parent selection schema today. Candidates are randomly assigned to the tournament with uniform probabilities, and the best performing one is selected.}
    \label{fig:background:tournament}
    \includegraphics[width=0.65\linewidth]{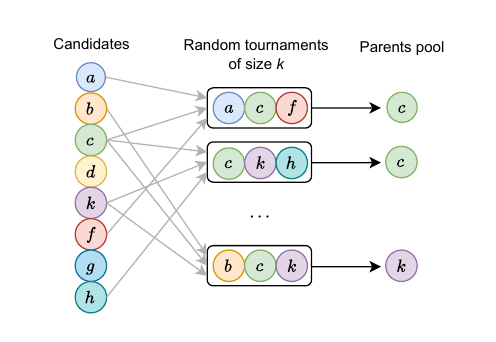}
    \legend{Source: Elaborated by the author.}
\end{figure}

Tournament selection \cite{miller1995genetic,brindle1980genetic} is a popular method, due to its simplicity. In this method, groups of $k \geq 2$ individuals compete as in tournaments, and the best in each group is selected as a parent (Figure~\ref{fig:background:tournament}). Notice that some individuals may never be selected, in the example, $d$ and $g$, although variants without replacement between tournaments exist \cite{goldberg1989messy}.

\textbf{Variation.}  
Variation introduces diversity and drives the search through the hypothesis space. Often the variation operators are the crossover and different types of mutation based on the model representation. Due to the populational aspect of EAs, the random variations helps explore several directions simultaneously in the hypothesis space.

\begin{figure}[htb]
    \centering
    \caption[Examples of applying the variation operators.]{Examples of applying the variation operators. The crossover recombines parts of the two parents, and the mutation performs local changes to one of the parents.}
    \label{fig:background:cx_mut}
    \includegraphics[width=0.75\linewidth]{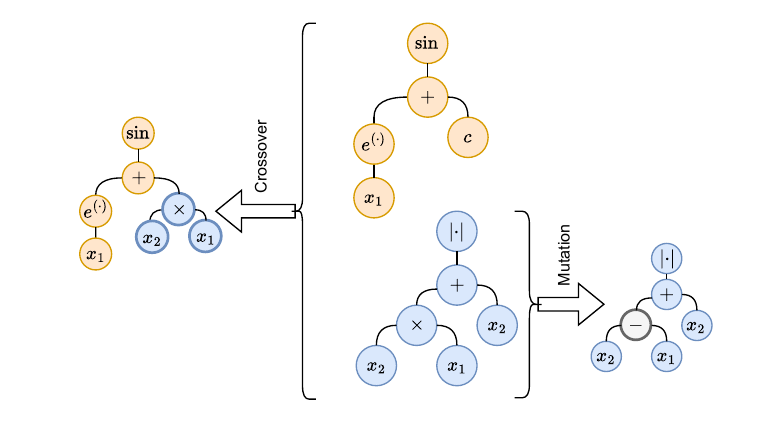}
    \legend{Source: Elaborated by the author.}
\end{figure}

\textit{Crossover} recombines parts of two parents to produce offspring, while \textit{mutation} introduces small perturbations to enable local exploration and novelty (Figure~\ref{fig:background:cx_mut}). Properly designing variation operators is challenging but critical for search efficiency. Implementing meaningful crossover operators is considered a hard task.

\textbf{Survival.}  
The survival step determines which individuals remain in the population, and it is applied after generating the offspring\footnote{Notice that selection and survival are different steps, often performed with different approaches. In this context, selection will pick individuals to mate, while survival will apply a selective pressure where individuals with better fitness are more likely to remain in the population.}. It can be used to balance accuracy and complexity, \textit{e.g.}, through NSGA-II's non-dominated sorting \cite{NSGA-II}. 
Tournament-based survival and elitism are common. Elitism ensures that the best solution found so far is retained for the next generation, regardless of not being selected in any step during the survival phase. Some implementations merge parent and offspring populations before survival, while others rely solely on offspring.

\section{Search and optimization}~\label{sec:background:search}

Beyond function structure search, symbolic models often require parameter optimization to fine-tune constants; good selection mechanisms to avoid selecting overfitted, non-generalizable models; and mechanisms to control bloat and redundancy to avoid overly complex expressions.
This section details the main strategies for optimization and search control that will be explored in this thesis, also pointing out current limitations of these strategies.

\subsection{Non-linear parameter optimization}

Mathematical expressions may contain numeric parameters that influence model predictions.
Optimizing these parameters is non-trivial \cite{Kommenda2019,PrioritizedGrammarEnumeration,kamienny_end--end_nodate}.
In early GP, constants were generated as ephemeral random constants \cite{GPOnTheProgrammingOfComputersByMeansOfNaturalSelection}, which often led to poor coverage of the parameter space.

Leaving the parameter optimization task solely to the evolutive process often results in slow convergence \cite{8393325,chen_relieving_2023} and suboptimal constants \cite{WANG20042453}, since it lacks the capability of obtaining the global optimum.
With that in mind, the practitioner can include parameter optimization steps into the loop.

Later strategies attempted to mitigate the burder of random parameters, such as linear scaling \cite{keijzer2003improving} or enforcing linear models \cite{arnaldo2014multiple,10.1162/evco_a_00285,tir,cava2018learning}. 
When creating linear representations, it is easy to use the minimum least squares method to find global optimum parameters, but the function gets restricted in shape and capabilities of modeling any algebraic function does not hold.

Modern SR systems frequently combine the search for the function structure with dedicated parameter optimization, often relying on non-linear methods such as Levenberg–Marquardt (LM) \cite{levenberg1944method, marquardt1963algorithm}, which has proven effective in many frameworks \cite{ITWithCoeffsOptimization,Kommenda2019,PrioritizedGrammarEnumeration}.
When many parameters are involved, this step can significantly increase the computational cost, as the optimization methods often relies on iterative steps, having to deal with dataset dimensionality and number of simultaneous parameters to tune.

\begin{figure}[htb]
    \centering
    \caption[Flowchart of evolutionary algorithms, highlighting the extra steps required for parameter optimization.]{Flowchart of evolutionary algorithms, highlighting the extra steps required for parameter optimization. Parameters must be optimized before selection or survival, ensuring fitness values reflect tuned parameters before selection and survival. }
    \label{fig:background:EAs_with_optimization}
    \includegraphics[width=0.525\linewidth,clip=true,trim={7.5cm 0cm 0cm 0cm}]{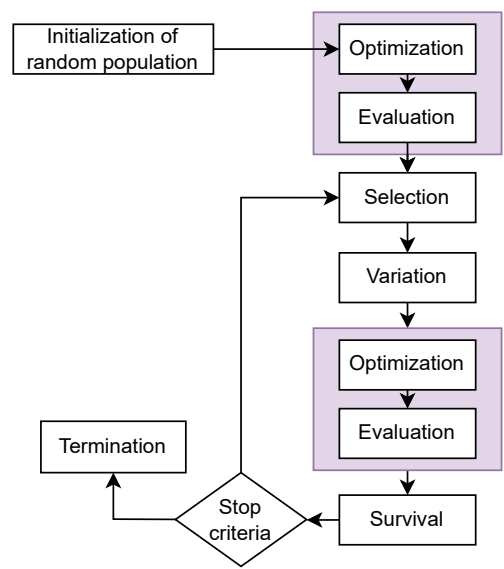}
    \legend{Source: Elaborated by the author.}
\end{figure}

Figure \ref{fig:background:EAs_with_optimization} shows how local parameter optimization can be embedded in the evolutionary cycle, before the evaluation steps.

Alternative strategies have also been proposed. For instance, some divide expressions into building blocks optimized locally before assembling the final model with global optimization \cite{CHEN20181973}, while others use specific heuristics like Low Dimensional Simplex Evolution \cite{LUO20121182}.
Deterministic approaches such as FFX \cite{McConaghy2011} employ regularized regression to directly fit parameters. 
Reinforcement learning–based SR methods also rely on numerical solvers to refine constants \cite{kamienny_end--end_nodate,pmlr-v139-biggio21a,UnifiedFrameworkForDeepSR,shojaee_transformer-based_nodate}, showing that parameter optimization is a universal challenge across paradigms, and cannot be generated in an end-to-end fashion using transformer architectures.

Benchmark studies such as SRBench \cite{srbench} have highlighted the impact of parameter optimization. Operon \cite{operon}, ranked among the most accurate algorithms, systematically associates a parameter with every terminal and optimizes them using a non-linear optimization method, although this often leads to over-parameterized expressions \cite{10.1145/3583131.3590346}. GP-GOMEA \cite{gpgomea,10.1162/evco_a_00278} showed the best trade-off between accuracy and complexity, but still relies on ERCs without subsequent optimization, indicating it may achieve a higher predictive power under parameter optimization. Hybrid frameworks such as PS-Tree \cite{ZHANG2022101061} build ensembles of small SR models fitted with least squares, but the resulting structures are often overly large.

Altogether, these examples illustrate that parameter optimization is a central bottleneck for SR, and iterative non-linear methods like LM have been successfully applied for parameter optimization in SR. Still, some methods relies on linear parameter optimization, and the trade-off between these different approaches are not well studied.

\subsection{Parent selection with \texorpdfstring{$\epsilon$-}{epsilon-}lexicase}

Selection indirectly optimize the solutions by driving the balance between exploration and exploitation in GP-based SR, and have minimum interference with other GP parts.

Classical schemes such as roulette wheel and the tournament selection choose parents based on aggregated fitness values over the training set. While simple and effective, these methods ignore case-specific performance and can lead to information loss due to aggregation, often discarding individuals that excel on difficult subsets of the data. Since less information is available when aggregating fitness values, it tends to underperform selection methods that relies on case-specific performances \cite{10.1145/2576768.2598288}.

\begin{figure}[htb]
    \centering
    \caption[Illustration of the lexicase parent selection process.]{Illustration of the lexicase parent selection process. The entire population enters the selection pool. The test cases are shuffled and chosen one by one, and individuals failing that test --- miss-classifying the sample, in classification problems --- are eliminated from the pool. The process repeats until only one individual remains. If all test cases are exhausted, a random individual from the remaining pool is selected.}
    \label{fig:background:clf_lexicase}
    \includegraphics[trim={0.5cm 0cm 0.5cm 0cm},clip=true,width=0.9\linewidth]{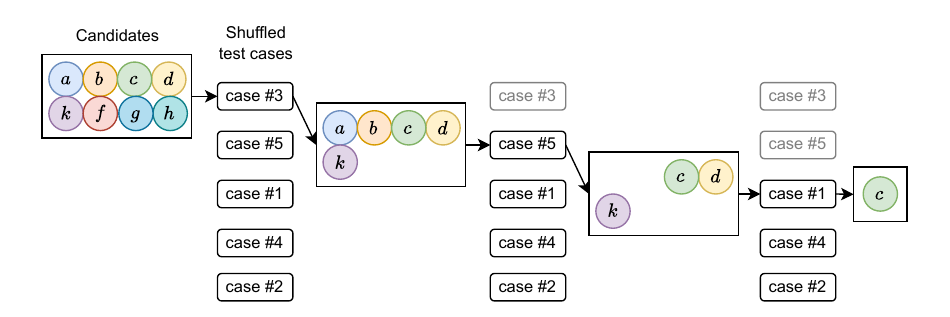}
    \legend{Source: Elaborated by the author.}
\end{figure}

To address this, \citeonline{10.1145/2330784.2330846}, introduced the lexicase selection, rewarding individuals that perform well on challenging test cases, assessing performance one case at a time. 
Figure \ref{fig:background:clf_lexicase} depicts the process of lexicase selection.

In lexicase selection, in order to select one parent, the entire population goes into a selection pool. In each round, candidate parents are filtered based on performance on a randomly ordered sequence of test cases. Then, a random test case is selected, and individuals who fail that test are eliminated from the pool. Randomly selecting the following test case is repeated until the pool contains only one individual. If the procedure runs out of test cases, a random individual from the pool is returned with uniform probability. 

For regression tasks, $\epsilon$-lexicase extends this idea by allowing tolerance around the best error per case, originally designed using the Median Absolute Deviation \cite{la_cava_epsilon-lexicase_2016}. This variant has shown state-of-the-art performance in program synthesis and symbolic regression \cite{ding_probabilistic_2023}.

The canonical $\epsilon$-lexicase selection introduced multiple variants for threshold calculation, including dynamic or semi-dynamic thresholds, and the properties of this parent selection method could be further explored, aiming to design variants with a more natural interpretation.

\subsection{Simplification of symbolic regression expressions}

\begin{figure}[htb]
    \centering
    \caption{Illustration of several equivalent expressions.}
    \label{fig:background:simp_examples}
    \includegraphics[width=\linewidth]{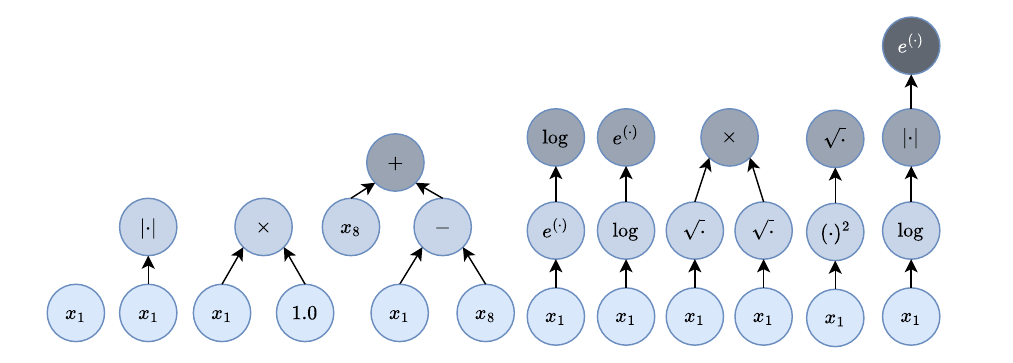}
    \legend{Source: Elaborated by the author.}
\end{figure}

A long-standing challenge in SR is the uncontrolled growth of expressions without improvements in accuracy, often referred to as \textit{bloat}.
Symbolic simplification can be a key to mitigating this, as many candidate models are functionally equivalent despite increasing structural complexity with redundant operations (as depicted in Figure \ref{fig:background:simp_examples}). 

Several approaches have been proposed to alleviate this through simplification of expressions. These include automatic simplification during evolution through random deletions \cite{helmuth_improving_2017}, algebraic rewrite rules \cite{kulunchakov_creation_2017}, constrained search spaces \cite{10.1162/evco_a_00285}, Bayesian model selection criteria \cite{bomarito_bayesian_2022}, and equality saturation \cite{10.1145/3583131.3590346}.
Simplification not only reduces model size but can also improve generalization: \citeonline{helmuth_improving_2017} showed that GP programs failing to generalize before simplification often did so afterward, likely due to the presence of unnecessary code that did not expressed effects in prediction during training, but failed to handle new test instances.

Interestingly, some researchers argue not all redundancy is harmful. Neutrality ---changes in genotype that leave phenotype and fitness unchanged--- may promote broad exploration of the search space \cite{neutrality}. Recent work argues that neutrality can be a decisive factor in search success, with linear GP exhibiting structural introns and tree-based GP showing more implicit neutrality \cite{winkler_how_2024}. From this perspective, what appears as bloating might actually support effective exploration, suggesting that simplification should not always be applied since it can harm the neutral space.

Another risk associated to a lack of simplification is overparameterization. For example, \citeonline{10.1145/3583131.3590346} argue that models with too many parameters are difficult to interpret and suffer from ill-conditioning, which slows down the parameter optimization. They propose simplifying expressions prior to optimization in order to avoid excessive parameterization, showing that this produces simpler models with fewer parameters, thus motivating the importance of simplification procedures.

Although all of the different simplification methods have their own benefits they often require special rules or correspond to simple heuristics, having limitations, mainly the additional computational cost and the need to manually write the algebraic simplification rules. Also, one needs to understand the relationships of simplification methods with the neutral space, as whether early simplification may hinder the search.

\section{Evaluation and benchmarking}~\label{sec:background:benchmarking}

The evaluation of symbolic regression methods has long been a challenge. Benchmarks are essential for comparing algorithms, understanding their strengths and weaknesses, and tracking progress in the field. However, running a benchmark requires standardized datasets, consistent protocols, and clear reporting; without these, comparisons risk being misleading or even counterproductive. At the same time, \citeonline{radwan_comparison_2024} warned that optimizing methods for benchmark performance, rather than for broader scientific utility, risks making benchmarks less useful over time.

Historically, benchmarking in SR has been limited. Early studies often compared new ideas against a few outdated baselines on small and overly simple problems. \citeonline{10.1145/3205455.3205539} surveyed SR papers up to $2018$, and acknowledged previous efforts to conduct benchmarks for SR methods, also highlighting the lack of rigor, showing that most benchmarks relied on trivial datasets with little standardization. This lack of rigor risked diminishing the credibility of SR results outside its immediate research community. To address this, they compared four SR algorithms against a suite of ML methods on over $100$ datasets from PMLB~\cite{Olson2017PMLB,romano2021pmlb}, calling for broader and more systematic evaluation.

Building on this, \citeonline{srbench} introduced SRBench, the first large-scale collaborative benchmark for SR, evaluating $14$ modern algorithms (GP-based, deterministic, and deep learning) across hundreds of datasets, and emphasized that benchmarks must remain living resources, continuously updated to reflect new challenges and use cases.
The authors emphasized that defining a ``state of the art'' should not be the sole goal of research, but argued that empirical evidence is nonetheless crucial for guiding future work.
Each algorithm was run multiple times with evaluation budgets defined by function evaluations, though this proved difficult to apply uniformly across methods with very different runtimes.

The release of SRBench have motivated further work. \citeonline{de2024srbenchplusplus} benchmarked a smaller set of recent algorithms on artificially generated datasets designed to probe specific capabilities: accuracy, robustness to noise, feature selection, extrapolation, and interpretability. Using runtime limits instead of evaluation counts, they encouraged more efficient implementations.

Several domain-specific benchmarks have also been proposed. For instance, \citeonline{egklitz2020} emphasized open-source principles, benchmarking four SR methods on nine datasets (five synthetic, four real-world). Similarly, \citeonline{esg1352} evaluated five SR algorithms in the context of civil and construction engineering, where equations are often derived via linear regression. More recently, \citeonline{thing2025cp3} developed a unified tool for benchmarking SR in cosmology, testing 12 methods on 28 datasets, finding that most methods performed poorly in this domain, highlighting the gap between generic benchmarks (\textit{e.g.}, PMLB) and specialized scientific applications. 

\subsection{SRBench: the most famous symbolic regression benchmark}

The SR community has broadly adopted SRBench as the standard for evaluating and benchmarking approaches. It enables comparisons with standard ML regression models, enforces a common API, and provides reproducible pipelines\footnote{\url{https://cavalab.org/srbench/}}. 

SRBench drew on PMLB and a ground-truth track based on more than $100$ physics equations from the Feynman lectures~\cite{feynman2006feynman,feynman2015feynman}, curated by \citeonline{udrescu_ai_2020}. 

Further discussions emerged in top-tier conferences related to SR (such as GECCO), with criticisms about aggregated views of the results~\cite{dick2022genetic,tir}, arguing that aggregated results obscure important differences between methods.
Similarly, \citeonline{matsubara2024rethinking} criticized SRBench's reliance on the Feynman dataset, noting that its uniform sampling poorly reflected real-world physics. They proposed alternative sampling strategies (normal or logarithmic) and tested six SR methods under this setup.

Reproducibility has historically been a challenge due to the stochastic nature of evolutionary algorithms and inconsistent experimental protocols. Recent initiatives such as SRBench provide standardized benchmarks, datasets, and metrics to enable fair comparisons between methods. Still, new algorithms were released since SRBench publication, and there are improvements recognized by the SR community that can be addressed, providing a next iteration of this benchmark.

\section{Recent trends}~\label{sec:background:recent_trends}

Recent research has explored neural and transformer-based approaches as alternatives or complements to GP. \citeonline{wu_evolutionary_2024} reviewed the intersection of large language models (LLMs) and Evolutionary Algorithms (EA), highlighting the potential for LLMs to guide intelligent searches and for EA to provide global optimization in black-box settings.
Some arguments in favor of using LLMs to perform the search is that it can perform sequential changes, better approximating changes that humans would make, overcoming the GP limitation of exploring changes by random \cite{lehman_evolution_2022}.

End-to-end neural approaches, \textit{e.g.}, NeSymRes~\cite{pmlr-v139-biggio21a}, E2E~\cite{kamienny_end--end_nodate}, have been explored, but still face limitations in robustness and generalization, often requiring additional optimization steps such as BFGS or Monte Carlo Tree Search~\cite{shojaee_transformer-based_nodate}. Hybrid frameworks, such as uDSR~\cite{UnifiedFrameworkForDeepSR}, attempt to integrate pre-training, transformers, GP, and linear models into a single pipeline.

These hybrid approaches suggest that evolutionary components remain intrinsically valuable for the types of problems where SR excels, as the most promising transformer-based alternatives are in fact hybridizations of GPs and transformers. It can be argued that GP retains a unique advantage: its ability to discover genuinely novel solutions rather than merely reproducing patterns already present in the training data. This advantage arises from GP's stochastic exploration of the search space. While LLM-based models sample according to prior probabilities learned through token processing ---thereby biasing their modifications toward patterns seen during training--- GP can introduce random changes across many individuals, enabling a broader and more exploratory search.

The rise of LLMs is particularly relevant since program synthesis has become a key application area for these models. However, transformer-based approaches still have many limitations familiar from earlier SR work: they often struggle with noise, generalization, and robustness. 
Further evidence for this view was provided by \citeonline{radwan_comparison_2024}: although recent advances have attempted to combine neural networks and SR, their study showed that Operon, used off-the-shelf, still outperformed two recently proposed SR methods based on end-to-end transformers. In the same line, different pre-trained models for SR were evaluated by \citeonline{voigt2025analyzinggeneralizationpretrainedsymbolic}, and it was found that they did not achieve competitive results and failed to generalize to data outside the training domain.

Some neural network-based SR methods, such as AIFeynman \cite{aifeynman}, require a large amount of data, with the authors using up to one million samples to train the model. Transformers require a large pre-training step, and the distribution over equations in the pre-training influences the prior during inference, which can play a negative role in generalization capabilities \cite{pmlr-v139-biggio21a}, for example \citeonline{constantin_statistical_2024} noticed bias influencing the formulas in physics, and sampling from this distribution may perpetuate established patterns or failing to explore novel formulas.
Meanwhile, GP-based SR has proven to learn from few observations \cite{wilstrup2021symbolic}, useful in problems where data is not widely available, rare, or expensive.
GPs remain robust, data-efficient, and relatively lightweight.

Despite the rapid emergence of neural and transformer-based methods, GP remains central to many SR systems. 
Well-established GP frameworks such as Operon~\cite{Kommenda2019}, PySR~\cite{cranmer2023interpretablemachinelearningscience}, and GP-GOMEA~\cite{10.1162/evco_a_00278} are still widely used, particularly because modern implementations increasingly combine GP with parameter optimization --- a step that has become almost indispensable for competitive predictive performance. Applications of GP-based SR continue to expand in domains such as physics and healthcare, where the need for interpretable models is especially critical.

\subsection{Taxonomy of contemporary symbolic regression}

\begin{figure}[htb]
    \centering
    \caption[Taxonomy of symbolic regression methods.]{
        Taxonomy of symbolic regression methods, based on their main search strategy. Bayesian methods use prior distributions to sample nodes. Genetic programming approaches use population-based heuristics that manipulate programs, commonly program trees, by mimicking evolutionary variation. Deterministic approaches return the same model given the same data, using heuristics to reduce the search space or enumeration. Neural and deep learning methods use neural networks to sample symbols. There are also hybrid approaches, combining elements of different paradigms.}
    \label{fig:background:taxonomy}
    \includegraphics[width=\linewidth]{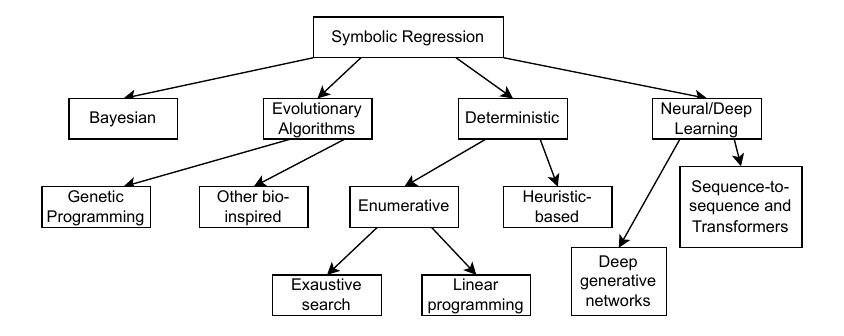}
    \legend{Source: Elaborated by the author.}
\end{figure}

A visual summary of these categories is shown in Figure~\ref{fig:background:taxonomy}\footnote{This taxonomy represents the author's perspective. A different point of view of the taxonomy of SR was presented by \citeonline{Makke2024}, grouping SR methods into regression-based, tree-based, physics inspired, and mathematically inspired. They group the methods by the model representation and application.}.
Among popular SR methods published in the past decade, several of them relies on using GP as the main search heuristics, such as EPLEX~\cite{eplex}, Genetic Engine~\cite{espada2022data}, GPGOMEA~\cite{10.1162/evco_a_00278}, GPZGD~\cite{10.1145/3377930.3390237}, Operon~\cite{Kommenda2019}, PSTree~\cite{ZHANG2022101061}, PySR~\cite{cranmer2023interpretablemachinelearningscience}, TIR~\cite{tir}, and FEAT~\cite{cava2018learning}.
GP remains a popular method, likely due to its closeness to the original idea proposed by Koza, although recent trends have also explored neural networks, such as EQL~\cite{pmlrv80sahoo18a}, deterministic approaches, such as FFX~\cite{McConaghy2011} and SymTree~\cite{SymTree}, iterative processes, such as RILS-ROLS~\cite{Kartelj2023}, transformer architectures, such as E2E~\cite{kamienny_end--end_nodate} and TPSR~\cite{shojaee_transformer-based_nodate}, and hybrids that combine neural networks with GPs, such as uDSR~\cite{LIU2024108986}.

Among the GP variants, the approaches differ in some particularities about their implementation. These differences can be in representation (\textit{e.g.}, TIR uses a different expression representation), constants optimization (\textit{e.g.}, Operon performs non-linear optimization of parameters and assigns weights to all features; GPZGD performs z-score standardization and stochastic gradient descent for parameter optimization; and PSTree relies on ERC), or selection methods (\textit{e.g.}, FEAT and EPLEX uses $\epsilon$-lexicase selection).

While GP implementations vary in aspects such as representation, optimization, or selection strategies, they share the same core principle of evolving symbolic expressions, whereas neural methods instead attempt to learn symbol distributions directly.
For instance, NeSymRes~\cite{pmlr-v139-biggio21a} generates equation skeletons that are later fitted using non-linear optimization, while E2E~\cite{kamienny_end--end_nodate} predicts full expressions including parameters, optimized with BFGS optimization, but limited to a subsampling up to $1024$ data points. TPSR~\cite{shojaee_transformer-based_nodate} enhances transformer decoding with Monte Carlo Tree Search, improving the results at the cost of longer inference. uDSR~\cite{LIU2024108986} unifies pre-training, transformers, GP, and linear models into a single modular framework.

Other methods include physics-inspired approaches such as AI Feynman~\cite{aifeynman}, which uses a neural network but has shown to generalize poorly outside physics~\cite{srbench}, and Bayesian models such as BSR~\cite{jin2020bayesiansymbolicregression}, which sample expressions from learned priors.

The FFX method has a unique property: there are no hyperparameters. Many of the hyperparameters, such as population size or number of generations, are long-term discussions in the field as well, and modern methods often have lots of hyperparameters, with few to none sensitivity analysis performed by their authors, and that may drastically affect the final results, since they may be problem-dependent. 

\subsection{Role of symbolic regression in the machine learning ecosystem}

An alternative to black-box models is the use of \emph{white-box} models, which are considered intrinsically interpretable.
Many GP papers highlight interpretability as an advantage ~\cite{GainingDeeperInsightsInSymbolicRegression,SymTree, ExplainingSRPredictions, ApplyingGeneticProgrammingToImproveInterpretability, Aldeia2022,radwan_comparison_2024}.

\citeonline{Angelis2023} argue that SR reduces the need for domain-specific assumptions, enabling data-driven discovery in the era of big data. While ML has shown great utility in this context, a major barrier remains the difficulty of interpreting models, particularly in fields such as physics.

SR occupies a niche similar to white-box ML models by delivering explicit equations, but with greater flexibility than other parametric regression methods. 
Examples of white-box ML models include linear models, decision trees, and decision rules~\cite{StopExplainingBlackBoxMachineLearningModels}.
The drawback of other white-box methods is that they may deliver lower predictive accuracy compared to black-box approaches~\cite{DoWeNeedHundredsOfClassifiers}, and SR has been proven to yield better results than common white-box regression methods while keeping the expressions relativelly small~\cite{srbench}.
The role of SR in the ML ecosystem is then to provide white-box, flexible models rather than just black-box predictions, and it is often considered a white- or gray-box method \cite{Aldeia2022}.

The assumption that more complex models are inherently more accurate does not always hold~\cite{StopExplainingBlackBoxMachineLearningModels}, and SR is most useful when it yields concise, intelligible expressions. Indeed, expressions exceeding roughly $10$ or $20$ nodes may already strain human interpretability, raising the question of whether SR is worthwhile for problems that can only be approximated by large expressions. As noted by ~\citeonline{radwan_comparison_2024}, SR \textit{holds potential} for interpretable models, but this is not always guaranteed.

\textit{Holding potential} implies that not all equations found by SR are interpretable, and without careful search control, GP often converges to overly complex solutions. Methodological advances are therefore crucial to ensure SR fulfills its promise of producing simple, interpretable models.
Regarding the enumerated challenges mentioned earlier in this chapter for symbolic regression, parameter optimization is important when it comes to interpretability, as the values can be meaningful and help interpreting the model.
Guiding the search through parent selection also plays a major role, as it may hinder the search if not done properly, and can speed up the convergence otherwise.
Finally, it is also important to ensure that the returned expression is as simple as it could be, although model simplification is hard and often overlooked.

\subsection{Applications of symbolic regression}

SR has found applications in many domains where interpretability is critical.
Across domains, SR is often valued not only for predictive performance but for providing scientific insight into the data-generating process.

A comprehensive review of SR applied to physics problems ~\cite{Angelis2023} identified over 30 SR methods and emphasized that explicit equations reduce the need for domain-specific assumptions, facilitating data-driven discovery in the era of big data. 
\citeonline{Reinbold2021} proposed incorporating physical constraints to SR (\textit{e.g.}, smoothness, symmetries) to recover of non-equillibrium systems. 
Similarly, \citeonline{KUBALIK2021115210} integrated prior knowledge to solve three physics problems.
Applications also extend to recovering physical laws from observational data. \citeonline{cranmer2023interpretablemachinelearningscience} curated data from famous physics discoveries and measured how well SR could find the analytical solutions, achieving a high rate of recovery.
\citeonline{Kronberger2022} proposed shape-constrained models to ensure properties such as monotonicity, showing better extrapolation behavior.
\citeonline{martinek44shape} proposed shape-constrained penalization to the parameter optimization step as a way to guide the search more efficiently

SR has also been applied to design new formulas for acoustic transmission~\cite{HRUSKA2025118821}, helping understanding the underlying principles for sonic cristals --- an artificial structure that modify the transmission of waves in a way that is not achieved by any natural medium, thus showing how interpretability helped getting new insights into that.
SR has been applied in astrophysics, where a multiview SR uncovered and validated new candidate equations without known ground-truth~\cite{multiviewsr}: the authors found equations that were validated by analyzing the expression, and the final models helped to grasp some understanding of the underlying physics by recovering some previous equations from the literature and showing promissing new ones.

In aerospace engineering, SR has been used for wind turbine modeling, producing equations that outperformed more complex alternatives in turbulent environments~\cite{LACAVA2016892}, and for predicting aircraft remaining useful life from sparse data, where structurally penalized SR achieved lower error than grey-box approaches~\cite{LIU2024108986}.
In dynamical systems, Schmidt and Lipson~\cite{schmidt_distilling_2009} famously derived equations for the double pendulum. In chemical catalysis, SR identified descriptors for oxide perovskites through multi-objective optimization of predictive error and size, even from small datasets~\cite{Weng2020}.

In medical informatics, SR has been applied to resistant hypertension phenotyping~\cite{la_cava_flexible_2023}, to discovering nonlinear feature transformations that improve survival prediction in heart failure patients~\cite{Wilstrup2022}, and to modeling anatomical dissimilarity in 3D abdominal reconstructions~\cite{10.1145/3205455.3205604}, where SR yielded interpretable solutions, with a smaller degree of overfit when compared to other ML approaches.

\section{Building an unified framework}~\label{sec:background:brush}

This thesis investigates different open challenges in symbolic regression. During the last three years of this thesis, the author has been an active contributor to the Brush framework, continuously integrating his developments into it.
During the development of this thesis, this framework was further extended to include parameter optimization for classification tasks, use $\epsilon$-lexicase parent selection and Pareto dominance survival, and to perform simplification of individuals with the inexact simplification method proposed in this thesis. These additions unify the various facets explored in this research into a cohesive framework.

This section describes the evolutionary backbone of Brush. Brush originally incorporated parameter optimization inspired by Operon, but introduced a different strategy for managing toggleable weights.

\subsection{Evolutionary algorithm}

The evolutionary backbone of Brush is a direct derivation of GP algorithms with parameter optimization (see Fig. \ref{fig:background:EAs_with_optimization} for reference). It performs parameter optimization as proposed by \citeonline{Kommenda2019}, parent selection with $\epsilon$-lexicase as proposed by \citeonline{la_cava_epsilon-lexicase_2016}, and uses non-dominated sorting survival \cite{NSGA-II}. The variation operators, \textit{i.e.}, crossover and mutation, are modified to allow failures and then provide more reliable generation of variations within pre-defined maximum size and depth of individuals.

The algorithm starts with a randomly initialized population $\mathcal{P}$ with $S$ individuals, where programs have sizes uniformly distributed between $1$ and $\text{max}_s$, using the PTC2 algorithm to create the initial expressions \cite{PTC2}.
Then, for a given number of generations $G$, it repeats the steps of selection, variation and survival over the individuals on the population.
Selection step is repeated until $S$ individuals are selected as parents $\mathcal{P}'$. 
After selection, the parents undergo variation to generate new individuals.

\begin{figure}[htb]
    \centering
    \caption[How individuals are created in Brush.]{How individuals are created in Brush. Given a set of pre-defined symbols (\textbf{a}), Brush uses the Probabilistic Tree-Creation 2 (PTC2) algorithm to create a set of random individuals to fill the initial population. Then, this population goes through a selection step, pairing individuals (\textbf{b}) to go together into the variation step. The variation step can apply one of $7$ possible variation operations (\textbf{c}). \textit{Crossover} randomly switches two sub-trees between the selected parents. \textit{Toggle weight on/off} will enable/disable a learnable weight into a random node. \textit{Subtree} will replace one node by a random sub-tree, generated with PTC2. \textit{Point} replaces one random node with a new one of same arity. \textit{Delete} removes one node, keeping its children. \textit{Insert} creates a new node. Best viewed in colors.}
    \label{fig:benchmarking:programs-and-variations}
    \includegraphics[width=\linewidth]{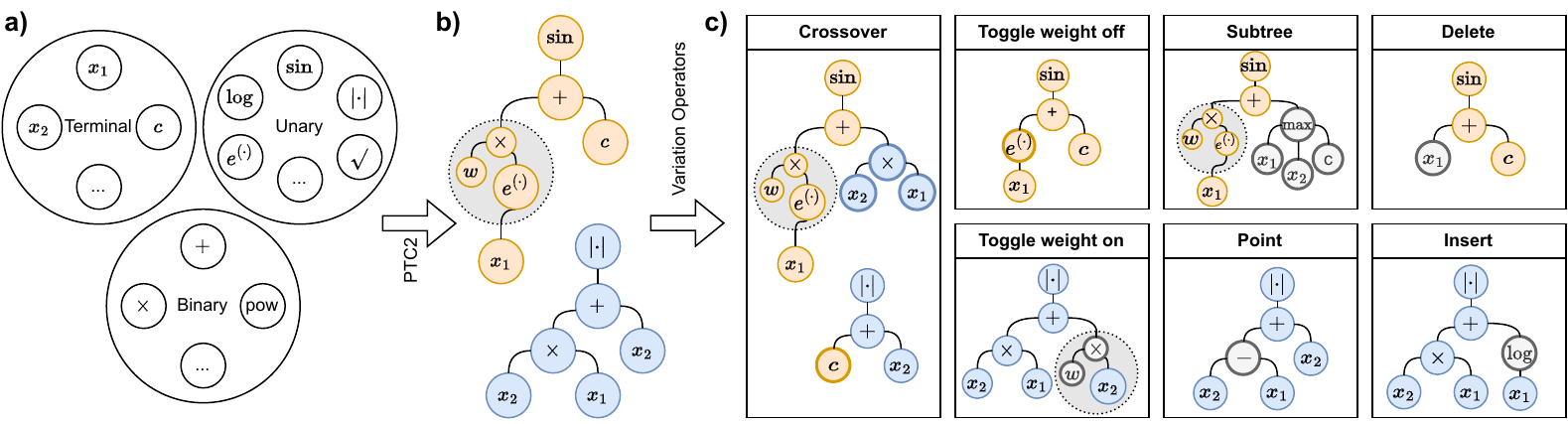}
    \legend{Source: Elaborated by the author.}
\end{figure}

The variation is applied until a new intermediate population $\mathcal{Q}$, so-called \textit{offspring}, reaches the same size as the original population. Finally, the original population and the offspring are merged into $\mathcal{R}$ and goes into the survival step based on the non-dominated sorting. The non-dominated sorting identifies multiple fronts of non-dominated individuals, and returns the first $S$ best-performing ones to replace the population in $\mathcal{P}$.
Figure \ref{fig:benchmarking:programs-and-variations} illustrates different programs that can be constructed using the symbols and tree structures, then, one possible result for each variation operator implemented. The shadowed area represents a node with its toggleable weight on. 

Brush is implemented to internally split the training data into an smaller training partition, and a validation partition. The inner training partition is used to tune the parameters and evaluate the prediction error for each test case in $\epsilon$-lexicase selection.
The test partition is not used in any step to avoid data leakage, and it comes into play only when selecting the final individual, asserting that the returned expression is not overfitted to the training data.
The final model is picked using a validation fraction $vs$ out of the original data, by choosing the individual with best prediction performance, although the framework was developed to support custom user-defined selection criteria based on any of the fitness measures implemented: the error metric on train/test, size, complexity, depth, or Pareto dominance rank.

\begin{algorithm}[htb]
    \caption{Brush's evolutionary backbone}\label{alg:background:nsga-ii}
    \begin{algorithmic}[1]
        \REQUIRE{\hfill\\
            \textbf{GA parameters:} problem data points ($\mathcal{X}, \mathbf{y}$), population size ($S$), generations ($G$), validation fraction ($vs$), list of concurrent objectives ($\mathcal{O}$), loss function to compute the errors ($\mathcal{L}$), tolerance for variation failures ($\tau$) \\ 
            \textbf{Tree constrains:} maximum size ($\text{max}_{s}$) and depth ($\text{max}_{d}$)
        }
        \ENSURE{ Best individual in validation partition, or any custom user-defined pick criteria }

        \STATE $\mathcal{X}^t$, $\mathbf{y}^t$ $\leftarrow$ $(1.0-vs)$ fraction of ($\mathcal{X}$, $\mathbf{y}$)
        
        \STATE $\mathcal{X}^v$, $\mathbf{y}^v$ $\leftarrow$ remaining $vs$ fraction of ($\mathcal{X}$, $\mathbf{y}$)

        \STATE $\mathcal{P} \leftarrow \{\mathtt{PTC2}(\text{max}_s, \text{max}_d) \textbf{ for } \_ \in \{1, 2, \ldots, S\}\}$

        \FOR[Initializing parameters]{$p \in \mathcal{P}$}
            \STATE $p.\mathtt{weights}$ $\leftarrow \mathtt{ParameterOptimization}(p, \mathcal{X}^t, \mathbf{y}^t, \mathcal{L})$
    
            \STATE $\text{loss}_t$ $\leftarrow \mathcal{L}(p.\mathtt{evaluate}(\mathcal{X}^t), \mathbf{y}^t)$
            \STATE $\text{loss}_v$ $\leftarrow \mathcal{L}(p.\mathtt{evaluate}(\mathcal{X}^v), \mathbf{y}^v)$

            \STATE $p.\mathtt{fitness}$ $\leftarrow \mathtt{UpdateFitness}([\text{loss}_t, p.\mathtt{size}, \text{loss}_v, p.\mathtt{complexity}, p.\mathtt{depth}], \mathcal{O})$
        \ENDFOR
        
        \FOR{$\text{gen} \in \{1, 2, \ldots, G\}$}
            \STATE $\mathcal{P}' \leftarrow \mathtt{Select}\epsilon\text{-}\mathtt{Lexicase}(\mathcal{P}, S)$
            \STATE $\mathcal{Q}$ $\leftarrow \emptyset$ \COMMENT{Initialize offspring}
            \FOR{$i \in \{1, 3, 5, \ldots, |\mathcal{P}'|-1\}$}
                \STATE $p_1, p_2 \leftarrow \mathcal{P}'[i], \mathcal{P}'[i+1]$
                
                \STATE $o_1, o_2 \leftarrow \mathtt{Variation}(\mathcal{X}^t, \mathbf{y}^t, p_1, p_2, \mathcal{L}, \tau)$    

                \STATE $\mathcal{Q} \leftarrow \mathcal{Q} \cup \{o_1, o_2\}$
            \ENDFOR
            
            \STATE $\mathcal{R} \leftarrow \mathcal{P} \cup \mathcal{Q}$
            \STATE $\mathcal{P} \leftarrow \mathtt{NSGA\text{-}II}(\mathcal{R}, S)$
        \ENDFOR

        \STATE \textbf{return} $\arg\min_{p \in \mathcal{P}} \mathtt{Dist}\big( (p.\mathtt{evaluate}(\mathcal{X}^v), \mathbf{y}^v), (0, 1)\big)$
    \end{algorithmic}
\end{algorithm}

Algorithm \ref{alg:background:nsga-ii} illustrates the evolutionary framework.
For convenience, let $p.\mathtt{weights}$ denote the tunable parameters of the individual $p$ from the population.
The parameter optimization relies on training data, as well as the loss function, to find values that minimizes the loss function on the data used. The loss can be defined different for classification and regression problems: when solving regression problems, Brush uses the MSE as loss function, whereas it uses the logistic loss function for classification.
The $\mathtt{UpdateFitness}$ function will take as argument all the fitness measures calculated to the individuals, but will update the fitness attribute by selecting only the metrics specified in the list of concurrent objectives $\mathcal{O}$.

\begin{algorithm}[htb]
    \caption{Variation}\label{alg:background:mutation}
    \begin{algorithmic}[1]
        \REQUIRE{Problem data points ($\mathcal{X}, \mathbf{y}$), parents ($p_1$, $p_2$), loss function to compute the errors ($\mathcal{L}$), and tolerance until returning the original parent ($\tau$)
            }
        \ENSURE{ Pair of variations over the parents $(o_1, o_2)$ }
        
        \STATE $o_1, o_2 \leftarrow p_1, p_2$ \COMMENT{Placeholders with the original parents}

        \FOR{(ind, otherInd) $\in [ (o_1, p_2), (o_2, p_1) ]$}
            \STATE $\mathtt{attempts} \leftarrow 0$
            \WHILE{$\mathtt{attempts} < \tau$}
                \STATE $\mathtt{variationToApply} \leftarrow \mathtt{SampleVariation}()$
                \STATE $\mathtt{spot} \leftarrow \mathtt{SelectVariationSpot}(ind, \mathtt{variationToApply})$

                \STATE $c \leftarrow \mathtt{nothing}$
                \IF{$\mathtt{variationToApply} = \mathtt{crossover}$}
                    \STATE $c \leftarrow \mathtt{crossover}(ind, otherInd)$
                \ELSE[Mutation requires only one argument]
                    \STATE $c \leftarrow \mathtt{mutation}(ind, \mathtt{variationToApply})$
                \ENDIF

                \IF[Initialize and get reward]{$c \neq \mathtt{nothing}$}
                    \STATE $c.\mathtt{weights} \leftarrow \mathtt{ParameterOptimization}(c, \mathcal{X}, \mathbf{y}, \mathcal{L})$
            
                    \STATE $c.\mathtt{fitness} \leftarrow \big( \mathcal{L}(c(\mathcal{X}), \mathbf{y}), c.\mathtt{size} \big)$ 

                    \STATE ind $\leftarrow c$
                    \STATE \textbf{break}
                \ENDIF
            \STATE $\mathtt{attempts} \leftarrow \mathtt{attempts} + 1$
            \ENDWHILE
        \ENDFOR
        \STATE \textbf{return} $[o_1, o_2]$
    \end{algorithmic}
\end{algorithm}

The variation operator work as follows.
Individuals goes in pairs to the variation operator, as crossover requires two individuals, whereas mutation may use only one.
The variation to be applied is selected first. Then, the location in the individual's tree where the mutation will occur is determined based on the mutation type. In this way, candidate spots can be preselected, excluding those where generating a valid individual within depth and size constraints is unlikely.
By choosing the spot according to the variation type, meaningless modifications are also avoided (\textit{e.g.}, attempting to toggle a weight in an already fully weighted individual).
For crossover, the procedure for selecting candidate spots is also implemented to ensure that size and depth constraints are not violated. First, a random spot is selected in one parent, provided it is not located at the maximum depth. Then, in the second parent (the donor tree), all spots that satisfy the constraints are listed as candidate locations. A final subtree is selected from the second parent from the candidate pool, and if no valid candidates exist, a new spot is selected in the first parent and the process is repeated.
The $\tau$ defines the number of times the algorithm will tolerate failures during variation, set by default to $3$, and returning a new randomly generated individual in case it fails. 
Finally, the procedure returns the new individuals to compose the offspring in the main loop. 
Algorithm \ref{alg:background:mutation} describes this procedure.

By default, the multiple objectives are the loss function and the complexity of expressions as in Eq. \ref{eq:background:complexity}. The complexity is calculated as the sum of the individual complexity of each node described in \ref{tab:background:complexity}.

\begin{table}[tbh]
    \centering
    \caption[Complexity of each operator in Brush, for regression problems.]{Complexity of each operator in Brush, for regression problems. This table reports only operations used in this thesis, although Brush supports additional operators not reported here.}
    \label{tab:background:complexity}
    \begin{tabular}{@{}cl@{}}
    \toprule\midrule
    \textbf{Complexity} & \textbf{Operators} \\ \midrule
    2 & $\mathtt{Constant}$ \\
    3 & $+$, $-$, $\mathtt{Terminal}$ \\
    4 & $|\cdot|$, $(\cdot)^2$, $\mathtt{Logistic}$,
        $\mathtt{Min}$, $\mathtt{Max}$, $\mathtt{Mean}$, $\mathtt{Median}$, $\mathtt{Sum}$, $\mathtt{Prod}$, 
        $\times$ \\
    5 & $\lceil \cdot \rceil$, $\lfloor \cdot \rfloor$, $e^{(\cdot)}$, $\log$, $\sqrt{\cdot}$, $\sqrt{|\cdot|}$, 
        $\div$, $(\cdot)^{(\cdot)}$\\
    6 & $\cos$, $\sin$, $\tan$, $\cosh$, $\sinh$, $\tanh$, $\cos^{-1}$, $\sin^{-1}$, $\tan^{-1}$ \\
    9 & $\log(1+\cdot)$ \\
    10 & $\mathtt{Logabs}$ \\
    \midrule\bottomrule   
    \end{tabular}
    \legend{Source: Elaborated by the author.}
\end{table}

\subsection{Parameter optimization}

In Brush, every node contains an innate node weight that can be toggled on or off during the search. If it is on, the parameter is optimized using LM. Initially, only terminal weights are toggled on.
This extends Operon~\cite{operon} mechanism that assigns weights exclusively to terminals.
Brush utilizes the large-scale Ceres solver \cite{Agarwal_Ceres_Solver_2022} for obtaining a fast estimation of the parameters.

\subsection{Parallel populations}

\begin{figure}[htb]
    \centering
    \caption[Brush island schema. ]{Brush island schema. On the left, the overview of the steps performed. The main loop will split the population into islands for faster processing by sending each island to one core. The population is in practive a single vector with all individuals, and each island is represented as a list of indexes from this population. The islands initially do not have overlapping in indexes, but migration can occur. A synchronization point is needed when generating the offspring, since at this point the algorithm is manipulating the size of the population vector by doubling the population to accommodate the offspring.}
    \label{fig:background:island_schema}
    \includegraphics[width=\linewidth]{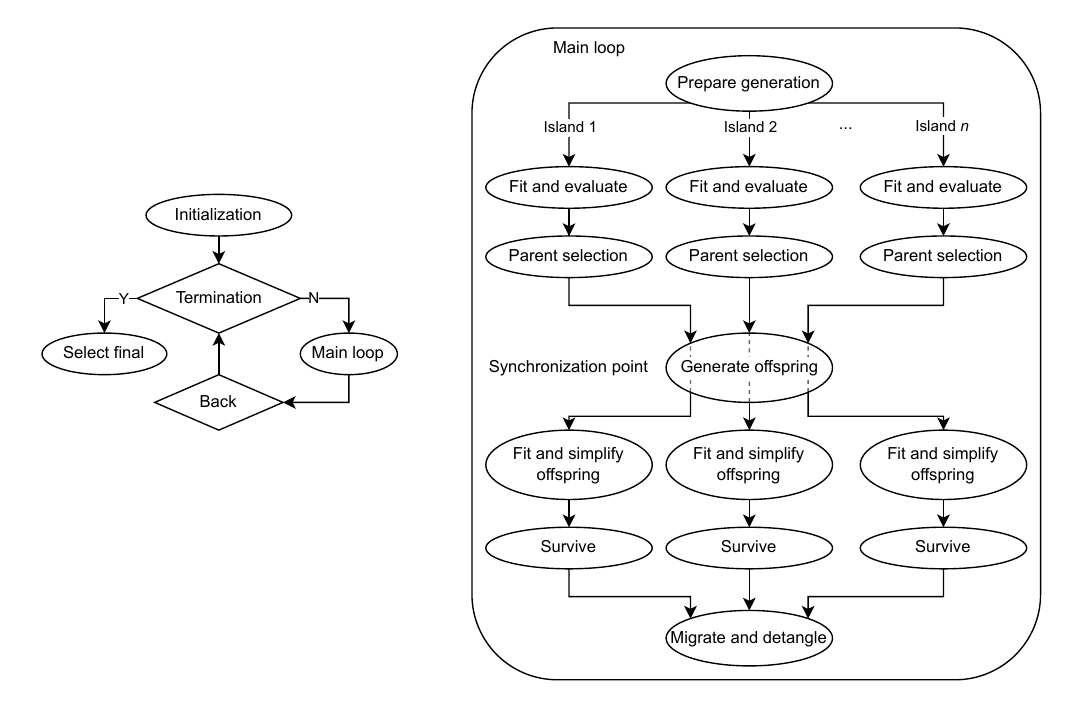}
    \legend{Source: Elaborated by the author.}
\end{figure}

In Brush, population management is designed for runtime efficiency by handling the population as if it were divided into multiple islands. The number of islands can be either specified by the user or automatically inferred from the number of available CPU cores. The island model was originally proposed to evolve solutions in parallel while preventing all subpopulations from converging in the same direction, thereby maintaining higher diversity across the overall population \cite{whitley1999island}.

Brush adopts this idea in a loose sense, primarily to exploit parallel computing. Steps that can be parallelized are executed independently within islands, while offspring generation remains centralized in a single thread, since the islands are not fully decoupled from a single population, but only abstracted for paralelism. Importantly, parent selection and survival are carried out independently within each subpopulation, meaning that multiple Pareto fronts may coexist. This approach improves scalability while remaining transparent to the user.

\section{Summary}~\label{sec:background:summary}

This chapter provided a historical and conceptual overview of symbolic regression. This chapter introduced its origins, taxonomy, problem formulation, and key components such as representation, optimization, and evaluation. It also reviewed common applications and highlighted open challenges, particularly in the areas of search, optimization, and simplification, which affect symbolic regression algorithms broadly. 

The problem formulation was presented in mathematical terms, which can vary across authors; here, the definitions are clarified to ensure consistency and to highlight the impact of the contributions presented in this thesis.  

Benchmarking SR methods, an issue of central importance, was also discussed. This topic will be revisited in Chapter~\ref{chap:benchmarking}, where the results are reported for a new benchmark that includes the algorithm used as the foundation for implementing several contributions of this thesis, such as expression simplification. In the benchmark, Brush achieved state-of-the-art performance.

The next four chapters present the contributions developed during this thesis, including a large-scale benchmark study. Finally, the concluding chapter synthesizes the findings, discusses their impact, and outlines future perspectives.  


%% file: Capitulos/Optimization/optimization.tex
\Chapter{Parameter optimization}{Linear and non-linear optimization influences on predictive accuracy and model size}
\chaptermark{Parameter optimization}~\label{chap:parameteropt}

\begin{laysummary}
    Coefficient optimization in symbolic regression is a challenging task.
    This chapter investigates linear and non-linear optimization methods in a symbolic regression algorithm with a constrained representation, formulated as a linear combination of non-linear terms. This design allows both optimization approaches to be applied independently or in combination.
    Linear optimization provides a closed-form solution with globally optimal parameters, while non-linear optimization is iterative and may converge to local optima or fail to improve results. To evaluate these strategies, a benchmark is performed by applying the symbolic regression method on six well-known datasets in the machine learning community. The goal is to assess how parameters can be effectively optimized and, in particular, to investigate whether separating the optimization of linear and non-linear parameters improves predictive accuracy and model size, thereby informing the development of future algorithms that rely on constants and can benefit from parameter optimization.
    Results show that separately optimizing linear and non-linear parameters produces the best-performing expressions, though with higher runtime and larger expression size. However, combining both optimization methods may require a constrained representation, thus limiting the hypothesis space to a narrower set, which can be undesirable in some contexts. Non-linear optimization alone performs worst for the given representation, likely related to overparameterization or the difficult to find feasible initial points.
\end{laysummary}

{\let\thefootnote\relax\footnotetext{Part of this chapter was published as: Guilherme Seidyo Imai Aldeia and Fabrício Olivetti de França. 2022. Interaction-Transformation Evolutionary Algorithm with parameters optimization. In Genetic and Evolutionary Computation Conference Companion (GECCO '22 Companion), July 9-13, 2022, Boston, MA, USA. ACM, New York, NY, USA, 8 pages. \url{https://doi.org/10.1145/3520304.3533987}}}

\vfill
\pagebreak

\section{Introduction}~\label{sec:optimization:introduction}

The search space of Symbolic Regression (SR) is vast and may include any algebraic function; thus, identifying optimal parameter values is a non-trivial task.
Parameter optimization\footnote{Also called \textit{coefficient learning} or \textit{parameter identification} in the SR context.} in SR has been recognized as challenging task \cite{Kommenda2019}, and leaving it entirely to the evolutionary process further increases the difficulty. In the first SR algorithm proposed by Koza \cite{GPOnTheProgrammingOfComputersByMeansOfNaturalSelection}, Ephemeral Random Constants (ERCs) were generated within a bounded range, in the expectation that evolution will prioritize those values that favored expressions. In this scenario, constants needed to suit the entire population globally, or else they were discarded during evolution. Moreover, crossover and mutation could drastically alter the behavior of functions ---as analyzed in \citeonline{SymTree}---, making existing parameters meaningless after the variation.

Relying exclusively on ERCs can hinder performance, as evolution must discover parameters independently. This not only lacks the ability to obtain global optima \cite{WANG20042453} but also results in slow convergence when relying solely on the evolutionary procedure \cite{8393325}. However, the burden of parameter adjustment can be alleviated with optimization methods. Real-valued parameter optimization in SR has been partially addressed with local methods \cite{PrioritizedGrammarEnumeration}, and it is now widely accepted in the GP community that parameter optimization is essential when running SR algorithms.

Finding optimal parameters for mathematical expressions is not new; linear models in statistics provide a classic example. Linear models admit a single solution ---given the number of sample exceeds the number of variables--- because the derivative of the error function with regard to the parameters can be explicitly calculated. This leads to a closed-form solution for minimizing error, the foundation of the linear least squares methods \cite{James2021}. 
Notice that this also generalizes to linear combinations of non-linear terms, such as transformations of the original variables. The challenge then becomes identifying suitable non-linear terms to serve in the linear combination. This may require either exhaustive search or black-box optimization. While the former guarantees global optimality, it is often intractable because the non-linear term may take the form of any algebraic function. The latter is fundamentally SR itself.

The linear composition idea underlies several SR methods. FEAT \cite{cava2018learning} constructs meta-features as trees and combines them with regularized linear optimization techniques such as Lasso (although Lasso does not have a closed-form and performs an iterative process that requires several steps and may not converge). The ITEA \cite{10.1162/evco_a_00285} employs the Ordinary Least Squares (OLS) over linear combination of constrained non-linear terms. The PS-Tree algorithm \cite{ZHANG2022101061} creates small trees with ERCs and combines them with Lasso, in a boosting-like approach. FFX \cite{McConaghy2011} performs a deterministic combination of a library of automatically generated interactions and combine them linearly with linear regression.

Nevertheless, restricting SR to linear combinations reduces the hypothesis space to a narrow subset. Addressing this limitation requires non-linear optimization, typically carried out via iterative processes such as gradient descent. These methods require gradient computation, which is difficult to provide automatically given that unconstrained SR can represent nearly any function. Consequently, existing approaches rely on black-box optimization, which requires only access to parameters, but are computationally expensive and provide no guarantees of reaching global optima. Operon \cite{Kommenda2019}, for example, uses the Levenberg-Marquardt (LM) algorithm \cite{levenberg1944method,marquardt1963algorithm} to optimize black-box non-linear functions iteratively by approximating the gradient with regard to the parameters.

Given the discussions above, ITEA stands out for its constrained representation, which facilitates adaptation to diverse search algorithms, demonstrated by the family of methods that also uses the same representation, such as Simulated Annealing~\cite{kantor2021simulated}, Neural Networks~\cite{de2021interaction}, and regression analysis through automatic differentiation~\cite{aldeia2021measuring}.
In \citeonline{srbench}, ITEA demonstrated competitive performance among state-of-the-art SR algorithms.
In this chapter, the ITEA algorithm is extended to incorporate non-linear parameters and used to investigate how different parameter optimization methods perform and interact.

This chapter therefore explores whether optimization methods can be combined in a constrained representation that has separable linear and non-linear parameters.
Because non-linear optimization is prone to local optima, four strategies are presented for parameter optimization: \textit{(i)} optimizing only linear parameters with OLS, \textit{(ii)} optimizing all parameters with LM, \textit{(iii)} first adjusting linear parameters with OLS and using them as the starting point for LM, and \textit{(iv)} optimizing each term individually with LM and then combining the results with OLS.

The primary objective is to investigate whether separating the optimization of linear and non-linear parameters improves predictive accuracy and model size, and to assess the trade-off in runtime when applying iterative non-linear optimization and closed-form linear optimization.
This analysis provides insights that may guide the development of future algorithms that rely on constants. The results shows that separately optimizing linear and non-linear parameters is the most effective strategy for producing high-performing expressions, though at the cost of increased runtime and larger expression size. In contrast, non-linear optimization alone yields the worst results, with higher normalized MSE in almost all benchmarked datasets. Finally, while combining linear and non-linear optimization can be beneficial, it requires a constrained representation that narrows the search space, which may be undesirable under certain circumstances.

The remainder of this chapter is organized as follows.
Section~\ref{sec:optimization:relatedwork} reviews related work on the role of parameter optimization in different SR algorithms.
Section~\ref{sec:optimization:linearnonlinear} introduces the background necessary to understand parameter optimization.
Section~\ref{sec:optimization:methods} describes the experimental setup used to assess the effects of parameter optimization, including the SR algorithm employed.
Section~\ref{sec:optimization:resultsanddiscussion} presents and discusses the experimental results.
Finally, Section~\ref{sec:optimization:conclusions} summarizes the chapter by outlining the main insights and conclusions.

\section{Related work}~\label{sec:optimization:relatedwork}

In the literature, numerous works have delved into linear or non-linear methods for local optimization of symbolic regression expressions.

\citeonline{reis_benchmarking_2024} benchmarked eight parameter optimization methods, emphasizing that constant optimization has a major impact on performance yet remains underexplored. Their results show that optimizing parameters greatly improves model accuracy and helps recover the correct mathematical expression. They also note that many methods follow a common pattern: parameters are initialized randomly and subsequently optimized. Easy problems tend to be solved efficiently when parameters are optimized, whereas harder problems exhibit strong dependence on both the SR method and the optimization strategy.

Other studies have emphasized the risks of overparameterization. For example, \citeonline{10.1145/3583131.3590346} argue that models with too many parameters are difficult to interpret and suffer from ill-conditioning, which slows down memetic approaches. Similarly, \citeonline{franca_improving_2025} point out that redundant parameters can hinder optimization in SR methods. This issue is analogous to multicollinearity in linear models, where OLS estimation may fail to produce reliable parameter estimates \cite{gujarati2003multicollinearity}. Kronberger et al. \cite{kronberger_effects_2025} further show that detecting and removing overparameterization can improve both the convergence speed and results of non-linear optimization, as fewer parameters make it easier to reach local optima.

Other approaches have explored dividing expressions into modular components. \citeonline{CHEN20181973} propose decomposing expressions into building blocks that are first optimized individually and then assembled using a global optimization procedure. \citeonline{LUO20121182} apply the Low Dimensional Simplex Evolution method to optimize parameters. The Fast Function Extraction (FFX) method \cite{McConaghy2011}, a deterministic approach to symbolic regression, employs regularized learning to identify parameters. Recent work has also investigated the use of the LM algorithm for non-linear parameter optimization in SR \cite{Kommenda2019, PrioritizedGrammarEnumeration}.

Finally, \citeonline{burlacu2025zobristhashbasedduplicatedetection} studied the impact of memoization on parameter optimization, specifically using caching to store and reuse previously fitted parameters. Their results demonstrate up to 34\% speedups without loss in accuracy, by detecting duplicates and retrieving fitness values without re-fitting the model. They tested different numbers of local search iterations (0, 1, 10, 50, and 100), showing the scalability of this approach.

Despite this variety of methods and experiments, none of the existing works have combined linear and non-linear optimization procedures. This integration requires that linear and non-linear parameters are explicitly defined, facilitating their identification and enabling a systematic study of their combination.

\section{Linear and non-linear optimization}~\label{sec:optimization:linearnonlinear}

The linear optimization method admits a closed-form solution with globally optimal parameters, whereas the non-linear method is iterative and may converge to local optima or fail to improve the solution if the starting point is not adequate. 

Let $\mathcal{X} = \left [ \mathbf{x}_{1}, \mathbf{x}_{2}, \ldots, \mathbf{x}_{d} \right ] \in \mathbb{R}^{d \times n}$ be a matrix where each column corresponds to an independent variable, and each row represents a sample from the training data. Each row is therefore an observation of $n$ variables, $\mathbf{x} = [ x_1, x_2, \ldots, x_n ]$. Let $\mathbf{y} = ( y_{i} )_{i=1}^{d} \in \mathbb{R}^{d}$ be the target variable vector.

Linear models (belonging to the category of parametric models) assume that the data can be explained by a linear combination of the input variables. Under this assumption, only the free parameters need to be estimated, with the goal of minimizing the distance between the linear function (or hyperplane, for $\mathbf{x} \in \mathbb{R}^n$ with $n > 1$) and the observed data points.

Consider the function $\hat{f}$ as the linear combination of the input variables:

\begin{equation}~\label{eq:optimization:regressaoLinear}
    \widehat{f}(\mathbf{x}) = \widehat{f}(x_{1}, x_{2}, \ldots, x_{n}) = \beta_{0} +  \sum^{n}_{i=1} \beta_{i} \cdot x_{i},
\end{equation}

\noindent where $\boldsymbol{\beta} = [\beta_{1}, \beta_{2}, \ldots, \beta_{n}]$ represents the parameters for each variable, and $\beta_{0}$ is the intercept. The intercept can be added to the input variables $\mathcal{X}$ by creating a dummy column $\mathcal{X}_{*,0} = (1_i)_{i=1}^{d}$.

In vector form, The Eq. \ref{eq:optimization:regressaoLinear} can be expressed as:

\begin{equation}
    \widehat{y} = \mathbf{x}^{\intercal} \boldsymbol{\beta}.
\end{equation}

The goal is to find values of $\boldsymbol{\beta}$ that minimize the distance between the function and the observed points. This is equivalent to minimizing the Residual Sum of Squares (RSS):

\begin{equation}
    \text{RSS} = \sum^{d}_{i=1} (y_i - \widehat{y}_{i})^2
    = \sum^{d}_{i=1} \left(y_i - \mathbf{x}^{\intercal}_{i} \boldsymbol{\beta}\right)^2,
\end{equation}

\noindent where $y_i - \widehat{y}_{i}$ is the residual. The optimization problem can be stated as:

\begin{equation}
    \widehat{\boldsymbol{\beta}} = \underset{\boldsymbol{\beta}}{\arg\min} \ \text{RSS}(\boldsymbol{\beta})
    \quad \equiv \quad \frac{\partial}{\partial \boldsymbol{\beta}}\text{RSS}(\boldsymbol{\beta}) = 0.
\end{equation}

The Ordinary Least Squares method provides an unbiased estimator with a closed-form solution for the parameters that minimize squared error \cite{doi:10.1080/00031305.1989.10475644}. 
Since RSS is a convex quadratic function, it admits a single global optimum, so it's partial derivative yields optimal $\boldsymbol{\beta}$ values.
The partial derivative with regard to any $\beta_j$ can be calculated by:

\begin{equation}
    \frac{\partial}{\partial\beta_j} \text{RSS}(\boldsymbol{\beta}) = 2 \sum^{d}_{i=1} x_{i,j} \left [ y_i - \sum^{n}_{k=1} \beta_k x_{i,k} \right ].
\end{equation}

The $2$ factor can be ignored. Rearranging:

\begin{equation}\label{eq:optimization:olsinalgebraicnotation}
    \sum^{d}_{i=1}\sum^{n}_{k=1} x_{i,j} x_{i,k} \beta_k = \sum^{d}_{i=1} x_{i,j} y_i
\end{equation}

In matrix notation, the Eq. \ref{eq:optimization:olsinalgebraicnotation} can be written as:

\begin{equation}\label{eq:optimization:olsadj}
    \left( \mathcal{X}^\intercal \mathcal{X} \right)\boldsymbol{\beta} =  \mathcal{X}^\intercal \mathbf{y} \Rightarrow \widehat{\boldsymbol{\beta}} = \left( \mathcal{X}^\intercal \mathcal{X} \right)^{-1} \mathcal{X}^\intercal \mathbf{y}.
\end{equation}

This method is fast to calculate, as it relies on matrix operations, and performs well when the relationship between variables and the target is linear.

To optimize non-linear parameters, iterative methods must be used. SR community has often applied the Levenberg-Marquardt algorithm \cite{levenberg1944method, marquardt1963algorithm}, as in \citeonline{Kommenda2019, PrioritizedGrammarEnumeration}. This approach generalizes linear optimization.

For convenience, let the residual error be defined as the difference between the model prediction and the target values, for an arbitrary function with $p$ parameters, $\textup{H}: \mathbb{R}^{p} \rightarrow \mathbb{R}^{d}$:

\begin{equation}
\textup{H}(\boldsymbol{\beta}) = \mathbf{y} - \widehat{f}(\mathcal{X}, \boldsymbol{\beta}).
\end{equation}

The optimization problem becomes:

\begin{equation}
\widehat{\boldsymbol{\beta}} = \underset{\boldsymbol{\beta} \in \mathbb{R}^{p}}{\arg\min} \ ||H(\boldsymbol{\beta})||_2,
\end{equation}

\noindent where $||\cdot||_2$ denotes the 2-norm, $||\mathbf{h}|| := \sqrt{h_1^2 + h_2^2 + \ldots + h_d^2}$.

This is solved iteratively by approximating a step in $\textup{H}$ through linearization around $\boldsymbol{\beta}$:

\begin{equation}
H(\boldsymbol{\beta} + \Delta \boldsymbol{\beta})
\approx H(\boldsymbol{\beta}) + J(\boldsymbol{\beta}) \Delta \boldsymbol{\beta},
\end{equation}

\noindent where $J(\boldsymbol{\beta})$ is the Jacobian matrix of first-order derivatives with regard to the parameters:

\begin{equation}
J(\boldsymbol{\beta}) =
\begin{bmatrix}
\frac{\partial h_1}{\partial \beta_1} & \frac{\partial h_1}{\partial \beta_2} & \hdots & \frac{\partial h_1}{\partial \beta_{p}}\\
\vdots & \vdots & \ddots & \vdots\\
\frac{\partial h_n}{\partial \beta_1} & \frac{\partial h_n}{\partial \beta_2} & \hdots & \frac{\partial h_n}{\partial \beta_{p}}\\
\end{bmatrix},
\end{equation}

\noindent with $h_i$, $i=1, 2, \ldots, |\text{expression}|$ denoting the non-linear component associated with the parameter $\beta_j$, $j=1, 2, \ldots, p$, treating parameters as the optimization parameters.

At each iteraction, the update rule for $\boldsymbol{\beta}$ is given by approximating a gradient descent step:

\begin{equation}\label{eq:optimization:lmadj}
\boldsymbol{\beta} \leftarrow \boldsymbol{\beta} + \Delta \boldsymbol{\beta}
= \underset{\Delta\boldsymbol{\beta}}{\arg\min} \ || H(\boldsymbol{\beta}) + J(\boldsymbol{\beta})\Delta\boldsymbol{\beta} ||_2.
\end{equation}

Although LM provides a general and flexible optimization scheme, it introduces several challenges. First, it is computationally expensive. Second, the choice of initialization for $\boldsymbol{\beta}$ strongly affects convergence and feasibility of the solution. Finally, as the number of parameters increases, convergence becomes more difficult due to the enlarged search space, which may contain many local minima or non-convex error landscapes.
In the context of SR methods, \citeonline{burlacu2025zobristhashbasedduplicatedetection} explored performance as the number of iterations increased, and concluded that a very small number of iterations (\textit{e.g.}, $10$) is sufficient to find proper parameter values.

\section{Methods}~\label{sec:optimization:methods}

This section proposes a modified version of the ITEA algorithm, and then presents how it will be used to investigate the effects of linear and non-linear optimization methods. Then, it explains the experimental setup.

The original ITEA algorithm employs a constrained representation inspired by physics equations, allowing only interactions between variables up to specified strength levels. This representation also enables the method to express ratios of two functions, where each function consists of interactions among the variables. As a result, the algorithm is capable of representing a wide range of equations commonly found in physics \cite{8477951}.

\subsection{Non-linear Interactrion-Transformation Evolutionary Algorithm}~\label{sec:optimization:itea}

The Interaction-Transformation Evolutionary Algorithm (ITEA) uses a restricted representation proposed by \citeonline{SymTree}, called \textit{Interaction\hyp{}Transformation} (IT) tuple: a tuple $(g, \mathbf{k})$ that describes the application of a transformation function $g:\mathbb{R} \rightarrow \mathbb{R}$ over the interaction of the $n$ independent variables $(x_{1}, x_{2}, \ldots, x_{n})$ to the power of their respective strengths $\mathbf{k} = k_1, k_2, \ldots, k_n$. Expressions are built by linearly combining $t$ IT tuples in the following form:

\begin{equation}\label{eq:optimization:it}
    \hat{f}(\mathbf{x}, \boldsymbol{\beta}) = \beta_0 + \sum_{i=1}^{t}{\beta_i \cdot g_i(\underbrace{\text{p}(\mathbf{x}, \mathbf{k}_i)}_{\textup{Interaction}})},
\end{equation}

\noindent with the interaction defined as:

\begin{equation}\label{eq:optimization:interaction}
    \text{p}(\mathbf{x}, \mathbf{k}) = \prod_{j=1}^{d}{x_j^{k_j}}.
\end{equation}

The linear parameters are adjusted using the OLS method, since the representation restricts the hypothesis space to linear combinations to facilitate precise parameter identification. Multiple pairs of IT terms are linearly combined to create a final expression in the form given by Eq. \ref{eq:optimization:it}. 

The ITEA algorithm uses only mutations to explore the search space, and tournament selection (with a default size of $3$) to generate the next generation at each iteration. The ITEA starts by creating random expressions with the restricted representation, where the user can also specify the complexity constraints of the expression (such as the maximum number of IT terms or strength range).
This is illustrated in Algorithm \ref{alg:optimization:itea}. 

In total, there are $3$ types of mutations: 

\begin{itemize}
    \item \textbf{Expand:} adds a new tuple $(\textup{g}, \mathbf{k})$ to the expression, which can be a random new tuple, or the positive/negative element-wise combination of two existing tuples on the expression;
    \item \textbf{Shrink:} removes a random existing tuple $(g, \mathbf{k})$ of the expression;
    \item \textbf{Local modification:} randomly changes one value in the interaction strengths $\mathbf{k}$ with a random integer.
\end{itemize}

Notice that since the ITEA only performs mutation operations, the evolutionary procedure can be interpreted as a local search in the structure, and the parameter optimization is a local search in the parameter.

\begin{algorithm}[htb]
\caption{Interaction-Transformation Evolutionary Algorithm}\label{alg:optimization:itea}
    \begin{algorithmic}[1]
        \REQUIRE {\hfill\\
            Function set $\mathbf{f}$, IT expression constrains $\Pi$, population size $s$, optimization function \textsc{optimize\_params}, train data $(\mathcal{X}, \mathbf{Y})$ }
        \ENSURE \textit{IT expression} $\widehat{f}$
        
        \STATE cache $\leftarrow$ \CALL{LRUCache}{MAX\_CACHE\_SIZE}\
        \STATE pop $\leftarrow$ $[$ \CALL{generate\_IT\_expressions}{$\mathbf{f}$, $\Pi$} for $\_ \in [1, 2, ..., s]$ $]$
        \WHILE{Criteria $\Omega$ is not meet}
            \STATE mutants $\leftarrow$ $[$ \CALL{mutate}{p, $\Pi$} for $\text{p} \in \text{pop}$ $]$
            \FOR{$\text{it} \in \text{pop ++ mutants}$}
                \IF{\CALL{hash}{it}$ \notin \text{cache.keys}$}
                    \STATE optm\_coeffs $\leftarrow$ \CALL{optimize\_params}{it, $\mathcal{X}$, $\mathbf{Y}$}
                    \IF{$\exists c \in \text{optm\_coeffs} \ni c = \pm \infty$}
                        \STATE optm\_coeffs $\leftarrow$ \CALL{get\_neutral\_elements}{it}
                    \ENDIF
                    \STATE cache[\CALL{hash}{it}] $\leftarrow$ optm\_coeffs
                \ENDIF
                \STATE \CALL{replace\_parameters}{it, cache[\CALL{hash}{it}]}
            \ENDFOR
            \STATE pop $\leftarrow$ \CALL{replace}{pop $++$ mutants, $\mathbf{X}$, $\mathbf{Y}$}
        \ENDWHILE
        \STATE \textbf{return} arg max $[$ \CALL{fitness}{p, $\mathbf{X}$, $\mathbf{Y}$} for $\text{p} \in \text{pop}$ $]$
    \end{algorithmic}
\end{algorithm}

The modified version of ITEA proposed here adds two inner parameters $\boldsymbol{\theta} \in \mathbb{R}^2$ to each IT tuple, replacing Eq. \ref{eq:optimization:interaction} with the following:

\begin{equation}
    \text{p}'(\mathbf{x}, \mathbf{k}_i, \boldsymbol{\theta}) = \theta_0 + \theta_1 \prod_{j=1}^{d}{x_j^{k_{j}}}.
\end{equation}

\begin{figure}[htb]
    \centering
    \caption[How non-linear IT expressions are constructed.]{How non-linear IT expressions are constructed. The final expression is the composition of several IT terms and a Linear offset. Each square in the figure represents a numeric parameter that must be adjusted. Between parenthesis, all applicable methods to optimize each square is represented: linear optimization (OLS), non-linear optimization (LM), and the evolutionary algorithm (EA).}
    \label{fig:optimization:nonlinearvisualexplanation}
    \includegraphics[width=0.9\linewidth,clip=true,trim={1cm 5cm 1cm 4.25cm}]{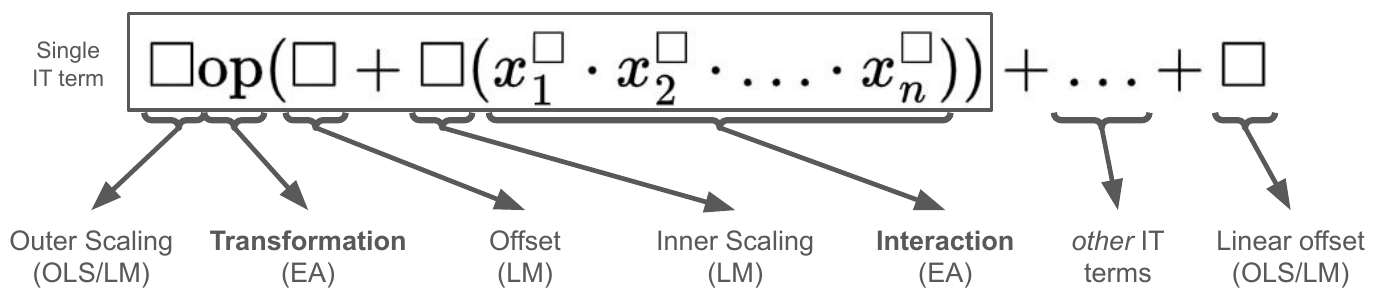}
    \legend{Source: Elaborated by the author.}
\end{figure}

The $\theta_0$ performs a shift in the domain of the transformation function argument, and $\theta_1$ allows to perform horizontal compression or stretch of the transformation function argument.
Figure \ref{fig:optimization:nonlinearvisualexplanation} provides an illustrative view of all components of IT terms that can now be adjusted --- some through the evolutionary process (EA), and some using an optimization method (OLS/LM).

\begin{figure}[htb]
    \centering
    \caption[Example of constructing an equation using the non-linear IT representation.]{Example of constructing an equation using the non-linear IT representation. Linear parameters are shown in red, and nonlinear parameters in blue. The operators and strengths are left for the evolutionary algorithm to optimize.}
    \label{fig:optimization:nonlinearexample}
    \includegraphics[width=0.9\linewidth,trim={0.2cm 0.2cm 0.2cm 0.5cm},clip=true]{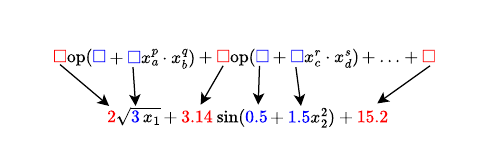}
    \legend{Source: Elaborated by the author.}
\end{figure}

To illustrate how expressions are generated using this representation, Figure~\ref{fig:optimization:nonlinearexample} shows the constrained representation structure, with linear parameters in red and non-linear parameters in blue, for a problem with two variables. Below the structure, the expression $2\sqrt{3 x_1} + 3.14 \sin{(0.5 + 1.5 x_2^2)} + 15.2$ is constructed using the representation. 
Identifying the operators ($\sqrt{\cdot}$ and $\sin$) and the strengths for each variable in the interaction corresponds to optimizing the function structure, while the linear and non-linear parameters can be optimized using different strategies.

Through an evolutionary process, the original ITEA algorithm would search for the best values for $(g, \mathbf{k})$ to build the expression structure. At the same time, it would optimize $\boldsymbol{\beta}$ with the OLS method. With the proposed modification, two non-linear parameters $\boldsymbol{\theta}$ are added to each expression's terms, and these cannot be adjusted with linear methods. 

Notice that the interaction of variables eliminates the $i$-th independent variables of the calculation when $k_i=0$, since $x_i^0=1$.
The original implementation does not limit the maximum number of variables in each interaction strength vector $\mathbf{k}$, which is extended with the possibility of setting a maximum number of variables to avoid overly complex expressions for the experiments. With this addition, the methods \CALL{generate\_IT\_expressions}{} and \CALL{mutate}{} were modified to respect this limit.
This is done because, in general, the manual modeling of a regression model with interactions of variables tends to avoid large interactions, which leads to less interpretable expressions. The idea of the IT representation is to restrict the search space to simpler expressions, and it could benefit from having control of how complex the interactions could get. While adding non-linear parameters increases the total number of nodes in the expressions, the mitigation is done by limiting the complexity of the interactions.

Regarding the optimization procedure, a Least Recently Used (LRU) Cache \cite{970573} is used to memoize previously seen parameters. The LRU Cache is used since it can restrict the number of elements and avoid an uncontrolled growth in memory consumption. While this can prevent the non-linear methods from further optimizing previous equations, it can also decrease the time consumption of the algorithms.
The parameters are memoized if the method successfully returns adjusted parameters without numeric error, and that results in an improvement on prediction accuracy. Otherwise, each parameter is set to the neutral element of the operation in which it takes part (zero for addition; one for multiplication). Similar expressions have the same hash to be used as keys in the cache (\textit{i.e.}, permutations of the IT tuples $(g, \mathbf{k})$ do not result in different expressions).

\subsection{Mixing linear and non-linear optimization}

To combine OLS and LM methods, the following heuristics are proposed:

\begin{description}
    \item[\textit{ITEA} with OLS adjust.] This is the original approach using OLS to adjust the parameters of the linear expression (Eq. \ref{eq:optimization:olsadj}).  This version of \textit{ITEA} has two purposes: \textit{i)} to serve as a baseline to compare the other adjustment heuristics and \textit{ii)} to be compared with the original ITEA implementation that does not limit the number of variables in the interaction.
    
    \item[\textit{ITEA} with LM adjust.] This approach optimizes all the parameters at once using at most ten iterations of the LM method, updating the parameters with Eq. \ref{eq:optimization:lmadj}. All parameters are initialized with random values taken from a uniform distribution in the interval $[-100, 100]$.
    
    \item[\textit{ITEA} with OLS + LM adjust.] This heuristic first optimizes the linear parameters using OLS, then randomly attributes values taken from a uniform distribution in the interval $[-100, 100]$ only for the non-linear parameters $\boldsymbol{\theta}$, and applies the LM method for at most ten iterations to the whole expression. The idea is to improve the starting point by partially adjusting it before the LM.
    
    \item[\textit{ITEA} with LM + OLS adjust.] This approach optimizes each $\boldsymbol{\beta}$ individually by applying the LM method to adjust only the non-linear parameters to fit the individual term to the dependent variable $\mathbf{y}$. Then, the individual parameters are updated in the expression, which has its linear parameters $\boldsymbol{\beta}$ globally adjusted using the OLS method. The idea is to stimulate local optimizations in each term, which will be the linear value from the OLSused. Also, individual optimization with LM can minimize the impact of errors during the iterative process.
\end{description}

Preliminary experiments were performed using a narrower interval of $[-1, 1]$ to initialize the parameters, but the results were unsatisfactory. These experiments shown that the interval between $[-1, 1]$ performs worse than the interval as in~\citeonline{Kommenda2019}, that is, sampling uniformly in the range $[-100, 100]$. 

\subsection{Experimental setup}

Six popular datasets from the machine learning community were used, with their names and dimensions reported in Table \ref{tab:optimization:datasets}. Since ITEA has stochastic behavior, $30$ independent runs were performed for each dataset. The datasets were divided into a $5$-fold setup (meaning $0.2$ test partition size), and for each fold, the training was done using the four remaining folds, and the fold-out was used to measure the final model performance on unseen data.
Results therefore are calculated from the distribution of the $30$ independent executions per dataset.

\begin{table}[htb]
    \caption{Datasets considered in the parameter optimization experiments.}
    \label{tab:optimization:datasets}
    \centering
    \begin{tabular}{@{}ccc@{}}
    \toprule\midrule
    \textbf{Dataset name} & \textbf{\# samples} & \textbf{\# variables} \\ \midrule
    Airfoil &  1503 & 5 \\
    Concrete & 1030 & 8 \\
    Energy Cooling & 768 & 8 \\
    Energy Heating & 768 & 8 \\
    Wine Red & 1599 & 11 \\
    Yacht & 308 & 6 \\ \midrule\bottomrule
    \end{tabular}
    \legend{Source: Elaborated by the author.}
\end{table}

All algorithms had the hyper-parameters fixed to the same configuration to allow an unbiased performance comparison, reported in Table \ref{tab:optimization:hyperparameters_config}, with corresponding descriptions reported in Table \ref{tab:optimization:hyperparameters_desc}.

\begin{table}[htb]
    \caption{Hyper-parameters of the non-linear ITEA and their descriptions.}
    \label{tab:optimization:hyperparameters_desc}
    \centering
    \begin{tabular}{@{}ll@{}}
    \toprule\midrule
    \textbf{Hyper-parameter} & \textbf{Description} \\ \midrule
    transf\_funcs & Set of allowed transformation functions \\
    popsize & Number of individuals in the population \\
    gens & Number of generations to perform the evolution (\textit{stop criteria}) \\
    strength\_bounds & Interval (inclusive) of allowed strength values \\
    terms\_bounds & Interval (inclusive) of the number of terms in each expression \\
    max\_nonzero\_strengths & Maximum number of strengths different of zero in the expressions\\ \midrule\bottomrule
    \end{tabular}
    \legend{Source: Elaborated by the author.}
\end{table}

\begin{table}[htb]
    \caption{Configuration used for the non-linear ITEA algorithm.}
    \label{tab:optimization:hyperparameters_config}
    \centering
    \begin{tabular}{@{}ll@{}}
    \toprule\midrule
    \textbf{Hyper-parameter} & \textbf{Configuration} \\ \midrule
    transf\_funcs & [identity, sin, cos, tan, $\sqrt{\cdot}$, log, exp, abs] \\
    popsize & 250 \\
    gens & 400 \\
    strength\_bounds & {[}-3, 3{]} \\
    terms\_bounds & {[}2, 15{]} \\
    max\_nonzero\_strengths & 2 \\ \midrule\bottomrule
    \end{tabular}
    \legend{Source: Elaborated by the author.}
\end{table}

The rationale for selecting the strength bounds  $[-3, 3]$ is that some existing physical equations exhibit relationships within this range. For example, the magnetic dipole field decreases with the inverse cube of the distance, and Kepler's third law states that the orbital period is proportional to the cube of the semi-major axis.

After each run, the final expression was saved, together with the total execution time, the Normalized Mean Squared Error (NMSE) for train and test partitions, and the number of nodes in the expression. The NMSE normalizes the mean squared error (MSE) by the variance of the dataset, allowing for the comparison of all results on the same scale. This performance measure is calculated by:

\begin{equation}
    \textup{NMSE} = \frac{\frac{1}{d}\sum_{i=1}^{d}{\left (\widehat{f}(\mathbf{x}_i) - y_i\right )^2}}{\textup{var}(\mathbf{y})},
\end{equation}

\noindent with $\textup{var}(\mathbf{y})$ being the variance of the dependent variable $\mathbf{y}$.

To assess model interpretability, the number of nodes is used as a proxy. Subsection~\ref{subsec:background:objective_functions} discussed several ways to measure model complexity, but due to the presence of multiple commutative linear IT terms in expressions generated with ITEA, computing complexity requires converting each expression into a parse tree. Because there are several valid ways to construct such a tree, different complexities can exist for the same expression, making it an unreliable metric. In contrast, counting the number of nodes offers a more stable and informative estimate of how complicated the equations are for the given representation.

The experiments were conducted on a 64-bit computer running Ubuntu 19.10, equipped with 16 GB of RAM and an Intel i7-9700 processor (8 cores @ 3.00 GHz). The LM method was allocated a budget of $10$ iterations. The implementation was developed in the Julia programming language, and the LRU cache was set to store at most $10{,}000$ expressions.

For boxplots, methods are grouped by dataset and represented as notched boxes. To determine the order of the methods in each figure, the methods were sorted them according to their average NMSE rank across all datasets, such that the legend reflects the overall ranking with respect to the error metric. The best overall method (leftmost in the legend) was statistically compared to the others. A bracket above each pair denotes whether there is statistical evidence of performance differences, assessed using a Wilcoxon signed-rank test with Bonferroni correction. The results depicted refer to the test partition performances.
Asterisks denotes p-values in range *: 1.00e-02 $<$ p $\leq$ 5.00e-02; **: 1.00e-03 $<$ p $\leq$ 1.00e-02 ***: 1.00e-04 $<$ p $\leq$ 1.00e-03 ****: p $\leq$ 1.00e-04., and $\textup{(ns)}$ represents a non-significant value.

For tables, each cell reports the median and the $95\%$ confidence intervals (CI), which provide a textual description to the plots. The median and CIs are used due to their robustness to outliers. Each column corresponds to a method, and the best value in each row is highlighted.

\subsection{Online resources}

A Github repository is available at \url{https://github.com/gAldeia/itea-julia} with all implementations and datasets. Results and post-processing scripts are also included.

\section{Results and discussion}~\label{sec:optimization:resultsanddiscussion}

The results are divided into three parts.
First, a comparison is performed using the original ITEA implementation and the OLS adjustment, comparing its performance under restricted and unrestricted numbers of variables in the interactions.
Second, a comparison of the optimization heuristics proposed in this chapter is reported.
Finally, the relationship between execution time and the different optimization heuristics is analyzed.

\subsection{Restricted \textit{vs} unrestricted number of variables in the interaction}

\begin{figure}[htb]
    \centering
    \caption{NMSE of the original ITEA when the number of variables in the interaction is restricted and unrestricted (U).}
    \label{fig:optimization:NMSE_ITEA_restricted_unrestricted}
    \includegraphics[width=0.9\textwidth]{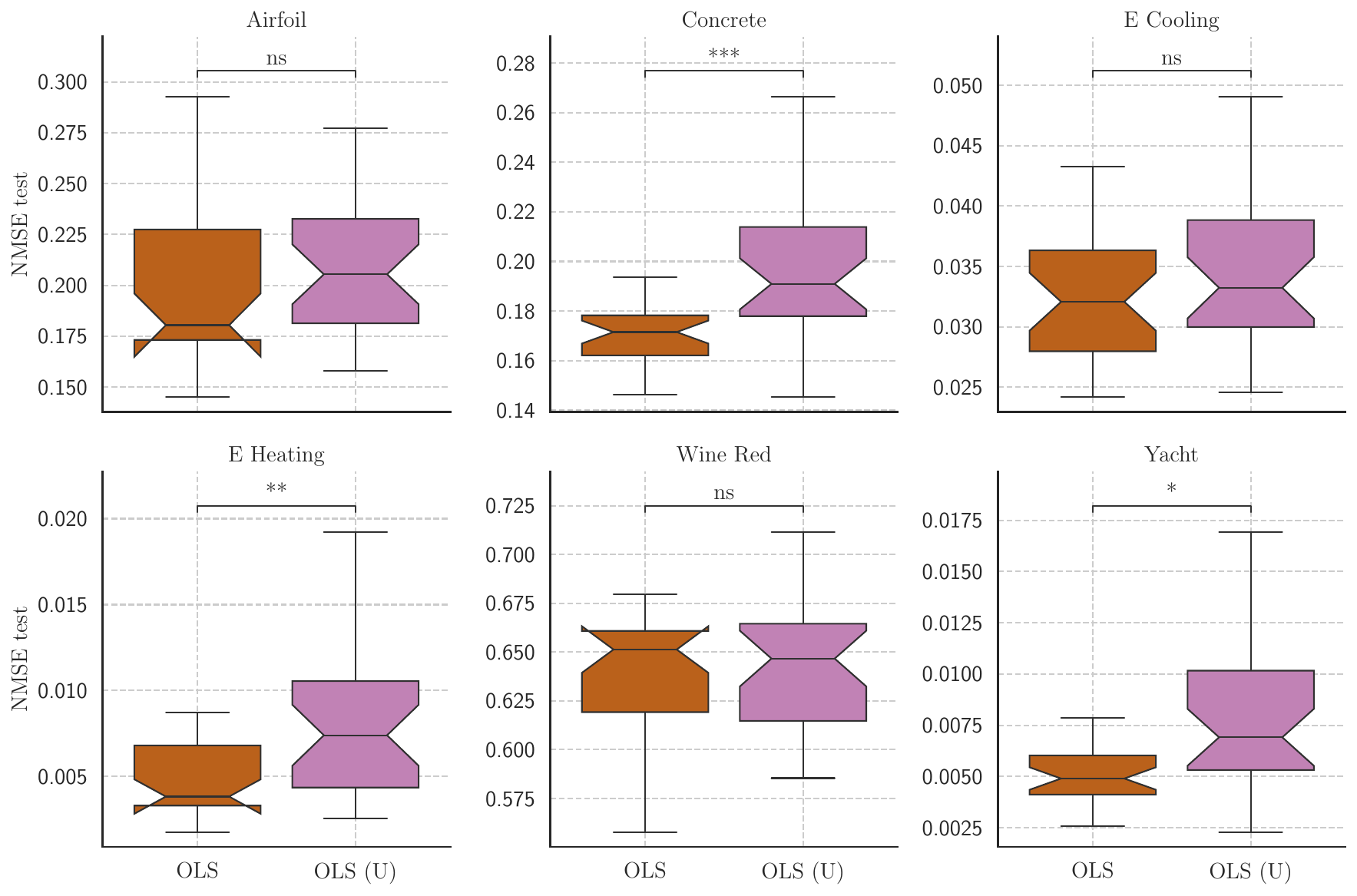}
    \legend{Source: Elaborated by the author.}
\end{figure}

\begin{figure}[htb]
    \centering
    \caption{Number of nodes of the original ITEA when the number of variables in the interaction is restricted and unrestricted (U).}
    \label{fig:optimization:N_Nodes_ITEA_restricted_unrestricted}
    \includegraphics[width=0.9\textwidth]{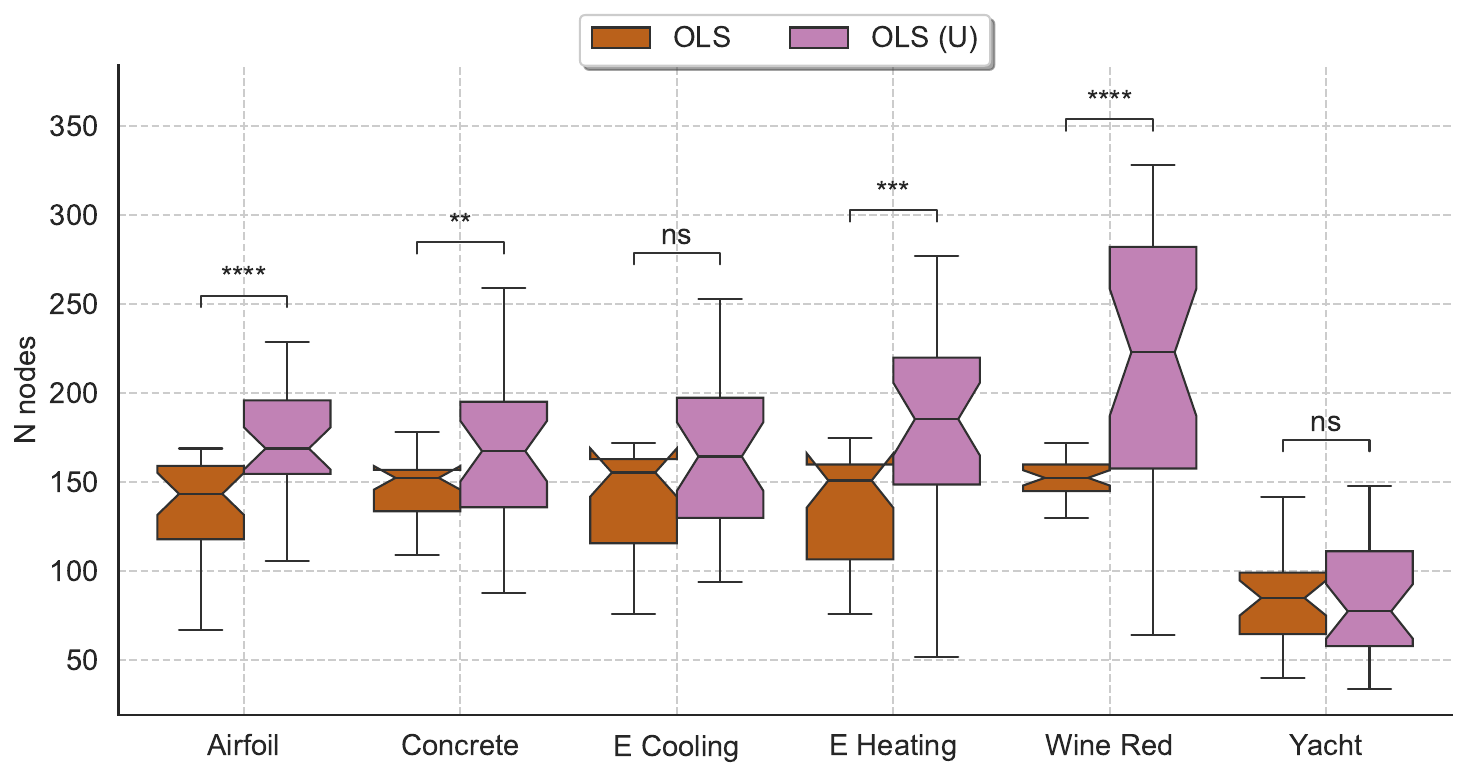}
    \legend{Source: Elaborated by the author.}
\end{figure}

The performance of the original ITEA algorithm is compared under two conditions: when the maximum number of variables in an interaction is restricted and when the interaction size is left unrestricted. In both cases, parameters are adjusted only through OLS, without consideration of inner parameters. The NMSE is reported in Figure~\ref{fig:optimization:NMSE_ITEA_restricted_unrestricted}, while the corresponding number of nodes is shown in Figure~\ref{fig:optimization:N_Nodes_ITEA_restricted_unrestricted}. The restricted version, with a maximum of two variables per interaction, is denoted as plain OLS, and the unrestricted version is denoted as OLS (U).

These results in Figure \ref{fig:optimization:N_Nodes_ITEA_restricted_unrestricted} reinforce the evidence that larger interaction terms do not consistently provides performance gains. Importantly, this figure illustrates that, despite the absence of accuracy improvements, the unrestricted version leads to considerable expression growth. This suggests that unconstrained interactions promote bloated models --- unnecessarily complex without justifiable improvements in predictive accuracy.
Regarding the number of nodes, the spread of the distributions in the unrestricted version is at least twice as big as the restricted one, with many datasets showing statistical differences.
This leads to the interpretation that larger interactions are more prone to overfit.

An interesting result emerges from Figures~\ref{fig:optimization:NMSE_ITEA_restricted_unrestricted} and ~\ref{fig:optimization:N_Nodes_ITEA_restricted_unrestricted}: allowing larger interactions does not statistically improve the NMSE, and sometimes hurts performance. This finding suggests that interpretability ---when measured in terms of the complexity of modeled interactions--- does not necessarily come at the expense of predictive accuracy. In other words, restricting interaction size can simplify models without degrading performance.

Moreover, there is no need to explore the interaction of all possible variables, as small interactions appear sufficient to capture the relevant structure. This points to the potential suitability of boosting-style approaches for SR, where multiple simple expressions are aggregated to form a more powerful model.

A comparison of NMSE on the test partition reveals that, for two out of the six datasets, there is no statistically significant difference between the restricted and unrestricted versions. For the remaining datasets, where the statistical test indicates significance, the restricted version consistently outperforms the unrestricted one. This trend is further supported by the median and confidence interval values in Table~\ref{tab:optimization:NMSE_ITEA_restricted_unrestricted}. Although the confidence intervals largely overlap, the restricted OLS variant generally achieves better performance than the unrestricted version.

\begin{table}[htb]
    \caption[Median NMSE of the ITEA when the number of variables in the interaction is restricted and unrestricted (U).]{Median NMSE of the ITEA when the number of variables in the interaction is restricted and unrestricted (U). $95\%$ confidence interval depicted in brackets. Best performance for each dataset depicted in bold.}
    \label{tab:optimization:NMSE_ITEA_restricted_unrestricted}
    \centering
    \begin{tabular}{rcc}
        \toprule\midrule
        Dataset & OLS & OLS (U) \\
        \midrule
    Airfoil & $\mathbf{0.18\, [0.16, 0.28]}$ & $0.21\, [0.16, 0.26]$\\
    Concrete & $\mathbf{0.17\, [0.15, 0.20]}$ & $0.19\, [0.15, 0.23]$\\
    Energy Cooling & $\mathbf{0.03\, [0.03, 0.04]}$ & $\mathbf{0.03\, [0.03, 0.06]}$\\
    Energy Heating & $\mathbf{0.00\, [0.00, 0.02]}$ & $0.01\, [0.00, 0.03]$\\
    Wine Red & $\mathbf{0.65\, [0.59, 0.75]}$ & $\mathbf{0.65\, [0.59, 0.78]}$\\
    Yacht & $\mathbf{0.00\, [0.00, 0.02]}$ & $0.01\, [0.00, 0.02]$\\
    \midrule\bottomrule
    \end{tabular}
    \legend{Source: Elaborated by the author.}
\end{table}

To enable a qualitative comparison of the algorithms, the smallest expressions obtained by ITEA in the restricted (Eq.~\ref{eq:optimization:itea_restricted}) and unrestricted (Eq.~\ref{eq:optimization:itea_unrestricted}) versions are reported for the airfoil dataset, which was selected for illustration due to its low dimensionality.
The features and their meanings are described in \citeonline{brooks1989airfoil} and presented in Table~\ref{tab:airfoil_features}. The goal is to predict the self-generated noise of an airfoil using the available features. The data were collected in a wind tunnel with multiple blades, where domain knowledge was used to characterize and predict self-noise behavior.

\begin{table}[htb]
    \centering
    \caption[Description of input variables for the Airfoil noise dataset.]{Description of input variables for the Airfoil noise dataset. These features are considered self-noise mechanisms, related to the phenomenon being predicted ($y$), the self-generated noise of different aircraft blades.}
    \label{tab:airfoil_features}
    \begin{tabular}{rl}
        \toprule\midrule
        Feature & Meaning \\ \midrule
        $f$ & Frequency, in hertz (Hz). \\
        $\alpha$ & Angle of attack, in degrees ($^\circ$). \\
        $c$ & Chord length, in meters (m). \\
        $U_{\infty}$ & Free-stream velocity, in meters per second (m/s). \\
        $\delta$ & Suction-side displacement thickness ($\delta$), in meters (m). \\
        $y$ & Scaled sound pressure level, in decibels (dB). \\
        \midrule \bottomrule
    \end{tabular}
    \legend{Source: Elaborated by the author.}
\end{table}

The Expression in Eq. \ref{eq:optimization:itea_restricted} took $84$ seconds to reach, and achieves an NMSE of $0.36$, with an effective expression size of $67$ nodes.


\begin{equation}\label{eq:optimization:itea_restricted}
    \begin{split}
    \textup{IT}_{\textup{OLS}} &= -283.83\cdot\sqrt{\frac{1}{f^{3}c^{3}}} -0.43\cdot \textup{cos}\left (\frac{1}{c^{3} U_\infty} \right ) -207.48\cdot \left (\frac{1}{U_\infty} \right ) \\
    & -2431.65\cdot \sqrt{\frac{\delta^3}{U_\infty}} -21.42\cdot \sqrt{c} -0.023\cdot \sqrt{f^2 \delta} + 143.0
    \end{split}
\end{equation}

The Expression found in the unrestricted scenario in Eq. \ref{eq:optimization:itea_unrestricted} took $140$ seconds to reach, and achieves an NMSE of $0.27$, with an effective expression size of $91$ nodes.

\begin{equation}\label{eq:optimization:itea_unrestricted}
    \begin{split}
    \textup{IT}_{\textup{OLS (U)}} &= 2.31 \cdot \textup{log}\left (\frac{U_\infty^3}{f^3 c \delta^2} \right ) + 19523.49\cdot \textup{sin} \left (\frac{c^3}{f^3 U_\infty \delta^3}\right ) \\
    & + 1718.74\cdot \textup{sin} \left (\frac{\alpha^2 c}{f^3 \delta^2} \right ) -10210.023\cdot \textup{sin}\left (\frac{1}{f^2 \delta}\right ) \\
    & -1.73 \cdot \left (\frac{\alpha}{f^2 c^2}\right ) -6996.82\cdot \textup{cos}\left (\frac{1}{U_\infty} \right ) + 7118.0
    \end{split}
\end{equation}

When the expressions are compared, the restricted version is observed to yield simpler and more interpretable equations, although not inherently interpretable. The former equation is shorter and involves simpler interaction terms; however, some exponents equal to three are noted, which hinders interpretability. In contrast, the latter equation is characterized by extensive use of trigonometric functions and larger parameter values.

Such findings are particularly compelling because they indicate opportunities to improve both objectives simultaneously: model size and predictive performance. In this case, the presumed trade-off between accuracy and model size does not hold, and constraining interaction size offers a principled way to mitigate bloat while preserving or even enhancing generalization.

\subsection{Parameter optimization heuristics}

\begin{figure}[htb]
    \centering
    \caption{NMSE of the four proposed parameter optimization heuristics for the benchmark datasets.}
    \label{fig:optimization:NMSE_optimization_heuristics}
    \includegraphics[width=0.9\textwidth]{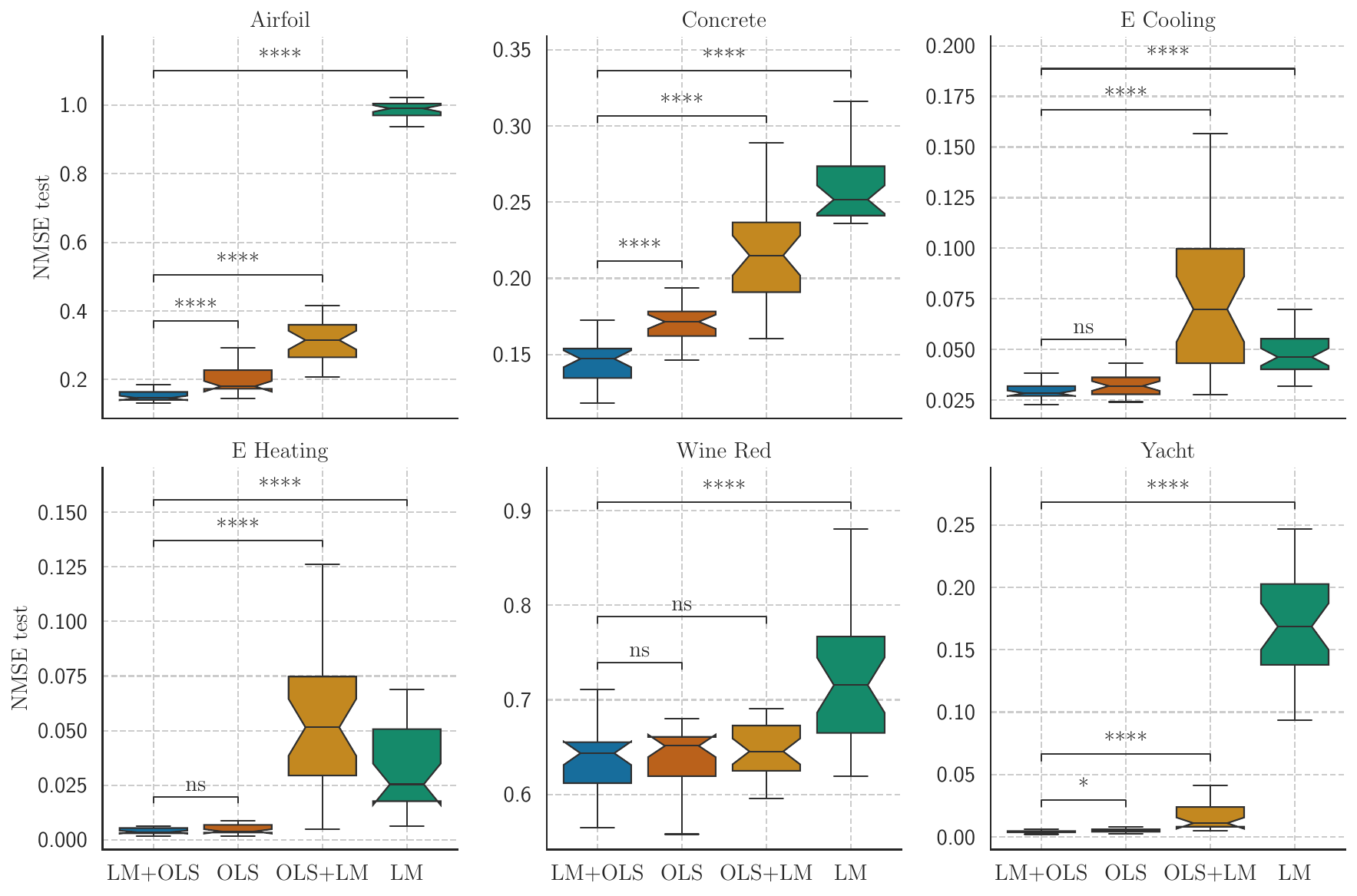}
    \legend{Source: Elaborated by the author.}
\end{figure}

To investigate the role of parameter optimization, four distinct heuristics were proposed and evaluated. Figure~\ref{fig:optimization:NMSE_optimization_heuristics} reports the NMSE on the test partitions across all datasets for each heuristic. In this version of ITEA, expressions were restricted to a maximum of two variables per interaction, based on findings from the previous section.
The boxplots show that the ITEA with LM was the worst alternative in almost all datasets, while the combination of LM+OLS was the best. Although LM+OLS was consistently the best-performing method, in four of the datasets, there is no statistical difference between the LM+OLS and the second-best method, OLS. The OLS+LM and LM methods usually present statistical significance compared with the best method.

The results highlight several important aspects. First, relying solely on linear parameters (OLS heuristic) consistently worse results compared to LM+OLS. This confirms that nonlinearities play a major role in improving predictive performance. While statistical significance was observed in two datasets between LM+OLS and OLS (datasets Airfoil and Concrete), the remaining four showed no significant difference, indicating that the benefits of non-linear optimization may be dependent on the dataset.
Secondly, LM+OLS does not degrade performance when compared to pure OLS, suggesting that non-linear optimization can provide improvements without hurting general performance.

The LM+OLS being the best performing method also suggests that local optimization of the small expressions, followed by a global combination, may show better results. This perspective suggests that hybrid optimization heuristics, which balance linear and non-linear adjustments, may provide a good direction for improving symbolic regression algorithms in terms of parameter optimization.
This is also particularly interesting because non-linear optimization is faster and has a less complex landscape when few parameters are optimized rather than several parameters at once.

Table~\ref{tab:optimization:nmse_table} reports the median and CI test NMSE for all datasets and optimization heuristics. The results confirm that LM+OLS consistently achieves the lowest or tied median NMSE when compared to OLS, reinforcing its effectiveness.

\begin{table}[htb]
    \caption[Median NMSE of the four proposed parameter optimization heuristics on the test partition.]{Median NMSE of the four proposed parameter optimization heuristics on the test partition. $95\%$ confidence interval depicted in brackets. Best performance for each dataset depicted in bold.}
    \label{tab:optimization:nmse_table}
    \centering
    \begin{tabular}{@{}rcccc@{}}
    \toprule\midrule
    Dataset & LM & LM+OLS & OLS & OLS+LM \\
    \midrule
    Airfoil &$\mathbf{0.15\, [0.13, 0.18]}$&$0.18\, [0.16, 0.28]$&$0.32\, [0.22, 0.41]$&$0.99\, [0.94, 1.02]$\\
    Concrete &$\mathbf{0.15\, [0.12, 0.17]}$&$0.17\, [0.15, 0.20]$&$0.22\, [0.17, 0.28]$&$0.25\, [0.24, 0.35]$\\
    Energy Cooling &$\mathbf{0.03\, [0.02, 0.04]}$&$\mathbf{0.03\, [0.03, 0.04]}$&$0.07\, [0.03, 0.12]$&$0.05\, [0.03, 0.07]$\\
    Energy Heating &$\mathbf{0.00\, [0.00, 0.01]}$&$\mathbf{0.00\, [0.00, 0.02]}$&$0.05\, [0.02, 0.10]$&$0.03\, [0.01, 0.07]$\\
    Wine Red &$\mathbf{0.64\, [0.58, 0.67]}$&$0.65\, [0.59, 0.75]$&$0.65\, [0.60, 0.69]$&$0.72\, [0.64, 0.86]$\\
    Yacht &$\mathbf{0.00\, [0.00, 0.01]}$&$\mathbf{0.00\, [0.00, 0.02]}$&$0.01\, [0.01, 0.04]$&$0.17\, [0.11, 0.24]$\\
    \midrule\bottomrule
    \end{tabular}
    \legend{Source: Elaborated by the author.}
\end{table}

Nevertheless, a trade-off must be considered. While LM+OLS yields superior predictive performance, its implementation is considerably more complex, involving non-linear optimization methods and good initialization. In contrast, OLS is straightforward to implement, computationally efficient, and still produces competitive results, especially in cases where non-linear optimization offers only marginal improvements. This balance between implementation effort and predictive gains should be taken into account when designing symbolic regression algorithms for practical applications.

The results show that non-linear adjustment is challenging, and the IT representation does not benefit substantially from applying the LM algorithm directly. The number of free parameters to optimize is large, and convergence is highly dependent on the starting point, which often leads the LM to suboptimal parameter vectors. 
This also suggests that the LM may be suffering from overparameterization, as the addition of the inner parameters to the IT expressions can quickly increase the total number of parameters in the models.

\begin{figure}[htb]
    \centering
    \caption{Number of nodes of the four proposed parameter optimization heuristics for the benchmark datasets.}
    \label{fig:optimization:N_nodes_optimization_heuristics}
    \includegraphics[width=0.9\textwidth]{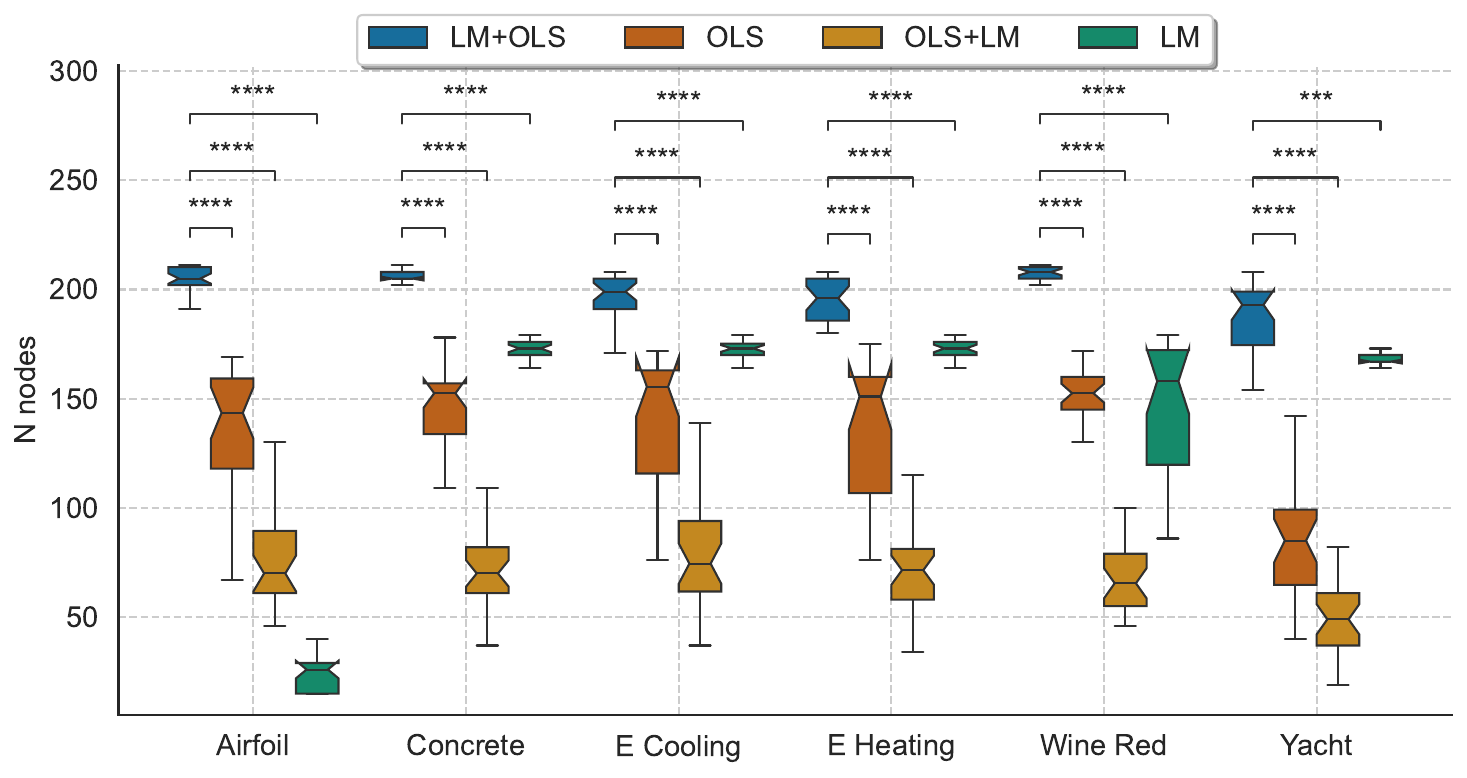}
    \legend{Source: Elaborated by the author.}
\end{figure}

However, this improvement comes with a cost in model size. Figure~\ref{fig:optimization:N_nodes_optimization_heuristics} shows that LM+OLS tends to generate significantly larger expressions, tending to overparameterize the expressions. Heuristics such as OLS+LM or OLS achieve a more balanced trade-off between accuracy and expression size, with overlapping distributions in many datasets. This suggests that while stronger parameter optimization can boost predictive accuracy, it also risks inflating model size unnecessarily. In practice, OLS alone emerges as a compelling choice, combining simplicity, computational efficiency, and robustness without sacrificing much predictive performance.

\begin{figure}[htb]
    \centering
    \caption{Number of nodes \textit{versus} NMSE of the four proposed parameter optimization heuristics for the airfoil dataset.}
    \label{fig:optimization:n_nodes_nmse}
    \includegraphics[width=0.6\textwidth]{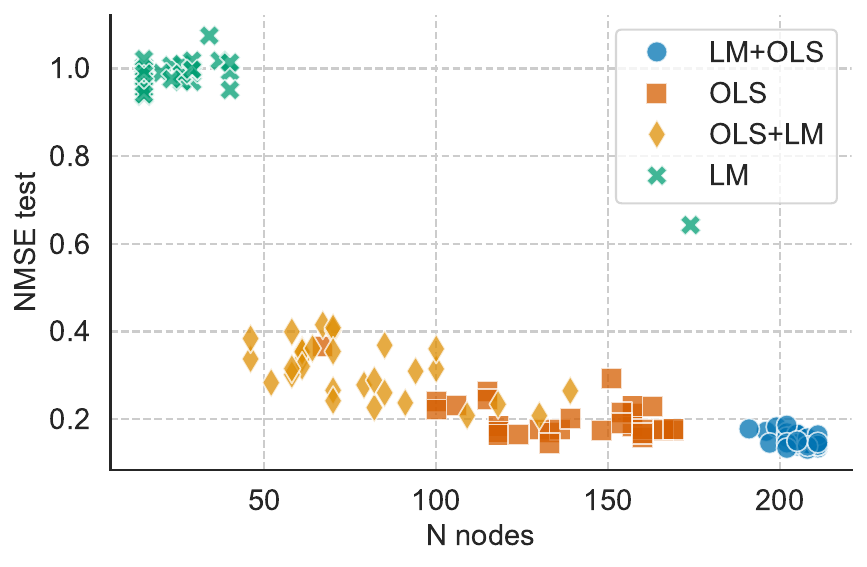}
    \legend{Source: Elaborated by the author.}
\end{figure}

Despite the LM+OLS being the best heuristic in terms of NMSE, it produces the largest expressions compared to the other approaches. The OLS was outperformed only by OLS+LM, but this heuristic did not present a better NMSE on the test partition. Fig. \ref{fig:optimization:n_nodes_nmse} shows the NMSE \textit{versus} the number of nodes for the airfoil dataset, which helps to highlight the compromise between the number of nodes and error.

LM, which produced smaller expressions, had the highest error. Meanwhile, the other heuristics have all close values in error but with a large spread in the number of nodes. The method that presented the best compromise between the number of nodes and error was the OLS+LM. The method with the smallest error, LM+OLS, was the one with the largest expressions.

An initial argument against adding nonlinear parameters to the original ITEA implementation is that doing so could reduce interpretability by producing larger models. However, the reported results show that the version using only linear parameters (OLS) actually produces models that are, on average, larger than those generated by the LM and OLS+LM variants. It is also worth noting that combining two optimization heuristics, as in OLS+LM, yields even smaller expressions --- although the resulting models include more sophisticated parameters that may, in practice, make their behavior harder to interpret.

To illustrate, the smallest expression found by the LM+OLS heuristic, which achieved the performance of $0.17$ in terms of NMSE for the airfoil dataset, is reported in Eq.~\ref{eq:optimization:lm_ols}, with an expression size of $191$ nodes.

\begin{equation}\label{eq:optimization:lm_ols}
    \begin{split}
    \textup{IT}_{\textup{LM+OLS}} &= -3.38\cdot \left ( 52.81-1415.57\cdot \frac{1}{f^3 c^3} \right ) \\
    &+ 4.00\cdot \left ( -105.46 - 4.26\cdot \frac{1}{\alpha^2 c^3}\right ) \\
    & + \ldots \textup{(12 terms ommited)} - 1411.0
    \end{split}
\end{equation}

\subsection{Execution time}

Regarding execution time, the median and CI are reported in Table~\ref{tab:optimization:exec_time}. It can be observed that all heuristics incorporating LM in the adjustment procedure required longer execution times than OLS alone. It is also evident that LM alone was the slowest method in many datasets, with a notably large CI range.

\begin{table}[htb]
    \caption[Median Execution time (in seconds, with no decimal places) of the four proposed parameter optimization heuristics.]{Median Execution time (in seconds, with no decimal places) of the four proposed parameter optimization heuristics. $95\%$ confidence interval depicted in brackets. Best performance for each dataset depicted in bold.}
    \label{tab:optimization:exec_time}
    \centering
    \begin{tabular}{@{}rcccc@{}}
    \toprule\midrule
        Dataset & LM & LM+OLS & OLS & OLS+LM \\
    \midrule
    Airfoil       & $630\,[443,1455]$ & $\mathbf{147\,[80,226]}$ & $389\,[269,3035]$ & $2066\,[1109,5567]$ \\
    Concrete      & $427\,[306,637]$  & $\mathbf{107\,[68,150]}$ & $258\,[184,819]$  & $351\,[262,553]$    \\
    Energy Cooling& $657\,[373,869]$  & $\mathbf{106\,[55,138]}$ & $718\,[341,7117]$ & $1133\,[311,6110]$  \\
    Energy Heating& $576\,[232,995]$  & $\mathbf{94\,[66,130]}$  & $734\,[306,4399]$ & $712\,[332,4175]$   \\
    Wine Red      & $1000\,[713,2339]$& $\mathbf{273\,[214,318]}$& $1157\,[709,4620]$& $5238\,[926,33219]$ \\
    Yacht         & $138\,[90,360]$   & $\mathbf{25\,[12,51]}$   & $574\,[185,2411]$ & $820\,[326,22713]$  \\
    \midrule\bottomrule
    \end{tabular}
    \legend{Source: Elaborated by the author.}
\end{table}

\begin{figure}[htb]
    \centering
    \caption{Execution time \textit{versus} Number of nodes of the four proposed parameter optimization heuristics for the airfoil dataset.}
    \label{fig:optimization:exectime}
    \includegraphics[width=0.65\textwidth]{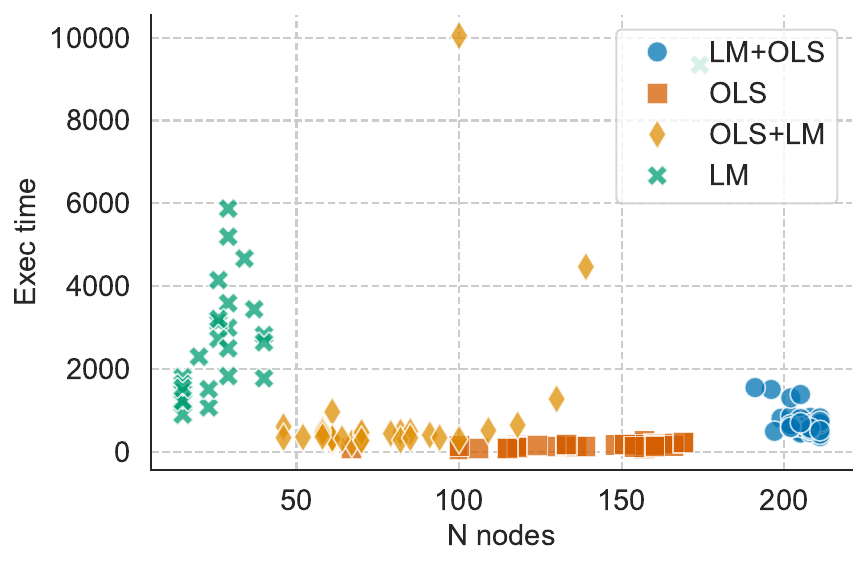}
    \legend{Source: Elaborated by the author.}
\end{figure}

To visualize the relationship between the number of nodes and execution time, Fig. \ref{fig:optimization:exectime} shows, for the airfoil dataset, the number of nodes of the final expression \textit{versus} the execution time it took.
Except for a few outliers, OLS+LM, LM+OLS, and OLS were observed to be faster than LM alone. The smallest spread in execution time was obtained with OLS.

The compromise between the number of nodes and execution time is counter-intuitive: smaller equations had the highest execution time. This shows that execution time is more related to the optimization method than to the complexity of the equations that can be generated.

When comparing OLS and OLS (U) execution times, they had no statistically significant difference for all datasets, indicating they have similar execution times.

\section{Conclusions}~\label{sec:optimization:conclusions}

In this chapter, the effect of adding inner parameters to Interaction-Transformation expressions was studied, using four different heuristics for non-linear optimization. The inner parameters, which are part of the interaction functions, shifts and scale the arguments of the transformation functions. This introduces a non-linear optimization problem while maintaining a linear structure with identifiable parameters. As a by-product, the ITEA algorithm was extended to support non-linear optimization, and the implementation was made available in Julia.

An initial experiment was conducted to impose a limit on the number of variables in the interactions. Empirical evaluation showed that restricting interactions produced results equal to or better than those obtained with the unrestricted version. This limitation was found to improve the interpretability and simplicity of the generated expressions.

Subsequently, four different heuristics for optimizing linear and non-linear parameters in IT expressions were compared: optimizing only the linear parameters with OLS; optimizing all parameters using LM; using OLS to optimize linear parameters as a starting point for LM optimization of all parameters; and optimizing the inner parameters of each individual term with LM before combining them with OLS.
It was observed that a combination of Levenberg-Marquardt (LM) and Ordinary Least Squares (OLS) methods produced the best predictive performance, albeit at the cost of generating larger expressions. Among the proposed heuristics, LM+OLS was identified as the most effective overall. The role of OLS in this combination was to regularize the influence of individual IT tuples by assigning low parameters to poorly performing terms, while guaranteeing convergence to the iterative least-squares solution.

The findings are expected to generalize to symbolic regression methods that rely on linear optimization and constraint the hypothesis space to functions expressed as linear combinations of non-linear terms. 
The OLS optimization is sufficiently simple to be incorporated into SR frameworks, and performance can be further enhanced by combining them with local non-linear optimization. Overparameterization was found to negatively affect the performance of non-linear optimization alone, and the choice of starting points was observed to be crucial.



%% file: Capitulos/Selection/selection.tex
\Chapter{Parent selection}{A Variance-based approach for \texorpdfstring{$\epsilon$-}{epsilon-}lexicase}~\label{chap:parentsel}
\chaptermark{Parent selection}

\begin{laysummary}
    Parent selection is a crucial component of evolutionary algorithms, and numerous strategies have been proposed to build the parent pool to create the next generation. Common approaches rely on the average error across the entire dataset as the selection criterion, which may result in a loss of information due to the aggregation of all test cases.
    Under $\epsilon$-lexicase selection, the population goes into a selection pool that is iteratively reduced by evaluating shuffled test cases individually, discarding individuals with errors greater than the elite error plus the Median Absolute Deviation (MAD) of errors for that specific case. To better capture differences in individual performance across test cases, the use of the Minimum Variance Threshold (MVT) was proposed. This method partitions errors into two groups by minimizing the within-partition variance, inspired by the splitting criterion of decision trees.
    The proposed criterion was embedded into the FEAT symbolic regression algorithm under static and dynamic heuristics. Two sets of experiments were conducted: the first evaluated convergence behavior, prediction error, and size distributions on six widely used real-world datasets; the second employed the SRBench framework, which includes $122$ synthetic and real-world regression tasks. The results indicated that dynamic heuristics achieved improved convergence and performance comparable to MAD, albeit requiring more test cases during selection.
    Empirical findings demonstrated that the proposed MVT-based approach is competitive with traditional $\epsilon$-lexicase with MAD on real-world datasets, yielding a superior trade-off between model size and $R^2$. On SRBench synthetic datasets, equivalent performance to MAD was achieved, with the proposed method ranking first more frequently.
\end{laysummary}

{\let\thefootnote\relax\footnotetext{Part of this chapter was published as: Guilherme Seidyo Imai Aldeia, Fabrício Olivetti de França, and William G. La Cava. 2024. Minimum variance threshold for \texorpdfstring{$\epsilon$}{epsilon}-lexicase selection. In Genetic and Evolutionary Computation Conference (GECCO '24), July 14-18, 2024, Melbourne, VIC, Australia. ACM, New York, NY, USA, 9 pages. \url{https://doi.org/10.1145/3638529.3654149}}}

\vfill
\pagebreak

\section{Introduction}

Symbolic Regression (SR) methods are commonly based on genetic programming. An essential aspect of any evolutionary algorithm is the selection of solutions that are recombined to generate new \emph{offspring} solutions. The main goal of the recombination is to mix different parts of promising parents, with the expectation that improvements over the parents are achieved in the offspring. For this reason, the parent selection process requires that chosen individuals perform well on different samples of the dataset.

Traditional approaches, such as tournament selection \cite{miller1995genetic, brindle1980genetic}, rely on aggregated fitness information for the selection of parents. Consequently, parents performing well on average across the entire dataset are favored, while individuals excelling in difficult subsets of the problem may be missed. Evidence has shown that selection based on aggregated fitness scores discards relevant cases, as less information is preserved \cite{10.1145/2576768.2598288}.

An alternative approach is based on distinguishing individuals in difficult test cases. This idea underlies lexicase selection \cite{10.1145/2330784.2330846}, a method demonstrated to achieve state-of-the-art performance in program synthesis and symbolic regression \cite{10.1145/3583131.3590375}. It has also been incorporated into SR algorithms such as EPLEX \cite{eplex} and FEAT \cite{cava2018learning}. Lexicase selection is based on the assumption that problem modularity is reflected in individual test cases, each representing a distinct circumstance under which the program must succeed. By avoiding the aggregation of all test cases, lexicase selection promotes diversity among selected parents, since any individual performing well on at least one test case retains a chance of being chosen.

To select a parent, the entire population is placed into a selection pool. A test case is then chosen at random, and all individuals failing that case are eliminated. Subsequent test cases are randomly selected, and eliminations continue until only one individual remains. If all test cases are exhausted before the pool is reduced to one candidate, an individual is chosen at random with uniform probability among the remaining individuals in the pool.

The original lexicase method was designed for binary outcomes, which made it well suited to classification problems. However, modifications were required for regression, where test cases yield continuous error values rather than Boolean outcomes. This led to the introduction of $\epsilon$-lexicase selection \cite{la_cava_epsilon-lexicase_2016}, in which a tolerance threshold $\epsilon$ is defined. For each test case, this threshold is determined from the Median Absolute Deviation (MAD) of errors, and all individuals within the tolerance are retained in the pool. This adaptation has been shown effective in symbolic regression compared to traditional selection techniques \cite{geiger_was_2025,geiger_performance_2025,helmuth_benchmarking_2020}.

In addition, the theoretical analysis of lexicase selection has revealed substantial challenges. The direct computation of selection probabilities under lexicase has been proven NP-hard \cite{dolson_calculating_2023}, and no polynomial-time algorithm for this estimation has been identified.

In this chapter, a new method for estimating $\epsilon$ is proposed. Instead of relying on the MAD, the threshold is defined such that test errors are partitioned into two groups while minimizing the total variance within them. This criterion corresponds to a form of information entropy, a concept commonly used in decision tree methods \cite{breiman1984classification}. The criterion works as a one-dimensional clustering, making it more sensitive to similarities in performance among individuals.

The proposed method is implemented into FEAT, a SR framework that has previously incorporated $\epsilon$-lexicase selection \cite{srbench} and has been successfully applied to real-world problems \cite{la_cava_flexible_2023}. The evaluation is twofold: \textit{(i)} an in-depth analysis of performance characteristics during evolution using six low-dimensional datasets, and \textit{(ii)} a large-scale assessment with SRBench \cite{srbench}, a unified benchmarking framework for SR algorithms.

Empirical results demonstrate that the proposed method achieves superior $R^2$ performance while maintaining solution complexity at levels comparable to MAD-based $\epsilon$-lexicase. Within SRBench, the new method improved the accuracy complexity trade-off of FEAT on real-world problems and improved its Pareto Front ranking, and on synthetic problems it increased the frequency of top $1$ performance of FEAT.

The remainder of the chapter is organized as follows. Section \ref{sec:selection:relatedwork} reviews related work. Section \ref{sec:selection:epsilon-lexicase} provides a detailed discussion of the canonical $\epsilon$-lexicase parent selection method. Section \ref{sec:selection:methods} introduces the symbolic regression framework, details the proposed modification, and outlines the experimental design. Results are analyzed in Section \ref{sec:selection:results-discussion}, followed by conclusions and future directions in Section \ref{sec:selection:conclusions}.

\section{Related work}~\label{sec:selection:relatedwork}

The $\epsilon$-lexicase selection method was introduced by \citeonline{la_cava_epsilon-lexicase_2016}, where it was argued that although lexicase is well suited for classification, it is not directly applicable to regression. Variants for computing $\epsilon$ thresholds, including static, semi-dynamic, and dynamic approaches, were later proposed by \citeonline{eplex}, highlighting the difficulty and relevance of stablishing a threshold criterion.

Since then, many benchmarks of $\epsilon$-lexicase variants have been performed. \citeonline{geiger_performance_2025} evaluated multiple versions, including batch and downsampled approaches, and reported that $\epsilon$-lexicase performs best under random downsampling. In contrast, tournament selection was shown to perform well only when combined with informed downsampling, in which redundant cases are removed and diverse subsets of training cases are used, a strategy that somehow mimics the behavior of $\epsilon$-lexicase of not using all test cases aggregated.
Another comparison of parent selection methods by \citeonline{helmuth_benchmarking_2020} revealed that lexicase and its variants consistently achieved top performance, although $\epsilon$-lexicase specifically did not appear among the best-performing methods in their benchmark. In addition, \citeonline{geiger_was_2025} also benchmarked $\epsilon$-lexicase and tournament selection and reported that $\epsilon$-lexicase selects individuals close to elite solutions when squared errors are considered, which can be considered a downside, although good results were obtained under this variant. 
Previous work suggested and evaluated downsampling the test cases, achieving similar or better performance for lexicase, \cite{10.1162/evco_a_00346, 10.1145/3319619.3326900, 10.1145/2908812.2908851}, and $\epsilon$-lexicase was also shown to hold these benefits \cite{geiger_down-sampled_2023}.

The exploration behavior of the threshold in $\epsilon$-lexicase has also been studied. \citeonline{hernandez_exploration_2021} used a fixed $\epsilon$ as absolute error, concluding that lexicase selection is sensitive to expansions of the test set, showing that exploration is improved under $\epsilon$-lexicase and investigated the influence of $\epsilon$ on exploration. \citeonline{la_cava_epsilon-lexicase_2016} found that larger $\epsilon$ values increase diversity, since the cut-off threshold is less strict.

Other approaches attempted to eliminate the threshold altogether by estimating probabilities corresponding to $\epsilon$-lexicase outcomes \cite{10.1145/3583131.3590375}. However, this problem was demonstrated to be NP-hard by \citeonline{dolson_calculating_2023}. Dolson has also been indicated that $\epsilon$-lexicase permits various subsets of solutions to tie under a given criterion, suggesting that empirical evaluation remains highly relevant.

Regarding computational complexity, the authors in \citeonline{10.1007/978-3-031-14721-0_34} showed that lexicase selection (and, by extension, $\epsilon$-lexicase in its dynamic and semi-dynamic variants) has a time complexity of $O(d \cdot p)$, with $p$ being the population size, which naturally increase to $O(d \cdot p^2)$ when selecting $p$ parents. However, the actual running time is typically much lower when the population is diverse. Their proof relies on clustering individuals based on their performances, and they observed that small clusters (\textit{i.e.}, few individuals with high similarity within $\epsilon$) result in shorter runtimes.

All these related work shows the importance of the determination of the threshold, as it plays a major role in $\epsilon$-lexicase, and it is still an open question.

\section{\texorpdfstring{$\epsilon$}{epsilon}-lexicase selection}~\label{sec:selection:epsilon-lexicase}

A test case is defined as a pair $(\mathbf{x}_t, y_t)$, $t = 1, 2, \ldots, d$, taken from the training data used during fitness evaluation. The maximum number of test cases is equal to the number of available samples in the training data. Any possible test case is denoted as $t \in \mathcal{T}$. The error of an individual $p$, denoted as $e_t(p)$, is defined as the absolute difference between its predicted value given $\mathbf{x}_t$ and the target value $y_t$. The vector of errors for a test case $t$, denoted as $\mathbf{e}_t$, is formed by the concatenation of errors for all individuals in the selection pool. The set $\mathcal{P}$ denotes the population, and the set $\mathcal{S}$ denotes the selection pool.

In the static scenario, the criterion for remaining in the pool is defined as $e_t^* + \epsilon$, where $e_t^*$ represents the best error from the entire population for that case. In the semi-dynamic and dynamic implementations, the threshold for remaining in the pool is calculated as the smallest error ($\mathtt{elite}$) plus $\epsilon$. The distinction between the semi-dynamic and dynamic versions is that, in the former, $\epsilon$ is calculated over the entire population, while in the latter, it is calculated over the pool, making the process more computationally expensive.
A comparison of the static, semi-dynamic, and dynamic approaches is characterized by distinct configurations regarding the population pool and the criterion applied for selection, and is summarized in Table \ref{tab:selection:comparison_e_lexicases}.

\begin{table}[htb]
    \centering
    \caption{Comparison of $\epsilon$-lexicase variations: Strategy, Pool, criterion, and Characteristic.}
    \label{tab:selection:comparison_e_lexicases}
    \footnotesize
    \begin{tabular}{@{}lllp{7.5cm}@{}}
    \toprule\midrule
    \textbf{Strategy} & \textbf{Pool} & \textbf{Criterion} & \textbf{Characteristic} \\ \midrule
    Static & pop ($\mathcal{P}$) & $e_t^* + \epsilon_t$ & The error vector is calculated previously with entire population. A binary mask matrix stores the global criterion to iterate through the random cases \\ \hline
    Semi-dynamic & pop ($\mathcal{P}$) & 
    \begin{tabular}[c]{@{}l@{}}$\texttt{elite} + \epsilon_t$,\\
    where $\mathbf{\epsilon} \leftarrow \lambda(\textbf{e}_t) \text{ for } t \in \mathcal{T}$\end{tabular} & The error vector is calculated previously, but the decision criterion is based on the individuals that are in the pool, as it uses the elite error in the criterion \\ \hline
    Dynamic & pool ($\mathcal{S}$) & 
    \begin{tabular}[c]{@{}l@{}}$\texttt{elite} + \epsilon_t$,\\
    where $\mathbf{\epsilon} \leftarrow \lambda(\textbf{e}_t(\mathcal{S}))$\end{tabular} &
    Criterion is dependent on the elite, and $\lambda$ is calculated with the selection pool \\ \midrule\bottomrule
    \end{tabular}
    \legend{Source: Elaborated by the author.}
\end{table}

In all $\epsilon$-lexicase scenarios, $\epsilon$ is calculated as the Median Absolute Deviation (MAD) \cite{PHAMGIA2001921}:

\begin{equation}
\epsilon_t = \lambda(\mathbf{e}_t) = \text{median}(|\mathbf{e}_t - \text{median}(\mathbf{e}_t)|).
\end{equation}

The MAD yields median-centered random variables with greater robustness than the mean, since the median is more representative of the center than the mean in asymmetrical distributions \cite{PHAMGIA2001921}. Within the $\epsilon$-lexicase selection context, the MAD is interpreted as a dispersion measure relative to the center of the distribution of errors in the current selection pool.

In the following subsections, pseudo-code is presented for the implementation of the static and dynamic strategies. The pseudo-codes are divided into two functions to illustrate in the first one any pre-calculation that can be performed before selecting the parents to emphasize where each implementation can take advantage of calcualting the information just once, or whether it have to perform computationally-intensive calculations every time a new parent must be selected.

\subsection{Static \texorpdfstring{$\epsilon$}{epsilon}-lexicase}~\label{subsec:selection:static-lexicase}

In static $\epsilon$-lexicase, the criterion for retaining individuals in the selection pool is based on a fixed threshold that remains constant throughout the selection process and is calculated before selection is executed. For each test case, individuals with errors less than or equal to the sum of the best error in the population ($e_t^*$) plus $\epsilon$ are kept in the pool.

The pseudo-code in Algorithm \ref{alg:selection:static_lexicase_selection} shows that the threshold is first defined based on the entire population, and then, for a number $\texttt{ns}$ of iterations, parents are selected using the procedure in Algorithm \ref{alg:selection:static_lexicase_get_parent}.

\begin{algorithm}[htb]
    \caption{Static $\epsilon$-lexicase selection}~\label{alg:selection:static_lexicase_selection}
    \begin{algorithmic}[1]
        \REQUIRE{ individuals $\mathcal{P}$, test cases $\mathcal{T}$, number of selections $\texttt{ns}$ }
        \ENSURE{ selected individuals $\mathcal{Q}$ }

        \STATE $\mathcal{Q} \leftarrow \emptyset$
        
        \STATE $\mathbf{\epsilon} \leftarrow \lambda(\textbf{e}_t) \text{ for } t \in \mathcal{T}$

        \STATE $f \leftarrow \text{ mapping function for } p \in \mathcal{P}$

        \FOR{$t \in \mathcal{T} \text{ and } p \in \mathcal{P}$}
            \IF{$e_t(p) \leq {e_t^* + \epsilon_t}$}
                \STATE $f_t(p) = 0$
            \ELSE
                \STATE $f_t(p) = 1$
            \ENDIF
        \ENDFOR
        
        \FOR{$\texttt{ns} \text{ times}$}
            \STATE $\mathcal{Q} \leftarrow \mathcal{Q} \cup \text{get\_parent\_static}(\mathcal{P}, \mathcal{T}, f)$
        \ENDFOR

        \STATE \textbf{return} $\mathcal{Q}$
    \end{algorithmic}
\end{algorithm}

\begin{algorithm}[htb]
    \caption{Static parent selection (get\_parent\_static)}~\label{alg:selection:static_lexicase_get_parent}
    \begin{algorithmic}[1]
        \REQUIRE{ individuals $\mathcal{P}$, test cases $\mathcal{T}$, pass condition function $f$ }
        \ENSURE{ individual from $\mathcal{P}$ }

        \STATE $\mathcal{T}' \leftarrow \mathcal{T}$
        \STATE $\mathcal{S} \leftarrow \mathcal{P}$
        \WHILE{$|\mathcal{T}'|>0 \text{ and } |\mathcal{S}|>1$}
            \STATE $t \leftarrow \text{ random choice from } \mathcal{T}'$
            \STATE $\texttt{elite} \leftarrow \text{min } f_t(p) \text{ for } p \in \mathcal{S}$
    
            \FOR{$p \in \mathcal{S}$}
                \IF{$f_t(p) \neq \texttt{elite}$}
                    \STATE $\mathcal{S} \leftarrow \mathcal{S} \setminus \{p\}$
                \ENDIF
            \ENDFOR
            \STATE $\mathcal{T}' \leftarrow \mathcal{T}' \setminus \{t\} $
        \ENDWHILE
        
        \STATE \textbf{return} \text{ random choice from } $\mathcal{S}$
    \end{algorithmic}
\end{algorithm}

It should be noted that preparation for static $\epsilon$-lexicase can take advantage of masking individuals with binary values that indicate, for each test case, whether the error is below or above the threshold. This allows runtime improvements by pre-calculation of errors, avoiding repeated evaluations during parent selection, since the order of test cases does not affect the outcome.

\subsection{Semi-dynamic \texorpdfstring{$\epsilon$}{epsilon}-lexicase}~\label{subsec:selection:semi-lexicase}

In semi-dynamic $\epsilon$-lexicase, the threshold for retaining individuals in the selection pool is dynamically adjusted based on the errors of the individuals currently present in the pool. The semi-dynamic scenario is differentiated by the use of the best error ($\texttt{elite}$) for the test case to update the threshold, while the $\epsilon$ value continues to be computed from the error distribution of the entire population using MAD. This approach allows the selection criterion to adapt to the performance of the current pool.
The implementations of this strategy are presented in Algorithms \ref{alg:selection:semi_dynamic_lexicase_selection} and \ref{alg:selection:semi_dynamic_lexicase_get_parent}.

\begin{algorithm}[htb]
    \caption{Semi dynamic $\epsilon$-lexicase selection}\label{alg:selection:semi_dynamic_lexicase_selection}
    \begin{algorithmic}[1]
        \REQUIRE{ individuals $\mathcal{P}$, test cases $\mathcal{T}$, number of selections $\texttt{ns}$ }
        \ENSURE{ selected individuals $\mathcal{Q}$ }

        \STATE $\mathcal{Q} \leftarrow \emptyset$
        
        \STATE $\mathbf{\epsilon} \leftarrow \lambda(\textbf{e}_t) \text{ for } t \in \mathcal{T}$

        \FOR{$\texttt{ns} \text{ times}$}
            \STATE $\mathcal{Q} \leftarrow \mathcal{Q} \cup \text{get\_parent\_semi\_dynamic}(\mathcal{P}, \mathcal{T}, \epsilon)$
        \ENDFOR

        \STATE \textbf{return} $\mathcal{Q}$
    \end{algorithmic}
\end{algorithm}

\begin{algorithm}[htb]
    \caption{Semi dynamic parent selection (get\_parent\_semi\_dynamic)}\label{alg:selection:semi_dynamic_lexicase_get_parent}
    \begin{algorithmic}[1]
        \REQUIRE{ individuals $\mathcal{P}$, test cases $\mathcal{T}$, thresholds $\epsilon$ }
        \ENSURE{ individual from $\mathcal{P}$ }

        \STATE $\mathcal{T}' \leftarrow \mathcal{T}$
        \STATE $\mathcal{S} \leftarrow \mathcal{P}$
        \WHILE{$|\mathcal{T}'|>0 \text{ and } |\mathcal{S}|>1$}
            \STATE $t \leftarrow \text{ random choice from } \mathcal{T}'$
            \STATE $\texttt{elite} \leftarrow \text{min } e_t(p) \text{ for } p \in \mathcal{S}$
    
            \FOR{$p \in \mathcal{S}$}
                \IF{$e_t(p) > \texttt{elite} + \epsilon_t$}
                    \STATE $\mathcal{S} \leftarrow \mathcal{S} \setminus \{p\}$
                \ENDIF
            \ENDFOR
            \STATE $\mathcal{T}' \leftarrow \mathcal{T}' \setminus \{t\} $
        \ENDWHILE
        
        \STATE \textbf{return} \text{ random choice from } $\mathcal{S}$
    \end{algorithmic}
\end{algorithm}

\subsection{Dynamic \texorpdfstring{$\epsilon$}{epsilon}-lexicase}~\label{subsec:selection:dynamic-lexicase}

In dynamic $\epsilon$-lexicase, both the threshold ($\epsilon_t$) and the $\texttt{elite}$ for retaining individuals in the selection pool are continuously adjusted based on the errors of the individuals currently in the pool. For each test case, individuals are retained if their errors are less than or equal to the sum of the smallest error among the individuals in the pool and a dynamically calculated $\epsilon$ value.
The implementation of this strategy is presented in Algorithms \ref{alg:selection:dynamic_lexicase_selection} and \ref{alg:selection:dynamic_lexicase_get_parent}.

\begin{algorithm}[htb]
    \caption{Dynamic $\epsilon$-lexicase selection}\label{alg:selection:dynamic_lexicase_selection}
    \begin{algorithmic}[1]
        \REQUIRE{ individuals $\mathcal{P}$, test cases $\mathcal{T}$, number of selections $\texttt{ns}$ }
        \ENSURE{ selected individuals $\mathcal{Q}$ }

        \STATE $\mathcal{Q} \leftarrow \emptyset$
        
        \FOR{$\texttt{ns} \text{ times}$}
            \STATE $\mathcal{Q} \leftarrow \mathcal{Q} \cup \text{get\_parent\_dynamic}(\mathcal{P}, \mathcal{T})$
        \ENDFOR

        \STATE \textbf{return} $\mathcal{Q}$
    \end{algorithmic}
\end{algorithm}

\begin{algorithm}[htb]
    \caption{Dynamic parent selection (get\_parent\_dynamic)}\label{alg:selection:dynamic_lexicase_get_parent}
    \begin{algorithmic}[1]
        \REQUIRE{ individuals $\mathcal{P}$, test cases $\mathcal{T}$}
        \ENSURE{ individual from $\mathcal{P}$ }

        \STATE $\mathcal{T}' \leftarrow \mathcal{T}$
        \STATE $\mathcal{S} \leftarrow \mathcal{P}$
        \WHILE{$|\mathcal{T}'|>0 \text{ and } |\mathcal{S}|>1$}
            \STATE $t \leftarrow \text{ random choice from } \mathcal{T}'$
            \STATE $\texttt{elite} \leftarrow \text{min } e_t(p) \text{ for } n \in \mathcal{S}$
            \STATE $\epsilon_t \leftarrow \lambda(\textbf{e}_t(\mathcal{S}))$
    
            \FOR{$p \in \mathcal{S}$}
                \IF{$e_t(p) > \texttt{elite} + \epsilon_t$}
                    \STATE $\mathcal{S} \leftarrow \mathcal{S} \setminus \{p\}$
                \ENDIF
            \ENDFOR
            \STATE $\mathcal{T}' \leftarrow \mathcal{T}' \setminus \{t\} $
        \ENDWHILE
        
        \STATE \textbf{return} \text{ random choice from } $\mathcal{S}$
    \end{algorithmic}
\end{algorithm}

Unlike the semi-dynamic approach, the $\epsilon$ value in the dynamic method is computed from the error distribution of the individuals currently in the pool rather than from the entire population. This design allows the selection criterion to be highly responsive to the performance of the current pool but requires repeated recalculations of the criterion. The method is therefore the most computationally expensive among the variants, since no opportunities for improving efficiency by precomputing information before execution exists.

\section{Methods}~\label{sec:selection:methods}

This section describes the proposed parent selection method, its integration into the FEAT algorithm, and the experimental design used to evaluate its performance. 
Finally, the experimental setup is described. Two sets of experiments were conducted: an in-depth evaluation using six benchmark datasets and a large-scale assessment with SRBench.

\subsection{Minimum Variance Threshold}~\label{sec:selection:split-threshold}

Here the Minimum Variance Threshold (MVT) is proposed to replace MAD as the selection criterion in static and dynamic $\epsilon$-lexicase algorithms. Since the proposed methods do not rely on the elite of the pool to calculate the threshold, no distinction exists between semi-dynamic and dynamic implementations; therefore, only the static (S-Split) and dynamic (D-Split) variants are considered.

With MVT, the errors are partitioned into two clusters, and the selection pool is reduced to those individuals whose errors fall below the identified threshold. This criterion is inspired by regression trees \cite{breiman1984classification}. In essence, the parent selection distribution is shifted toward specific cases, and it is hypothesized that clustering good- and poor-performing individuals can both improve the overall performance of $\epsilon$-lexicase selection and provide a different mechanism for determining the $\epsilon$ value.

At each step of MVT lexicase, the threshold $\tau^*$ for retaining individuals in the pool is defined as the value that minimizes the sum of the variances of the errors on either side of the split. To estimate $\tau^*$, the following calculation is performed:

\begin{equation}~\label{eq:selection:split_threshold}
    \tau^* = \underset{\text{min}(\mathbf{e}_t) < \tau < \text{max}(\mathbf{e}_t)}{\text{min}} \left ( \frac{\text{Var}(\mathbf{l})}{|\mathbf{l}|} + \frac{\text{Var}(\mathbf{r})}{|\mathbf{r}|} \right ),
\end{equation}

\noindent where $\mathbf{l} = [e_t : \mathbf{e}_t | e_t < \tau]$ and $\mathbf{r} = [e_t : \mathbf{e}_t | e_t \geq \tau]$ are the left-sided (good performing) and right-sided (bad performing) partitions, respectively. 

\begin{figure}[htb]
    \centering
    \caption[Process of consecutively splitting the pool of individuals into two clusters based on their error.]{Process of consecutively splitting the pool of individuals into two clusters based on their error. The process consists of randomly picking a test case, estimating $\tau^*$ by solving Eq. \ref{eq:selection:split_threshold}, and removing individuals with errors higher than the pool threshold. This is repeated until only one individual remains in the pool, or all training data was already used as test cases --- returning a random individual among the remaining.}
    \label{fig:selection:feat_splis}
    \includegraphics[width=0.725\linewidth]{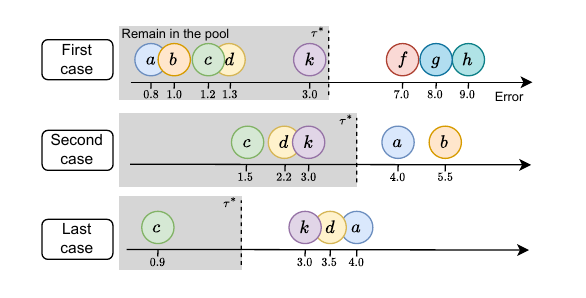}
    \legend{Source: Elaborated by the author.}
\end{figure}

Figure \ref{fig:selection:feat_splis} illustrates this process. At each split, the remaining pool consists of all individuals that consistently perform better on the test cases, \textit{i.e.}, those positioned on the left side of the split. This procedure can be seen as a more intuitive way of partitioning the pool. Since the elite error is not used in the definition of the $\tau$ threshold calculation, parent selection is relaxed, reducing the bias toward selecting only near-elite individuals.

\subsection{Feature Engineering Automation Tool}~\label{sec:selection:feat}

The Feature Engineering Automation Tool (FEAT) was proposed as an application of SR for feature engineering \cite{cava2018learning} and has been previously benchmarked in SRBench with compelling results \cite{srbench}. This choice is motivated by the fact that semi-dynamic $\epsilon$-lexicase is the default selection method in FEAT and by the model's prior study with different lexicase variants \cite{cava2018learning}. Moreover, FEAT provides the most faithful implementation of $\epsilon$-lexicase currently available, as it was implemented by the same author who originally proposed the method. 

The framework evolves a set of trees that are subsequently used as features in another machine learning (ML) model and follows the standard steps of an evolutionary algorithm, with specific adaptations beyond $\epsilon$-lexicase for parent selection. First, multi-objective survival is performed with non-dominated sorting algorith \cite{NSGA-II} to select from the Pareto front of non-dominated individuals in terms of error and complexity. Second, a backpropagation algorithm is applied to optimize the parameters of the models by linearly combining the symbolic regression trees --- a representation that relates, but does not directly translate, to the ITEA algorithm \cite{10.1162/evco_a_00285} studied in the parameter optimization chapter.

\begin{figure}[htb]
    \centering
    \caption{An individual in FEAT is a collection of symbolic regression trees as meta-features for any machine learning model.}
    \label{fig:selection:feat_individual}
    \includegraphics[width=.5\linewidth]{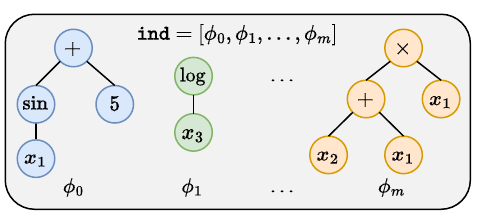}
    \legend{Source: Elaborated by the author.}
\end{figure}

Each individual consists of a combination of $\phi_0, \phi_1, \ldots, \phi_m$ expression trees. These trees are used as input features for any ML model, with linear regression and $l_2$ regularization \cite{ridge} used as default. Figure \ref{fig:selection:feat_individual} illustrates an example individual.

The fitness of each individual is calculated by training a ML model using the collection of symbolic trees as input features and then evaluating the mean squared error between the model predictions, $\mathbf{\widehat{y}} = \text{ML}([\phi_0, \phi_1, \ldots, \phi_m])$, and the observed values, $\mathbf{y}$.

The complexity of a single tree is defined recursively for a node $n$ with $k$ arguments as the combination of the complexity of its children with the complexity of the node itself (the same presented in the Background Chapter, Eq. \ref{eq:background:complexity}), where the complexity for each operator is depicted in Table \ref{tab:selection:complexity}, with values defined by the method's authors.
In FEAT, the final complexity of an individual is calculated as the sum of the complexities of each feature $\phi$ it contains, and the model size is defined as the total number of nodes across all features.

\begin{table}[htb]
    \centering
    \caption{Complexity of each operator in FEAT.}
    \label{tab:selection:complexity}
    \begin{tabular}{@{}cc@{}}
    \toprule\midrule
    \textbf{Complexity} & \textbf{Operators} \\ \midrule
    1          & $+$, $-$       \\
    2          & $\frac{\cdot}{\cdot}$, $*$, $(\cdot)^2$, $\sqrt{\cdot}$      \\
    3          & $\text{cos}$, $\text{sin}$, $(\cdot)^3$     \\
    4          &  $e^{(\cdot)}$, $\text{log}$         \\ \midrule\bottomrule   
    \end{tabular}
    \legend{Source: Elaborated by the author.}
\end{table}

Crossover either swaps two subtrees between individuals or exchanges two features (entire trees). Mutations are performed using \textit{point}, \textit{insert}, and \textit{delete} operations, which randomly replace, insert, or remove a node, respectively. Additionally, \textit{insert/delete dimension} mutations create or remove an entire tree representing a new feature dimension.

FEAT applies two mechanisms to split the data internally. The first, called \textit{validation split}, partitions the training data into two subsets: one for fitting the parameters and evaluating the model, and another for validation, used for logging and selecting the final individual from the last population. The second, called \textit{batch learning}, generates a random batch from the inner training partition to perform the fit and can be used to implement a downsampling strategy \cite{geiger_down-sampled_2023}.

\vspace{3.0em}
\subsection{Experimental setup}

The experiments were conducted in two stages. In the first stage, the FEAT algorithm was evaluated using the proposed parent selection schema with $30$ runs on $6$ datasets. Longer runs were performed to obtain larger sample sizes for statistical analysis and in-depth evaluation.

For the first stages, the NSGA2 survival step was toggled off, and survival was defined to replace the entire original population with the offspring, in order to isolate the effect of parent selection, since NSGA2 is an elitist algorithm. The batch size was also disabled to assess the number of test cases processed by each variation.

In the second stage, the method was evaluated using a comprehensive benchmark of symbolic regression algorithms, SRBench \cite{srbench}, which contains $122$ black-box regression problems drawn from the Penn Machine Learning Benchmarks (PMLB) \cite{romano2021pmlb}, including $62$ Friedman problems \cite{4a848dd1-54e3-3c3c-83c3-04977ded2e71}. The Friedman problems consist of synthetic datasets specifically designed to be challenging for regression models, whereas the remaining datasets contain real-world data. For SRBench, $10$ runs were performed per dataset in accordance with benchmark specifications. All hyper-parameters were maintained as specified by the benchmark to allow direct comparison with the latest published results. Runtime comparisons were not performed with previously published SRBench results, as the experiments were conducted on different hardware.

\begin{figure}[htb]
    \centering
    \caption{Dimensionality of the original SRBench datasets for the black-box track.}
    \label{fig:selection:pmlb_size}
    \includegraphics[width=.55\linewidth]{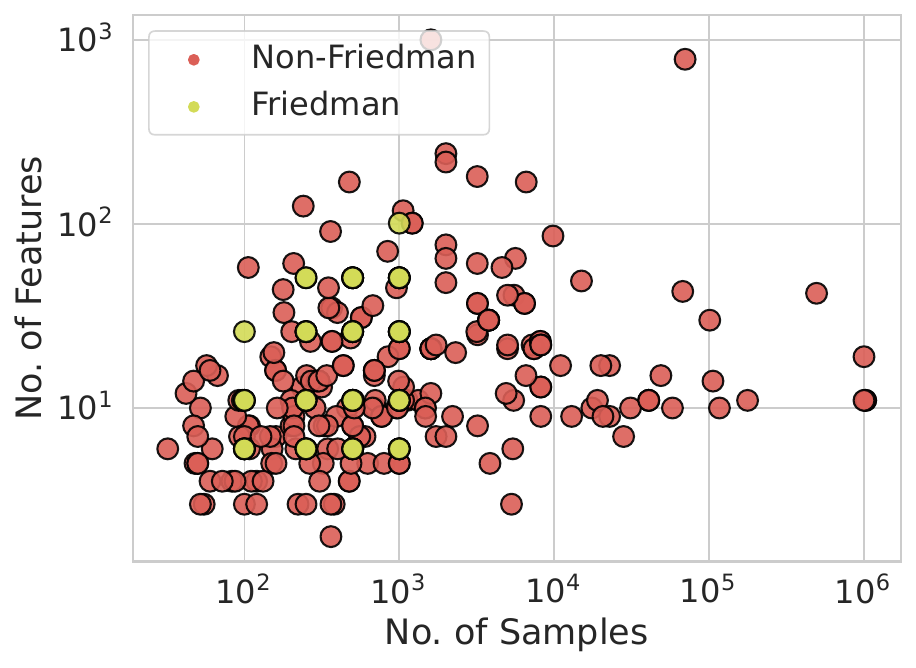}
    \legend{Source: Elaborated by the author.}
\end{figure}

For all experiments, whether from the first or second stage, the data were split into training and test partitions using a $0.75/0.25$ ratio. Table \ref{tab:selection:datasets} and Figure \ref{fig:selection:pmlb_size} report the dimensionality of each dataset used in the experiments.

\begin{table}[htb]
    \centering
    \caption{Dimensionality of the six datasets used in the parent selection experiments.}
    \label{tab:selection:datasets}
    \begin{tabular}{lll}
    \toprule\midrule
    \textbf{Dataset}        & \textbf{\# samples} & \textbf{\# features} \\ \midrule
    airfoil        & $1503$       & $5$           \\
    concrete       & $1030$       & $8$           \\
    energy cooling & $768$        & $8$           \\
    energy heating & $768$        & $8$           \\
    Housing        & $506$        & $13$          \\
    Tower          & $3135$       & $25$          \\ \midrule\bottomrule
    \end{tabular}
    \legend{Source: Elaborated by the author.}
\end{table}

Table \ref{tab:selection:feat-hyperparameters} presents the hyper-parameter configuration used in the SRBench experiments, with values in parentheses indicating the settings applied to the first six datasets. The batch size for the second stage was determined based on the results obtained from the first stage of experiments.

\begin{table}[htb]
    \centering
    \caption[FEAT hyper-parameters shared between all evaluated variations.]{FEAT hyper-parameters shared between all evaluated variations. Values in parenthesis indicates a configuration used just with the $6$ datasets.}
    \label{tab:selection:feat-hyperparameters}
    \begin{center}
    \begin{tabular}{ll}
        \toprule\midrule
        \textbf{Parameter} & \textbf{Value} \\
        \midrule
        objectives & ["fitness","complexity"] \\
        pop\_size & $100$ \\
        gens & $100$ ($350$) \\
        cross\_rate & $0.5$ \\
        ml & Linear ridge regression \\
        max\_depth & $6$ ($3$) \\
        backprop & True \\
        iters & $10$ \\
        validation split & $0.25$ \\
        selection & NSGA2 (offspring) \\
        batch\_size & $200$ (unlimited) \\
        functions & [$+$,$-$,$*$,$\frac{\cdot}{\cdot}$,$(\cdot)^2$,$(\cdot)^3$,$\sqrt{\cdot}$,$\text{sin}$,\text{cos},$e^{(\cdot)}$,$\text{log}$] \\
        \midrule\bottomrule
    \end{tabular}
    \end{center}
    \legend{Source: Elaborated by the author.}
\end{table}

\subsection{Online Resources}

FEAT with the MVT selection is available at \url{https://github.com/gAldeia/feat}. All data and the source code for experiments and post-processing analysis are available at \url{https://github.com/gAldeia/srbench/tree/feat_split_benchmark}.

\section{Results and  discussion}~\label{sec:selection:results-discussion}

First, the convergence characteristics of the different MVT $\epsilon$-lexicase algorithms are analyzed, with a focus on their impact on search convergence and the number of test cases utilized. Second, the proposed method is benchmarked using the SRBench framework, which includes $23$ machine learning methods (originally $21$, plus the two proposed FEAT variants), of which $16$ ($14$ plus the two variants) are symbolic regression algorithms. Finally, SRBench is used to assess how the proposed method scales with respect to the number of features and the number of samples in each dataset.

\subsection{Behavior during the run}

\begin{figure}[H]
    \centering
    \caption[Convergence loss of the best individual in the parent selection experiments for the six datasets.]{Convergence loss of the best individual in the parent selection experiments for the six datasets: (a) training partition and (b) validation partition.}
    \label{fig:selection:min_loss_group}
    \subfloat[Training partition\label{fig:selection:min_loss}]{%
        \includegraphics[width=0.95\linewidth]{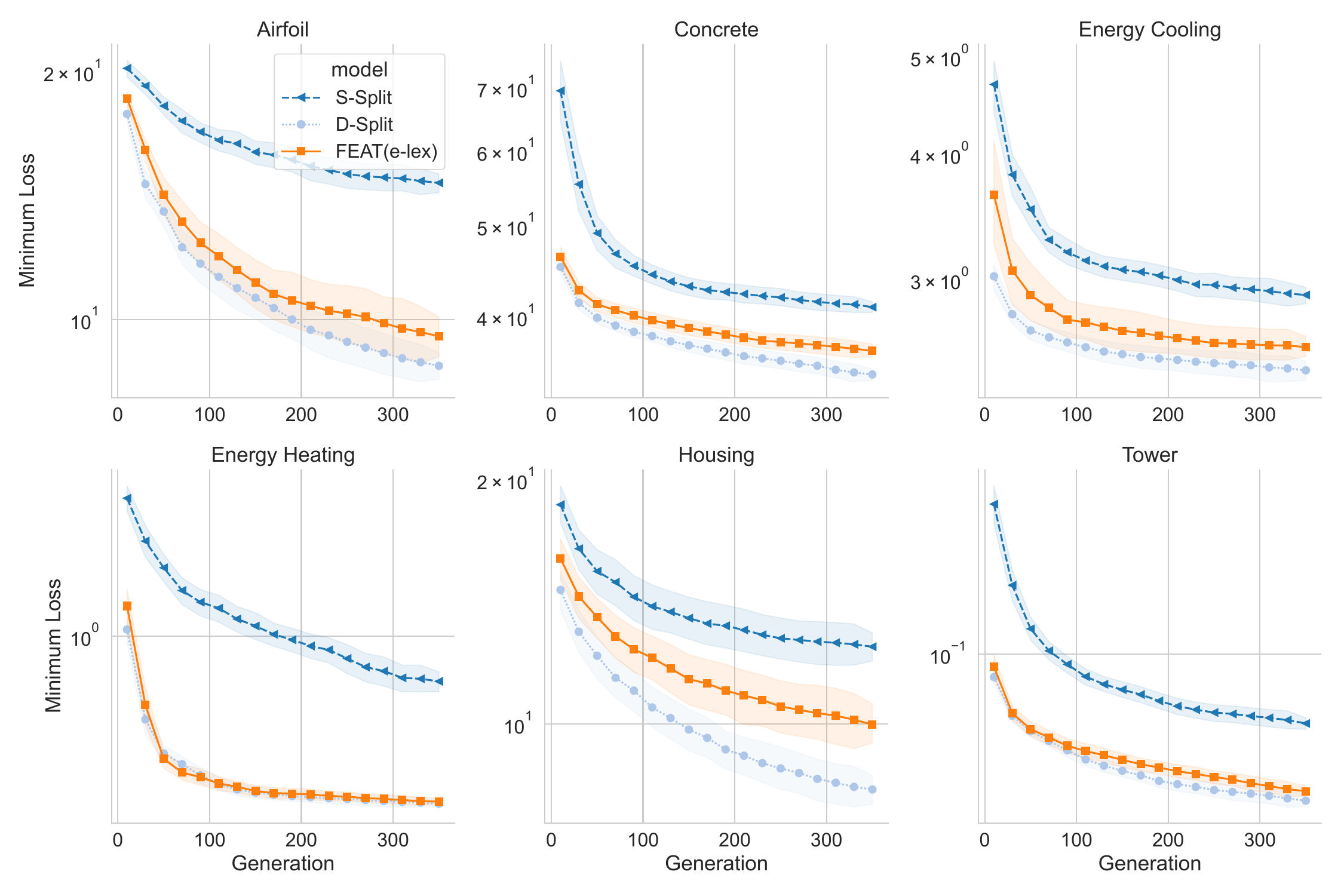}}\\[1ex]
    \subfloat[Validation partition\label{fig:selection:min_loss_val}]{%
        \includegraphics[width=0.95\linewidth]{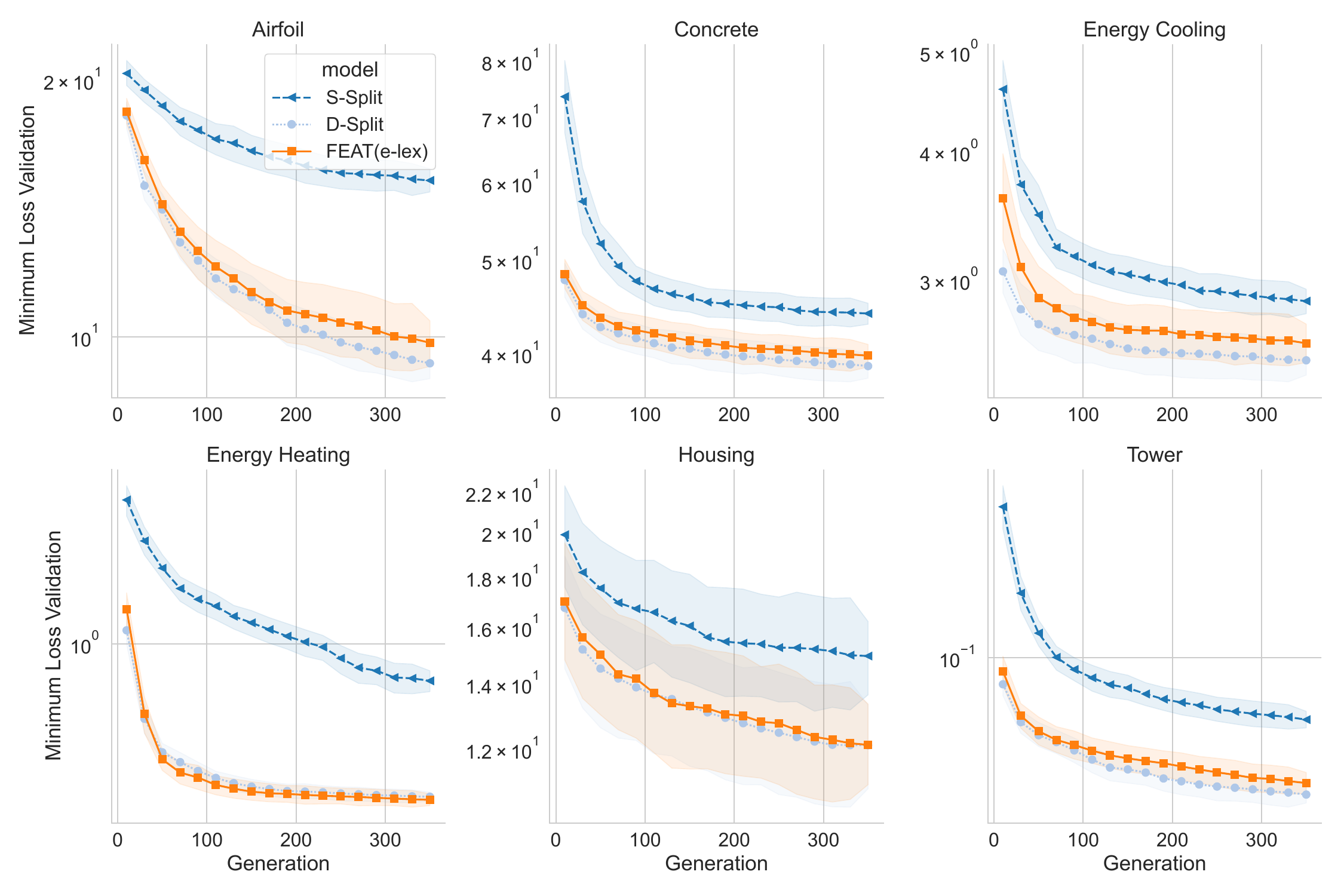}}
    \legend{Source: Elaborated by the author.}
\end{figure}

Figure \ref{fig:selection:min_loss} reports the minimum loss on the training data, while Figure \ref{fig:selection:min_loss_val} shows the minimum loss on the validation split for each dataset.

It can be observed that, although improvements in training loss are greater for D-Split compared to the original FEAT, the differences in the convergence curves are subtle when examining the validation partition. While this suggests overfitting, the generalization is maintained, evidenced by the validation loss. The effect is particularly noticeable when comparing the convergence curves of the training and validation partitions for the Housing dataset.

D-Split exhibits superior validation convergence curves on two datasets (Airfoil and Energy Cooling) compared to the other approaches, supporting the idea that the MVT can influence convergence behavior. S-Split demonstrates the poorest convergence across all datasets, even worse than the original FEAT. In certain cases, such as the Concrete, Energy Cooling, and Tower datasets, the S-split curves appear to reach a \textit{plateau} at a higher error level.

\begin{figure}[htb]
    \centering
    \caption{Median number of test cases used to pick each parent for the six datasets.}
    \label{fig:selection:max_tests_used}
    \includegraphics[width=0.95\linewidth]{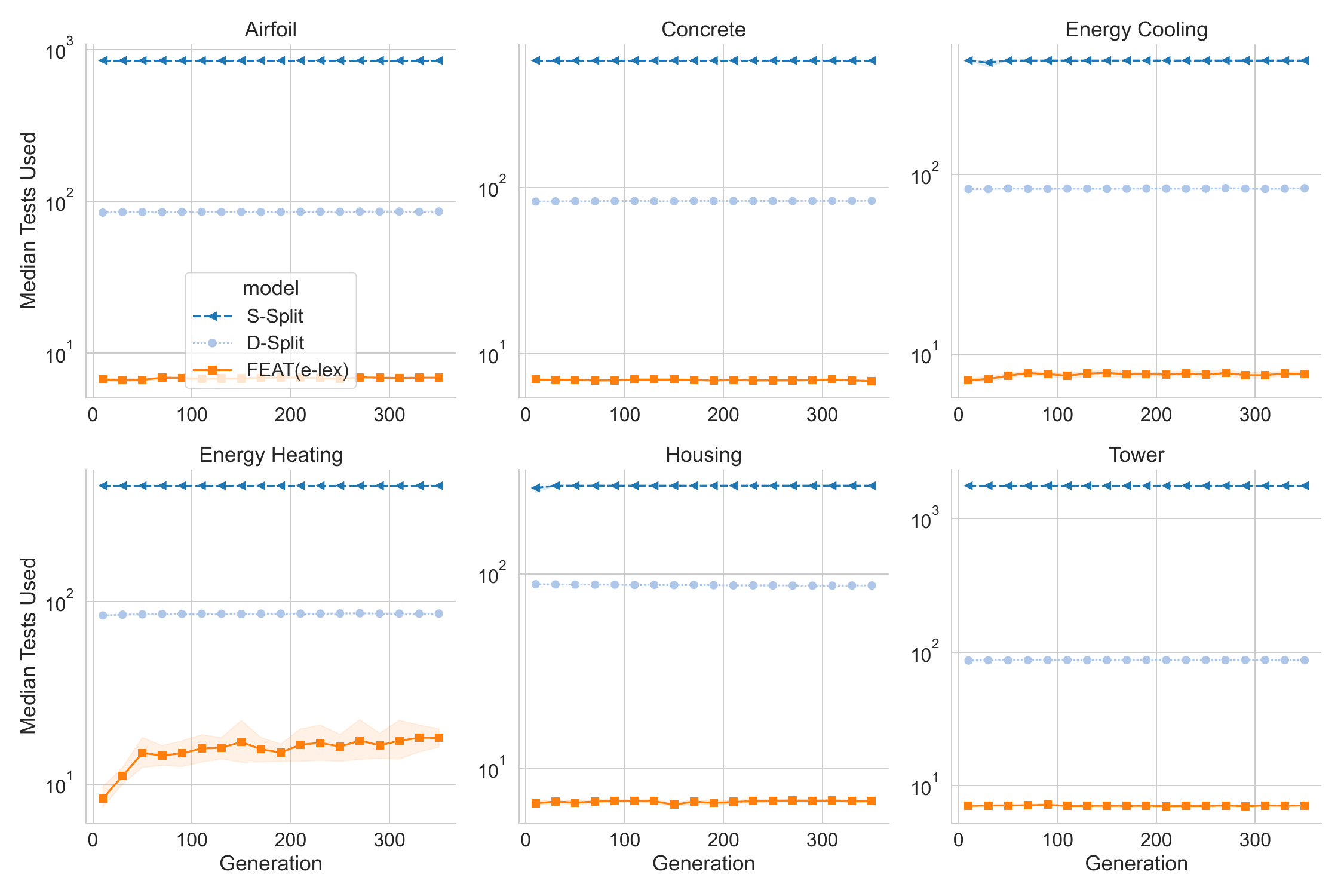}
    \legend{Source: Elaborated by the author.}
\end{figure}

Figure \ref{fig:selection:max_tests_used} reports the median number of test cases utilized to select the parent pool in each generation.
When examining the number of test cases used in each generation, it is observed that the original $\epsilon$-lexicase consistently requires fewer test cases across all datasets, implying faster execution times (as will be discussed later). Both D-Split and S-Split utilize more test cases, with S-Split requiring approximately twice as many as D-Split. This suggests that several test cases contains large clusters of well-performing individuals that would be ignored under selection using MAD, but are retained longer in the pool when MVT is applied. This extended retention can be associated to produce a more robust selection, resulting in improvements over the original FEAT with the MAD threshold.

This suggests the runtime performance of S-Split could potentially be enhanced by implementing a down-sampling strategy that selects a subset of test cases where significant differences between individuals are more likely \cite{10.1162/evco_a_00346} or used batch learning. D-Split, which calculates the threshold using the current pool, demonstrates better performance because the pool is reduced by removing individuals with subpar overall goodness-of-fit, yielding more stable results.

The number of test cases used supports the idea that the one-dimensional clustering into only two clusters produces a large cluster of well-performing individuals. This suggests that the approach may be too permissive in capturing similarities due to a high threshold.
Consequently, exploring alternative numbers of clusters may be a promising direction for future work.

\subsection{Performance on small datasets}

\begin{figure}[htb]
    \centering
    \caption{$R^2$, size, and complexity, for the final FEAT models.}
    \label{fig:selection:metrics_group}
    \subfloat[Median $R^2$ rank\label{fig:selection:barplot_r2_rank}]{%
       \includegraphics[width=0.25\linewidth]{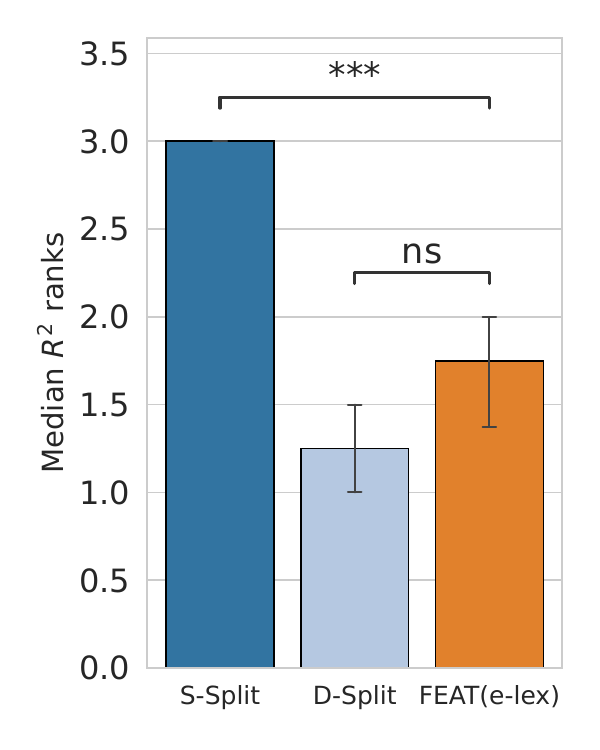}}
    \subfloat[$R^2$ score\label{fig:selection:boxplot_r2_score}]{%
       \includegraphics[width=0.25\linewidth]{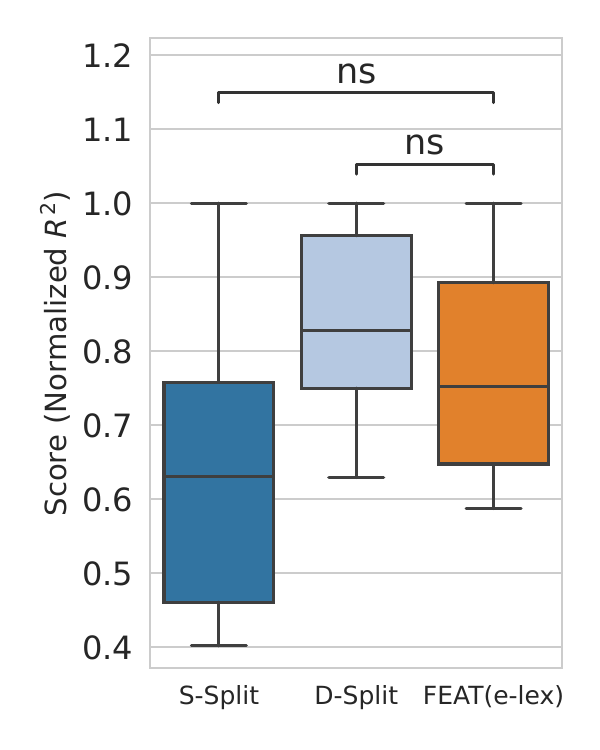}}
    \subfloat[Sizes\label{fig:selection:boxplot_size}]{%
       \includegraphics[width=0.25\linewidth]{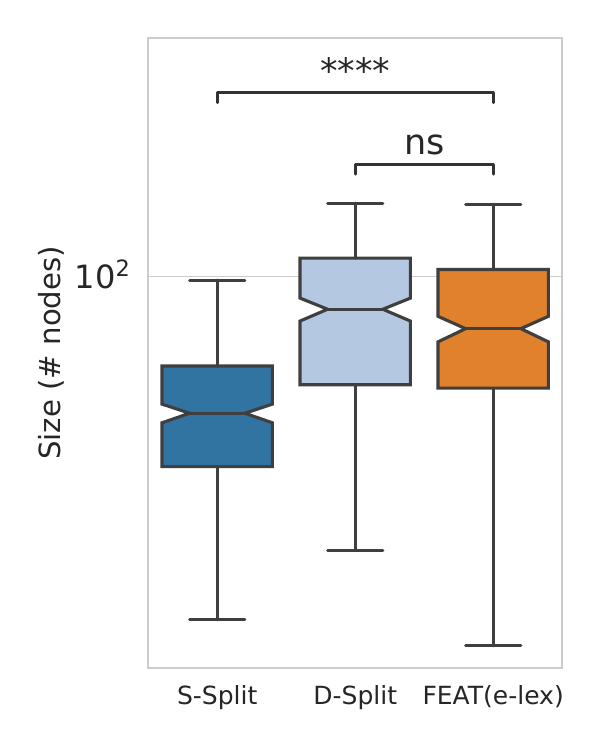}}
    \subfloat[Complexites\label{fig:selection:boxplot_complexity}]{%
       \includegraphics[width=0.25\linewidth]{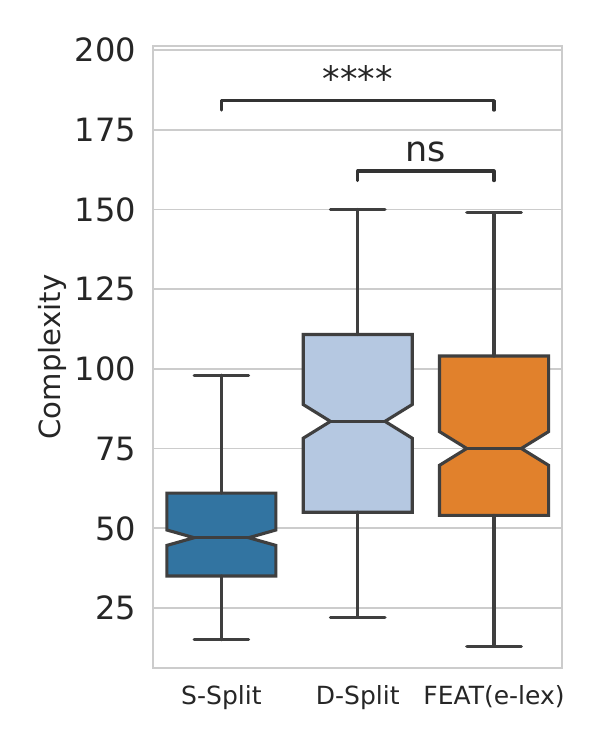}}
    \legend{Source: Elaborated by the author.}
\end{figure}

To assess the final performance on the first six datasets, the median $R^2$ ranking for each random seed used in the experiments is reported in Figure \ref{fig:selection:barplot_r2_rank}, with the raw values reported as a boxplot in Figure \ref{fig:selection:boxplot_r2_score}.
The size and complexity of the final models are shown in Figures \ref{fig:selection:boxplot_size} and \ref{fig:selection:boxplot_complexity}.

When examining the $R^2$ rankings (Figure \ref{fig:selection:barplot_r2_rank}), S-Split is observed to perform worst among the variants, with three asterisks indicating a $\text{p-value} \leq 1\times 10^{-3}$. 
When the grouped $R^2$ values across all datasets are considered (Figure \ref{fig:selection:boxplot_r2_score}), no statistically significant differences are observed; however, $75\%$ of the D-Split distribution lies above the median of the original $\epsilon$-lexicase. The lowest performance is exhibited by S-Split, with a wide spread ranging from $0.3$ to $0.8$.
The relatively poor performance of S-Split in terms of $R^2$ is accompanied by a better trade-off regarding model size and complexity (Figures \ref{fig:selection:boxplot_size} and \ref{fig:selection:boxplot_complexity}). 

\begin{figure}[htb]
    \centering
    \caption{Critical Differences diagrams of grouped metrics performances for the six datasets.}
    \label{fig:selection:cd}
    \subfloat[$R^2$ critical difference diagram\label{fig:selection:r2_cd}]{%
       \includegraphics[width=0.61\linewidth]{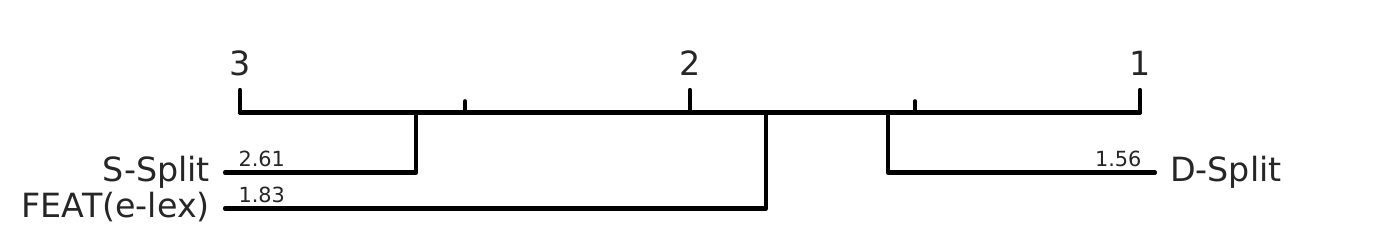}}\\ 
    \subfloat[Complexity critical difference diagram\label{fig:selection:complexty_cd}]{%
       \includegraphics[width=0.61\linewidth]{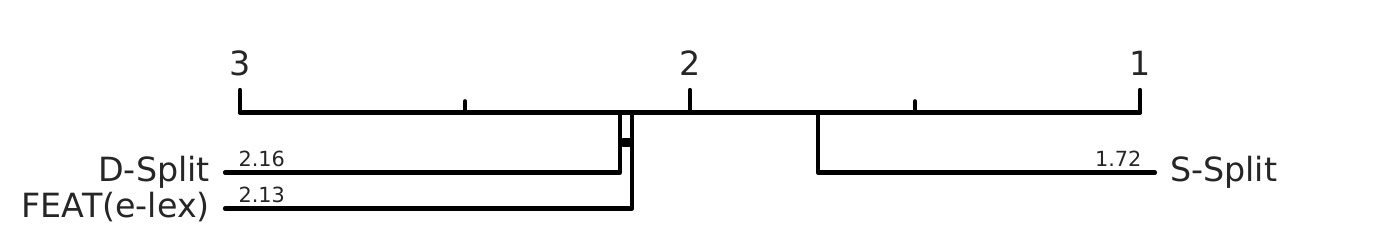}}
    \legend{Source: Elaborated by the author.}
\end{figure}

A critical differences (CD) diagram \cite{demvsar2006statistical} is also reported in Figure \ref{fig:selection:cd}, illustrating the median ranking of each algorithm, with horizontal lines connecting algorithms for which no statistically significant differences were observed.

The CD diagrams for $R^2$ performances (Figure \ref{fig:selection:r2_cd}) shows that D-Split is predominantly ranked first among the other variants. 
Regarding complexity, the CD diagram in Figure \ref{fig:selection:complexty_cd} indicates that S-Split achieves the best results in size and complexity with statistical significance, although no statistically significant differences are observed between the other two variants, D-Split and $\epsilon$-lexicase.

\subsection{Benchmarking with SRBench}

\begin{figure}[htb]
    \centering
    \caption{Pareto fronts of benchmarked parent selection variants, including previous results from the original SRBench.}
    \label{fig:selection:pareto_plot_srbench}
    \includegraphics[width=0.7\linewidth]{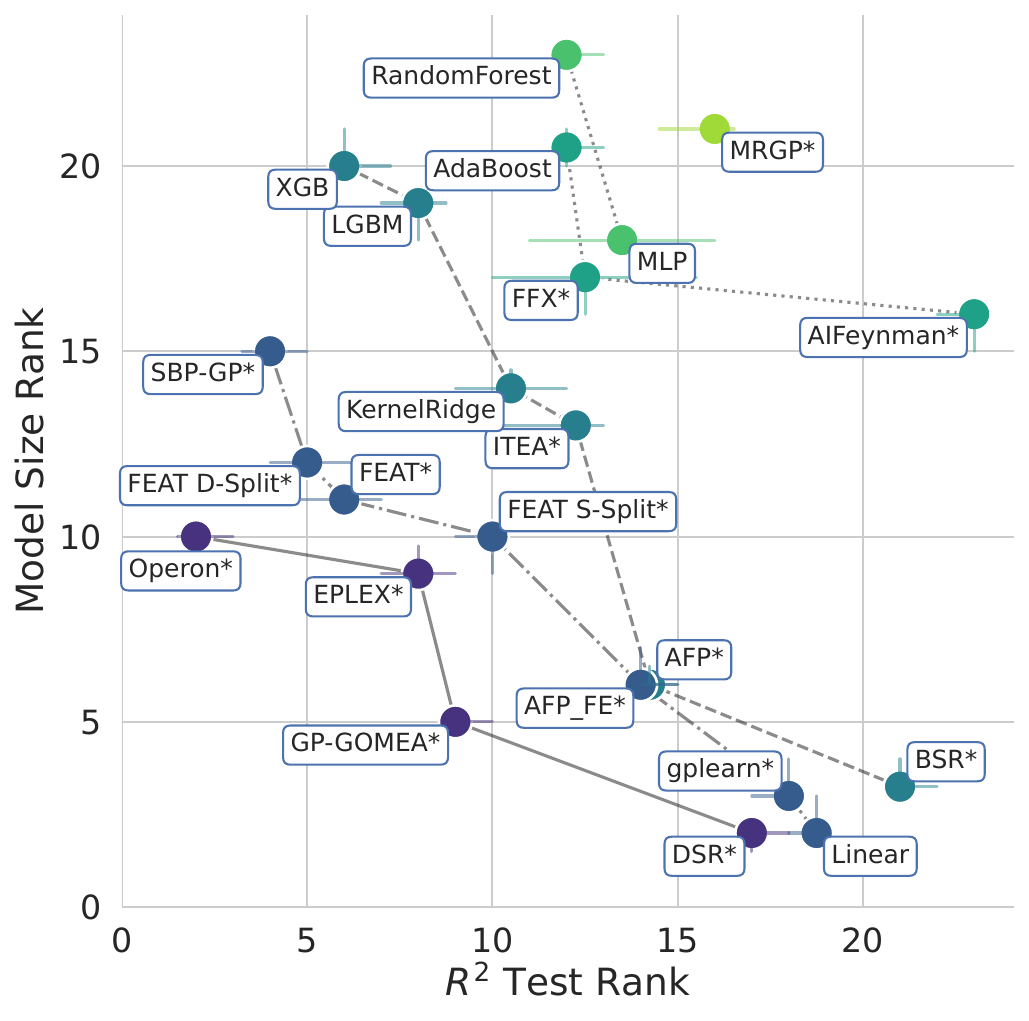}
    \legend{Source: Elaborated by the author.}
\end{figure}

The SRBench framework includes a variety of machine learning models. 
From the most recent analysis of state-of-the-art symbolic regression methods, it is observed that FEAT was among the top-performing algorithms, originally ranked third in terms of the root $R^2$ on the test partition, behind Operon \cite{operon} and SBP-GP \cite{10.1145/3321707.3321758}. 
The rank for model size and $R^2$ across all algorithms in SRBench is plotted in Figure \ref{fig:selection:pareto_plot_srbench}, using results from the original publication, plus experiments ran with FEAT using S- and D-split. For both axes, smaller ranks indicate better overall performance. The lines in the figure represent the Pareto fronts corresponding to each rank.

It can be observed that the two variations and FEAT lie on the same Pareto front, exhibiting different trade-offs between model size and accuracy. 
FEAT D-Split is observed to outperform the original FEAT by a modest margin, showing approximately similar $R^2$ and model size ranks. FEAT S-Split performs worse than the other two FEAT variants but still achieves performance to stay in the second Pareto front, surpassing more than half of the evaluated methods in terms of rank dominances.

\begin{figure}[htb]
    \centering
    \caption[Number of times each parent selection variant was ranked among the best for the Friedman datasets.]{Number of times each parent selection variant was ranked among the best for the Friedman datasets: (a) top 1, and (b) top 5 or less.}
    \label{fig:selection:histfri_group}
    \subfloat[Top 1 ranks\label{fig:selection:hist1fri}]{%
        \includegraphics[trim={0 0.7cm 0 0},clip=true,width=0.49\linewidth]{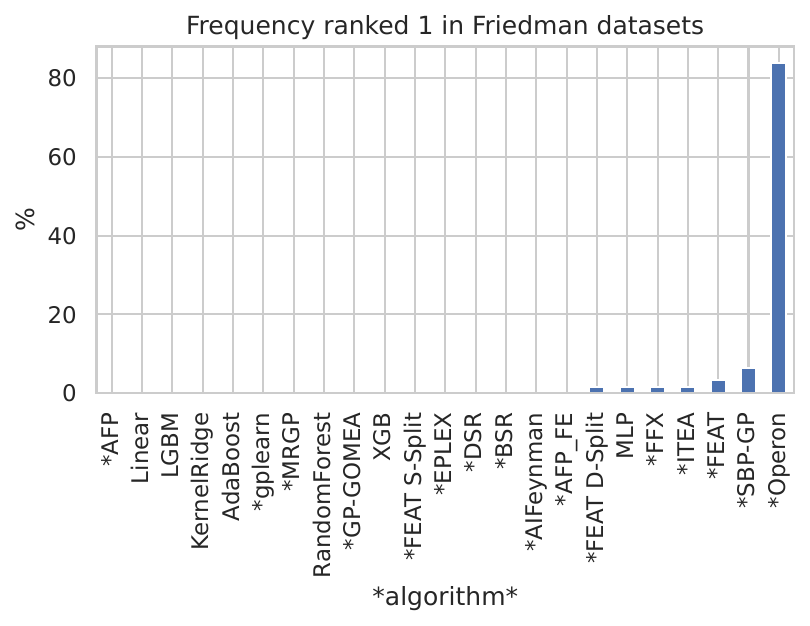}}
    \hfill
    \subfloat[Top 5 or better ranks\label{fig:selection:hist5fri}]{%
        \includegraphics[trim={0 0.7cm 0 0},clip=true,width=0.49\linewidth]{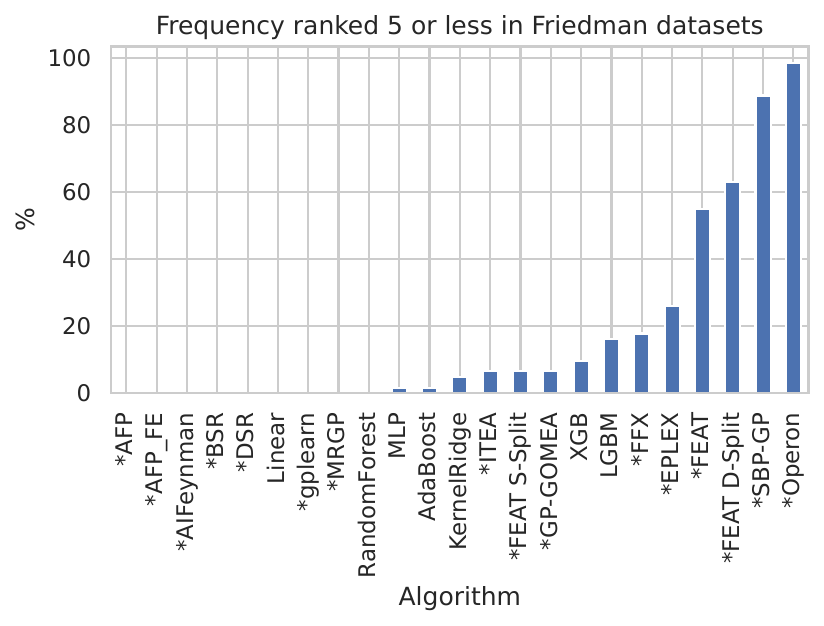}}
    \legend{Source: Elaborated by the author.}
\end{figure}

\begin{figure}[htb]
    \centering
    \caption[Number of times each parent selection variant was ranked among the best for the non-Friedman datasets.]{Number of times each parent selection variant was ranked among the best for the non-Friedman datasets: (a) top 1, and (b) top 5 or less.}
    \label{fig:selection:histnonfri_group}
    \subfloat[Top 1 ranks\label{fig:selection:hist1nonfri}]{%
        \includegraphics[trim={0 0.7cm 0 0},clip=true,width=0.49\linewidth]{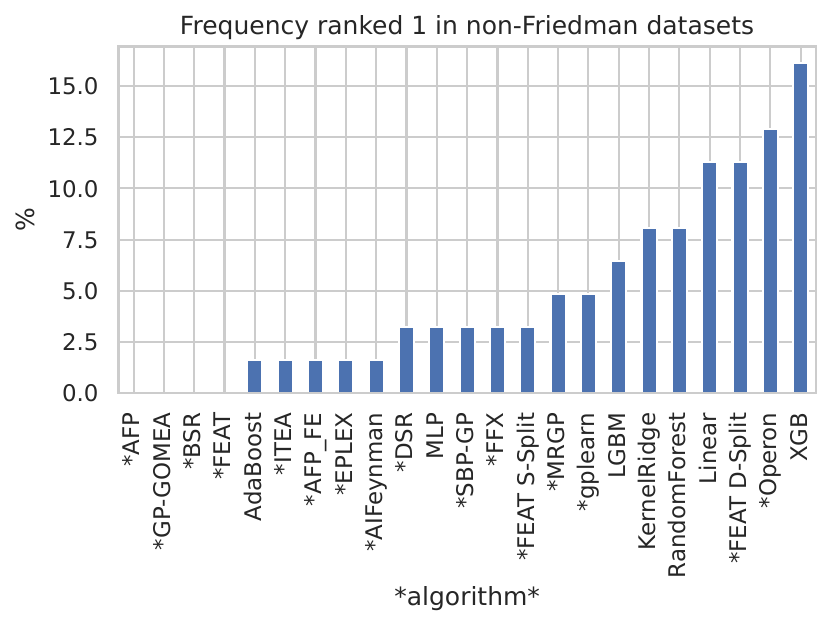}}
    \hfill
    \subfloat[Top 5 or better ranks\label{fig:selection:hist5nonfri}]{%
        \includegraphics[trim={0 0.7cm 0 0},clip=true,width=0.49\linewidth]{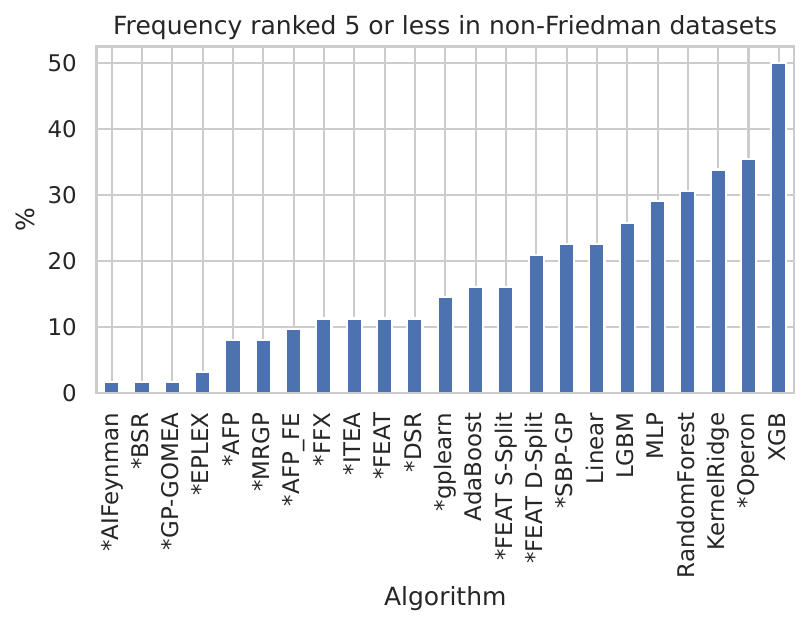}}
    \legend{Source: Elaborated by the author.}
\end{figure}

As noted by \citeonline{tir}, aggregated SRBench results can obscure differences among the top-performing algorithms, as there is also an overrepresentation of Friedman problems. When the benchmarks are divided into Friedman and non-Friedman datasets (roughly half each), it is observed that the best-ranked algorithms exhibit noticeable differences on the Friedman benchmarks while behaving similarly on the non-Friedman datasets. Following this approach, Figures \ref{fig:selection:hist1fri} and \ref{fig:selection:hist5fri} report the percentage of times each algorithm was ranked first and within the top one and top five for the Friedman synthetic problems, whereas Figures \ref{fig:selection:hist1nonfri} and \ref{fig:selection:hist5nonfri} report the corresponding percentages for the real-world, non-Friedman problems.

For all Friedman datasets, it is observed that all algorithms are dominated by Operon when only the top-$1$ percentage of success is considered (Figure \ref{fig:selection:hist1fri}). Given that approximately half of the black-box problems are Friedman datasets, this suggests that the overall results may be biased in favor of Operon, which is particularly efficient at finding solutions for challenging synthetic datasets.
Only $14$ of the $23$ methods are observed to be within the top-$5$ for the Friedman datasets. While FEAT achieves a rank of $5$ or lower in approximately $53\%$ of the runs, the proposed FEAT D-Split increases this percentage to over $60\%$. FEAT with S-Split selection is ranked among the top five algorithms in less than $10\%$ of the runs.

D-Split outperforms the other variants, achieving a total of $10\%$ of wins relative to the other algorithms. This represents a notable improvement, as the original FEAT performs relatively poorly on these problems. It is also observed that S-Split performs better on the non-Friedman datasets compared to the original FEAT. Given that S-Split tends to produce models with smaller size and complexity, this indicates that solutions with a better trade-off between $R^2$ and complexity were identified.

Overall, these histograms demonstrate that FEAT D-Split improved its rank on the Friedman benchmarks, although not enough to reach the top-$1$ positions frequently. Both D-Split and S-Split substantially increase the proportion of solutions at the top-$1$, with D-Split ranking among the top three algorithms for the non-Friedman benchmarks. 


\subsection{Scalability}

\begin{figure}[htb]
    \centering
    \caption[Training time for the parent selection experiments as a function of dataset characteristics.]{Training time for the parent selection experiments as a function of dataset characteristics: (a) number of features, and (b) number of samples in the training partition.}
    \label{fig:selection:training_time_group}
    \subfloat[Training time vs. number of features\label{fig:selection:training_time_dataset_nfeatures}]{%
        \includegraphics[width=0.475\linewidth]{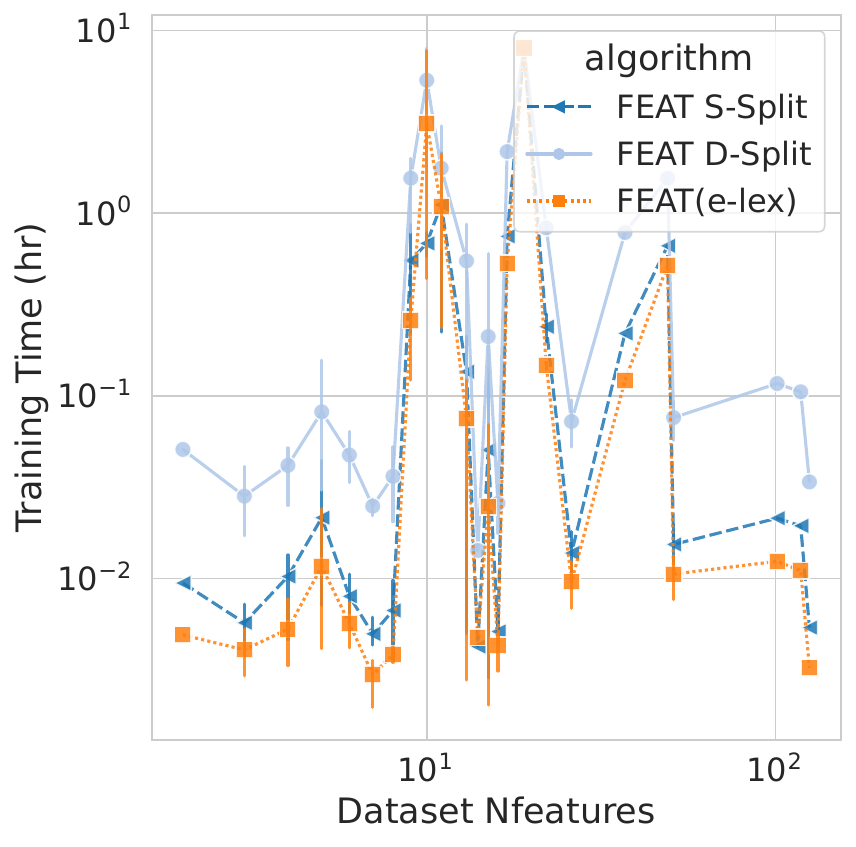}}
    \hfill
    \subfloat[Training time vs. number of samples\label{fig:selection:training_time_dataset_nsamples}]{%
        \includegraphics[width=0.475\linewidth]{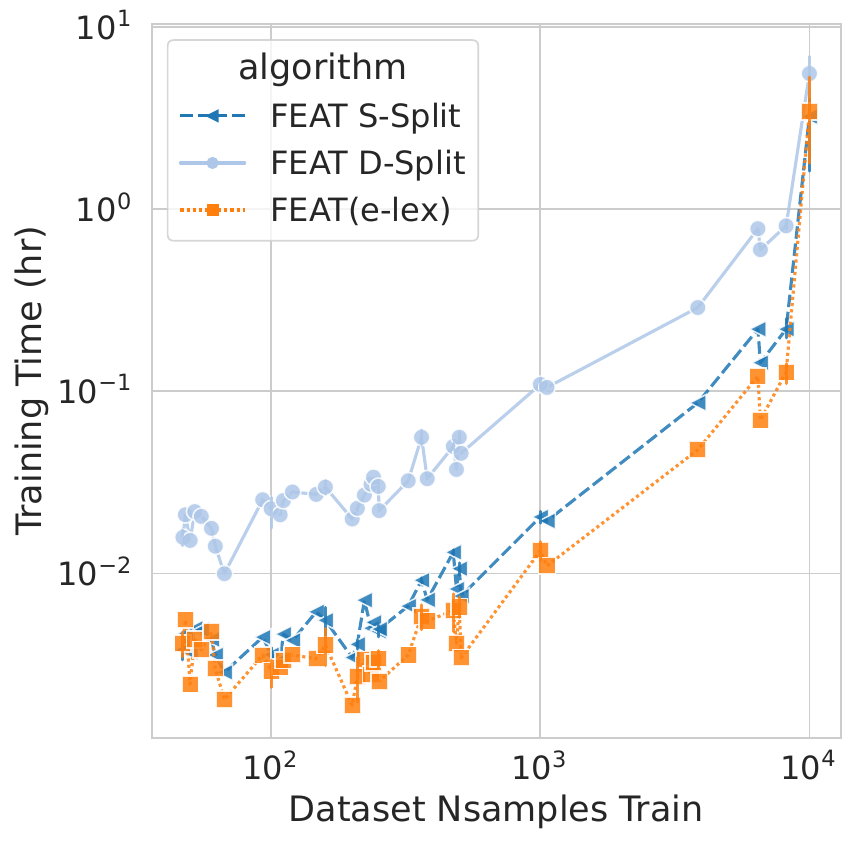}}
    \legend{Source: Elaborated by the author.}
\end{figure}

To assess scalability, execution time was plotted against the number of samples and features for the SRBench datasets. The FEAT algorithm was re-run on all datasets to eliminate hardware-related differences. Figure \ref{fig:selection:training_time_dataset_nfeatures} shows the training time as a function of the number of features, while Figure \ref{fig:selection:training_time_dataset_nsamples} shows it as a function of the number of samples.
Execution times between other models were not compared, as they are heavily hardware-dependent and could provide a biased estimate of performance, as the hardware used for the MVT experiments may differ.

All three FEAT variants with different parent selection strategies exhibit similar scaling trends, albeit with different offsets. The results suggest that scalability is more strongly influenced by the number of samples rather than the number of features. This is likely because an increase in samples raises the number of test cases, and the MVT selection method uses a higher number of test cases than the original $\epsilon$-lexicase.

Regarding the number of features, the size of individuals is constrained by the maximum allowable size of symbolic regression trees. Like many symbolic regression algorithms, FEAT implicitly performs feature selection due to these size constraints. Consequently, the number of features does not substantially impact training time, although the maximum expression sizes do. Nevertheless, competitive performance was achieved on SRBench with relatively small models, indicating that increasing the maximum expression size is unnecessary and that FEAT performs well under the current configuration.

An interesting observation is that S-Split uses more test cases than D-Split but exhibits faster execution. This can be explained by the fact that D-Split recalculates the threshold for each test case until a parent is selected, whereas S-Split requires only a single threshold calculation per selection.

\section{Conclusions}~\label{sec:selection:conclusions}

In this chapter, a novel threshold estimation method for $\epsilon$-lexicase selection was proposed, aimed at improving lexicase selection for regression problems, by using the Minimum Variance Threshold (MVT) approach. The threshold criterion splits the candidate pool in two clusters based on individual test case performances, providing a more intuitive interpretation of how parent selection with $\epsilon$-lexicase works.
This chapter also presents an in-depth review of parent selection methods, specially on lexicase selection, explaining the main differences between the variants and highlighting previous efforts on investigating its benefits.
As a by-product, the FEAT algorithm was extended to support new variants of $\epsilon$-lexicase selection, and the implementation was made available in \CPP{}.

Two experiments were conducted to evaluate the new threshold criterion. First, on a small batch of six small datasets and with the survival toggled off, it was shown that the new criterion can improve convergence and yield higher median values.
In the second set of experiments, $122$ regression datasets were used, comparing the proposed static and dynamic variations with results of $21$ other algorithms, where the MVT with dynamic estimation showed to improve original FEAT performance on synthetic problems.

The proposed method achieved competitive results relative to the original $\epsilon$-lexicase while demonstrating improvements on real-world datasets. Performance on synthetic datasets was preserved. A limitation of the approach is the increased number of test cases required, resulting in higher execution times. Nevertheless, the use of additional computational resources was shown to yield improved performance, as demonstrated by the final comparisons.




%% file: Capitulos/Simplification/simplification.tex
\Chapter{Model Simplification}{Fast Inexact Simplification with Locality-sensitive Hashing}~\label{chap:inexactsimp}
\chaptermark{Model Simplification}

\begin{laysummary}
    Symbolic regression (SR) aims to discover parametric models that accurately fit a dataset while prioritizing interpretability. Despite this secondary objective, the models are often unnecessarily complex due to redundant operations, introns, and bloat that accumulate during the iterative search process. Such inefficient expressions can hinder exploration by repeatedly revisiting redundant areas of the hypothesis space.
    Traditional heuristic simplifications are insufficient to fully reduce expressions, while exact methods often become computationally infeasible as expression size and complexity increase. To address this, this chapter introduces a novel, domain-agnostic simplification technique for SR that leverages efficient memoization with locality-sensitive hashing (LSH).
    The key idea is to store expressions and their sub-expressions encountered during iterative simplification in a dictionary indexed with LSH. Because LSH naturally introduces uncertainty, expressions with nearly identical semantics (\textit{i.e.}, similar predictions) can be mapped to the same hash key, relaxing the need for exact --—and often expensive—-- matching. This facilitates fast retrieval while allowing approximate behaviors to be treated as equivalent. During simplification, subtrees are iteratively replaced with one equivalent sharing the same hash whenever the replacement yields a smaller expression.
    Empirical results show that incorporating this simplification during evolution achieves equal or superior error minimization compared to unsimplified runs, while significantly reducing the use of nonlinear functions. This approach effectively learns simplification rules that generalize across problems or adapt to specific tasks, thereby improving convergence and reducing model complexity.
\end{laysummary}

{\let\thefootnote\relax\footnotetext{Part of this chapter was published as: Guilherme Seidyo Imai Aldeia, Fabrício Olivetti de França, and William G. La Cava. 2024. Inexact Simplification of Symbolic Regression Expressions with Locality-sensitive Hashing. In Genetic and Evolutionary Computation Conference (GECCO '24), July 14-18, 2024, Melbourne, VIC, Australia. ACM, New York, NY, USA, 9 pages. \url{https://doi.org/10.1145/3638529.3654147}}}

\vfill
\pagebreak

\section{Introduction}

Symbolic regression (SR) jointly optimize the parameters and structure of a function and is commonly implemented using Genetic Programming (GP).
This problem has been proven to be NP-hard \cite{SymbolicRegressionIsNPHard}, and the search space is vast, containing equivalent mathematical expressions \cite{burlacu2019hash}, introns (parts of the model that do not influence predictions) \cite{GainingDeeperInsightsInSymbolicRegression}, bloat (growth in model size without improvement in loss) \cite{luke_comparison_2006}, and model overparameterization (excessive number of parameters to tune) \cite{10.1145/3583131.3590346}.
Although these elements do not directly affect model accuracy, they inflate model size, thereby reducing simplicity and interpretability.
In addition, it has been shown that models better generalize when simplification is performed \cite{helmuth_improving_2017}, likely due to the presence of unnecessary code that did not expressed effects during training, but fails to handle new test instances.

\begin{figure}[htb]
    \centering
    \caption[Illustration of several equivalent expressions without any simplification.]{Illustration of several equivalent expressions without any simplification. The left-most expression is the simplest, and all others can be reduced to this form. Although this reduction is intuitive for humans, the semantics of each node are not known by the algorithm, and algebraic rules for simplification can be complex to implement.}
    \label{fig:simplification:simp_examples}
    \includegraphics[width=\linewidth]{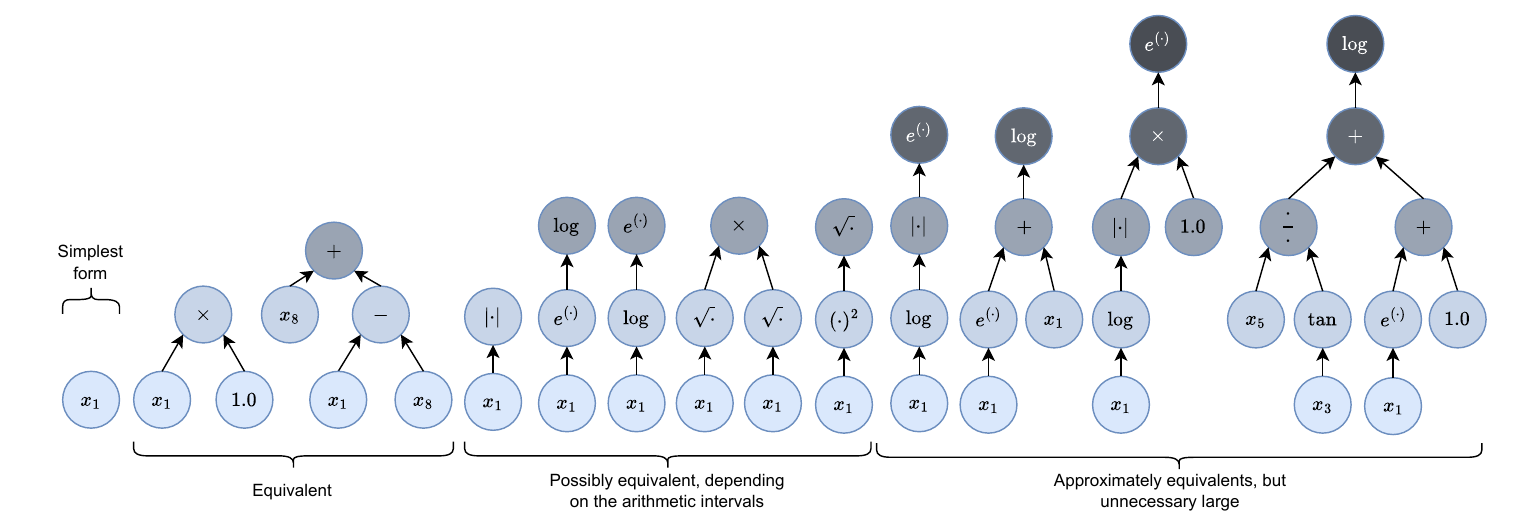}
    \legend{Source: Elaborated by the author.}
\end{figure}

As an illustrative example, the complexity of the search space in SR is demonstrated in Figure \ref{fig:simplification:simp_examples}. Different mathematical expressions, despite not all being equivalent ---that can be reduced to the same expression—-- contribute to this complexity. Introns, bloat, and overparameterization are all included within this complexity, reducing the simplicity and interpretability of the resulting models.

Mechanisms to alleviate these issues are therefore necessary.
Several approaches have been proposed in the literature, including heuristics for automatic simplification of models \cite{helmuth_improving_2017}, parametric rules for rewriting expressions \cite{kulunchakov_creation_2017}, Bayesian loss metrics for model selection \cite{bomarito_bayesian_2022}, constrained search spaces through structural restrictions \cite{10.1162/evco_a_00285}, and exact simplification procedures such as equality saturation \cite{10.1145/3583131.3590346}.
Each approach offers benefits, but limitations also exists, chiefly the additional computational cost and the need for manually defined algebraic simplification rules.

To address these limitations, a technique for dynamically building simplification rules was proposed, in which observed equivalences are memoized through hashing.
In contrast to previous implementation-dependent approaches that require extensive manual coding of rules, this idea introduces a data-driven strategy by leveraging similarity and memoization to facilitate simplification and bloat control.
A hash table is constructed in which the key is generated using Locality-Sensitive Hashing (LSH) \cite{gionis_similarity_nodate} based on the prediction vector, and the value is defined as the smallest observed subtree corresponding to that vector.
LSH is widely employed for efficient retrieval of nearest-neighbor points in database systems \cite{jafari_survey_2021}.
By analogy, expressions are transformed into smaller equivalents within the phenotype (\textit{i.e.}, prediction) space.

Some researchers argue not all redundancy is harmful. Neutrality ---changes in genotype that leave phenotype and fitness unchanged--- may promote broad exploration of the search space \cite{neutrality}, suggesting that simplification can harm the neutral space. This suggests that simplification should be also evaluated in terms of its effects on convergence to assess whether it interferes the neutral space. 
When the proposed simplification is integrated and applied to every expression throughout the generations, benefits such as faster convergence and reduced mean squared error are observed.
The size of the returned expressions remains unchanged, but their complexities are significantly lower due to less use of non-linear functions deep chains of nodes.
Additionally, expressions identified as equivalent reveal that many known algebraic identities are successfully captured by the algorithm.

The remainder of this chapter is organized as follows.
Section \ref{sec:simplification:relatedwork} provides an overview of related work on bloat control, hashing, and simplification methods in SR.
Section \ref{sec:simplification:lsh} introduces the application of Locality-Sensitive Hashing (LSH) in the proposed simplification method.
Section \ref{sec:simplification:methods} outlines the experimental methods, detailing how expressions are simplified through memoization with LSH and describing the SR framework used for testing.
Section \ref{sec:simplification:results} presents and discusses the empirical findings, focusing on the impact of the approach on solution quality, expression size, and elimination of unnecessary operations.
Finally, Section \ref{sec:simplification:conclusions} provides concluding remarks.

\section{Related work}~\label{sec:simplification:relatedwork}

Techniques for simplification to simpler models have been proposed in previous works through different approaches.
Among heuristics for automatic simplification, \citeonline{helmuth_improving_2017} proposed an heuristic in which subtrees were randomly removed and the change was accepted if the final prediction was not affected.
Indirect simplification through mutation was explored by \citeonline{nguyen_semantic_2020}, in which a percentage of the population was pruned by replacing subtrees with newly generated small random trees of similar semantics.
To produce simpler models, in \citeonline{la_cava_flexible_2023} proposed to perform delete mutations several times, keeping the change if it results in a model where the error metric does not drop more than a tolerance threshold.

Other works have focused on sets of rules for simplification.
\citeonline{kulunchakov_creation_2017} defined equivalent algebraic models as those producing the same outputs over a limited range and evolved expressions with GP to be used as replacements.
\citeonline{burlacu2019hash} proposed the hashing of SR trees using basic algebraic rules to handle commutative operations, ensuring that equivalent subtrees with regard to commutative rearrangements were hashed to the same value.
In \citeonline{10.1145/3583131.3590346}, equality saturation ---an enumerative simplification algorithm--- was applied to remove redundant parameters from symbolic expressions.

In contrast to these approaches, sets of rules and expensive computational resources are not relied upon in the proposed method.
A data-driven and computationally efficient form of inexact simplification is introduced, which can be embedded into any SR framework.
The method is agnostic, as it requires only the slicing of expressions and the replacement of subtrees.

\section{Locality-Sensitive Hashing}~\label{sec:simplification:lsh}

\begin{figure}[htb]
    \centering
    \caption[How Locality-Sensitive Hashing maps different predictions to the same hash.]{How Locality-Sensitive Hashing maps different predictions to the same hash. A set of vectors (\textit{i.e.}, predictions) is given, and similar predictions are mapped to the same hash. In this example, the two highlighted vectors present close values and are mapped into the same hash.}
    \label{fig:simplification:lsh_mapping}
    \includegraphics[width=.6\linewidth]{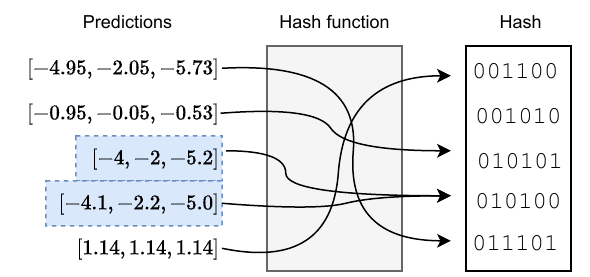}
    \legend{Source: Elaborated by the author.}
\end{figure}

Locality-Sensitive Hashing (LSH) \cite{gionis_similarity_nodate} was first introduced as a technique for efficiently finding approximate neighbors in high-dimensional spaces by employing hash functions that preserve local proximity \cite{jafari_survey_2021}.
The idea is that input vectors are hashed in such a way that similar vectors are more likely to be mapped to the same bucket, while dissimilar vectors are less likely to be. This process is illustrated in Figure \ref{fig:simplification:lsh_mapping}.

In this chapter, the SimHash LSH method \cite{charikar2002similarity} is employed, owing to its straightforward implementation for incorporation into existing frameworks.
A trade-off between computational efficiency and accuracy is thereby produced, which is advantageous in scenarios where exact similarity search is rendered impractical \cite{jafari_survey_2021} due to the curse of dimensionality \cite{10.1145/502807.502808}. 

Particularly related to the dimensionality, \citeonline{10.1145/502807.502808} says that as the volume of the space increases they tend to become more sparse and dissimilar, hindering data organization, because the dimensionality defines a metric space and distances are expensive to compute between pairs, highlighting the importance of minimizing distance calculations.

Given a set of $d$ $n$-dimensional samples $\{\mathbf{x}_j \in \mathbb{R}^n\}_{j=1}^d$, the hash size is first fixed at an arbitrary number of $b$ bits.
A plane $\mathbf{P} \in \mathbb{R}^{b \times d}$ is generated, where each entry $\mathbf{P}_{(i,j)} \sim \mathcal{N}(0, 1)$.
For any queried data point $\mathbf{x}_j$, the matrix multiplication $\mathbf{P} \cdot \mathbf{x}_j$ is computed, producing the vector $\mathbf{q} \in \mathbb{R}^{b \times 1}$.
The hash $h(\mathbf{x}_j)$ is then defined as:

\begin{equation}~\label{eq:simplification:hash}
    h(\mathbf{x}_j)_i = \begin{cases} 1 & q_i > 0 \\ 0 & \text{otherwise} \end{cases}.
\end{equation}

This bit string serves as the key in the hash table.
The main property of this function is that the probability of two hashes being equal, \textit{i.e.}, two vectors being mapped to the same hash, is proportional to the cosine similarity, which in turn is proportional to the normalized $l_2$-Euclidean distance:

\begin{equation}
    \text{P}[h(\mathbf{x}) = h(\mathbf{y})] = 1 - \frac{\theta(\mathbf{x},\mathbf{y})}{\pi}.
\end{equation}

The resulting bit string is used as a dictionary key, with the corresponding values stored as lists ordered by size, containing all entries sharing the same hash key.
Every entry in the hash table is used to store all queried objects sharing the same hash value, making similar objects clustered around the same keys.
As the number of bits used to generate the hash key increases, the number of clusters also increases while becoming less dense, thereby improving the accuracy of similarity estimation.

In addition to queries, indexing of vectors is also supported by the LSH table, in which pairs of the form $(\mathtt{hash} : \mathtt{vector})$ are stored.
Whenever a query is made, two values are returned: the hash of the input vector and the Euclidean distance to the closest previously indexed vector.
This property is useful for deciding simplification not only based on the hash but also on distance.

As a result, the prediction of an expression ---or part of it--- can be employed to retrieve the closest previously stored expressions with similar predictions, without requiring pairwise comparison of vectors.

The difference between using LSH and approaches such as token embeddings is that LSH is data-driven and it is designed to intentionally create collisions in a controlled, probabilistic manner. 
Moreover, token embeddings may fail to capture similarities between syntactically different expressions that share similar semantics.

\section{Methods}~\label{sec:simplification:methods}

This section details how the LSH is applied to memoize equivalent expressions and replace larger sub-trees by their simplest equivalents. Then, it presents the experimental setup used to evaluate the proposed method.

\subsection{Simplifying expressions by memoization}~\label{sec:simplification:memoization}

The concept of inexact simplification is based on the use of a simplification table in which the key is defined as the hash given by Eq.~\ref{eq:simplification:hash}, and the values correspond to a list of expressions mapped to that particular key.

\begin{figure}[htb]
    \centering
    \caption[Illustration of the inexact simplification method.]{Illustration of the inexact simplification method. In the first stage ($1$), a simplification table is created with only the problem variables and the constant. Every node is treated as the root of a subtree and generates a prediction vector. In the second stage ($2$), the predictions are used to obtain hash values for each node, updating the simplification table. Finally, the simplified tree is obtained at the final stage ($3$) by replacing the nodes with the smallest subtree corresponding to the same hash in the simplification table.}
    \label{fig:simplification:lsh_simplification}
    \includegraphics[width=\linewidth]{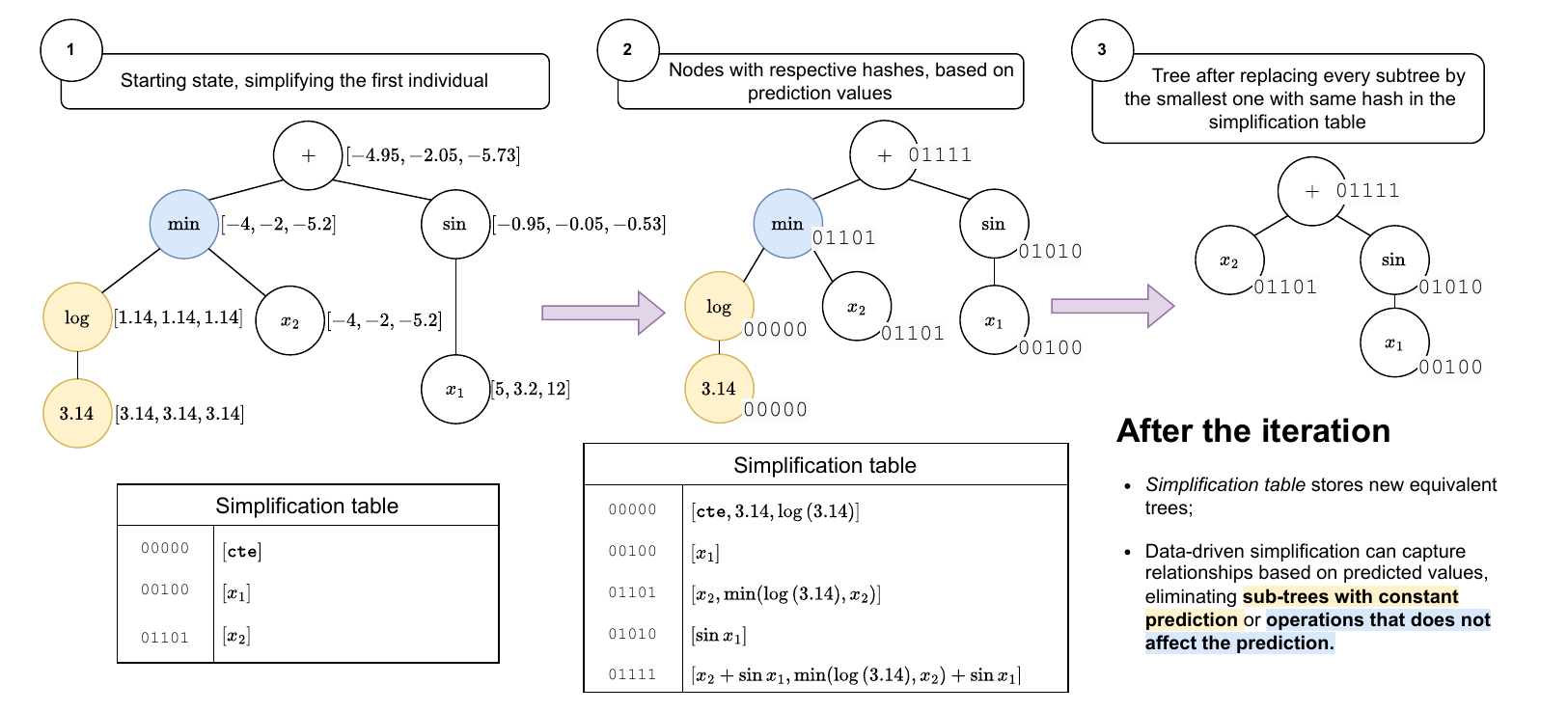}
    \legend{Source: Elaborated by the author.}
\end{figure}

In the first step, the expression is evaluated while maintaining a trace of the evaluation at every node. At this stage, each node contains a vector of predictions corresponding to the evaluation of its subtree. In the second step, the tree is traversed again\footnote{In practice, only a single traversal is required.}, and these vectors are hashed into binary string keys. Finally, if the key already exists in the hash table and the closest value to the subtree lies within a specified threshold, the subtree is replaced by the smallest tree in that table entry. If no entry exists for the key, a new entry is created with this subtree. The list of equivalent expressions is ordered by size, guaranteeing that the recovery of the smallest equivalent can be done instantly, and the comparison made is just if the equivalent expression is indeed smaller. If the smallest equivalent does not decrease the size of the tree, then the subtree itself is indexed as the new first element. The proposed algorithm for using LSH to simplify symbolic expressions is illustrated in Figure \ref{fig:simplification:lsh_simplification}.

Given a hash size of $\mathtt{hs}$ bits and a training data partition $(\mathcal{X}, \mathbf{y})$ consisting of $d$ samples, a plane $\mathcal{P}$ of dimensions $\mathtt{hs} \times d$ is first created, with values sampled from a standard normal distribution.
For any candidate solution $\hat{f}$, the prediction $\hat{\mathbf{y}} = \hat{f}(\mathcal{X})$ of size $d$ is then used as the vector for hash generation.
A query is performed by calculating the dot product between the plane $\mathcal{P} \in \mathbb{R}^{\mathtt{hs} \times d}$ and the prediction $\hat{\mathbf{y}} \in \mathbb{R}^{d \times 1}$, producing $\mathbf{q} \in \mathbb{R}^{\mathtt{hs} \times 1}$.
The corresponding hash is then generated as $[1 \ \text{if} \ \mathbf{q}_i > 0 \ \text{otherwise} \ 0 \ | \ i=1,2,\ldots, \mathtt{hs}]$.

First, the simplification table $\mathtt{st}$ is initialized by iterating over every variable and indexing the values of the terminals in the training partition, initializing the table with the simplest possible elements.
When a constant is evaluated, they are forced to have a prediction vector filled with zeroes (implying in a null hash), preventing unnecessary growth of the hash table by inserting constants as simplification candidates to other hashes.
This also mitigates the fact that non-linear equations may have non-unique solutions.
The prediction vector is indexed into the $\mathtt{lsh}$ table, and its hash is used to initialize a list of equivalent expressions.
The initialization process is described in Algorithm \ref{alg:simplification:init_table}.

\begin{algorithm}[htb]
    \caption{Initialize hash table (initialize\_table)}\label{alg:simplification:init_table}
    \begin{algorithmic}[1]
        \REQUIRE{ Training data points $(\mathcal{X}, \mathbf{y})$, single node constant tree $\texttt{cte}$ }
        \ENSURE{ simplification table $\texttt{st}$ }

        \STATE $\texttt{st} \leftarrow \text{ empty mapping of (key : value)}$

        \FOR{$\texttt{terminal} \in \{\texttt{cte}\} \cup \{\mathbf{x}_i \in \mathcal{X}\}$}
            \STATE $\texttt{pred} \leftarrow \texttt{terminal}.\text{values}$
            \STATE $\texttt{lsh}.\text{index}(\texttt{pred})$
            \STATE $\texttt{hash}, d \leftarrow \texttt{lsh}.\text{query}(\texttt{pred})$
            \STATE $\texttt{st}[\texttt{hash}] = [\texttt{terminal}]$
        \ENDFOR
        
        \STATE \textbf{return} $\texttt{st}$
    \end{algorithmic}
\end{algorithm}

Each time an individual is simplified, the steps described in Algorithm~\ref{alg:simplification:hash_simplify} are followed.
Every subtree of the individual is iterated over, and its prediction vector is assessed.
The prediction vector is then used as a query in the existing LSH table, which returns the equivalent hash and the distance to the vector of that hash, if any exists.
If the query result is found in the simplification table, the distance is checked against a tolerance theshold.
When the tolerance is satisfied, the subtree is inserted into the corresponding collection in the simplification table and replaced by the smallest element from that collection.
Otherwise, the prediction vector is indexed, and a new entry is created in the simplification table.

\begin{algorithm}[htb]
    \caption{Calculate hash for simplification (hash\_simplify)}\label{alg:simplification:hash_simplify}
    \begin{algorithmic}[1]
        \REQUIRE{ individual $\texttt{ind}$, simplification table $\texttt{st}$, lsh instance $\texttt{lsh}$, tolerance $\tau$ }
        \ENSURE{ simplified version of the individual $n$ }

        \FOR{$\texttt{subtree} \in \texttt{ind}$}
            \STATE $\texttt{pred} \leftarrow \texttt{subtree}.\text{predict}(\mathcal{X})$
            
            \IF{$\text{Var}(\texttt{pred}) = 0$}
                \STATE $\texttt{pred} \leftarrow \texttt{pred} \times 0.0$
            \ENDIF
            
            \STATE $\texttt{hash}, d \leftarrow \texttt{lsh}.\text{query}(\texttt{pred})$
    
            \IF{$\texttt{hash} \in \texttt{st} \text{ and } d \leq \tau$}
                \STATE $\texttt{st}[\texttt{hash}] \leftarrow \texttt{st}[\texttt{hash}] \cup \{\texttt{subtree}\}$
                \STATE $\texttt{subtree} \leftarrow \text{argmin} \text{ size}(\texttt{tree}) \text{ for } \texttt{tree} \in \texttt{st}[\texttt{hash}]$
            \ELSIF{$\texttt{hash} \not\in \texttt{st}$}
                \STATE $\texttt{lsh}.\text{index}(\texttt{pred})$
                \STATE $\texttt{st}[\texttt{hash}] = [\texttt{subtree}]$
            \ENDIF
        \ENDFOR
        
        \STATE \textbf{return} $\texttt{ind}$
    \end{algorithmic}
\end{algorithm}

\begin{figure}[htb]
    \centering
    \caption[Illustration of top-down and bottom-up traversal approaches for expression simplification.]{Illustration of top-down and bottom-up traversal approaches for expression simplification. The top-down approach prunes large subtrees first, potentially stopping earlier if a hash entry is found at the upper levels. The bottom-up approach may involve more steps, since starting at lower levels can trigger further simplifications higher up the tree, but may allow the method to find more equivalent expressions.}
    \label{fig:simplification:traversal_approaches}
    \includegraphics[width=0.5\linewidth]{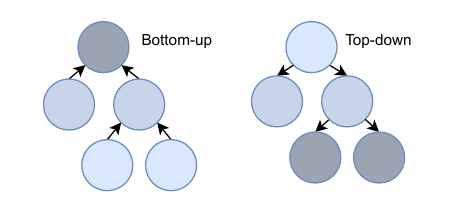}
    \legend{Source: Elaborated by the author.}
\end{figure}

When an expression is simplified, the tree can be traversed either top-down or bottom-up, as replacements are applied on the fly (Figure~\ref{fig:simplification:traversal_approaches}).
Different simplifications can be generated depending on the traversal order.

When traversing a tree for the purpose of simplification, the expectation is that branches, once visited, may reveal smaller equivalents suitable for simplification.
The top-down approach implies iterating through fewer nodes than a bottom-up strategy, as large subtrees are pruned first, anticipating encountering previously explored, potentially complex, also reducing the number of visited nodes, since traversal can stop earlier.
In contrast, the bottom-up traversal may require more steps, as a simplification at a lower level can trigger a new simplification higher in the tree, but may ultimately produce more effective simplifications by slowly ascending the tree, as equivalences are more likely to exist between smaller than large subtrees.

Notice that the simplification does not always produce an algebraically equivalent expression, but instead provides an approximation.
By imposing a maximum distance threshold, this issue is mitigated, as subtrees are guaranteed to be replaced only with those that yield similar semantics.
This behavior is advantageous for practical applications, since simpler expressions are favored at the expense of minor differences in predictions.

\subsection{Experimental setup}

To evaluate the simplification method, an evolutionary SR algorithm was implemented using the DEAP framework \cite{DEAP_JMLR2012}.
In this algorithm, the individuals are randomly initialized using the PTC2 method \cite{PTC2}.
Evolution runs for a fixed number of generations, using tournament selection with a tournament size of $3$.
The variation operators are \textit{crossover}, and the mutations \textit{insert node}, \textit{remove node}, \textit{replace node}, and \textit{replace subtree}.
Three variants were implemented: \textit{without simplify}, \textit{bottom-up}, and \textit{top-down}.

Simplification was applied to the initial population to learn an initial set of equivalences, and to every offspring generated by the variation operators, ensuring that each individual was simplified at least once.
The individuals were also processed using the nonlinear parameter optimization method Levenberg-Marquardt \cite{levenberg1944method, marquardt1963algorithm} implemented in SciPy \cite{2020SciPy-NMeth}. 
After simplification, parameter optimization was repeated.

To determine the hash size, it was initially set to $2^5$ and the strenght was increased until no collisions occurred during the initialization of the simplification table for all datasets ---\textit{i.e.}, the constant and input variables have no colisions---, resulting in the final value of $2^8$.
During initial experiments, it was observed that if the initial table contained a hash collision between features and constants, simplification could incorrectly replace features, which is undesirable unless a feature is nearly constant.

The threshold was set to a fixed value based on the smallest training MSE observed in preliminary experiments across all datasets, $0.01$, one order of magnitude below all training errors.
Population size and the number of generations were chosen so that most runs would complete within approximately $3$ wall-clock hours.
The hyperparameters used in the experiments are summarized in Table \ref{tab:simplification:sr-hyperparameters}, prioritizing the number of generations over population size to investigate the effects on convergence.

\begin{table}[htb]
    \centering
    \caption{Symbolic regression algorithm hyper-parameters used for the simplification experiments.}
    \label{tab:simplification:sr-hyperparameters}
    \begin{tabular}{ll}
    \toprule\midrule
    \textbf{Parameter}     & \textbf{Value}                                      \\ \midrule
    pop size ($S$)     & $80$                                        \\
    max gen  ($G$)     & $200$                                       \\
    max depth ($\text{max}_d$)    & $7$                                         \\
    max size  ($\text{max}_s$)    & $128$ ($2^7)$                                        \\
    tolerance ($\tau$) & $1e-2$ \\
    hash\_len  & $256$ bits                                        \\
    probabilities & $1/5$ for each variation operator \\
    objectives    & [error (MSE), size (\# nodes))] \\ 
    Function set  & [$+$, $-$, $*$, $\frac{\cdot}{\cdot}$, $|\cdot|$, $\cos^{-1}$, $\sin^{-1}$, $\tan^{-1}$, $\cos$, $\sin$, $\tan$, \\& $e^{(\cdot)}$, $\mathtt{min}$, $\mathtt{max}$, $\log$, $\log{(1+ \cdot)}$, $\exp{(1+\cdot)}$, $\sqrt{|\cdot|}$, $(\cdot)^2$] \\ \midrule\bottomrule
    \end{tabular}
    \legend{Source: Elaborated by the author.}
\end{table}

A selection of datasets was used to analyze the effect of simplification throughout the evolutionary process.
The names and dimensionalities of the datasets are listed in Table \ref{tab:simplification:datasets}.

\begin{table}[htb]
    \centering
    \caption{Dimensionality of the six datasets used in the simplification experiments.}
    \label{tab:simplification:datasets}
    \begin{tabular}{lll}
    \toprule\midrule
    \textbf{Dataset}        & \textbf{\# samples} & \textbf{\# features} \\ \midrule
    Airfoil        & $1503$       & $5$           \\
    Concrete       & $1030$       & $8$           \\
    Energy Cooling & $768$        & $8$           \\
    Energy Heating & $768$        & $8$           \\
    Housing        & $506$        & $13$          \\
    Yacht          & $308$        & $6$           \\ \midrule\bottomrule
    \end{tabular}
    \legend{Source: Elaborated by the author.}
\end{table}

As a measure of complexity, the same as used by \citeonline{cava2018learning} included as a measure.
For a node $n$ with $k$ arguments, complexity is defined as the recursive combination of the complexities of its children with that of its root/head node (the same presented in the Background Chapter, Eq. \ref{eq:background:complexity}).
The complexity values assigned to the operators are presented in Table \ref{tab:simplification:complexity}.

\begin{table}[htb]
    \centering
    \caption{Complexity of each operator used for the simplification experiments.}
    \label{tab:simplification:complexity}
    \begin{tabular}{@{}ll@{}}
    \toprule\midrule
    \textbf{Complexity} & \textbf{Operators} \\ \midrule
    
    2          & $+$, $-$, $\mathtt{cte}$       \\
    
    3          & $*$, $\mathtt{max}$, $\mathtt{min}$, $(\cdot)^2$, $|\cdot|$      \\
    
    4          & $\frac{\cdot}{\cdot}$, $\sqrt{|\cdot|}$,  $e^{(\cdot)}$     \\ 
    5 &  $\exp{(1+\cdot)}$, $\cos$, $\sin$, $\tan$ \\
    6 &  $\cos^{-1}$, $\sin^{-1}$, $\tan^{-1}$ \\
    8 & $\log{(1+ \cdot)}$ \\
    9 & $\log$ \\\midrule\bottomrule   
    \end{tabular}
    \legend{Source: Elaborated by the author.}
\end{table}

Each method was executed $30$ times using different split seeds for each dataset on the same hardware.
The data were divided into three partitions: $50\%$ for training, which was visible to the algorithm for parameter and function optimization; $25\%$ for validation, which was used to assess the loss during evolution but not for training; and $25\%$ for testing, which was held out to obtain the final experimental results.
The validation partition was also used to select the final model returned by the algorithm. This is inspired by how FEAT \cite{cava2018learning} handles the data internally for optimizing and selecting the final individual.

Statistical comparisons were performed using the non-parametric Wilcoxon test with Holm-Bonferroni correction, and all comparisons are explicitly depicted in the figures.
An annotation shown as ``* (ns)'' indicates a statistical significance corresponding to a single asterisk, which becomes non-significant after the correction.
The p-value annotations used to indicate statistical significance in the plots, after applying the alpha correction, are as follows. An annotation of `ns' corresponds to $5\times 10^{-2} < \text{p} \leq 1.0$, indicating non-significance. A single asterisk `*' represents $1\times 10^{-2} < \text{p} \leq 5\times 10^{-2}$, `**' corresponds to $1\times 10^{-3} < \text{p} \leq 1\times 10^{-2}$, and `***' represents $1\times 10^{-4} < \text{p} \leq 1\times 10^{-3}$. Finally, four asterisks`****' indicate $\text{p} \leq 1\times 10^{-4}$.

\subsection{Online resources}

All data and the source code for implementations, experiments, and post-processing analysis are available at \url{https://github.com/gAldeia/hashing-symbolic-expressions}.

\section{Results and discussion}~\label{sec:simplification:results}

In this section, several aspects of the experimental results are analyzed.
First, the impact of simplification on GP convergence and the number of simplifications per traversal strategy are examined.
Distributions of final solution size, complexity, and goodness-of-fit are presented, along with paired comparisons of runs with and without simplification.
The number of equivalent expressions, runtime differences, and selected simplification rules observed during the runs are also analyzed.

\subsection{Convergence}

\begin{figure}[htb]
    \centering
    \caption[Convergence curves of the validation error of the best individual during the evolution for the simplification experiments.]{Convergence curves of the validation error of the best individual during the evolution for the simplification experiments. $y$ axis is in log-scale.}
    \label{fig:simplification:convergence_error}
    \includegraphics[width=\linewidth]{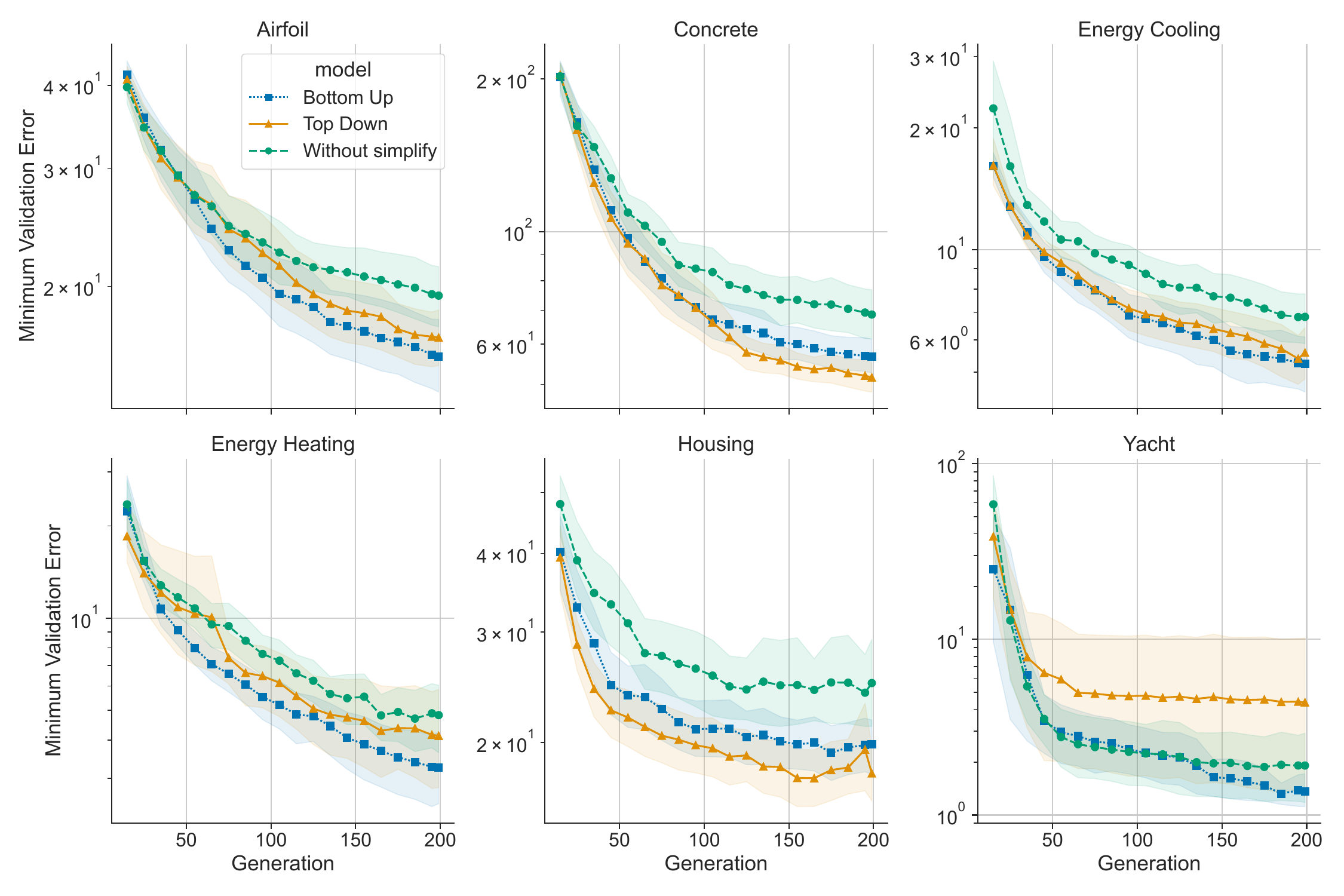}
    \legend{Source: Elaborated by the author.}
\end{figure}

Figure \ref{fig:simplification:convergence_error} reports the minimum validation error during the evolution, estimated using the mean of $30$ runs.
The first generation is discarded, since individuals were only randomly initialized and have not yet undergone selective pressure, and invalid or extremely poor individuals can be present in the population and bias the estimation.

These results show that simplification often outperforms the traditional GP throughout the run, leading to a better local optimum. The only exception is the Yacht dataset, in which the top-down strategy performs worse than the bottom-up and the version without simplification. In every other dataset, both traversal strategies present equiparable performance.
For the six datasets, worse performance was observed in the first five when simplification was not applied.
Overlap between the estimated confidence intervals was present during earlier generations but became less common in later generations.
In particular, for the Housing dataset, the validation error increased when simplification was not used, indicating overfitting to the training data.
For the Yacht dataset, no difference was observed between the absence of simplification and the bottom-up approach, while the top-down approach appeared to reach a \textit{plateau} in the earlier generations.

\begin{figure}[htb]
    \centering
    \caption{Number of simplifications performed in each generation by the top down and bottom up strategies.}
    \label{fig:simplification:n_simplifications}
    \includegraphics[width=\linewidth]{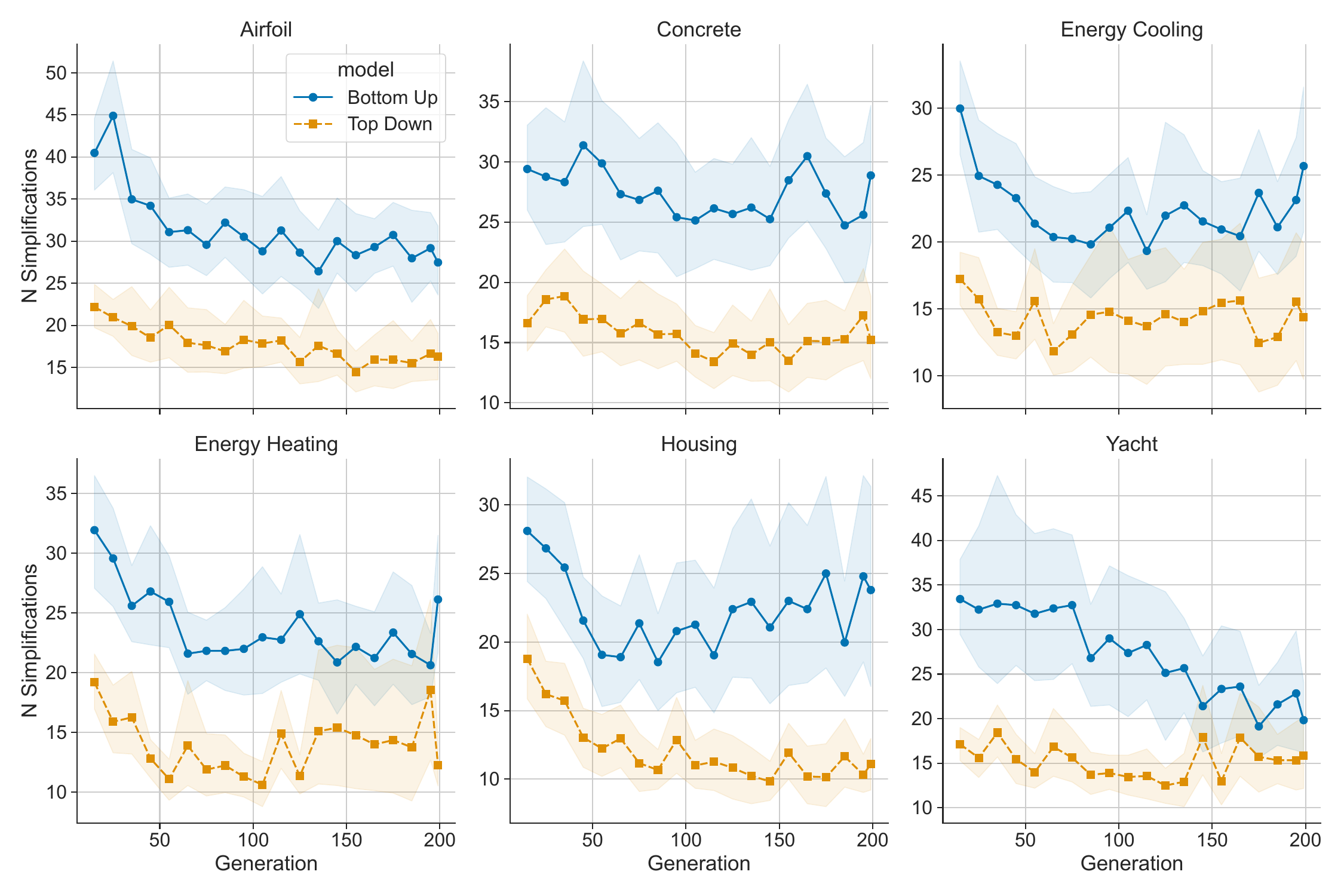}
    \legend{Source: Elaborated by the author.}
\end{figure}

As previously conjectured, the two traversal strategies perform different numbers of simplifications during the evolutionary process, as shown in Figure \ref{fig:simplification:n_simplifications}.
Approximately $50\%$ more simplifications are performed by the bottom-up strategy, with only a small overlap in the estimated confidence intervals.

The different simplification strategy shows only small differences and no clear better option in terms of behavior during the run, except for the number of simplifications.
The decrease in the number of simplifications performed by the bottom up strategy observed in later generations for some datasets, such as Airfoil, Energy Heating, and Yacht, is attributed to the effective removal of bloated and redundant subtrees; however, models also become more specialized and complex, making further simplification more difficult, albeit new random subtrees are being generated with the variation operators.
However, notice that the decrease of simplifications is not a global trend, as this behavior does not happen in all datasets. 

The observed decrease in the number of simplifications does not necessarily imply undesirable behavior; rather, it reflects that models in later generations are less likely to be simplified. However, it may indicate that, for some problems, the computational cost introduced by the simplification method becomes less beneficial as evolution progresses. This observation suggests opportunities for future work exploring heuristics to determine when and to which individuals simplification should be applied.


\subsection{Goodness-of-fit and size trade-off}

\begin{figure}[htb]
    \centering
    \caption[Comparison of the final solutions under different simplification methods]{Comparison of the final solutions under different simplification methods: (a) solution size and (b) solution complexity.}
    \subfloat[Final size of the solutions\label{fig:simplification:size_boxplot}]{%
        \includegraphics[width=0.5\linewidth]{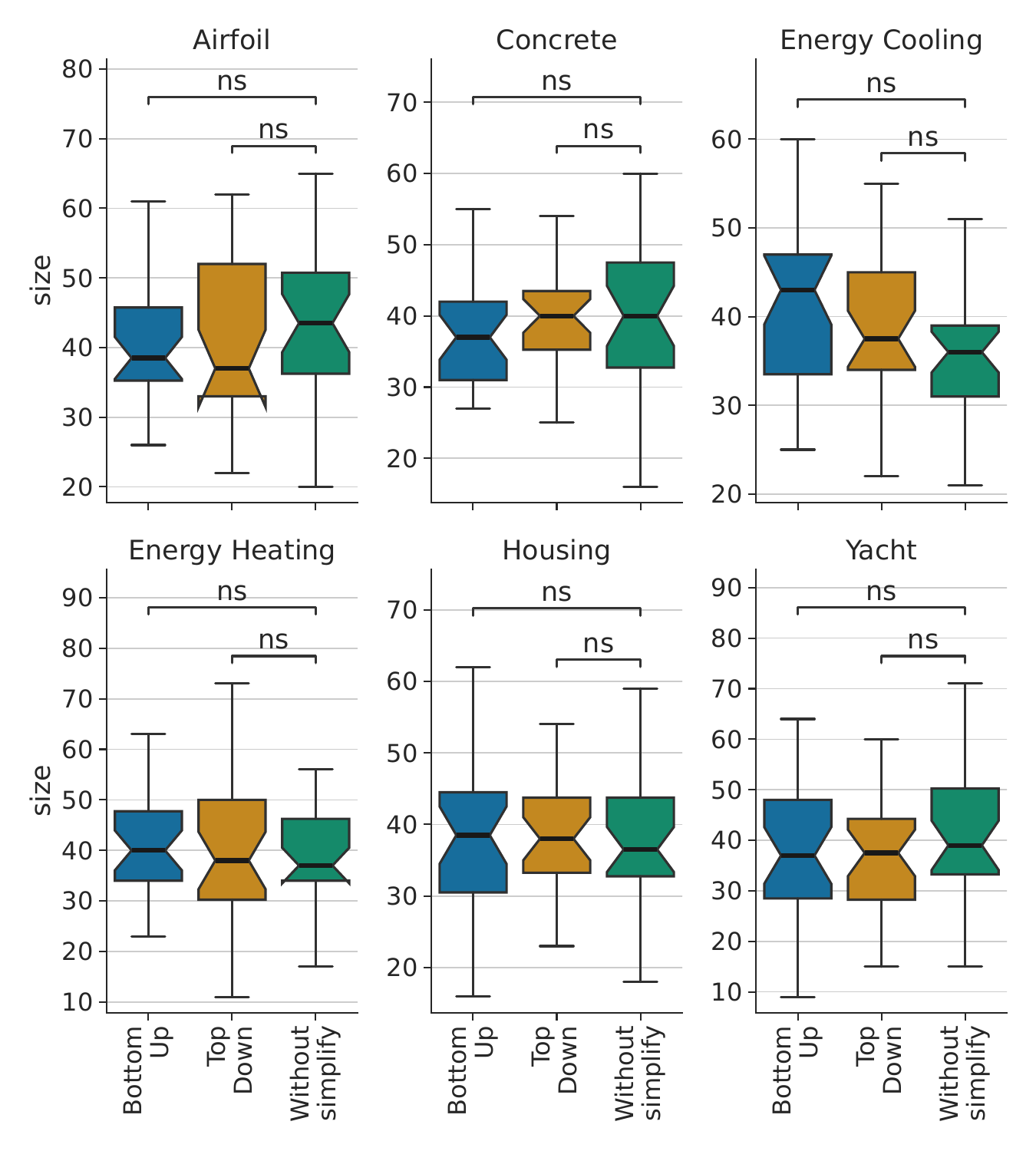}}
    \hfill
    \subfloat[Final complexity of the solutions\label{fig:simplification:complexity_boxplot}]{%
        \includegraphics[width=0.5\linewidth]{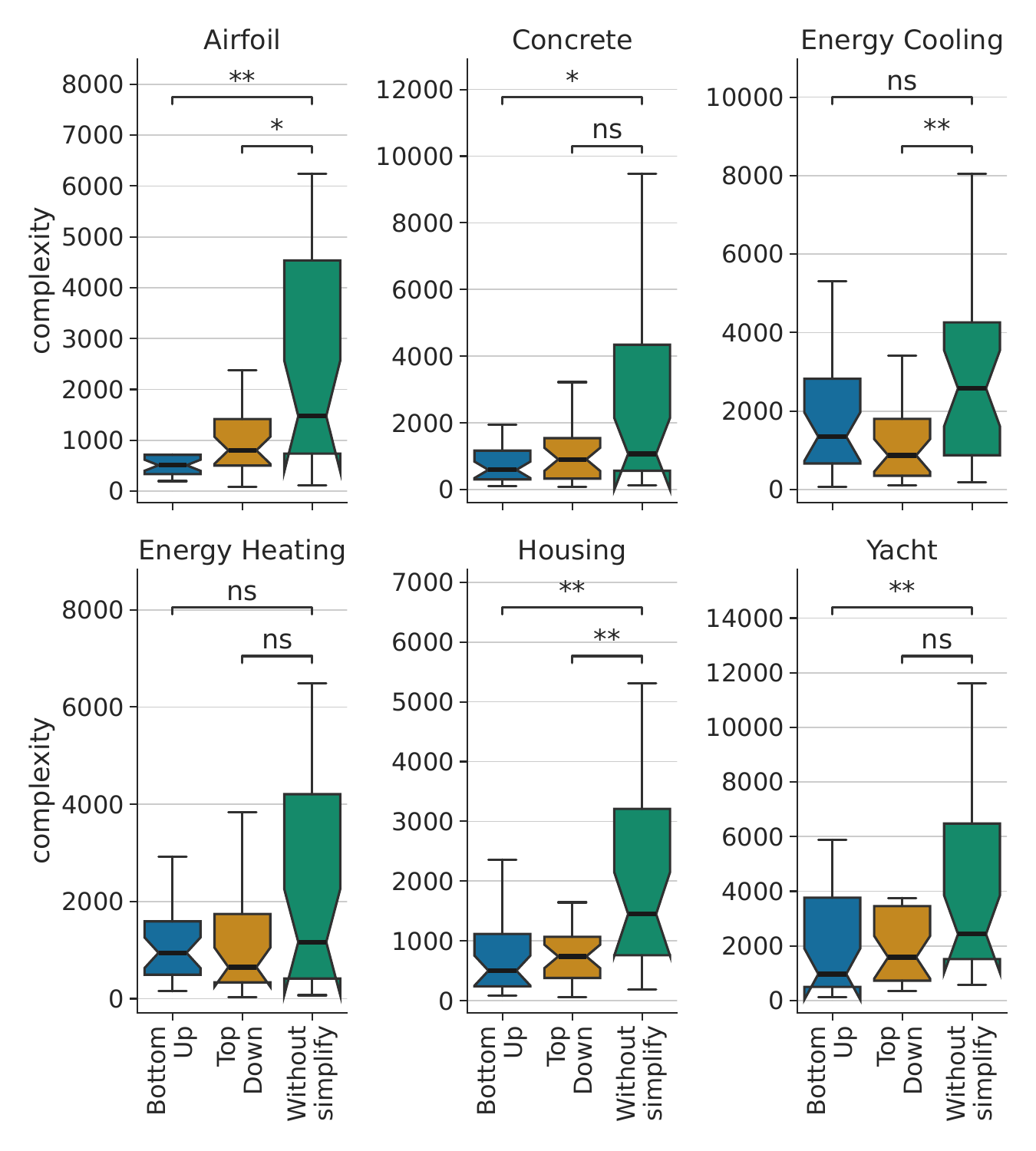}}
    \legend{Source: Elaborated by the author.}
\end{figure}

\begin{figure}[htb]
    \centering
    \caption[Final MSE on the test partition using different simplification strategies. ]{Final MSE on the test partition using different simplification strategies. Simplification led to better performance in error, with less complex models.}
    \label{fig:simplification:mse_boxplot}
    \includegraphics[width=0.5\linewidth]{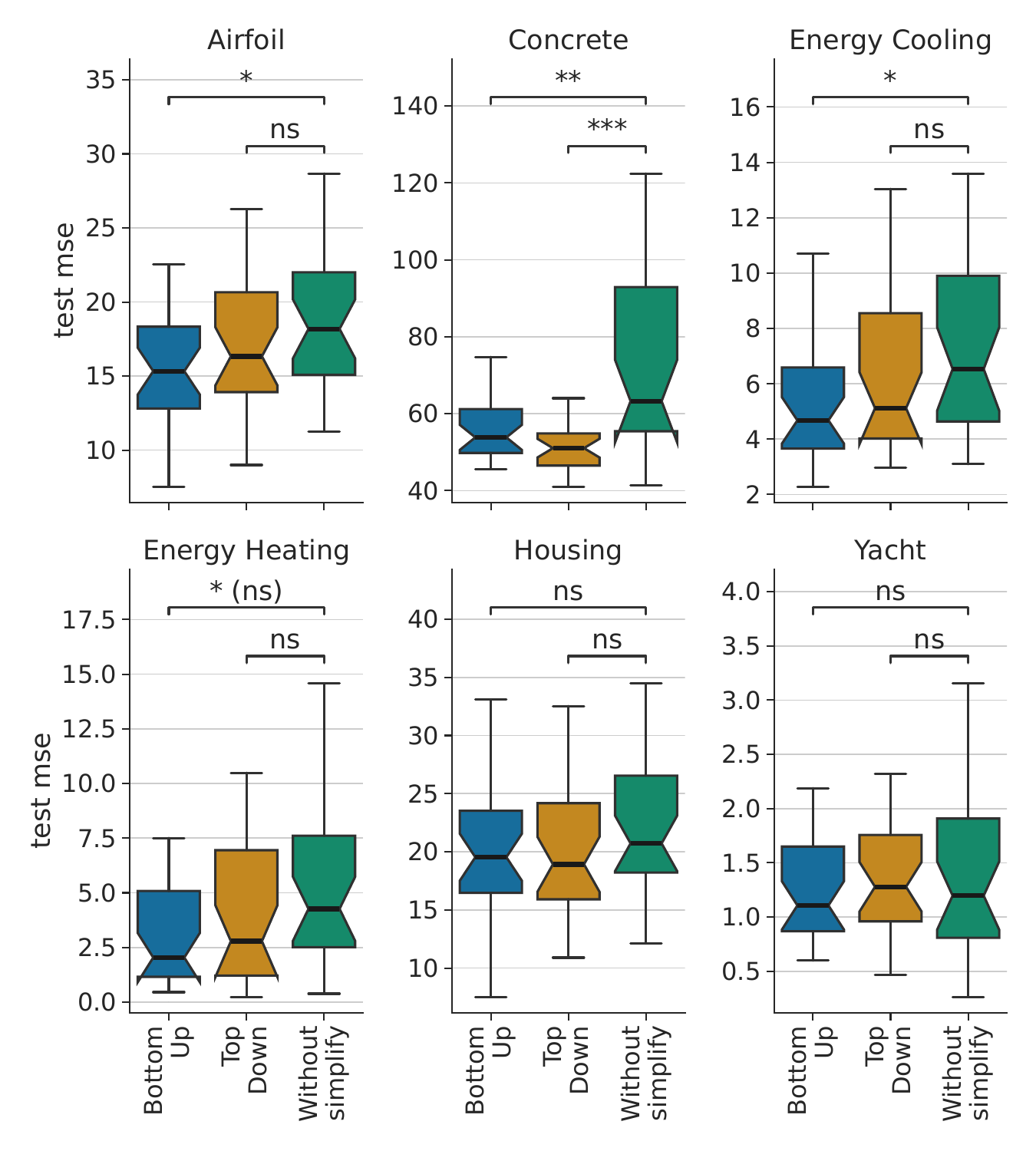}
    \legend{Source: Elaborated by the author.}
\end{figure}

At the end of the run, the algorithm picks the final model based on the performance on the validation split. Figures \ref{fig:simplification:size_boxplot}, \ref{fig:simplification:complexity_boxplot}, and \ref{fig:simplification:mse_boxplot} report the size, complexity, and MSE on the test partition, respectively.

Regarding solution size, the null hypothesis that no differences exist between the approaches cannot be rejected --- a counter-intuitive result, since simplification is expected to reduce expression size.
However, the complexity boxplots reveal smaller variation for the best solutions when simplification strategies are applied, with the null hypothesis being rejected at a p-value between $[10^{-3}, 10^{-2}]$.
Size and complexity are related, but not equivalent. A large model built with simple arithmetic operations will have a smaller complexity than a model with non-linear operations; and a model that performs non-linear operations close to the root will have a higher complexity than models that perform these transformations close to the leaves.

It is hypothesized that simplification temporarily decreases size, creating room for further improvements, which the evolutionary process exploits by producing expressions of similar size but improved performance, allowing more useful sub-expressions to be accommodated.
As a consequence, the generated expressions are less complex according to the recursive complexity.
Differences in complexity were observed for the Airfoil, Concrete, Energy Cooling, Housing, and Yacht datasets.
Even when solutions were equivalent in size, simplification led to expressions with better complexity without explicitly minimizing it.
The simplified expressions did not exhibit higher errors than the original ones.
Regarding test set MSE, statistically significant improvements were observed in the bottom-up strategy for the Airfoil, Concrete, and Energy Cooling datasets, while the top-down strategy showed improvements only for Concrete.

Yacht was considered the most challenging dataset due to the test MSE scale shown in Fig.~\ref{fig:simplification:mse_boxplot}, which is closer to the threshold than for any other dataset.
Parameter optimization, particularly the adjustment of the inexact simplification threshold, appears to be important based on the error scale and will be investigated further.

The observed difference in performance while maintaining solution sizes is hypothesized to occur because simplification creates space for further model improvements and removes overly complex subtrees that could drastically alter overall predictions when a variation operator is applied.
As a consequence, the resulting models are more robust to local changes.

\subsection{Relative change}

\begin{figure}[htb]
    \centering
    \caption[Percentage variation of size, complexity, and MSE on test compared to without any simplification]{Percentage variation of size (\ref{delta_size}), complexity (\ref{delta_complexity}) and MSE on test (\ref{delta_test_mse}) compared to without any simplification.}
    \label{fig:simplification:deltas}
    \subfloat[\label{delta_size}]{%
       \includegraphics[width=0.2\linewidth]{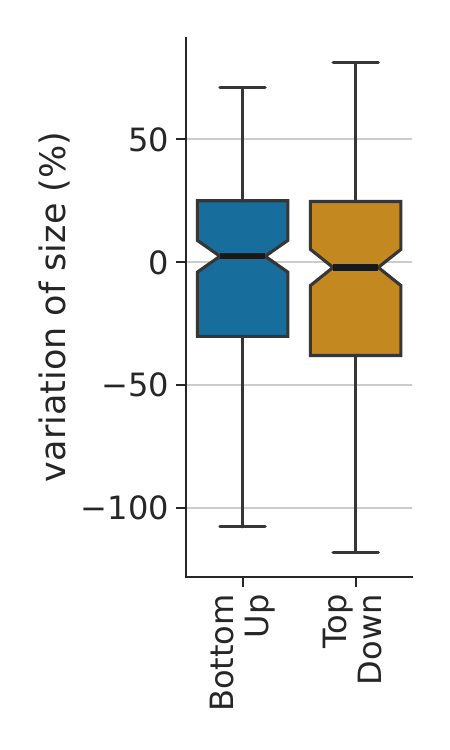}}
    \subfloat[\label{delta_complexity}]{%
       \includegraphics[width=0.2\linewidth]{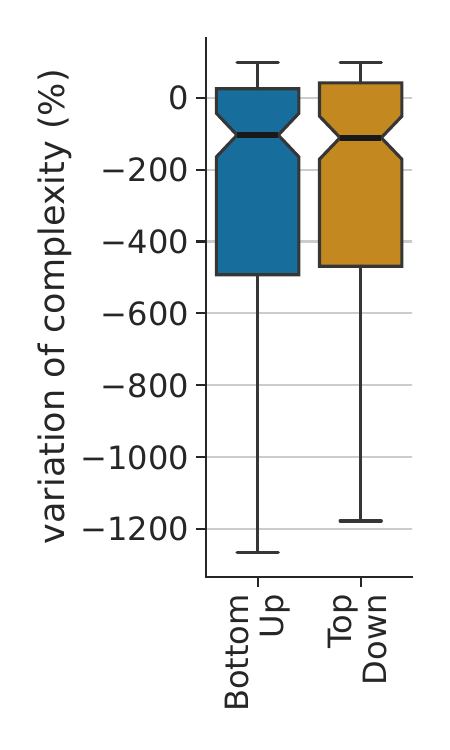}}
    \subfloat[\label{delta_test_mse}]{%
       \includegraphics[width=0.2\linewidth]{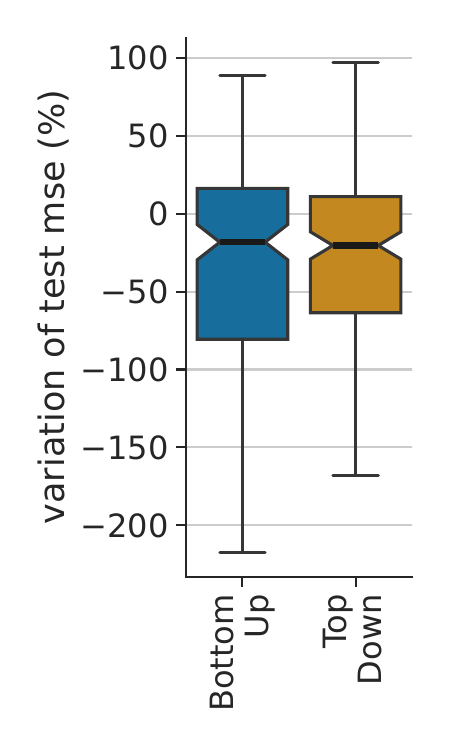}}
    \legend{Source: Elaborated by the author.}
\end{figure}

Even though the differences in MSE seem modest, those are considering the mean and median of the distribution. A better way to verify the benefits of the simplification strategy is to pair the $30$ runs based on their seed and calculate the percentage of variation between the simplification methods and without any simplification (\textit{i.e.}, the baseline). This way, the variation between datasets can be aggregated, and an overall measurement and a one-sided $t$-test can be applied to verify whether the mean of the relative change is different from zero. Figure \ref{fig:simplification:deltas} reports the variations for the bottom-up and top-down strategies.
A t-test for the mean of the distributions was performed for each subfigure with $180$ degrees of freedom ($6$ datasets and $30$ runs each).

Regarding size, the distributions show a median change of $2.43\%$ for the bottom-up strategy and $-2.13\%$ for the top-down strategy, but both yielded p-values greater than $0.05$, so the null hypothesis cannot be rejected.
Complexity exhibits median reductions of $-104.06\%$ and $-111.34\%$ for bottom-up and top-down, respectively, with p-values of $2.94 \times 10^{-7}$ and $1.59 \times 10^{-6}$, concluding that the complexity is decreased by $100\%$ on average when simplification is performed.
This indicates that a reduction in complexity is observed for each individual run compared to the baseline.
For MSE, median reductions of $-18.21\%$ and $-20.32\%$ were observed, with p-values of $8.165 \times 10^{-7}$ and $1.85 \times 10^{-5}$.
Thus, the null hypothesis can be rejected, and it can be concluded that simplification reduces the error by approximately $20\%$ on average.

\subsection{Number of equivalent expressions}

\begin{figure}[htb]
    \centering
    \caption[Number of equivalent expressions stored in each hash, for the first $30$ entries.]{Number of equivalent expressions stored in each hash, for the first $30$ entries. The same data is replicated with logarithmic scale on the $y$-axis on the top right corner of the plots.}
    \label{fig:simplification:n_equivalent_expressions}
    \includegraphics[width=\linewidth]{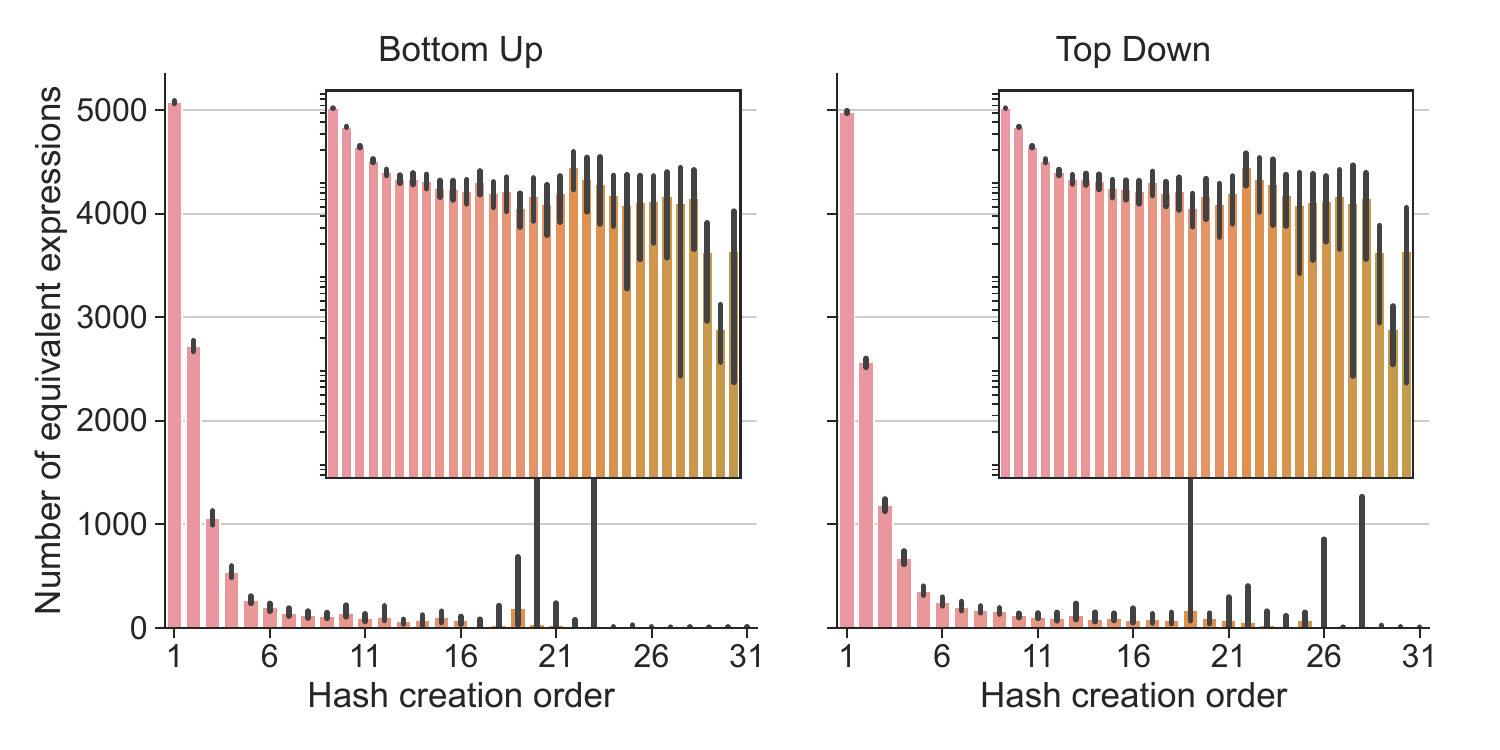}
    \legend{Source: Elaborated by the author.}
\end{figure}

Fig. \ref{fig:simplification:n_equivalent_expressions} 
reports the number of different expressions stored for each hash entry, where the $x$ axis refers to each hash based on it's order of creation. This means that the very first hash to be created is the constant, followed by hashes that stores each of the single features, altogether created during the table initialization. After that, the next entries stores more complex trees and are created after the initialization of the simplification table, during the evolution.


It can be observed that a very small number of hashes are mapped to almost all equivalent expressions, indicating that several equivalences are mapped to the terminals, mainly the constant, meaning the evolutionary process is creating bloated sub-trees that behaves as the original input features, or as constant branches that can be replaced with a single constant terminal.

The majority of hash entries, especially those involving trigonometric functions or large expressions, lack replacements but still consume memory to be kept in the table, also increasing time to run operations over it. A possible solution is to reduce computational and memory overhead by restricting simplification to expressions below a certain complexity threshold, such as those strictly smaller than $20$ nodes. Notice that this can be problem-dependent, or related to the operator set used.

\subsection{Measuring inexactness}

The differences in MSE between the original and simplified expressions can be directly inferred from the defined tolerance threshold, which is used to filter candidates and prevent large deviations in prediction error when performing the inexact simplification.
To further analize inexactness of simplifications, the angle between the prediction vectors before and after simplification is proposed as an additional measure. This angle quantifies the change in prediction semantics and indicates how the predictions are related.
As discussed in Section~\ref{sec:simplification:lsh}, the probability of collisions is proportional to the angle between the vectors. Large angles may indicate that distinct expressions are being mapped to the same hash, whereas close to zero angles suggest that simplification is capturing only exact matches rather than similar ones.
An angle of zero indicates perfect alignment and thus no variations in prediction --- but a small degree of variation is desirable, as it allows the simplification process to identify near-equivalent expressions with reduced complexity.

\begin{figure}[htb]
    \centering
    \caption[Distribution of the angle between the original and simplified expression predictions obtained through inexact simplification.]{Distribution of the angle between the original and simplified expression predictions obtained through inexact simplification, shown for different tolerance levels used to accept the simplification. Sorting the datasets by their number of features reveals that the angle is proportional to the number of features.}
    \label{fig:simplification:angle_between}
    \includegraphics[width=\linewidth]{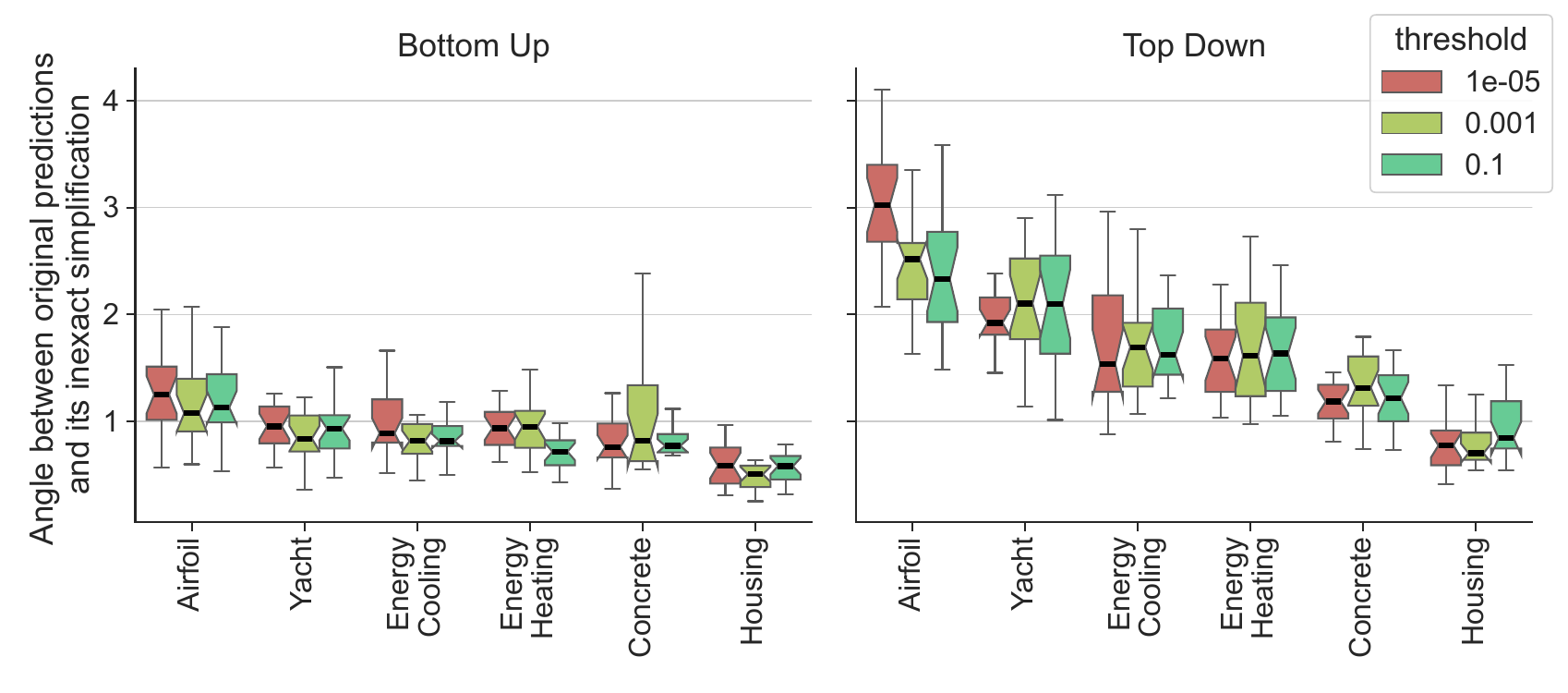}
    \legend{Source: Elaborated by the author.}
\end{figure}

Figure \ref{fig:simplification:angle_between} shows the distribution of the angle between the prediction of the original expression and that of its replacement, obtained through inexact simplification.  
The angle $\theta$ between the original prediction vector $\mathbf{p}$ and the simplified prediction vector $\mathbf{p}_{\text{replace}}$ was computed as:

\begin{equation}
    \theta = \arccos \Bigg( \underbrace{\frac{\mathbf{p} \cdot \mathbf{p}_{\text{replace}}}{\|\mathbf{p}\| \cdot \|\mathbf{p}_{\text{replace}}\|}}_{\cos{\theta}} \Bigg),
\end{equation}

\noindent where $\cdot$ denotes the dot product and $\|\cdot\|$ denotes the Euclidean norm. The angle was clipped between $[-1,1]$ to avoid numerical errors.

This result shows that the inexactness is expressed with few regard to the simplification threshold used, and the angle is of at most $4$, while this represents an extreme case. On average, the angle is between $1$ and $2$, and is problem-dependent, as Airfoil holds the highest angles, and Housing the smallest. 
In addition, the Bottom-up shows smaller angles, thus smaller inexactness, than the Top-Down approach. This is likely due to performing simplification through several steps from bottom-up, instead of directly cutting off a tree such as how Top-down is likely to do on larger sub-trees.

The angle between the original predictions and their inexact simplifications appears to correlate with the number of features in each dataset. Airfoil, which has five features, exhibits the highest median values for both the bottom-up and top-down strategies, whereas Housing, with thirteen features, shows the smallest median values for both strategies. The remaining datasets fall in between, with either six or eight features. Because the table is always initialized without collisions among input features, this suggests that having more initial entries in the table reduces the chances of collisions between expressions with distinct semantics.

\subsection{Analysis of hashes and individuals}

A single run was performed on the Yacht dataset with the bottom-up approach to gain insights into how simplification replaces nodes.
This dataset was chosen because no statistically significant differences in size or error were observed, and it is low-dimensional, making it suitable for demonstration.
At the end of the run, a total of $5,144$ hash entries were created from $7,324$ expressions.

Some entries of the simplification table are reviewed in this subsection.
Each entry consists of a tuple containing a hash and a list of equivalent trees.
The hash is truncated for clarity, and the expressions are shown in string format, ordered as the algorithm store them (from smallest to largest), so that any subtree would be replaced by the first in the list.
The first case is examined as follows:

\begin{Verbatim}[frame=single]
1010110111001101...
 - square(x_5)
 - multiply(x_5, x_5)
 - absolute(square(x_5))
 - maximum(square(x_5), x_0)
 - maximum(add(-15.455, x_1), square(x_5))
 - ... (5 more)
\end{Verbatim}

It is first observed that, without explicit rules, the method learned to replace the corresponding subtree \verb|multiply(x_5, x_5)| with \verb|square(x_5)|, providing a shorter representation.
It also learned that the absolute value of a square is always positive, so \verb|absolute(square(x_5))| can similarly be simplified to \verb|square(x_5)|.
Some simplifications are data-driven; for example, in \verb|maximum(square(x_5), x_0)|, the term $x_5^2$ was later confirmed in the data to always dominates the term $x_0$ in the $\mathtt{max}$ function, which would not be captured by algebraic rules.

Additional interesting cases are observed, including a permutation of arguments in a $4$-ary commutative operation, as well as a chain of commutative operations:

\begin{Verbatim}[frame=single]
1110001011011111...
 - multiply(x_1, x_7, x_0, x_4)
 - multiply(x_0, x_4, x_1, x_7)
 - multiply(x_1, x_0, multiply(x_4, x_7))
\end{Verbatim}

In particular, notice that rewriting \verb|multiply(x_1, x_0, multiply(x_4, x_7))| as \verb|multiply(x_1, x_7, x_0, x_4)| decreases by one the number of nodes, meaning the simplification table is also helping improving the aggregation of chained $2$-ary operations as a $4$-ary multiplication. This, however, is only possible because the algorithm is set to allow multiple arguments to operations with commutative properties.

Additional cases were observed in which redundant operations were simplified, such as \verb|minimum(x_2, x_2)|, the identity value of an operation was used as an argument, as in \verb|add(0.0, x_2)|, and functions were chained with their inverses, as in \verb|log(exp(x_2))|:

\begin{Verbatim}[frame=single]
1010101111001001...
 - x_2
 - absolute(x_2)
 - minimum(x_2, x_2)
 - minimum(x_2, x_6)
 - minimum(x_2, x_5)
 - add(0.0, x_2)
 - log(exp(x_2))
 - sqrtabs(square(x_2))
 - maximum(-523.249, x_2)
 - ... (34 more)
\end{Verbatim}

From the previous example, one inverse relationships were automatically learned, such as $\log{(\exp{(a)})} = a$. The addition of $a$ with the neutral element of sum is also identified by the simplification method, since $0.0 + a$ is being simplified to just $a$.

Expressions that are decomposable into the sum of two terms can also be observed, where one term dominates the other by several orders of magnitude, rendering the contribution of the smaller term irrelevant to the final predictions.
In these cases, the simplification method learned several trees in which only the dominant term needs to be retained:

\vspace{2.0em}

\begin{Verbatim}[frame=single]
1100000011100001...
 - exp_1_plus(x_3)
 - div(exp_1_plus(x_3), 0.0)
 - maximum(exp_1_plus(x_3), square(x_2))
 - add(tan(x_4), arctan(x_3), exp_1_plus(x_3))
 - add(exp_1_plus(x_3), add(x_1,x_2,x_6,x_5), tan(x_4))
 - maximum(exp_1_plus(x_3), multiply(multiply(77.001, x_4, x_2), 
                                     square(x_7),
                                     cdiv(x_2, x_6),
                                     add(x_6, x_6, -77.991)))
\end{Verbatim}

The power from the proposed simplification method is that all these simplifications were made without defining any simplification rule, and learned with small overhead of implementation, through an interative process that feeds a library of equivalent expressions and performs inexact approximations for fast retrieval of information.

\begin{figure}[htb]
    \centering
    \caption[Smallest expression found by bottom-up (left) and without simplification (right) for the Yacht dataset]{Smallest expression found by bottom-up (left) and without simplification (right) for the Yacht dataset. The MSE for each expression was $1.98$ using the bottom-up simplification and $2.08$ without simplification.}
    \label{fig:simplification:example_expressions}
    \includegraphics[width=0.6\linewidth]{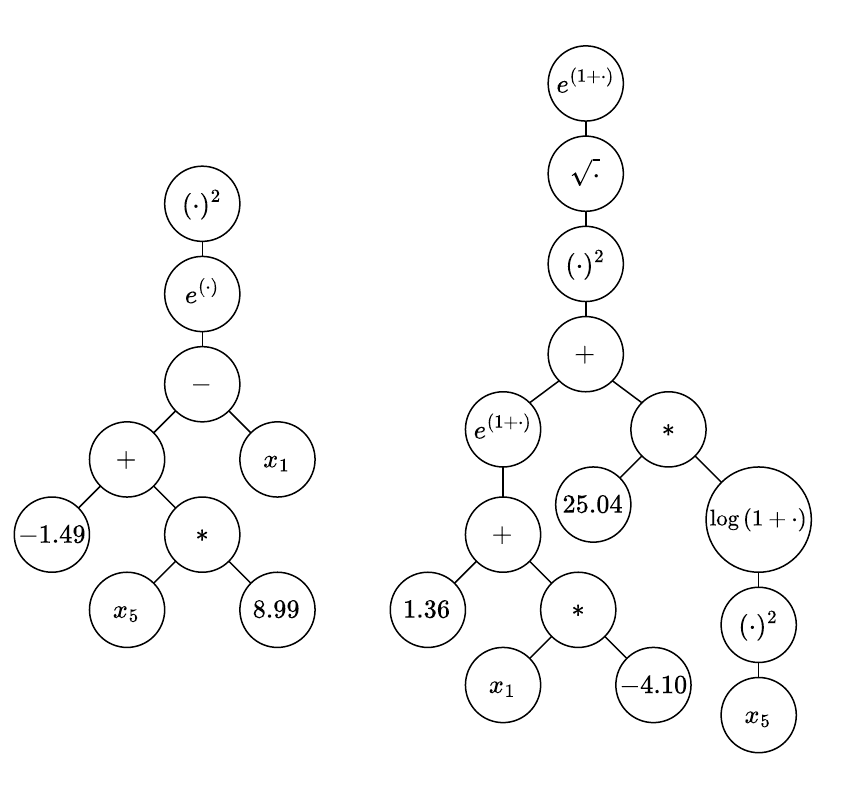}
    \legend{Source: Elaborated by the author.}
\end{figure}

Figure \ref{fig:simplification:example_expressions} shows the smallest expression found by the bottom-up strategy and by GP without simplification for the Yacht dataset.
An MSE of $1.98$ on the test set was obtained by the bottom-up model, using $9$ nodes, while GP without simplification produced a test error of $2.08$ using $15$ nodes.
More complex constructs were observed when simplification was not applied, including the chaining of $\sqrt{\cdot}$ and $(\cdot)^2$.

The Yacht dataset exhibits errors that are an order of magnitude greater than the simplification threshold. Although no improvement in error was observed, improvements in complexity were achieved.
By analyzing a single run, it was demonstrated that the benefits of the simplification method extend beyond size and error, as complexity was minimized without being explicitly targeted.

\subsection{Execution time}

\begin{figure}[htb]
    \centering
    \caption[Variation of execution time for each method in the simplification experiments.]{Variation of execution time (in seconds) for each method in the simplification experiments. The subplots are sharing the $y$ axis.}
    \label{fig:simplification:time_boxplots}
    \includegraphics[width=0.525\linewidth]{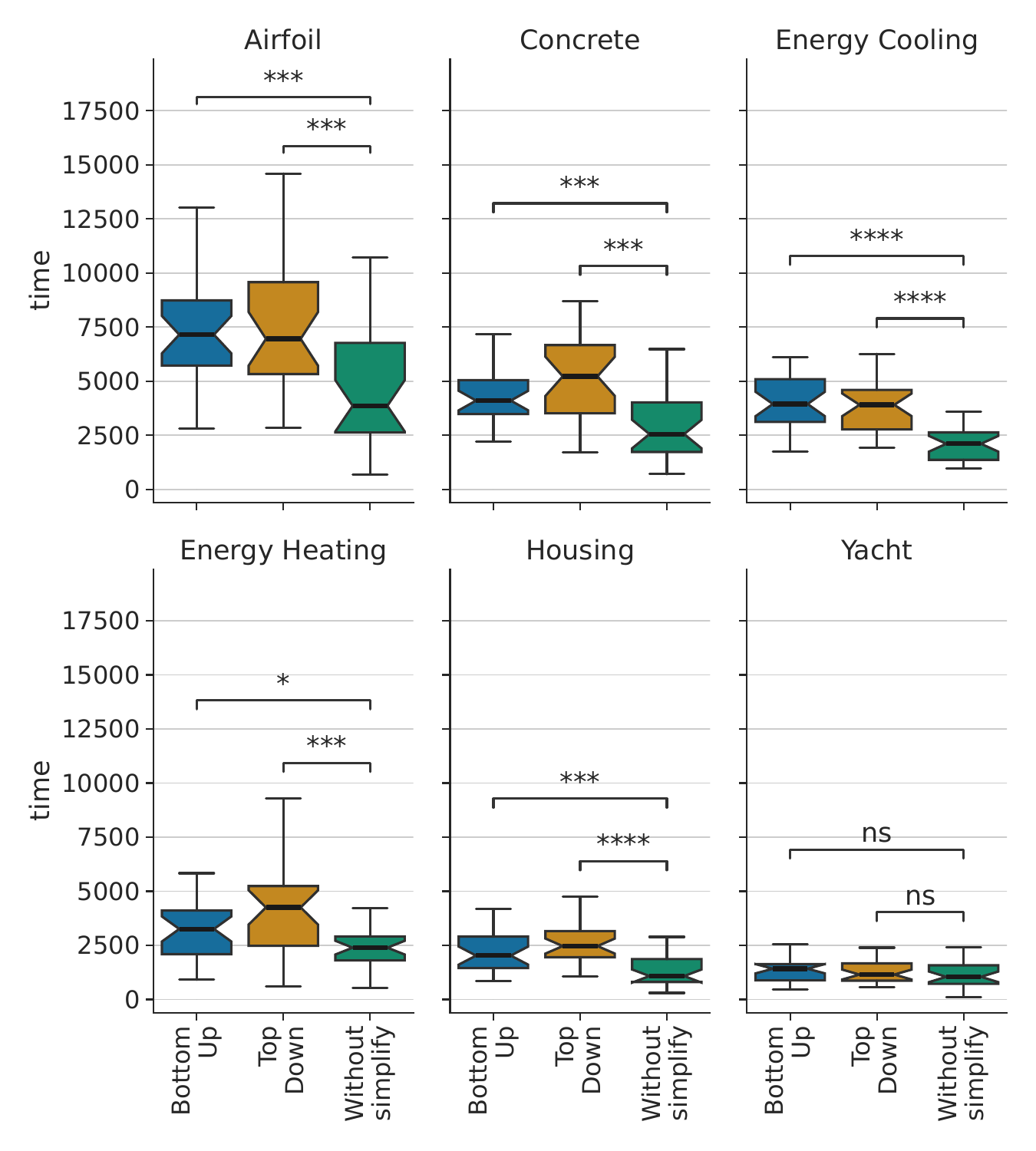}
    \legend{Source: Elaborated by the author.}
\end{figure}

Figure \ref{fig:simplification:time_boxplots} reports the execution time for each algorithm.

It is observed that the simplification methods require more time to run than when no simplification is applied, with statistically significant differences for all but the last dataset.
This occurs because non-linear optimization is performed twice for every individual, as they are re-optimized after simplification, indicating that the method could benefit from caching recent information to avoid re-optimizing the entire tree after each simplification, but this cannot be done with inexactness since prediction is highly sensitive to parameter values.

\section{Conclusions}~\label{sec:simplification:conclusions}

In this chapter, an inexact simplification method using LSH is proposed to learn substitution rules, which can correspond either to algebraic identities or to simplifications specific to the dataset domain. Expressions that behave similarly in the phenotypical space are efficiently stored and queried using LSH, enabling the simplification process to be applied to every sampled expression throughout the evolution.
Using this approach, the influence of simplification and bloat control on the evolutionary process can be experimentally analyzed. 
As a by-product, an implementation of an evolutionary algorithm with non-linear parameter optimization and inexact simplification was made available in Python.

The experiments were conducted with well-known datasets from the machine learning community, logging the performance on a validation partition during the run. The final results and convergences were analyzed.
Empirical results indicate that, on average, versions with simplification returned expressions of the same size but with less complex constructs and higher accuracy. When comparing runs originating from the same seed, a median reduction of $20\%$ in mean squared error was observed when simplification was applied.

Notice that because the simplification is inexact, some substitutions may slightly alter the behavior of the function or be exact only on the training data.
On the other hand, since this is a data-driven approach, it can learn the rules on the fly without any need to pre-determine the algebraic identities. It can also learn rules specific to the dataset (\textit{e.g.}, $x_1$ is always smaller than $x_2$ in all samples). 

The simplification does not hurt performance overall, but often yields better results than without any simplification, and has little burden to implement into any symbolic regression framework.
It can be implemented into any symbolic regression algorithm, as long as there is a pragmatic way of transversing the expressions and replacing parts of it.




%% file: Capitulos/Benchmarking/benchmarking.tex
\Chapter{Benchmarking}{Assessing the Symbolic Regression landscape}~\label{chap:benchmarking}
\chaptermark{Benchmarking}

\begin{laysummary}
    Benchmarking symbolic regression (SR) methods remains challenging due to the diversity of algorithms, datasets, and evaluation criteria. In this chapter, a comprehensive SR benchmark is introduced, inspired in SRBench. 
    A total of $25$ methods were assessed across two problem tracks: black-box and phenomenological/first-principles modeling. Each experiment was executed with $30$ independent runs to ensure robust statistical evaluation. Energy consumption and execution time were reported. A new way of visualizing the performance, as probabilities of observing specific performances, is presented and discussed as an alternative way of aggregating results. In addition, trade-offs among model complexity and predictive accuracy were systematically analyzed, focusing on non-aggregated reporting of results. The results reveal that no single algorithm consistently outperforms all others, underscoring the importance of context in method selection.
    Best practices for improving SR algorithms were also distilled from the benchmarking process, highlighting the need for adaptive hyperparameter tuning and energy-efficient implementations. Excessive and difficult-to-tune hyperparameters were identified as a recurring drawback for practitioners. Common patterns among the best-performing algorithms were observed, particularly the use of parameter optimization, boosting, and GP-based approaches.
    Finally, the Brush framework, integrating parameter optimization, $\epsilon$-lexicase selection, and inexact model simplification, was benchmarked. Brush achieved strong performance across multiple evaluation metrics, demonstrating its potential as a unifying approach to SR.
\end{laysummary}

{\let\thefootnote\relax\footnotetext{Part of this chapter was published as: Aldeia, G. S. I.; Zhang, H.; Bomarito, G.; Cranmer, M.; Fonseca, A.; Burlacu, B.; La Cava, W. G.; de França, F. O. 2025. Call for Action: towards the next generation of symbolic regression benchmark. In Proceedings of the Genetic and Evolutionary Computation Conference Companion (GECCO '25), July 14-18, 2025, Málaga, Spain. ACM, New York, NY, USA, 8 pages. \url{https://doi.org/10.1145/3638529.3654149}}}

\vfill
\pagebreak

\section{Introduction}~\label{sec:benchmarking:introduction}

Running a benchmark is a laborious yet essential task. Beyond simply comparing methods, it serves as a tool for measuring progress in the field. Despite the long-standing history of Symbolic Regression (SR) and recent advances, \citeonline{srbench} observed a lack of consensus regarding standardized datasets and benchmarking methodologies, making it difficult to accurately assess the state-of-the-art (SotA). Another challenge is the absence of a unified interface and installation/use instructions, which creates barriers to experimentation and reproducibility.

To address some of these challenges, La Cava et al. published SRBench\footnote{\url{https://cavalab.org/srbench/}} as a living benchmark for SR, comparing different SR implementations against standard ML regression models. This initiative introduced a common Python API and installation scripts within a \textit{sandbox} environment, simplifying installation and evaluation across algorithms.
SRBench represented a significant step toward comparing and understanding the SotA in SR. However, designing effective benchmarks often requires continuous refinements and consideration of new factors to ensure their usefulness and longevity. 

SRBench originally evaluated $14$ SR methods and $7$ traditional ML models (\textit{e.g.}, random forests, linear regression) across $122$ black-box problems ---including $62$ Friedman problems, a collection of synthetic datasets with varying noise levels that are challenging due to multiple local minima--- and $100$ datasets derived from Feynman equations with uniformly sampled data. Each algorithm was run $10$ times per dataset, using hyperparameters selected through grid search on the black-box problems, with the most frequent configuration fixed for all final evaluations.

While SRBench broadened the scope by including diverse datasets (PMLB and Feynman equations) and more algorithms, discussions on limitations and improvements were noted by researchers, regarding the interpretability of aggregated results \cite{dick2022genetic}, overrepresentation of Friedman problems \cite{tir}, and the adequacy of synthetic data \cite{matsubara2024rethinking}, which failed to mimic real-world distributions in the ground-truth track. These shortcomings highlighted the need for a new iteration of the benchmark with improved visualizations, more robust statistical evaluations, and domain-specific datasets better aligned with SR's practical appeal in scientific discovery.

An effective benchmark should contain diverse subclasses of problems, identifying which algorithms perform best under specific conditions, thereby mapping problems to algorithms rather than prescribing a single SotA solution. It must also challenge contemporary SR methods while remaining computationally feasible and representative of real-world applications, ensuring that conclusions extend beyond artificial test cases. SRBench falls short in this respect: it relies on over $200$ datasets without categorizing their inherent challenges and presents final results only through aggregated metrics, potentially obscuring finer performance differences.

In this chapter, several improvements to SRBench are proposed. Key changes include expanding the benchmark to $25$ SR algorithms; increasing the number of independent runs to $30$ to enhance statistical power; selecting $24$ datasets divided into two tracks: \textit{(i)} black-box, drawn from the original SRBench, and \textit{(ii)} phenomenological and first-principles, drawn from~\citeonline{cranmer2023interpretablemachinelearningscience}; and refining reporting practices to provide nuanced comparisons among top-performing algorithms.

This chapter also highlights the practical challenges of benchmarking and proposes features to streamline the process. Progress in this direction requires active collaboration with method developers and maintainers to extend and consolidate the common API, ensuring that SRBench can evolve as a flexible and robust framework.

The chapter is structured as follows. Section~\ref{sec:benchmarking:related_work} reviews previously published SR benchmarks. Section~\ref{sec:benchmarking:methods} presents the proposed update to SRBench, with subsections addressing each key aspect of the revision. Section~\ref{sec:benchmarking:results} reports and discusses results using different aggregation levels to extract deeper insights into algorithm performance. Finally, Section~\ref{sec:benchmarking:conclusions} summarizes the findings and presents a call for a collective effort to advance SR benchmarking.

\section{Related work}~\label{sec:benchmarking:related_work}

In 2018, \citeonline{10.1145/3205455.3205539} surveyed many SR papers showing that methods were being benchmarked on simple problems with a lack of standardization, which could diminish the perception of SR achievements outside of its field. They also acknowledged previous efforts to conduct benchmarks for SR methods, then conducted an experiment with four SR algorithms and different ML methods across more than $100$ datasets of up to $1000$ samples.

\citeonline{srbench} followed up the benchmark efforts and proposed SRBench, a joint-effort to achieve a large scale comparison of $14$ modern SR algorithms comprising GP-based, deterministic, and deep learning based approaches. In this paper, the authors mention that determining the SotA should not be the sole focus of research --- however, promising avenues of investigation cannot be well-informed without empirical evidence. In SRBench, the authors increased the number of problems and their dimensionalities, using datasets from Penn Machine Learning Benchmarks (PMLB)~\cite{Olson2017PMLB,romano2021pmlb}, an open-source library for benchmarking ML methods. A new ground-truth track was also added with more than $100$ physics equations from Feynman lectures~\cite{feynman2006feynman,feynman2015feynman} with data generated synthetically. They  provided a more informative view of the current state of the field. Given the large number of problems and algorithms, they performed $10$ individual runs for each dataset, and the budget was based on evaluations --- which is hard to measure uniformly across different approaches.

\citeonline{de2024srbenchplusplus} benchmarked a smaller selection of SR algorithms, including more recent approaches, on a proposed set of artificially generated datasets to measure the performance in specific tasks: overall accuracy performance, robustness to noise, capabilities of selecting the relevant features, extrapolation performance, and interpretability. With a smaller selection of algorithms and datasets it was possible to make precise statements of the current state of SR and their weaknesses. The authors also introduced a runtime limit instead of using the number of evaluations, stimulating the efficient implementation of the algorithms. The authors noticed that even though SR shares many shortcomings with traditional ML models, they have a potential to overcome it using domain knowledge and allowing a customized experience to the user~\cite{Ingelse2023}. 

\citeonline{matsubara2024rethinking} noticed that the Feynman dataset~\cite{aifeynman} used in the SRBench's ground-truth track did not adequately depict the original physical phenomena because of how the values were sampled --— the data generation was originally done by uniformly sampling standardized values. They proposed to replace the synthetic data generation process with normal or logarithmic distributions that mimic what would be empirically observed if the data was collected in practice, and evaluated six different SR methods.

Regarding problem-specific benchmarks, \citeonline{egklitz2020} proposed a benchmark with open-source and ease-of-use principles, evaluating four SR methods on five synthetic and four real-world problems.
\citeonline{esg1352} also benchmarked five SR methods focusing on applying SR for learning equations in civil and construction engineering, a field where equations are traditionally derived using linear regression.

\citeonline{thing2025cp3} proposed an unified tool for benchmarking SR in cosmology, including $12$ SR methods and $28$ datasets representing cosmological and astroparticles physics problems, finding that most methods performed poorly in the benchmark. They argue that using standard datasets (\textit{i.e.} PMLB) is prudent for academic comparisons, but these ensembles of datasets do not provide a representative measure for specific problems. They conducted off-the-shelf evaluations, which are useful because they do not require extra effort from the user, and, to improve reproducibility, they package the $12$ methods into Docker containers.

\section{Methods}~\label{sec:benchmarking:methods}

This section presents the datasets, benchmarked methods, experimental setup, and evaluation protocol. It also describes the computational infrastructure and online resources to ensure reproducibility.

\subsection{Dataset selection}

For this benchmark, two distinct tracks were defined: \textit{black-box} and \textit{phenomenological \& first-principles}. Table~\ref{tab:benchmarking:datasets} summarizes the datasets used in both tracks, including the number of samples (\# Rows), features (\# Cols), and target codomains. For the phenomenological \& first-principles track, all codomains consist of real values, and data sources are provided instead. All datasets were standardized prior to training and contain only numeric features without missing values.

\begin{table}[htb]
    \caption[Metadata for datasets in each benchmark track, for the proposed benchmark.]{
        Metadata for datasets in each benchmark track, for the proposed benchmark. Dataset names match their corresponding PMLB entries.
    }
    \label{tab:benchmarking:datasets}
    \centering
    \begin{tabular}{@{}rrrl@{}}
        \toprule\midrule
        \textbf{Black-box} & \textbf{\# Rows} & \textbf{\# Cols} & \textbf{Codomain} \\ \midrule
        1028\_SWD & 1000 & 11 & $\mathbb{Z}^{+}$ \\
        1089\_USCrime & 47 & 14 & $\mathbb{Z}^{+}$ \\
        1193\_BNG\_lowbwt & 31104 & 10 & $\mathbb{R}^{+}$ \\
        1199\_BNG\_echoMonths & 17496 & 10 & $\mathbb{R}$ \\
        192\_vineyard & 52 & 3 & $\mathbb{R}^{+}$ \\
        210\_cloud & 108 & 6 & $\mathbb{R}^{+}$ \\
        522\_pm10 & 500 & 8 & $\mathbb{R}^{+}$ \\
        557\_analcatdata\_apnea1 & 475 & 4 & $\mathbb{Z}^{+}$ \\
        579\_fri\_c0\_250\_5 & 250 & 6 & $\mathbb{R}$ \\
        606\_fri\_c2\_1000\_10 & 1000 & 11 & $\mathbb{R}$ \\
        650\_fri\_c0\_500\_50 & 500 & 51 & $\mathbb{R}$ \\
        678\_visualizing\_environmental & 111 & 4 & $\mathbb{R}^{+}$ \\ \midrule
        \textbf{Phenomenological \& first-principles} & \textbf{\# Rows} & \textbf{\# Cols} & \textbf{Data source} \\ \midrule
        first\_principles\_absorption & 14 & 2 & \citeonline{multiviewsr} \\
        first\_principles\_bode & 8 & 2 & \citeonline{bonnet1764contemplation} \\
        first\_principles\_hubble & 32 & 2 & \citeonline{Hubble1929} \\
        first\_principles\_ideal\_gas & 30 & 4 & Generated, 10\% noise \\
        first\_principles\_kepler & 6 & 2 & \citeonline{keplerHarmonicesMundi1619} \\
        first\_principles\_leavitt & 26 & 2 & \citeonline{leavittPeriods25Variable1912} \\
        first\_principles\_newton & 30 & 4 & Generated, 10\% noise \\
        first\_principles\_planck & 100 & 3 & Generated, 10\% noise \\
        first\_principles\_rydberg & 50 & 3 & Generated, 1\% noise \\
        first\_principles\_schechter & 27 & 2 & Generated, 20\% noise \\
        first\_principles\_supernovae\_zr & 236 & 2 & \citeonline{multiviewsr} \\
        first\_principles\_tully\_fisher & 18 & 2 & \citeonline{tullyNewMethodDetermining1977} \\ \midrule\bottomrule
    \end{tabular}
    \legend{Source: Elaborated by the author.}
\end{table}

The final black-box track comprises $12$ regression datasets selected among all regression datasets available from PMLB 1.0~\cite{romano2021pmlb}. Datasets were taken from the full collection of regression problems used in the original SRBench ($122$) and iteratively filtered based on the original SRBench performances: if all previously benchmarked algorithms achieved $R^2>0.99$, or if simple linear regression could reach the same threshold.
Metadata vectors for each dataset ---including number of samples, features, and algorithm performance in the original SRBench--- were reduced to two dimensions via t-SNE, followed by $k$-means clustering, with $k=12$, to have the same number of problems than the Phenomenological \& first-principles track, which is limited due to availability of open datasets in the literature that are related to the goal of the benchmark. 
After clustering, the dataset closest to each centroid was retained, ensuring diversity in horizontal and vertical dimensionality, as well as codomain distribution. To mitigate bias introduced by multiple Friedman dataset variants in the original SRBench, no more than $25\%$ of selected datasets were allowed to be from this collection of problems.
The goal is to identify a representative subset of the black-box problems from the original SRBench and to ensure that the selected datasets provide diverse coverage of SR applications, while avoiding datasets that are trivially solved by all previously benchmarked methods. This enables fine-grained analyses to better understand why certain methods outperform others.


The phenomenological \& first-principles track includes datasets from \citeonline{multiviewsr} and \citeonline{cranmer2023interpretablemachinelearningscience}. These datasets provide real-world measurements and underlying first-principles equations. They are particularly challenging due to small sample sizes and diverse, unknown distributions.

No level of synthetic noise was added for the black-box track, as the Friedman problems are already challenging. Datasets with real-world data from the phenomenological \& first-principles track already include noisy observations --- this is later observed in the Results and Discussion Section, where it is shown that the ground-truth equations for these problems do not achieve a perfect $R^2$ of $1.0$. Some particular cases for the phenomenological \& first-principles datasets were generated synthetically, and for these cases, Table \ref{tab:benchmarking:datasets} indicates the level of noise, which was sampled from a normal distribution based on each variable's range.

\subsection{Symbolic regression methods and hyperparameter tuning}

The benchmark extends the original SRBench by including recently published SR methods (Table~\ref{tab:benchmarking:methods_and_descriptions}) and detailing algorithm characteristics (Table~\ref{tab:benchmarking:methods_and_details}). Methods implemented in other languages were provided with Python bindings. Brush, a unified framework developed with the contributions presented in this thesis, was also included.

\begin{table}[h!]
    \caption[Algorithms evaluated and overall description of its main mechanisms.]{
        Algorithms evaluated and overall description of its main mechanisms. The methods are lexicographically ordered.
    }
    \label{tab:benchmarking:methods_and_descriptions}
    \scriptsize
    \centering
    \input{Capitulos/Benchmarking/floats/algorithm_descriptions}
    \legend{Source: Elaborated by the author.}
\end{table}

\begin{table}[h!]
    \caption{
        Algorithms evaluated, their original references, and relevant characteristics pertinent to benchmarking.
    }
    \label{tab:benchmarking:methods_and_details}
    \footnotesize
    \centering
    \input{Capitulos/Benchmarking/floats/algorithm_details}
    \legend{Source: Elaborated by the author.}
\end{table}

Grid search was performed for all methods, except Bingo, Brush, and GP-GOMEA, due to interface incompatibilities. The default (off-the-shelf) configuration was always included as a candidate, since preliminary results indicated that it occasionally outperformed the sets of hyperparameters defined for gridsearch. Hyperparameter optimization was performed using $3$-fold cross-validation on $75\%$ of the training data, selecting the configuration maximizing average $R^2$. Models were then trained on the full training set and evaluated on the held-out $25\%$. The tuning was performed before each run.
For the hyperparameter space, at most nine different combinations were allowed, with most combinations taken from the original SRBench paper. For the newly added methods, the hyperparameters were examined, focusing on those related to the search duration (\textit{e.g.}, number of generations) and search space size (\textit{e.g.}, maximum depth). These were adjusted so that the algorithms ran between $30$ minutes and $1$ hour, ensuring sufficient opportunity for search and convergence within the allowed budget, defined as wallclock time.

\subsection{Performance analysis}

Each method was executed with $30$ independent runs for each dataset, sharing the same $30$ random seeds across methods.

For this benchmark, a new visualization of aggregated results is proposed. The \textit{performance profile plot}~\cite{dolan2002benchmarking} is used, describing the empirical distribution of the obtained results. This plot illustrates the probability of achieving a performance greater than or equal to a given $R^2$ threshold for all possible thresholds. 
The choice of $R^2$ as the primary metric for predictive performance is motivated by its characteristic of placing results from different datasets on a comparable scale. This allows for both problem-specific analyses and broader evaluations across datasets with heterogeneous characteristics, with the latter offering an estimate that better reflects overall performance without being tied to any single problem.

In performance profile plots, the $x$-axis represents a threshold value of the $R^2$ and the $y$-axis the percentage of runs that a particular algorithm obtained that value or higher. This plot can give a broader view of the likelihood of successfully achieving a high accuracy for each algorithm, while keeping all the information about each algorithm in the same plot. Given the performance curve, it is possible to calculate the area under the curve (AUC) for each algorithm's profile as an aggregated measure: a value of $1.0$ indicates that all $30$ runs achieved the maximum $R^2=1.0$, whereas a value of $0.0$ indicates that all runs resulted in $R^2 \leq 0$. The AUC provides an estimate of the method's potential to achieve high performance across multiple runs.
The performance plot can be generated for each individual dataset as a result of the $30$ independent runs, or have the aggregated performance over all datasets using an aggregation function (\textit{e.g.}, mean, median, maximum, minimum) for each dataset. To report the results, the maximum value was used as the aggregation function, which reflects the best-case performance of each algorithm across the $30$ runs.

To measure model size, the final models are converted to a SymPy~\cite{10.7717/peerj-cs.103} compatible expression and count the number of nodes. This is necessary since the internal complexity measures of each algorithm may differ from the standard way of counting nodes.
The expression was not simplified (except for trivial simplifications such as constant merging) before counting the nodes, as simplification using SymPy can increase the model size in the process~\cite{10.1145/3583131.3590346}, being used only for assessing equivalence of expressions.


\subsection{Benchmark infrastructure}

Conducting extensive experiments across multiple algorithms and datasets requires substantial computational resources. Variability in hardware workload or non-homogeneous cluster nodes can make it challenging to establish a fixed computational budget or ensure that all methods are evaluated under identical conditions. Using fixed resources can help mitigate the ``hardware lottery'' effect~\cite{hooker2020hardwarelottery}, where an approach succeeds due to its compatibility with the available software and hardware rather than its inherent merit.

Every algorithm was implemented as a Docker image, meaning it can be built pragmatically and will always produce the same image ---specified by a installation script--- that can later be used in any operating system.

Unlike the original SRBench, which used the number of evaluations as the main budget criterion, wallclock hours were proposed as the budget criterion. The number of evaluations is difficult to standardize and does not allow for fair comparisons across different methods. By adopting wallclock hours as the primary budget control, preliminary e runs were conducted for all methods. Those with very short execution times had their hyperparameters related to search complexity increased (\textit{e.g.}, population size and number of generations).

Experiments were executed on a High-Performance Computing (HPC) Cluster. Each job was allocated $10$ GB of RAM, and GPUs were provided for methods that support them (see the \textit{runs on} column in Table \ref{tab:benchmarking:methods_and_details}). Jobs were not allowed to spawn subprocesses or utilize multiple cores. The time budget was set to $6$ hours for hyperparameter tuning and $1$ hour for training the algorithm with the optimal configuration, after which a \texttt{SIGALRM} signal was sent to terminate the process.
Energy consumption was estimated after hyperparameter tuning to avoid additional overhead, using the \textit{eco2AI} library~\cite{Budennyy2022}, which derives estimates from standard Linux commands monitoring CPU and memory usage. While not exact, this provides a rough yet informative profile of resource usage.

\subsection{Online Resources}~\label{sec:benchmarking:online_resources}

All datasets were submitted to PMLB and are available in the library and GitHub at \url{https://github.com/EpistasisLab/pmlb}. 
Experiment scripts and parameters were released to ensure transparency and reproducibility.
The SRBench repository was restructured into containerized environments for simplified standalone use and consistent execution, and all the images are publicly available on the author's DockerHub profile at \url{https://hub.docker.com/u/galdeia}.
The project is publicly available at \url{https://github.com/cavalab/srbench/tree/srbench_2025}.

\section{Results and discussion}~\label{sec:benchmarking:results}

The presentation of results and discussion is organized progressively, moving from more aggregated perspectives to increasingly granular analyses. The section begins with global aspects of energy consumption and execution time, before addressing algorithmic performance in the \textit{black-box} track. Subsequently, attention is directed to the phenomenological \& first-principles problems, where the ability of SR methods to approximate governing equations is examined.

\subsection{Energy consumption and execution time}

\begin{figure}[htb]
    \centering
    \caption[Median energy consumption (kWh) and training time for each algorithm.]{
        Median energy consumption (kWh) and training time for each algorithm. The color gradient, from red to blue, indicates the worst to best algorithm in each plot. Bar colors in the training time plot also range from red to blue based on rank, meaning that algorithms with different colors differ in ranking, but the $x$-axis are ordered by median energy consumption in both subplots. Note that the $y$-axis in the bottom plot increases in reverse.
    }
    \label{fig:power_consumption}
    \includegraphics[width=0.75\linewidth]{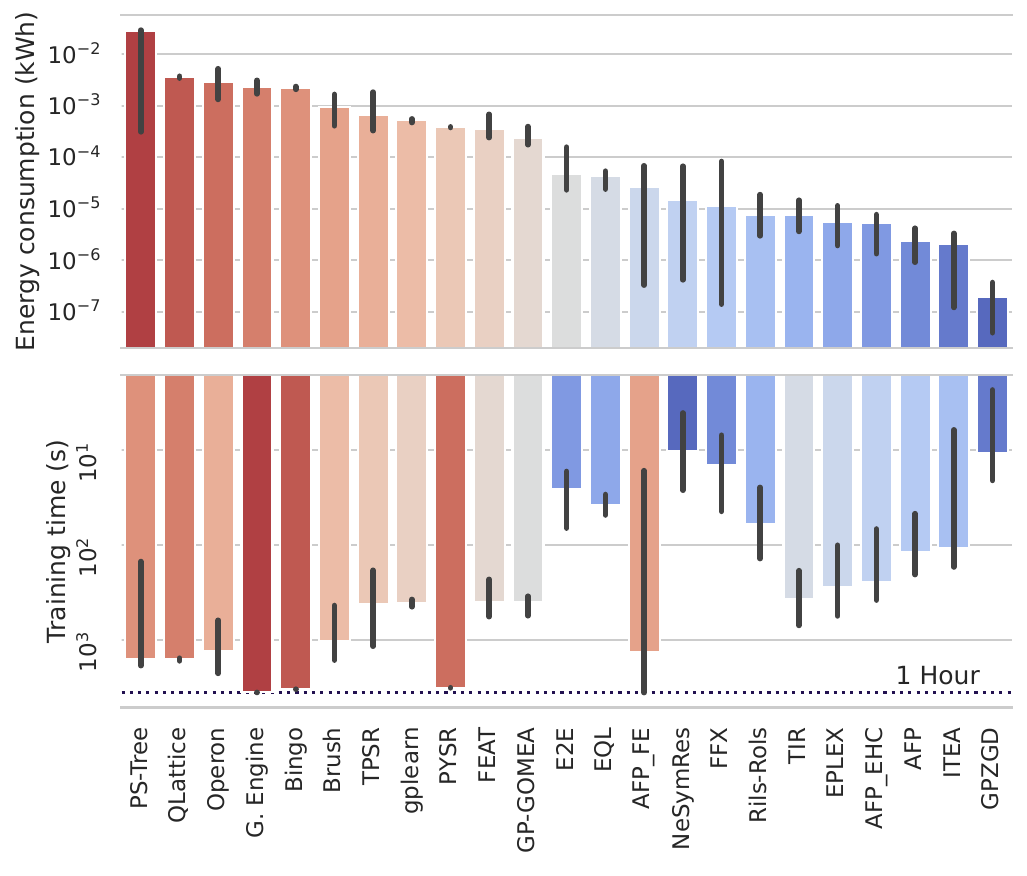}
    \legend{Source: Elaborated by the author.}
\end{figure}

Within the specified computational budget, the total runtime needed to run the experiments on a single core was estimated at $1$ year, $43$ weeks, and $5$ days.
Using eco2AI to monitor the final training phase, the energy and time consumption is reported in Figure \ref{fig:power_consumption} for each independent run, aggregating all datasets.

The entire benchmark was executed twice, as the first attempt presented technical issues related to saving certain results, requiring another run. Consequently, the eco2AI information was collected twice, producing a very robust estimate of energy consumption. This robustness was particularly relevant, since the heterogeneous environment of an HPC cluster and potential bias introduced by the job scheduler had been a concern prior to running the experiments. The consistent results confirmed that the energy consumption measurements are reliable.

Overall, energy consumption and execution time measurements were influenced not only by the choice of programming language and implementation details but also by the implementation aspects of the algorithms.
Since energy is the integral of power over time:

\begin{equation}
    E = \int P(t) \, dt \approx P_\mathrm{avg} \cdot t,
\end{equation}

\noindent with $P(t)$ representing instantaneous power, $P_\mathrm{avg}$ the average power for the task, and $t$ the time. The eco2AI measures energy $E$ using Linux profiling libraries, which in turn can give insights about resource usage between algorithms.
The results shows that some of the fastest algorithms demands more energy than others. This is a counterintuitive result, given that only a single core was provided. It indicates that these algorithms are faster but require higher CPU utilization, more aggressive computation, or greater memory resources to perform the calculations --- that is, their operations demand more power.

The top two most energy-hungry algorithms are Python-based, while the third, Operon, is implemented in \CPP{} but includes an inner hyperparameter tuning step. Some methods, such as FEAT and Brush, implement a \texttt{max\_stall} parameter, which halts execution if no improvement is observed in an inner validation partition after a number of iterations, but FEAT and Brush still shows increased training time than half of the methods. Although the stall feature does not necessarily correlate with better performance in terms of energy consumption, it could be further explored for more computationally demanding algorithms.

FFX performs comparably to the fastest algorithms, likely due to its deterministic nature. PS-Tree, the method that creates the largest expressions (see the original SRBench and the new results in the following subsection), is also among the slowest. AFP\_FE exhibits the largest confidence interval in training time.
GPZGD performs well on both evaluated criteria, achieving low energy consumption and short training time.

GPU-based algorithms such as TPSR, E2E, and NeSymRes do not exhibit large differences in energy consumption compared to CPU-based algorithms. In particular, TPSR, which uses traditional parameter optimization methods, demonstrated higher execution times than other GPU-based methods. uDSR was not compatible with the library during the experiments.
It should also be noted that some methods relying on pre-trained transformers (\textit{i.e.}, uDSR, TPSR, E2E) did not take into account training time and energy consumption associated with pre-training the transformer, which is known to be far more costly than inference. The pre-trained model was made available by the authors.

A high training time was observed for PySR, the only method implemented in Julia, due to the runtime compilation behavior of the language. Although this issue can be mitigated, it requires modifications to the Docker image.

These results suggest that energy-aware development should be taken into account when developing new SR algorithms, since optimizing runtime does not necessarily optimize energy consumption. In the era of large neural networks, models that are both fast and energy-efficient are highly promising alternatives.

This result also highlights another important observation: the number of evaluations is not the most suitable metric, since some of the fastest programs consumed large amounts of energy. This indicates that they are computationally intensive and perform more complex computations than others. The challenge then becomes finding a hyperparameter space in which all methods utilize nearly the entire available time while running on comparable fixed hardware.

\subsection{Performance analysis of the \textit{black-box} track}

In Figure \ref{fig:perfplot_max}, the performance plot is presented with AUC values based on the highest (\textit{i.e.} using the $\texttt{max}$ as aggregation function) empirically observed $R^2$ on the test set across $30$ runs. This the represents an optimistic perspective on model performance, under the assumption that a user performs the same number of runs.
The plot can be interpreted as the empirically observed likelihood of obtaining a maximum $R^2$ score for black-box problems under sufficient repetitions.
The probability on the $y$-axis, $\text{P}[R^2 \geq x]$, was estimated by calculating the proportion of runs in which the maximum $R^2$ exceeded a given threshold $x$.

\afterpage{%
\begin{landscape}
\begin{figure}[H]
    \centering
    \caption[Performance plots for the black-box track.]{
        Performance plots for the black-box track. The lines represent the probability of obtaining a given empirically observed $R^2$ value when the experiments are run multiple times (\textit{i.e.}, max aggregation).
        The methods were divided into intervals to better visualize all information. They were ordered by AUC and plotted into four figures, while methods not included in the legend are presented as gray lines for overall reference.
    }
    \label{fig:perfplot_max}
    \includegraphics[width=\linewidth]{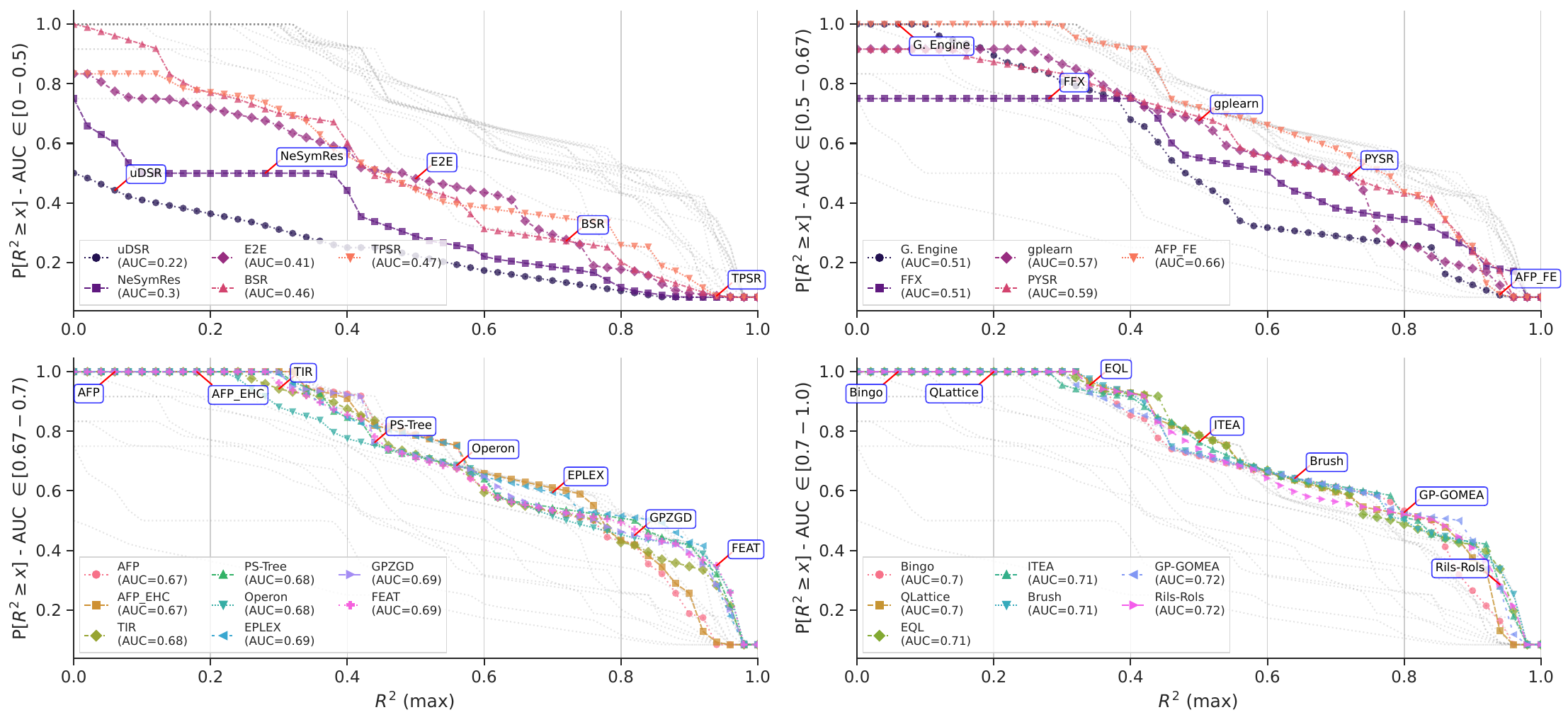}
    \legend{Source: Elaborated by the author.}
\end{figure}
\begin{figure}[H]
    \centering
    \caption[Cluster map of the Area Under the Curve (AUC) of Expected Performances across the $30$ independent runs is segregated by algorithm and dataset.]{
        Cluster map of the Area Under the Curve (AUC) of Expected Performances across the $30$ independent runs is segregated by algorithm and dataset. Higher values indicate better performance, while larger cells represent worse model size.
    }
    \label{fig:perfplot_clustermap_with_sizes}
    \includegraphics[width=0.9\linewidth]{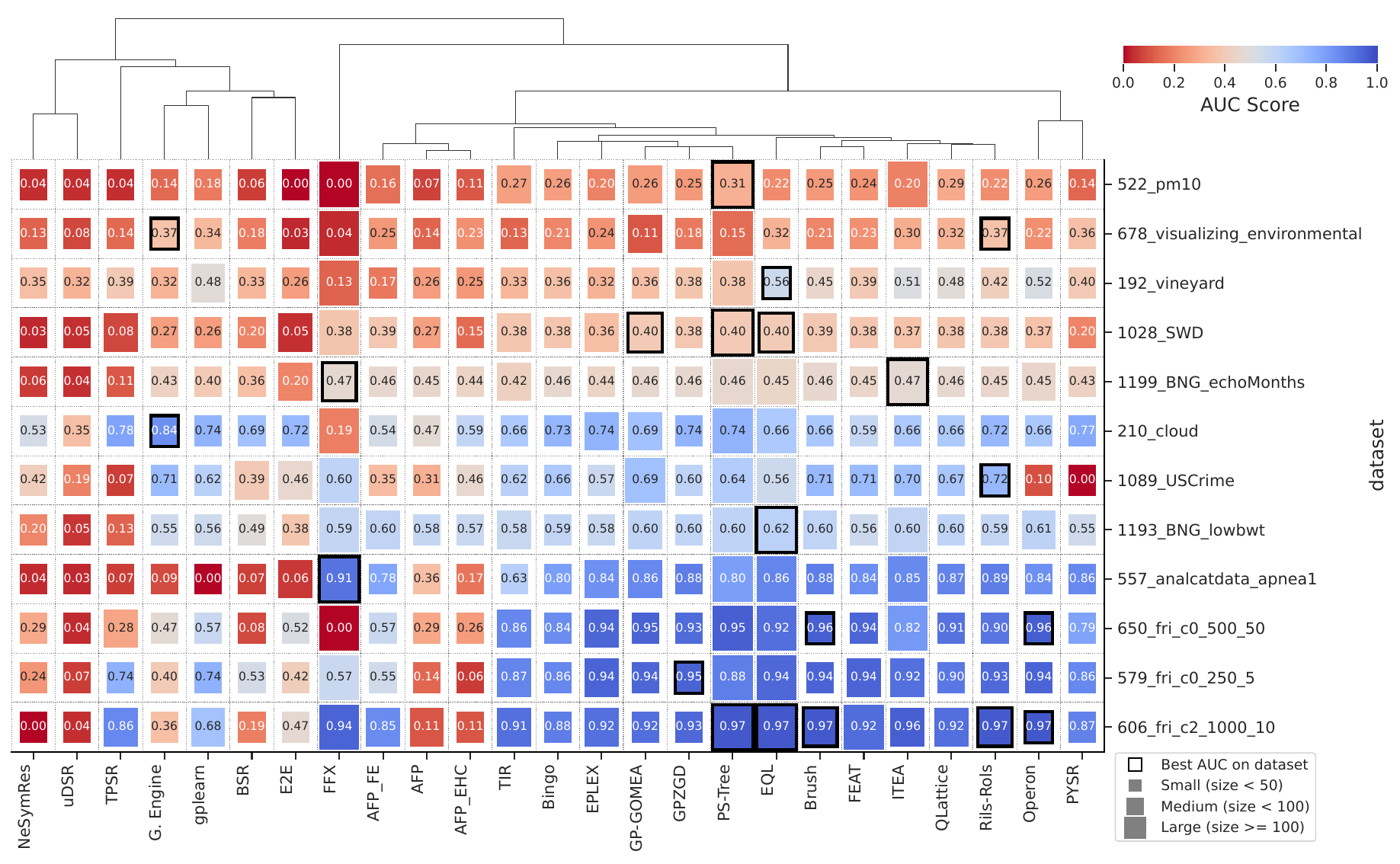}
    \legend{Source: Elaborated by the author.}
\end{figure}
\end{landscape}}

The performance curves in Figure \ref{fig:perfplot_max} display the expected probability of achieving maximum performance at different $R^2$ thresholds across multiple problems. Some methods (shown in the top-left subplot, with AUC below $0.5$) exhibited a probability below $1.0$ of achieving a strictly positive $R^2$, suggesting difficulties in generalization, failure to capture intrinsic data structures, or production of low-quality models.

Surprisingly, GPLearn performed well, despite being a canonical method. Interestingly, methods with performance below GPLearn are the neural network-based approaches, which suffered from downsampling ---since they lack scalability to more than a thousand samples for parameter optimization--- and incorporated different strategies for parameter optimization. For example, E2E attempted an end-to-end estimation with parameters included. This indicates that parameter optimization plays a major role, as TPSR, one of the best approaches, consolidates non-linear parameter optimization with transformers.

Regarding parameter optimization, GPLearn relied on ephemeral random constants. FFX is deterministic but included parameter optimization. Genetic Engine (G. Engine) also relied on random constants; it is based on strongly typed grammar, mainly focused on program synthesis, employing simple generation rules to handle numbers.
Above an AUC of $0.67$, all methods displayed very similar curves. The two bottom plots in Figure \ref{fig:perfplot_max} highlight methods that perform better in achieving $R^2 \geq 0.8$, as observed on the rightmost side of the plots.

Methods with performance superior than Genetic Engine performs parameter optimization at some stage.
Most top-performing algorithms incorporated constant optimization as a local search step—using linear, gradient-based, or non-linear methods—highlighting the importance of parameter optimization for achieving high performance. Among the methods with the highest AUC scores were GP-based algorithms such as Brush, GP-GOMEA, and GPZGD. Notably, GPZGD standardized the features using Z-scores and performed parameter optimization. Other methods, such as Rils-Rols, also performed constant optimization.

Figure \ref{fig:perfplot_clustermap_with_sizes} presents a cluster map of the AUC from the performance plot for each pair of dataset and algorithm on the test set. The size of each cell is proportional to the size of the best-performing final expression across the $30$ runs. Higher values and smaller cells indicate better performance. At the top of the plot, hierarchical clustering of the algorithms is shown, grouping methods based on performance and datasets based on their difficulty. The best algorithm for each dataset is highlighted with a black edge color on its cell. 

The clustering was performed using average linkage, meaning that clusters are formed by iteratively merging the two most similar algorithms, with similarity measured as the average distance between their AUC values across datasets. Clustering was applied to both algorithms and datasets, although the dendrogram for the datasets is omitted here. It is evident that the first seven methods form a cluster of underperforming algorithms, while the methods from TIR to RILS-ROLS form a cluster of stronger performers.

A closer inspection of Figure \ref{fig:perfplot_clustermap_with_sizes} provides some understanding of the limitations faced by existing algorithms.
Interestingly, some methods that performed poorly overall still achieved the best results on specific datasets. This highlights that no single algorithm dominates across all objectives; each presents distinct strengths depending on the problem characteristics.
The selection of datasets in Figure \ref{fig:perfplot_clustermap_with_sizes} revealed the ``\textit{no free lunch}'' aspect of the current state of SR: no algorithm performed exceptionally well on all datasets.
Likewise, no algorithm failed to solve all problems. Hierarchical clustering revealed two main clusters of good- and poor-performing algorithms, based on the AUC of the performance curves.

FFX and PS-Tree, methods resembling bagging or boosting heuristics, consistently produced overly large models, as denoted by their larger squares in the plot. Overall, most methods produced solutions below $50$ nodes, although some often generated expressions between $50$ and $100$.

Furthermore, the blue cells revealed a cluster of datasets in which the top-performing algorithms achieved similarly strong results. At the same time, the trade-off between performance AUC and model size was clearly observed. For example, in the \emph{557} dataset, FFX achieved the best AUC score, while Rils-Rols obtained a slightly lower AUC but with a smaller expression.

\subsection{Statistical comparisons}

An omnibus test was applied to determine whether overall differences existed among the groups. For this purpose, the Friedman rank-sum test was used, yielding a p-value of $1.07\times10^{-13}$ for $R^2$ and $5.29\times10^{-13}$ for model size. Since these p-values are several orders of magnitude smaller than the standard threshold for rejecting the null hypothesis (\textit{i.e.}, $0.05$), it can be safely assumed that the groups have different distributions.

\begin{figure}[htb]
    \centering
    \caption[Critical Differences diagrams comparing $25$ methods across datasets and random seeds.]{
        Critical Differences diagrams comparing $25$ methods across datasets and random seeds.
        The $x$-axis indicates the average ranks (maximum rank = $25$), with smaller values indicating better ranks.
        Horizontal crossbars connect methods with no statistically significant differences. Average ranks indicated between parenthesis.
    }
    \label{fig:benchmarking:cd_plots}
    \subfloat[$R^2$ test ranks. A higher rank means the algorithm achieves superior performance in $R^2$ across different datasets.\label{fig:benchmarking:cd_r2_test}]{%
        \includegraphics[width=\linewidth]{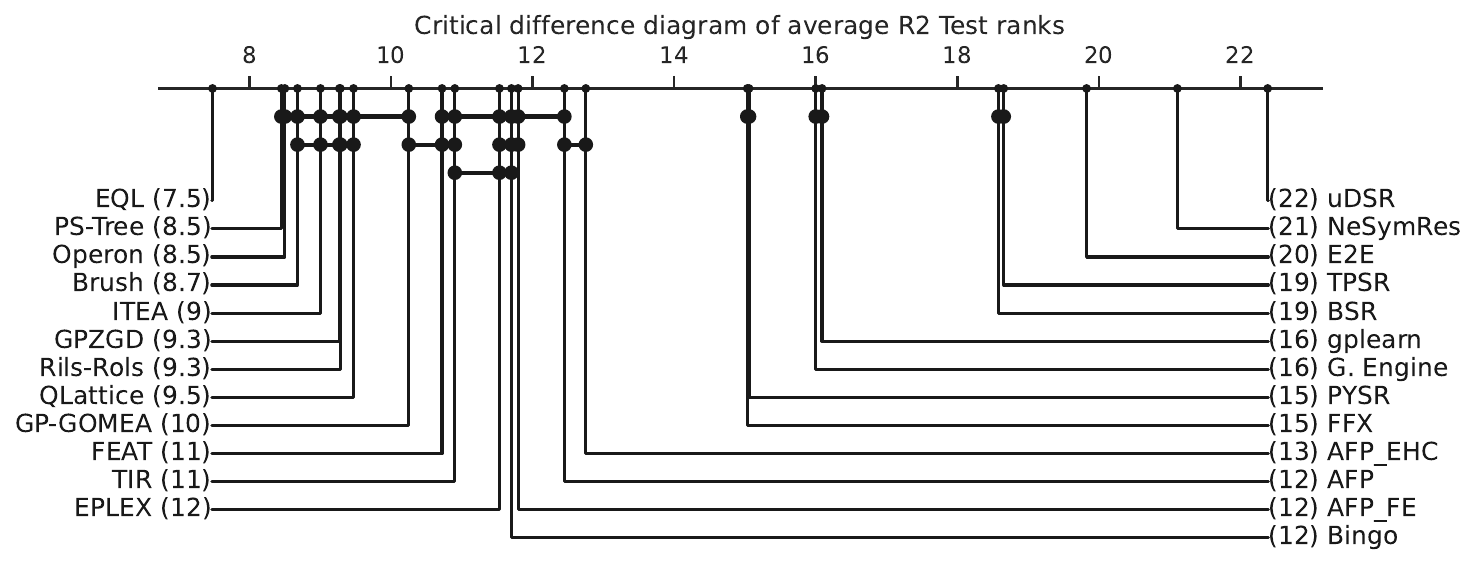}}\\[1em]
    \subfloat[Model size ranks. A higher rank means the algorithm produces smaller models across different datasets.\label{fig:benchmarking:cd_size}]{%
        \includegraphics[width=\linewidth]{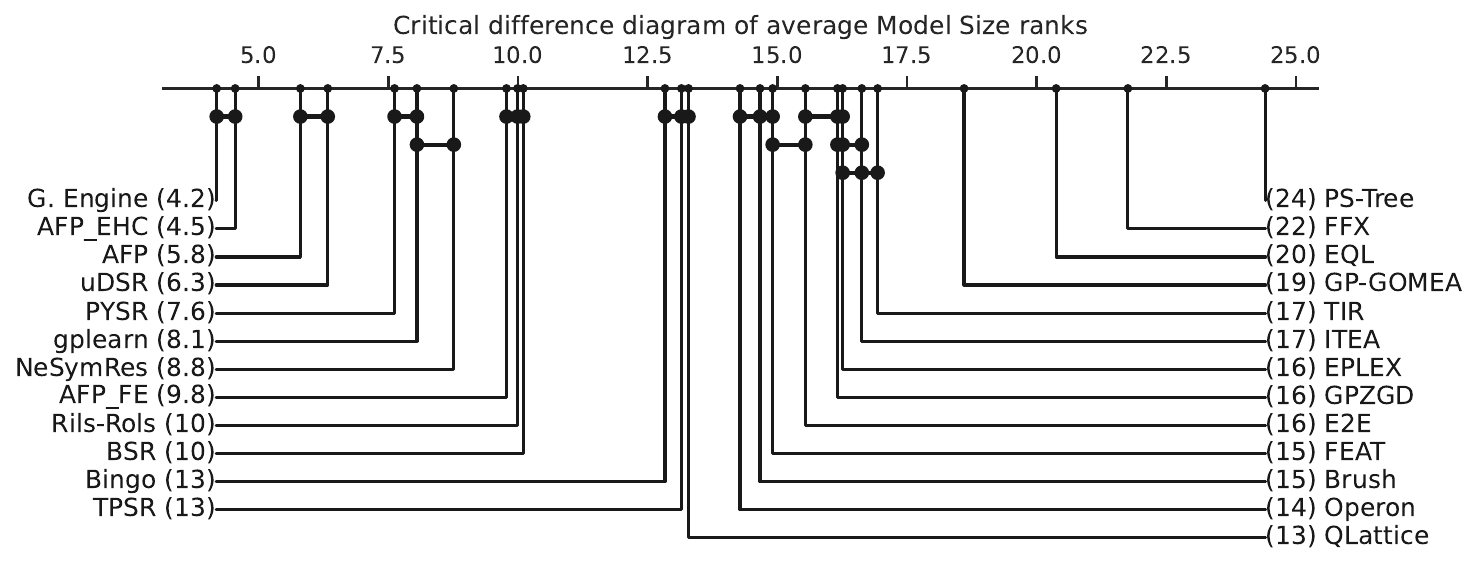}}
    \legend{Source: Elaborated by the author.}
\end{figure}

To assess the statistical significance of $R^2$ and model size performances, Critical Differences (CD) diagrams \cite{demvsar2006statistical} were employed to facilitate comparisons between methods.
Figure \ref{fig:benchmarking:cd_plots} presents the CD diagrams for the $R^2$ score on the test partition (Figure \ref{fig:benchmarking:cd_r2_test}) and for model sizes (Figure \ref{fig:benchmarking:cd_size}).

This diagram arranges algorithms based on their average ranks along the $x$-axis. Groups of algorithms that cannot be statistically distinguished from each other are connected by horizontal crossbars between them.
To generate the diagram, the methods are first ranked by pairing every independent run across all algorithms, ranking each algorithm within those specific combinations of dataset and random seed.
The CD diagram then calculates the minimum distance required for differences to be considered statistically significant, given the data's degrees of freedom (\textit{i.e.}, number of samples). Methods whose average ranks fall within this critical distance are linked together.
For example, in Figure \ref{fig:benchmarking:cd_size}, the leftmost crossbar indicates that Genetic Engine and AFP\_EHC cannot be statistically differentiated. However, AFP and uDSR are significantly lower ranked than both Genetic Engine and AFP\_EHC.

When analyzing the first CD diagram (Figure \ref{fig:benchmarking:cd_r2_test}), EQL achieves the best average $R^2$ test rank, followed by a large cluster of well-performing algorithms ranging from PS-Tree to GP-GOMEA. Algorithms ranked 15th and above exhibit poor average performance.
Regarding model size, the methods appear less concentrated and more dispersed along the axis, with fewer non-significant differences among the top-performing methods. Genetic Engine, which ranked 16th in $R^2$ test performance, achieves an average rank of $4.2$ in model size. The best-performing methods in terms of $R^2$ ---EQL (with an average rank of $7.5$) and PS-Tree (with an average rank of $8.5$)--- are now ranked 20th and 24th in model size, respectively, indicating that they incur a high complexity cost to achieve their observed superior $R^2$ performance.
The only algorithm that ranks among the top 10 in both $R^2$ and model size is Rils-Rols, which is placed 7th in $R^2$ (with an average rank of $9.3$) and 9th in model size (with an average rank of $10$). All other methods fail to demonstrate a balance between $R^2$ and size, achieving either low $R^2$ scores and high size ranks, or low size ranks and high $R^2$ scores.

\subsection{Towards a parameter-less experience}

Evolutionary Algorithms already computationally expensive, as they provide little to no parallelization options and operate as population-based heuristics.
Some hyperparameters have historically been present in these algorithms, such as population size or number of generations. 
The setting of hyperparameters in SR algorithms often requires careful experimentation with multiple configurations that may affect the final result.
For general users, this creates a significant barrier: default values must either be relied upon, or extensive experimentation must be conducted to identify the best configurations.

The current methods showed to be difficult to tune, and the papers proposing these algorithms often fail to conduct sensitivity analyses or evaluate the robustness and interaction among parameters, leaving this burden to users.
During the preparation for running the benchmark, it was observed that, when attempts were made to provide improved configurations, the default parameters often produced better results.
Hyperparameters appear to exert a large influence, yet they remain difficult to tune.

Fair comparisons were attempted by increasing the hyperparameters so that all methods had a similar runtime. However, some methods did not benefit significantly from the adjustments, and parameter tuning was intentionally kept simple to avoid deviating from the original purpose of benchmarking. Simple parameter tuning also matches with most users would do when using different SR algorithms.
Although performance improved in some cases, many methods did not utilize the maximum allowed time, suggesting that additional improvements may be possible when wallclock time alone is considered as the budget, regardless of consumption.

The field should move toward a parameter-less experience, with algorithms designed to require as few hyperparameters as possible, unless they are intuitive and domain-related.
Good practices involve the careful selection of default parameters and, when possible, the integration of internal hyperparameter tuning to make it transparent to the user (as in Operon with Optuna) or the adaptation of parameters (as in PySR, which includes strong default values but still exposes more than one hundred parameters).

For future iterations of SRBench, hyperparameter tuning should be standardized by requiring each method to define a small set (\textit{e.g.}, four) of configurations.
For user experience, the number of tunable parameters should be minimized.
A simple grid search could then be applied across these configurations, each limited to a maximum runtime (\textit{e.g.}, one hour).
The off-the-shelf configuration should also be automatically included in the experiment scripts, as some methods were observed to perform best with their default settings.

\subsection{Solving empirical and theoretical problems}

The final equations closest to the governing models for the phenomenological \& first-principles track are shown in Table \ref{tab:benchmarking:Pareto_first_principles_equations}, along with their corresponding $R^2$ scores and sizes. 
This track provides a detailed view of the trade-off between model size and accuracy, with established hypothesis models used as reference points.
The Pareto fronts and the correct ground-truth are depicted in Figure \ref{fig:Pareto_first_principles}.

\begin{table}[h]
    \caption[Equations from the Pareto front closest to the governing models for the phenomenological \& first principles track.]{Equations from the Pareto front closest to the governing models for the phenomenological \& first principles track. These equations correspond to the best solutions across all runs and all algorithms.}
    \label{tab:benchmarking:Pareto_first_principles_equations}
    \footnotesize
    \centering
    \begin{tabular}{@{}rrrl@{}}
        \toprule\midrule
        \textbf{Dataset} &  $R^2$ & \textbf{Size} & \textbf{Symbolic model} \\ \midrule
        absorption & $1.00$ & $6$ & $0.24 + 1.76\tanh{x}$ \\ \midrule
        bode & $1.00$ & $8$ & $0.34 \, e^{1.51\, n} - 0.87$ \\ \midrule
        hubble & $0.86$ & $3$ & $0.090+D$ \\ \midrule
        ideal\_gas & $0.99$ & $14$ & $0.69\, \log{n + 1.64} - 0.89 + 0.48\, e^{-V}$ \\ \midrule
        kepler & $1.00$ & $10$ & $1.98*\tanh{(0.66\, a - 0.56)} + 0.78$ \\ \midrule
        leavitt &$ 0.97$ & $3$ & $-0.94 \, \log{P}$ \\ \midrule
        newton & $0.90$ & $16$ & $0.34\, \frac{m_1}{0.83\, m_2 - \frac{0.09}{m_1}} + 1.13$ \\ \midrule
        planck & $1.00$ & $24$ & $0.023\, \log{(\sqrt{nu + 0.38} - 0.04)} - 0.31 \, \frac{nu + 0.3}{T + 0.94} + 0.41$ \\ \midrule
        rydberg & $1.00$ & $21$ & $1.01\, e^{0.1 \, e^{1.73 \, n_1 -1.31 \, n_2}} - e^{-0.69 \, n_1}$ \\ \midrule
        schechter & $1.00$ & $13$ & $0.748 - 0.274 \, (\log{(163.992 \, L + 106.737)} + 1.414 \, L)$ \\ \midrule
        supernovae\_zr & $1.00$ & $29$ & $(\sin{(0.2 \, x \, e^{-x})} - 0.04) \, (x + 0.72 \, \sin{(5.46 \, x + 0.76)} - 0.16) - \sin{(x + 0.18)}$ \\ \midrule
        tully\_fisher & $0.93$ & $3$ & $-0.93\, \Delta V$ \\ \midrule\bottomrule
    \end{tabular}
    \legend{Source: Elaborated by the author.}
\end{table}

Only ITEA successfully retrieved one of the first-principles equations (Leavitt). In other cases, SR methods produced either a more complex but more accurate expression or a simpler but less precise model. 
Although the correct ground-truth was not found, likely due to high noise, expressions with $R^2$ equal to or close to one were obtained. In $7$ out of $12$ problems, small expressions (fewer than $29$ nodes) with perfect scores were found.

This outcome indicates overfitting to noise but also demonstrates the strong capacity of SR to minimize error. It suggests that minimizing error alone while leaving size to be managed by the evolutionary process may not be optimal. Multi-objective scenarios with effective heuristics for selecting the final model from the Pareto front, balancing predictive accuracy and size, appear to be more suitable.

The discouragement of nested operations or hard-to-interpret functions becomes evident when these expressions are examined, as several complex functions such as $\log$, $\tanh$, and $\sin$ are incorporated. The chaining in the smallest solution found for the Rydberg equation illustrates this. 

This challenge is further illustrated in Figure \ref{fig:Pareto_first_principles}, where the Pareto front on the test set of the most accurate solutions was displayed, and SR algorithms were compared against the currently accepted hypothesis (\textit{star}).

\afterpage{%
\begin{landscape}
\begin{figure}[H]
    \centering
    \caption[Pareto plots for the phenomenological \& first-principles track, with model sizes on the $y$-axis and $R^2$ on the $x$-axis.]{
        Pareto plots for the phenomenological \& first-principles track, with model sizes on the $y$-axis and $R^2$ on the $x$-axis. The \textit{star} marker denotes the performance of the ground-truth expression, also reported in the box inside each subplot. Only the first front is plotted for clarity, meaning that only non-dominated algorithms are shown. To select one model for each algorithm, the smallest model among the $30$ runs was chosen, and Pareto fronts were then generated by excluding dominated solutions. The remaining methods represent a good trade-off between model size and predictive complexity.
        }
    \label{fig:Pareto_first_principles}
    \includegraphics[width=\linewidth]{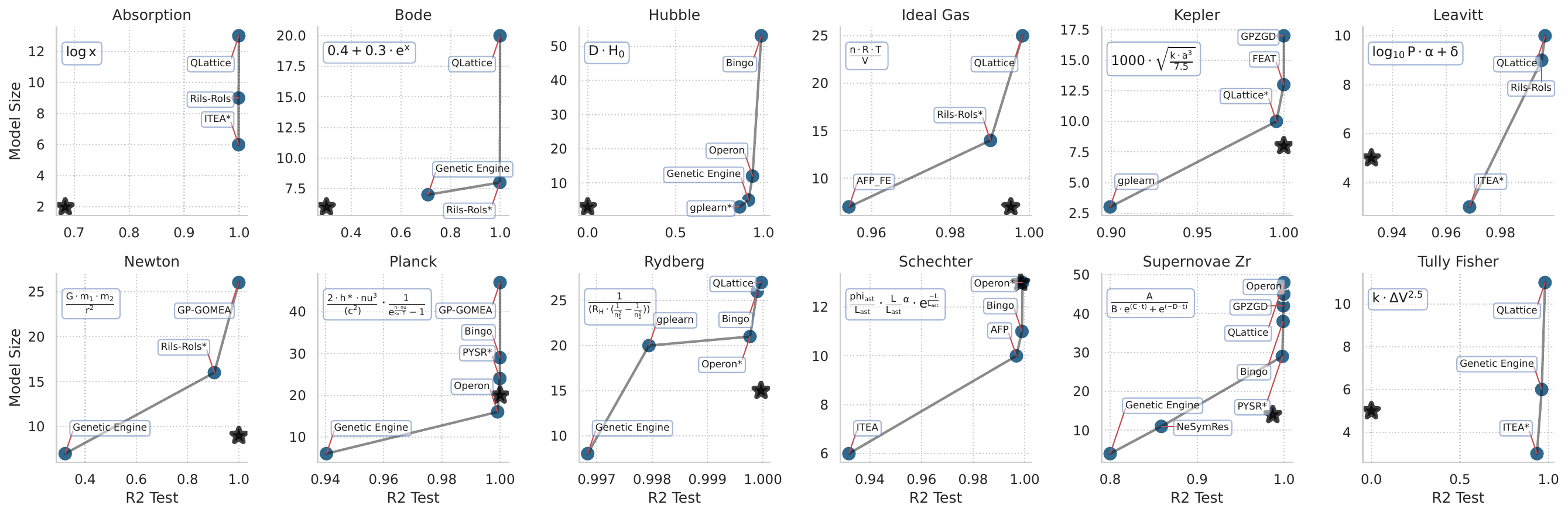}
    \legend{Source: Elaborated by the author.}
\end{figure}
\end{landscape}}

As shown in Figure \ref{fig:Pareto_first_principles}, for almost every problem, no SR method was able to discover an expression that strictly dominated the original equation. Larger and more accurate expressions or smaller and less accurate ones were typically generated instead.
As an example, for the \emph{Schechter} dataset, reasonable alternatives were found by Operon without significantly deviating from the complexity-accuracy balance of the original equation.
The Genetic Engine, which performed poorly on the black-box track, appeared in almost all subplots here, as it was able to identify simpler models.

It is observed that noise, whether inherent in real-world data or synthetically introduced in generated datasets, can cause ground-truth hypotheses to fall short of achieving a perfect $R^2$.
Asterisks next to algorithm names indicate those closest (based on Euclidean distance over the normalized axes) to the ground-truth.

It is also observed that some first-principles equations failed to achieve perfect $R^2$ due to noise in the original data—notably in the Hubble and Tully-Fisher datasets.
In such cases, the attainment of higher $R^2$ with a more complex model often indicated that noise was being fitted rather than the underlying relationship. This can be seen in the equations found for Hubble and Tully-Fisher, which deviated from the ground-truth by only a small edit distance.

It is also notable that the ground-truth equations generally required fewer than $20$ nodes, raising the question of whether the SR search space should be vast and encompass numerous methods, or restricted to smaller or medium-sized expressions.

By contrast, the Absorption and Bode datasets represented phenomenological problems for which no ground-truth equations are known, and SR successfully found models performing better than the current hypothesis, although domain experts are required to evaluate the models to confirm it's relationships with theoretical basis.

\subsection{Current difficulties and limitations}

During the benchmarking process, several bugs were fixed in most evaluated algorithms.
Only a few corresponding repositories are actively maintained, requiring joint decisions on which algorithms should be deprecated and excluded from further benchmarking.

The establishment of deprecation rules is proposed to maximize benchmarking efforts. A simple criterion could be applied: if a repository is no longer actively maintained and the algorithm does not perform well in the black-box datasets, it should be deprecated.
Such a guideline may also encourage the development of more robust algorithms capable of consistent performance across diverse problems.

Another important issue is the fair comparison of GPU-based and CPU-based approaches, as well as single-core and multi-core implementations. Although every algorithm that supports a \texttt{max\_time} hyperparameter respects its limit, some terminate much earlier (\textit{e.g.}, NeSymRes), potentially biasing performance evaluations. A more precise implementation of time management would allow computational efficiency to be compared more fairly across different methods, but this requires author's engagement with the benchmark.

Benchmarking could also yield more data-driven insights for algorithm development if appropriate metadata were incorporated (\textit{e.g.}, a set of keywords), directly linking methodologies to results.
Similarly, algorithms could generate post-run metadata (\textit{e.g.}, internal grid search results, number of local search iterations) as part of their output. Such information would facilitate a deeper understanding of algorithmic behavior, enable more precise fine-tuning, and ultimately support more informed improvements in SR techniques.

\section{Conclusions}~\label{sec:benchmarking:conclusions}

This chapter updated SRBench with the aim of improving the benchmark of symbolic regression as a community-accepted standard. The number of SR algorithms was nearly doubled compared to the previous version, and alternative, more detailed visualizations were introduced to depict the current state of SR.

The main contributions included the proposal of novel datasets, which were successfully integrated into the official PMLB via pull request on GitHub; new methods of visualization and the execution of additional runs to provide more robust statistical evidence. The overrepresentation of the Friedman datasets was also alleviated by restricting their share to at most $25\%$ of the benchmark.

It was shown that having a parameter optimization step is a common strategy among the best-performing algorithms, as well as GP-based implementations, although some employ scaling and others rely on iterative searches of the function structure without a population-based heuristic.

Performance analyses, along with difficulties encountered during the benchmark related to parameter tuning and computational budget, also highlighted the importance of moving toward a parameterless experience, and ideally, also toward a energy-aware development. 

It was further discussed that SR methods are highly effective at improving model accuracy, but this is often achieved at the cost of overfitting to noise, producing very small yet accurate models that fail to capture the governing laws of empirical data. This demonstrates that progress is still required to achieve proper scientific discovery.

To ensure continuous improvement of SRBench, active participation from the community is required to maintain compatibility with ongoing experiments and ensure correctly functioning implementations. Contributions from researchers—through discussions, new ideas, constructive criticism, and proposed corrections—are crucial. A continuously evolving benchmark will help drive progress in SR.
The achievement of a mature state for the field should be a shared priority, especially given the growing emphasis on transparency in machine learning and the increasing interest in scientific ML.





%% file: Capitulos/Benchmarking/floats/algorithm_descriptions.tex
\begin{tabular}{@{}rp{9cm}@{}}
\toprule\midrule
\textbf{Algorithm} & \textbf{Description} \\ \midrule
    
AFP~\cite{Schmidt2010} & Age-fitness Pareto (AFP) optimization, meaning model age is used as an objective, with constants randomly changed \\ \midrule

AFP\_FE & AFP with co-evolved fitness estimates \\ \midrule

AFP\_EHC~\cite{la2016inference} & AFP with epigenetic hill climbing for constants optimization as local search  \\ \midrule

Bingo~\cite{randall2022bingo} & Evolves acyclic graphs with non-linear optimization, using islands for managing parallel populations\\ \midrule

Brush~\cite{brush_multi_armed_bandits} & GP with multi-armed bandits for controlling search space exploration\\ \midrule

BSR~\cite{jin2020bayesiansymbolicregression} & Bayesian model with priors for operators and parameters is used to sample expression trees \\ \midrule

E2E~\cite{kamienny_end--end_nodate} & Generator using pre-trained transformers, using BFGS and subsampling for tuning parameters\\ \midrule

EPLEX~\cite{eplex} & GP with $\epsilon$-lexicase parent selection \\ \midrule

EQL~\cite{pmlrv80sahoo18a} & Shallow neural network using mathematical operators as activation functions, and performs a pruning to refine the network to an expression \\ \midrule

FEAT~\cite{cava2018learning} & GP algorithm with $\epsilon$-lexicase selection and linear combination of expressions using L1-OLS\\ \midrule

FFX~\cite{McConaghy2011} & Non-evolutionary, deterministic approach, that generates a set of base functions and fits a regularized OLS to combine them\\ \midrule

Genetic Engine~\cite{espada2022data} & GP using Context-Free Grammars to guide the generation process in an efficient manner \\ \midrule

GPGomea~\cite{10.1162/evco_a_00278}  & GP with linkage learning used to propagate patterns and avoid their disruption\\ \midrule

GPlearn~\cite{Stephens2025trevorstephens} & Canonical GP implementation \\ \midrule

GPZGD~\cite{10.1145/3377930.3390237}  & GP with Z-score standardization and stochastic gradient descent for parameter optimization \\ \midrule

ITEA~\cite{10.1162/evco_a_00285}  & Mutation-based algorithm with constrained representation and OLS parameter optimization\\ \midrule

NeSymRes~\cite{pmlr-v139-biggio21a} & Pre-trained encoder-decoders generate equation skeletons, optimized with non-linear optimization\\ \midrule

Operon~\cite{Kommenda2019} & GP algorithm with weighted terminals and non-linear OLS parameter optimization \\ \midrule

Ps-Tree~\cite{ZHANG2022101061} & GP algorithm that evolves Piecewise trees with SR expressions as leaves\\ \midrule

PySR~\cite{cranmer2023interpretablemachinelearningscience} & Evolve-simplify-optimize loop with islands to manage parallel populations \\ \midrule

Qlattice~\cite{broløs2021approachsymbolicregressionusing} & Uses a learned probability distribution updated over iterations to sample expressions, with parameter optimization. Closed source software \\ \midrule

Rils-rols~\cite{Kartelj2023} & Iterative generation of perturbations and parameter optimization with OLS and local search selection of next candidates\\ \midrule

TIR~\cite{tir} & GP with crossover and mutation that uses a constrained representation capable of tuning non-linear parameters with OLS \\ \midrule

TPSR~\cite{shojaee_transformer-based_nodate} & Monte-Carlo Tree Search planning with non-linear optimization wrapper for generative models E2E and NeSymRes\\ \midrule

uDSR~\cite{UnifiedFrameworkForDeepSR} & Unification of pre-trained transformers, GP, and linear models, into a framework that decomposes the problem \\ 
\midrule\bottomrule
\end{tabular}

%% file: Capitulos/Benchmarking/floats/algorithm_details.tex
\begin{tabular}{@{}rccccc@{}}
\toprule\midrule
\textbf{Algorithm} & \makecell{\textbf{Const.}\\\textbf{Opt.}} & \makecell{\textbf{Time}\\\textbf{limit}} & \makecell{\textbf{Multiple}\\\textbf{solutions}} & \makecell{\textbf{Runs}\\\textbf{on}} & \textbf{Language} \\ \midrule
    
AFP~\cite{Schmidt2010} & \xmark & \checkmark & \xmark & CPU & \CPP{} \\ \midrule
AFP\_FE & \xmark & \checkmark & \xmark & CPU & \CPP{} \\ \midrule
AFP\_EHC~\cite{la2016inference} & \checkmark & \checkmark & \xmark & CPU & \CPP{} \\ \midrule
Bingo~\cite{randall2022bingo} & \checkmark & \checkmark & PF & CPU & Python \\ \midrule
Brush~\cite{brush_multi_armed_bandits} & \checkmark & \checkmark & PF & CPU & \CPP{} \\ \midrule
BSR~\cite{jin2020bayesiansymbolicregression} & \xmark & \checkmark & \xmark & CPU & Python \\ \midrule
E2E~\cite{kamienny_end--end_nodate} & \xmark & \xmark & \xmark & GPU & Python \\ \midrule
EPLEX~\cite{eplex} & \xmark & \checkmark & \xmark & CPU & \CPP{} \\ \midrule
EQL~\cite{pmlrv80sahoo18a} & \xmark & \xmark & \xmark & CPU & Python \\ \midrule
FEAT~\cite{cava2018learning} & \checkmark & \checkmark & PF & CPU & \CPP{} \\ \midrule
FFX~\cite{McConaghy2011} & \checkmark & \xmark & \xmark & CPU & Python \\ \midrule
Genetic Engine~\cite{espada2022data} & \xmark & \checkmark & \checkmark & CPU & Python \\ \midrule
GPGomea~\cite{10.1162/evco_a_00278} & \checkmark & \xmark & \checkmark & CPU & \CPP{} \\ \midrule
GPlearn~\cite{Stephens2025trevorstephens} & \xmark & \xmark & \checkmark & CPU & Python \\ \midrule
GPZGD~\cite{10.1145/3377930.3390237} & \checkmark & \checkmark & \xmark & CPU & C \\ \midrule
ITEA~\cite{10.1162/evco_a_00285} & \checkmark & \xmark & \xmark & CPU & Haskell \\ \midrule
NeSymRes~\cite{pmlr-v139-biggio21a} & \checkmark & \checkmark & \xmark & GPU & Python \\ \midrule
Operon~\cite{Kommenda2019} & \checkmark & \checkmark & PF & CPU & \CPP{} \\ \midrule
Ps-Tree~\cite{ZHANG2022101061} & \xmark & \xmark & \xmark & CPU & Python \\ \midrule
PySR~\cite{cranmer2023interpretablemachinelearningscience} & \checkmark & \checkmark & \xmark & CPU & Julia \\ \midrule
Qlattice~\cite{broløs2021approachsymbolicregressionusing} & \checkmark & \checkmark & \xmark & CPU & Python \\ \midrule
Rils-rols~\cite{Kartelj2023} & \checkmark & \checkmark & \xmark & CPU & \CPP{} \\ \midrule
TIR~\cite{tir} & \checkmark & \checkmark & PF & CPU & Haskell \\ \midrule
TPSR~\cite{shojaee_transformer-based_nodate} & \checkmark & \checkmark & \xmark & GPU & Python \\ \midrule
uDSR~\cite{UnifiedFrameworkForDeepSR} & \checkmark & \xmark & \xmark & GPU & Python \\ 
\midrule\bottomrule
\end{tabular}

%% file: Capitulos/Conclusions/conclusions.tex
\chapter{Closing Remarks}~\label{chap:Conclusions}

The doctoral research was specialized in Symbolic Regression (SR) and its applications, with a primary focus on investigating and optimizing subroutines common across several SR methods. This positioned the work within the broader context of algorithmic advancements in SR.

The research was conducted between February $2022$ and November $2025$, spanning $46$ months, including six months abroad at the Computational Health Informatics Program (CHIP) at Boston Children's Hospital. After the time at CHIP, regular participation in research meetings and collaborative activities was maintained, better detailed in this chapter.

Throughout the doctorate, involvement with the international SR community was gradually established as a consequence of efforts to publish contributions. The student participated in two editions of the Genetic and Evolutionary Computation Conference (GECCO), where papers were presented orally and feedback was received from the audience. These exchanges progressively clarified community needs and helped define subsequent research directions in the broader sense of SR.

That being said, the thesis initially aimed at improving SR algorithms in general, and its trajectory was shaped by the practical needs of collaborators applying SR and by insights gained from discussions at conferences. 

Several directions were also explored but not pursued further. Examples include the adaptive use of variation operators through multi-armed bandits, inspired by their previous applications in \citeonline{dacosta_adaptive_2008,thanh_multi-armed_2021,belluz_operator_2015}; simplification strategies based on commutative chains, related to \citeonline{burlacu2019hash}; and deterministic heuristics for constructing expressions without stochastic variation operators, following ideas from \citeonline{McConaghy2011,ExhaustiveSR}. While relevant to improvements in SR, these ideas did not yield results sufficiently conclusive are therefore only briefly mentioned.

By contrast, three directions produced successful improvements and were included as contributions: parameter optimization, parent selection, and model simplification. These became the core of the thesis and were published at GECCO.
Benchmarking SR algorithms also emerged as a significant contribution, both as a means to evaluate the proposed improvements and as a tool to promote collaboration within the community. 

The relevance of the thesis lies in its focus on open problems affecting multiple SR methods. Although studied in specific algorithmic contexts, the conclusions and lessons learned were shown to generalize more broadly. A unified framework was also developed as part of this effort, incorporating many of the published contributions.

This chapter closes the thesis by restating current challenges and objectives, summarizing the main contributions, and presenting the key takeaways, limitations, and future perspectives. 

\section{Recapitulation}

This thesis addresses challenges common to SR algorithms. It begins with a review of the necessary background and then presents contributions that target specific problems in SR.
Instead of proposing an entirely new SR method, emphasis was placed on identifying and improving critical subroutines that influence predictive accuracy and model complexity. 
The contributions focused on: \textit{(i)} parameter optimization strategies that combine linear and non-linear methods, \textit{(ii)} improving adaptive threshold in $\epsilon$-lexicase parent selection method, \textit{(iii)} fast symbolic simplification through inexact estimation of equivalences, and \textit{(iv)} a systematic benchmarking of several SR methods. The main objective was therefore to improve SR algorithms by addressing different challenges, then assessing the SR landscape through robust benchmarking.

For parameter optimization, modern frameworks typically combine structural search with dedicated optimization steps, often relying on non-linear methods, which have proven effective \cite{Kommenda2019}, and implemented on high-performing SR methods such as Operon \cite{operon}. However, different high-performing SR methods use linear optimization instead, such as FEAT \cite{cava2018learning} and ITEA \cite{10.1162/evco_a_00285}. The challenge lies in effectively combining linear and non-linear methods, as their interaction remains poorly understood.

For parent selection, $\epsilon$-lexicase has demonstrated state-of-the-art performance in program synthesis and symbolic regression \cite{10.1145/2330784.2330846,la_cava_epsilon-lexicase_2016}. However, its dynamics remain discussed, and open questions persist regarding the impact of different thresholding strategies and dynamic variants.

For simplification, effective implementation ---such as through rewrite rules \cite{kulunchakov_creation_2017}--- is burdened by the need for extensive algebraic rule sets. Heuristic or restricted approaches also have clear limitations, for example replacing subtrees with randomly generated ones and checking for unchanged predictions \cite{nguyen_semantic_2020}. A simple, efficient, and fast simplification method would be valuable, particularly given that simplification has been shown to improve generalizability \cite{helmuth_improving_2017}.

Altogether, these challenges also highlight the importance of systematic benchmarking. Benchmarking is not only necessary to evaluate improvements, but also to identify which techniques are currently effective under standardized conditions. The last large-scale comprehensive benchmark for SR was published by \citeonline{srbench}, named SRBench. That study has since been revisited by other researchers, who identified gaps and opportunities for further improvements \cite{dick2022genetic,tir,matsubara2024rethinking}. This created a timely opportunity to contribute by reassessing the current landscape of symbolic regression with updated protocols and recently proposed algorithms.

The objectives of the thesis, as defined in the introduction, were the following:

\begin{description}
    \item[Parameter optimization.] To investigate whether separating the optimization of linear and non-linear parameters improves predictive accuracy and model size, and to assess the trade-off in runtime when applying iterative non-linear optimization and closed-form linear optimization.
    \item[Parent selection.] To use a more intuitive criteria and evaluate the effect of new variants of $\epsilon$-lexicase parent selection on the convergence of symbolic regression algorithms, and to determine in which problem domains the new criteria enhance overall performance.
    \item[Simplification.] To develop and assess a new method for automatically simplifying symbolic regression expressions, reducing unnecessary complexity while preserving predictive accuracy, and to explore whether approximate simplification strategies can reduce implementation burden without harming performance.
    \item[Benchmarking.] To perform a comprehensive benchmarking study of contemporary symbolic regression algorithms, establishing standardized evaluation criteria and clarifying the broader landscape of SR research. 
\end{description}

Each of these objectives is associated with a dedicated chapter. In each case, related work was reviewed, a novel approach was proposed, and extensive empirical experiments were conducted to demonstrate that the improvements effectively enhance algorithmic performance.

\section{Limitations, takeaway messages, and future works}

This section recapitulates the conclusions, discusses their limitations, and presents the main takeaways for each contribution. Each conclusion is related to the corresponding objective, and its achievement is indicated. Future work for each contribution is also identified.

\subsection{Parameter optimization}

The inclusion of non-linear parameters in Interaction-Transformation representation \cite{SymTree} expressions was proposed and included in the ITEA algorithm \cite{10.1162/evco_a_00285}, and four heuristics for combining linear and non-linear optimization were evaluated. Levenberg-Marquardt (LM) \cite{levenberg1944method,marquardt1963algorithm} was applied for non-linear optimization, and Ordinary Least Squares (OLS) \cite{doi:10.1080/00031305.1989.10475644} for linear optimization. The LM+OLS combination achieved the best predictive performance, though at the cost of larger expressions. Restricting interaction variables improved interpretability without reducing accuracy. 
The objective of separating the optimization of linear and non-linear parameters was achieved. Predictive accuracy improved with the LM+OLS combination, and trade-offs in expression size and convergence were visualized.
These findings may generalize to SR methods that rely on linear optimization over non-linear terms.

Several limitations were identified. LM is sensitive to starting points, making initialization crucial. Initialization was explored using ERCs, but more informed strategies could improve performance. Only one algorithm was studied, even though other SR algorithms share many similarities in representation form, but generalization across these algorithms was not empirically tested. Finally, the benchmark datasets were of small dimensionality, limiting generalization of insights to higher-dimensional settings.

Key takeaways are the following. Non-linear optimization alone suffers from overparameterization and high computational cost, resulting in longer runtimes. OLS should always be used whenever linear decomposition is available, as it provides a closed-form solution. For mixed cases, local non-linear optimization (LM) should be combined with global linear optimization (OLS), applying LM to isolated parts of the tree. ERCs should be always avoided if possible, as they reduce flexibility and interpretability. Restricting the number of variables in interactions improves interpretability at no cost of predictive performance for the given representation.

Future work should extend the study with larger datasets to assess the cost-benefit of non-linear optimization of inner parameters. Certain transformations, such as the identity, gain no benefit from the addition of non-linear parameters, while others (\textit{e.g.}, exponential, logarithm) admit only partial optimization; accounting for this may further improve search efficiency. Other optimization methods with regularization, such as OLS with L1 penalty, could reduce expression size by producing sparse solutions, alleviating the downside of ITEA with LM+OLS. Improved initialization strategies should also be explored.

\subsection{Parent selection}

A new criterion for $\epsilon$-lexicase selection was proposed, named Minimum Variance Threshold (MVT), inspired by decision tree splits \cite{breiman1984classification}, and was implemented in the FEAT algorithm \cite{cava2018learning}, which originally performs $\epsilon$-lexicase selection with semi-dynamic calculation of the threshold via Median Absolute Deviation (MAD) \cite{PHAMGIA2001921}. This approach yielded competitive results with original FEAT, and particularly improved FEAT performance on synthetic datasets when evaluated on the SRBench framework. Performance on real-world tasks was preserved but required more evaluations.
The calculation of the threshold with MVT alleviates one problem that can be oserved in the MAD approach, that is defining a threshold that prioritizes only near-elite solutions \cite{geiger_was_2025}.
The experiments were two fold, performing an in-depth analysis of small dimensionality datasets, and a robust evaluation on more than $100$ regression problems.
The objective was achieved by introducing MVT as a more intuitive criterion for $\epsilon$-lexicase. Convergence improved on synthetic and real-world datasets, and the computational trade-offs were clarified.

The following limitations were identified. Benefits were problem-dependent and more noticeable on synthetic datasets. The in-depth analysis was focused on problems of small dimensionality and a slowdown was observed, but no in-depth evaluation was made on large benchmarks. The work focused on FEAT due to its original implementation be based on $\epsilon$-lexicase, but the results may not generalize to other algorithms, although they could benefit from just including $\epsilon$-lexicase parent selection.

Key takeaways are the following. $\epsilon$-lexicase with dynamic or semi-dynamic thresholds should be preferred. MVT can look beyond near-elite individuals in each test case, which improves performance but requires higher computational cost. Even without MVT, $\epsilon$-lexicase remains superior to alternatives. Parent selection, though indirect, has a strong impact on model quality. Computational cost increased due to the additional test cases.

Future work should study how new criteria change the probability distribution of selected individuals. Down-sampling strategies could be evaluated to reduce test case usage. Finally, integration into other SR algorithms should be explored, with possible improvements through MVT.

\subsection{Model simplification}

An inexact simplification method was introduced, enabling automatic learning of purely data-driven substitution rules that reduce expression complexity with small overhead. Locality-sensitive hashing (LSH) \cite{gionis_similarity_nodate} using SimHash \cite{charikar2002similarity} was used to approximate semantics and identify equivalent subtrees. Experiments showed a median of $20\%$ error reduction when compared to results without the simplification method, and the method can be integrated into any SR framework.
The objective of designing a lightweight simplification method was achieved. The proposed LSH-based method reduced unnecessary complexity while maintaining predictive accuracy, introduced low implementation overhead, and was able to automatically learn dataset-specific equivalences.

Some limitations were noted. Inexact rules may slightly alter function behavior, which is undesirable in applications such as precise scientific discovery. The simplification was tested with SimHash to keep implementation simple and easy to reproduce, but other hash methods may affect the occurence of colisions thus affect the simplification performance. The benchmark considered small to medium dimensionality datasets, and the impact on runtime was not analyzed to large datasets.

The key takeaways are the following. Simplification should always be applied, at least to the final model, to improve interpretability and generalizability to unseen data. Inexact simplification is lightweight and often beneficial, and in cases it does not improve the objectives, it also does not harm execution. Data-driven simplification can reveal problem-specific equivalences. Simplification reduces bloat while preserving or improving accuracy.

Future work should investigate a ``warm-up'' phase in which a library of small expressions is generated to initialize the table. Simplification could also be applied more sporadically --- such as through a dedicated mutation operator or only during the final generation. Additional hash functions with stronger guarantees than SimHash should be evaluated. Finally, because hash collisions play a key role in controlling the behavior of inexact simplification, the impact of hash size should also be examined. The first two directions may reduce runtime, while the latter two may improve the overall performance of the framework.

\subsection{Benchmarking}

Much progress in SR arises from informal exchanges at conferences, where ideas are discussed during coffee breaks as much as in formal sessions. 
The benchmarking initiative emerged not only as a technical contribution but also as a community effort. The benchmark thus served as a rallying point, inviting collaboration, and maintaining SRBench \cite{srbench} as a living benchmark.
Achieving maturity for the field should be a shared goal, especially given the growing emphasis on transparency in machine learning and the rising interest in scientific ML.
The objective of conducting a large-scale comparative study was achieved. Evaluation practices across SR algorithms were standardized, also situating the proposed contributions within the broader landscape, and highlighting strengths and limitations of existing methods.

It was shown that parameter optimization is common to the best-performing algorithms, many relying on GP-based implementations. On the other hand, some optimization approaches rely on scaling, while others perform iterative structural searches without population-based heuristics.
Performance analyses revealed difficulties with hyperparameter tuning and defining effective computational budgets, highlighting the need for parameterless experience when using SR algorithms. SR methods were shown to improve model accuracy but often at the cost of overfitting noise, producing small accurate models that fail to capture the underlying ground-truth equation, indicating that progress is still needed for true scientific discovery.

Several limitations were identified. Hyperparameter tuning was difficult and not done exhaustively, particularly due to the restricted computational budget of an already expensive-to-run benchmark. Equation analysis with SymPy could not be performed, so raw equations were presented instead. Benchmark datasets, though comprehensive, can never fully generalize to all problem classes. Some methods that support hardware acceleration were not able to take full advantage of that, hindering their performances.

Key takeaways are the following. Running a benchmark is highly labor-intensive but essential for measuring progress and guiding the development of new methods. Parameter optimization was shown to be crucial for achieving high performance on black-box problems. GP-based SR remains competitive, even given recent transformer-based trends. Aggregated visualizations obscure specific performance, and no free lunch behavior was observed, showing that no algorithm consistently dominates.
Sustainability of SRBench requires continuous community participation. Contributions through discussions, ideas, criticism, and corrections are crucial.

Future work should consider dataset selection guided by informed metrics such as instance space analysis \cite{10.1145/3572895}. Hyperparameter tuning could be improved with Optuna \cite{10.1145/3292500.3330701}, which avoids exhaustive search. Emissions could be measured with tools such as CodeCarbon \cite{codecarbon2024}, which tracks CPU, GPU, and RAM usage. Model equivalence could be assessed with equality saturation approaches \cite{10.1145/3712256.3726385} and should evaluate the extent of recovery of ground-truth equations. These suggestions were later proposed by the audience at the oral presentation during the conference. 

\section{Building an unified framework}

Since $2023$, collaborations have been carried out with researchers from the University of Pennsylvania, active users of the Brush package. These interactions highlighted the importance of engaging with real users, who often present specific requirements that must be implemented in the backend of the library, a high-performing, template-based \CPP{} framework. Weekly meetings were held, where different needs were presented by the collaborators.
While the framework itself was not the central focus, many of the contributions proposed in this thesis were implemented within it, making its mention appropriate. Brush was not included as a primary thesis contribution because it was not conceived and developed under sole authorship, and many of its applications involved protected datasets that required IRB approval for access.
Among the contributions to Brush were the implementation of the evolutionary algorithm using island models, the development of extensive Python bindings for ease of use, and the application of the framework to healthcare problems, such as modeling computable phenotypes for treatment-resistant hypertension and modeling clinical scores with symbolic regression.

Several specific improvements were also incorporated into the framework: non-linear parameter optimization using the LM method for classification problems by minimizing log-loss; inexact simplification on a strongly typed SR algorithm ---with separate simplification tables for each data type---; parent selection with $\epsilon$-lexicase while keeping compatibility to a variety of loss metrics; final model selection heuristics based on inner validation partitions; initialization of populations using PTC2 within a strongly typed SR method; auxiliary functions to visualize Pareto fronts and explore trade-offs between complexity and predictive accuracy; recursive and linear complexity measures integrated into Brush; benchmarking within the updated SRBench framework; and the ability to lock parts of expressions and perform partial fitting starting from a previous population, enabling model transfer across different healthcare sites by optimizing both functions and parameters with new data.

\section{Impact and future perspectives}

The dilemma between advancing theoretical developments and supporting practical applications was confronted: while Brush was not the thesis focus, it demonstrated how theoretical contributions can acquire purpose when directed by practical demands. Bridging the gap between theory and practice is essential. From the perspective of end-users, the exact implementation details of SR methods are less relevant than their ability to deliver reliable and interpretable results.

Developing this thesis also required the acquisition of a broad range of skills beyond the technical contributions. These included learning several programming languages and their nuances, studying the mathematics of parameter optimization, working with Linux and SSH, and remotely accessing and managing high-performance computing clusters. Equally important were the skills related to academic and professional communication: preparing posters and oral presentations, responding to reviewers, presenting at conferences, networking with peers, organizing personal and professional tasks, and actively participating in research meetings.

As for future perspectives, in the short/mid term, the intention is to continue pursuing an academic career, combining teaching and research. Theoretical development will remain a central focus, whether by further exploring the problems raised in this thesis or by investigating new challenges that may emerge from ongoing engagement with the research community. At the same time, applied research will be pursued more actively, with the aim of grounding theoretical contributions in practical impact.
The trajectory of this thesis reinforced that algorithmic innovation and interpretability must advance in parallel to serve both the scientific community and applied domains.

The role of the thesis committee was essential in shaping this work into a more cohesive and didactic text. Their feedback complemented the thoughtful comments from anonymous reviewers, helping refine the manuscript in both structure and style. Important improvements included the expansion of the background section, the detailed explanations provided for the methods behind each contribution, and the effort to frame the results from a broader perspective. Further suggestions also enhanced the quality of the writing, \LaTeX{} formatting, figures, tables, and diagrams. Initially, the document was heavily based on the original publications, but following the committee's recommendations, it was revised to reduce redundancy, integrate additional related work, and present all contributions in a unified structure.

\section{Online resources}~\label{sec:conclusions:online_resources}

This section compiles all online resources referenced throughout the thesis, providing direct access to source code, datasets, and experiment scripts for each contribution.

The experiments described in Chapter~\ref{chap:parameteropt}, which focus on parameter optimization, are available in a GitHub repository at \url{https://github.com/gAldeia/itea-julia}, containing all implementations and datasets. The results and post-processing scripts are also included.

For Chapter~\ref{chap:parentsel}, which addresses parent selection, the following resources are provided.
FEAT with MVT selection is available at \url{https://github.com/gAldeia/feat}. All data, source code for experiments, and post-processing analyses are available at \url{https://github.com/gAldeia/srbench/tree/feat_split_benchmark}.

For the simplification study in Chapter~\ref{chap:inexactsimp}, all data and source code for implementations, experiments, and post-processing analyses are provided at \url{https://github.com/gAldeia/hashing-symbolic-expressions}.

For the benchmarking study in Chapter~\ref{chap:benchmarking}, all datasets were submitted to PMLB and are available both in the library and on GitHub at \url{https://github.com/EpistasisLab/pmlb}.
Experiment scripts and parameter configurations were released to ensure transparency and reproducibility.
The SRBench repository was restructured into containerized environments to enable simplified standalone use and consistent execution, and all Docker images were made publicly available on the student’s DockerHub profile at \url{https://hub.docker.com/u/galdeia}.
The complete project is publicly available at \url{https://github.com/cavalab/srbench/tree/srbench_2025}.

\section{List of publications}~\label{sec:conclusions:publications}

Below is a list of all sub-products of this thesis that have been published or submitted for review anytime that overlaps the period of redaction of this thesis.
This list follows the chronological order of publication.
The citation style used is defined by \textit{Brazilian National Standards Organization} (ABNT), following the directions under which this thesis is written.

\begin{flushleft}
     ALDEIA, G. S. I.; de FRANÇA, F. O. Interaction-Transformation Evolutionary Algorithm with parameters optimization. \textit{Genetic and Evolutionary Computation Conference Companion (GECCO '22)}, ACM New York, 2022. p. 1-8. Available from Internet \url{https://doi.org/10.1145/3520304.3533987}.
\end{flushleft}

\begin{flushleft}
     ALDEIA, G. S. I.; de FRANÇA, F. O.; La CAVA, W. G. Minimum variance threshold for \texorpdfstring{$\epsilon$}{epsilon}-lexicase selection. \textit{Genetic and Evolutionary Computation Conference (GECCO '24)}, ACM New York, 2024. p. 1-9. Available from Internet \url{https://doi.org/10.1145/3638529.3654149}.
\end{flushleft}

\begin{flushleft}
     ALDEIA, G. S. I.; de FRANÇA, F. O.; La CAVA, W. G. Inexact Simplification of Symbolic Regression Expressions with Locality-sensitive Hashing. \textit{Genetic and Evolutionary Computation Conference (GECCO '24)}, ACM New York, 2024. p. 1-9. Available from Internet \url{https://doi.org/10.1145/3638529.3654147}.
\end{flushleft}


\begin{flushleft}
    ALDEIA, G. S. I.; ZHANG, H.; BOMARITO, G.; CRANMER, M.; FONSECA, A.; BURLACU, B.; La CAVA, W. G.; de FRANÇA, F. O. Call for Action: towards the next generation of symbolic regression benchmark. \textit{Proceedings of the Genetic and Evolutionary Computation Conference Companion (GECCO '25)}.
    ACM New York, 2025. 
    Available from Internet \url{https://doi.org/10.1145/3712255.3734309}.
\end{flushleft}

\begin{flushleft}
    ALDEIA, G. S. I.; HERMAN, D. S.; La CAVA, W. G.
    Iterative Learning of Computable Phenotypes for Treatment Resistant Hypertension using Large Language Models.
    \textit{PMLR, Volume 298, Machine Learning for Healthcare}.
    2025. 
    Available from Internet \url{https://proceedings.mlr.press/v298/aldeia25a.html}.
\end{flushleft}

\begin{flushleft}
    ALDEIA, G. S. I.; ROMANO, J. D.; de FRANÇA, F. O.; HERMAN, D. S.; La CAVA, W. G. Towards symbolic regression for interpretable clinical decision scores.
    \textit{Phylosophical Transactions A, Royal Society}. (accepted, under production).
\end{flushleft}


%% file: Bibliografia.bib
@String{Computing = "Computing" }

@String{Computer = "{IEEE} Computer" }

@String{Springer = "Springer-Verlag" }

@article{HRUSKA2025118821,
title = {Analytical formulae for design of one-dimensional sonic crystals with smooth geometry based on symbolic regression},
journal = {Journal of Sound and Vibration},
volume = {597},
pages = {118821},
year = {2025},
issn = {0022-460X},
doi = {https://doi.org/10.1016/j.jsv.2024.118821},
url = {https://www.sciencedirect.com/science/article/pii/S0022460X24005832},
author = {Viktor Hruška and Aneta Furmanová and Michal Bednařík}
}

@inproceedings{10.1145/3583131.3590375,
author = {Ding, Li and Pantridge, Edward and Spector, Lee},
title = {Probabilistic Lexicase Selection},
year = {2023},
isbn = {9798400701191},
publisher = {Association for Computing Machinery},
address = {New York, NY, USA},
url = {https://doi.org/10.1145/3583131.3590375},
doi = {10.1145/3583131.3590375},
booktitle = {Proceedings of the Genetic and Evolutionary Computation Conference},
pages = {1073–1081},
numpages = {9},
location = {Lisbon, Portugal},
series = {GECCO '23}
}

@misc{hernandez_exploration_2021,
	title = {An {Exploration} of {Exploration}: {Measuring} the ability of lexicase selection to find obscure pathways to optimality},
	shorttitle = {An {Exploration} of {Exploration}},
	url = {http://arxiv.org/abs/2107.09760},
	doi = {10.48550/arXiv.2107.09760},
	urldate = {2025-09-16},
	publisher = {arXiv},
	author = {Hernandez, Jose Guadalupe and Lalejini, Alexander and Ofria, Charles},
	month = jul,
	year = {2021},
	note = {arXiv:2107.09760 [cs]},
}

@inproceedings{la_cava_epsilon-lexicase_2016,
	title = {Epsilon-{Lexicase} {Selection} for {Regression}},
note="Available at: \url{http://arxiv.org/abs/1905.13266}",
	doi = {10.1145/2908812.2908898},
	language = {en},
	urldate = {2024-01-26},
	booktitle = {Proceedings of the {Genetic} and {Evolutionary} {Computation} {Conference} 2016},
	author = {La{ }Cava, William and Spector, Lee and Danai, Kourosh},
	month = jul,
	year = {2016},
	pages = {741--748},
}

@inproceedings{geiger_down-sampled_2023,
author = {Geiger, Alina and Sobania, Dominik and Rothlauf, Franz},
year = {2023},
month = {07},
pages = {1109-1117},
title = {Down-Sampled Epsilon-Lexicase Selection for Real-World Symbolic Regression Problems},
doi = {10.1145/3583131.3590400}
}

@inproceedings{ding_probabilistic_2023,
	title = {Probabilistic {Lexicase} {Selection}},
note="Available at: \url{http://arxiv.org/abs/2305.11681}",
	doi = {10.1145/3583131.3590375},
	language = {en},
	urldate = {2024-01-26},
	booktitle = {Proceedings of the {Genetic} and {Evolutionary} {Computation} {Conference}},
	author = {Ding, Li and Pantridge, Edward and Spector, Lee},
	month = jul,
	year = {2023},
	pages = {1073--1081},
}

@article{romano2021pmlb,
  title = {PMLB v1.0: an open-source dataset collection for benchmarking machine learning methods},
  volume = {38},
  ISSN = {1367-4811},
note="Available at: \url{http://dx.doi.org/10.1093/bioinformatics/btab727}",
  DOI = {10.1093/bioinformatics/btab727},
  number = {3},
  journal = {Bioinformatics},
  publisher = {Oxford University Press (OUP)},
  author = {Romano,  Joseph D and Le,  Trang T and La{ }Cava,  William and Gregg,  John T and Goldberg,  Daniel J and Chakraborty,  Praneel and Ray,  Natasha L and Himmelstein,  Daniel and Fu,  Weixuan and Moore,  Jason H},
  editor = {Kelso,  Janet},
  year = {2021},
  month = oct,
  pages = {878–880}
}

@inproceedings{srbench,
 author = {La{ }Cava, William and Orzechowski, Patryk and Burlacu, Bogdan and de{ }Fran{\c{c}}a, Fabricio and Virgolin, Marco and Jin, Ying and Kommenda, Michael and Moore, Jason},
 booktitle = {Proceedings of the Neural Information Processing Systems Track on Datasets and Benchmarks},
 editor = {J. Vanschoren and S. Yeung},
 pages = {},
 publisher = {Curran},
 title = {Contemporary Symbolic Regression Methods and their Relative Performance},
 volume = {1},
 year = {2021}
}

@misc{FDA,
    author = "Food and Drug Administration",
    title  = "Clinical Decision Support Software - Guidance for Industry and Food and Drug Administration Staff",
    year   = "2022",
    note   = "https://www.fda.gov/regulatory-information/search-fda-guidance-documents/clinical-decision-support-software"
}

@book{breiman1984classification,
  title={Classification and Regression Trees},
  author={Breiman, L. and Friedman, J. and Stone, C.J. and Olshen, R.A.},
  isbn={9780412048418},
  lccn={83019708},
note="Available at: \url{https://books.google.com.br/books?id=JwQx-WOmSyQC}",
  year={1984},
  publisher={Taylor \& Francis}
}

@inproceedings{10.1145/2908812.2908851,
author = {Helmuth, Thomas and McPhee, Nicholas Freitag and Spector, Lee},
title = {The Impact of Hyperselection on Lexicase Selection},
year = {2016},
isbn = {9781450342063},
publisher = {Association for Computing Machinery},
address = {New York, NY, USA},
note="Available at: \url{https://doi.org/10.1145/2908812.2908851}",
doi = {10.1145/2908812.2908851},
booktitle = {Proceedings of the Genetic and Evolutionary Computation Conference 2016},
pages = {717–724},
numpages = {8},
location = {Denver, Colorado, USA},
series = {GECCO '16}
}

@inbook{complexity,
author = {Kommenda, Michael and Kronberger, Gabriel and Affenzeller, Michael and Winkler, Stephan and Burlacu, Bogdan},
year = {2016},
month = {12},
pages = {1-19},
title = {Evolving Simple Symbolic Regression Models by Multi-Objective Genetic Programming},
isbn = {978-3-319-34221-4},
doi = {10.1007/978-3-319-34223-8\_1}
}

@article{ridge,
 ISSN = {00401706},
note="Available at: \url{http://www.jstor.org/stable/1267351}",
 author = {Arthur E. Hoerl and Robert W. Kennard},
 journal = {Technometrics},
 number = {1},
 pages = {55--67},
 publisher = {[Taylor \& Francis, Ltd., American Statistical Association, American Society for Quality]},
 title = {Ridge Regression: Biased Estimation for Nonorthogonal Problems},
 urldate = {2024-01-31},
 volume = {12},
 year = {1970}
}

@article{aifeynman,
author = {Silviu-Marian Udrescu  and Max Tegmark },
title = {AI Feynman: A physics-inspired method for symbolic regression},
journal = {Science Advances},
volume = {6},
number = {16},
pages = {eaay2631},
year = {2020},
doi = {10.1126/sciadv.aay2631},
note="Available at: \url{https://www.science.org/doi/abs/10.1126/sciadv.aay2631}",
eprint = {https://www.science.org/doi/pdf/10.1126/sciadv.aay2631}}

@inproceedings{multiviewsr,
author = {Russeil, Etienne and de{ }Fran{\c{c}}a, Fabricio Olivetti and Malanchev, Konstantin and Burlacu, Bogdan and Ishida, Emille and Leroux, Marion and Michelin, Cl\'{e}ment and Moinard, Guillaume and Gangler, Emmanuel},
title = {Multiview Symbolic Regression},
year = {2024},
isbn = {9798400704949},
publisher = {Association for Computing Machinery},
address = {New York, NY, USA},
url = {https://doi.org/10.1145/3638529.3654087},
doi = {10.1145/3638529.3654087},
booktitle = {Proceedings of the Genetic and Evolutionary Computation Conference},
pages = {961–970},
numpages = {10},
location = {Melbourne, VIC, Australia},
series = {GECCO '24}
}

@inproceedings{10.1145/2330784.2330846,
author = {Spector, Lee},
title = {Assessment of problem modality by differential performance of lexicase selection in genetic programming: a preliminary report},
year = {2012},
isbn = {9781450311786},
publisher = {Association for Computing Machinery},
address = {New York, NY, USA},
note="Available at: \url{https://doi.org/10.1145/2330784.2330846}",
doi = {10.1145/2330784.2330846},
booktitle = {Proceedings of the 14th Annual Conference Companion on Genetic and Evolutionary Computation},
pages = {401–408},
numpages = {8},
location = {Philadelphia, Pennsylvania, USA},
series = {GECCO '12}
}

@article{PHAMGIA2001921,
title = {The mean and median absolute deviations},
journal = {Mathematical and Computer Modelling},
volume = {34},
number = {7},
pages = {921-936},
year = {2001},
issn = {0895-7177},
doi = {https://doi.org/10.1016/S0895-7177(01)00109-1},
note="Available at: \url{https://www.sciencedirect.com/science/article/pii/S0895717701001091}",
author = {T. Pham-Gia and T.L. Hung}
}

@article{10.1162/evco_a_00346,
    author = {Boldi, Ryan and Briesch, Martin and Sobania, Dominik and Lalejini, Alexander and Helmuth, Thomas and Rothlauf, Franz and Ofria, Charles and Spector, Lee},
    year = {2024},
    month = {01},
    pages = {1-32},
    title = {Informed Down-Sampled Lexicase Selection: Identifying productive training cases for efficient problem solving},
    journal = {Evolutionary computation},
    doi = {10.1162/evco\_a\_00346}
}

@inproceedings{10.1145/3319619.3326900,
author = {Hernandez, Jose Guadalupe and Lalejini, Alexander and Dolson, Emily and Ofria, Charles},
title = {Random subsampling improves performance in lexicase selection},
year = {2019},
isbn = {9781450367486},
publisher = {Association for Computing Machinery},
address = {New York, NY, USA},
note="Available at: \url{https://doi.org/10.1145/3319619.3326900}",
doi = {10.1145/3319619.3326900},
booktitle = {Proceedings of the Genetic and Evolutionary Computation Conference Companion},
pages = {2028–2031},
numpages = {4},
location = {Prague, Czech Republic},
series = {GECCO '19}
}

@article{brindle1980genetic,
  title={Genetic algorithms for function optimization},
  author={Brindle, Anne},
  year={1980}
}

@article{miller1995genetic,
  title={Genetic algorithms, tournament selection, and the effects of noise},
  author={Miller, Brad L and Goldberg, David E and others},
  journal={Complex systems},
  volume={9},
  number={3},
  pages={193--212},
  year={1995},
  publisher={[Champaign, IL, USA: Complex Systems Publications, Inc., c1987-}
}

@inproceedings{10.1145/2576768.2598288,
author = {Krawiec, Krzysztof and O'Reilly, Una-May},
title = {Behavioral programming: a broader and more detailed take on semantic GP},
year = {2014},
isbn = {9781450326629},
publisher = {Association for Computing Machinery},
address = {New York, NY, USA},
note="Available at: \url{https://doi.org/10.1145/2576768.2598288}",
doi = {10.1145/2576768.2598288},
booktitle = {Proceedings of the 2014 Annual Conference on Genetic and Evolutionary Computation},
pages = {935–942},
numpages = {8},
location = {Vancouver, BC, Canada},
series = {GECCO '14}
}

@article{Kommenda2019,
  title = {Parameter identification for symbolic regression using nonlinear least squares},
  volume = {21},
  ISSN = {1573-7632},
note="Available at: \url{http://dx.doi.org/10.1007/s10710-019-09371-3}",
  DOI = {10.1007/s10710-019-09371-3},
  number = {3},
  journal = {Genetic Programming and Evolvable Machines},
  publisher = {Springer Science and Business Media LLC},
  author = {Kommenda,  Michael and Burlacu,  Bogdan and Kronberger,  Gabriel and Affenzeller,  Michael},
  year = {2019},
  month = dec,
  pages = {471–501}
}

@inproceedings{10.1145/3321707.3321758,
    author = {Virgolin, Marco and Alderliesten, Tanja and Bosman, Peter A. N.},
    title = {Linear scaling with and within semantic backpropagation-based genetic programming for symbolic regression},
    year = {2019},
    isbn = {9781450361118},
    publisher = {Association for Computing Machinery},
    address = {New York, NY, USA},
note="Available at: \url{https://doi.org/10.1145/3321707.3321758}",
    doi = {10.1145/3321707.3321758},
    booktitle = {Proceedings of the Genetic and Evolutionary Computation Conference},
    pages = {1084–1092},
    numpages = {9},
    location = {Prague, Czech Republic},
    series = {GECCO '19}
}

@article{past_present_future_phd,
    title={The past, present and future of the PhD thesis},
    journal={Nature},
    year={2016},
    month={Jul},
    day={01},
    volume={535},
    number={7610},
    pages={7-7},
    issn={1476-4687},
    doi={10.1038/535007a},
note="Available at: \url{https://doi.org/10.1038/535007a}"
}

@article{Gould2016,
  title = {What’s the point of the PhD thesis?},
  volume = {535},
  ISSN = {1476-4687},
note="Available at: \url{http://dx.doi.org/10.1038/535026a}",
  DOI = {10.1038/535026a},
  number = {7610},
  journal = {Nature},
  publisher = {Springer Science and Business Media LLC},
  author = {Gould,  Julie},
  year = {2016},
  month = jul,
  pages = {26–28}
}

@article{https://doi.org/10.1111/area.12779,
author = {Hulme, Mike},
title = {Reflections on the afterlives of a PhD thesis},
journal = {Area},
volume = {54},
number = {2},
pages = {280-289},
doi = {https://doi.org/10.1111/area.12779},
note="Available at: \url{https://rgs-ibg.onlinelibrary.wiley.com/doi/abs/10.1111/area.12779}",
eprint = {https://rgs-ibg.onlinelibrary.wiley.com/doi/pdf/10.1111/area.12779},
year = {2022}
}

@article{BurroughBoenisch2016,
  title = {Being more open about PhD papers},
  volume = {536},
  ISSN = {1476-4687},
note="Available at: \url{http://dx.doi.org/10.1038/536274b}",
  DOI = {10.1038/536274b},
  number = {7616},
  journal = {Nature},
  publisher = {Springer Science and Business Media LLC},
  author = {Burrough-Boenisch,  Joy},
  year = {2016},
  month = aug,
  pages = {274–274}
}

@misc{kotary2021endtoend,
      title={End-to-End Constrained Optimization Learning: A Survey}, 
      author={James Kotary and Ferdinando Fioretto and Pascal Van Hentenryck and Bryan Wilder},
      year={2021},
      eprint={2103.16378},
      archivePrefix={arXiv},
      primaryClass={cs.LG}
}

@book{James2021,
  title = {An Introduction to Statistical Learning: with Applications in R},
  ISBN = {9781071614181},
  ISSN = {2197-4136},
  url = {http://dx.doi.org/10.1007/978-1-0716-1418-1},
  DOI = {10.1007/978-1-0716-1418-1},
  journal = {Springer Texts in Statistics},
  publisher = {Springer US},
  author = {James,  Gareth and Witten,  Daniela and Hastie,  Trevor and Tibshirani,  Robert},
  year = {2021}
}

@article{levenberg1944method,
  title={A method for the solution of certain non-linear problems in least squares},
  author={Levenberg, Kenneth},
  journal={Quarterly of applied mathematics},
  volume={2},
  number={2},
  pages={164--168},
  year={1944}}

@article{marquardt1963algorithm,
  title={An algorithm for least-squares estimation of nonlinear parameters},
  author={Marquardt, Donald W},
  journal={Journal of the society for Industrial and Applied Mathematics},
  volume={11},
  number={2},
  pages={431--441},
  year={1963},
  publisher={SIAM}}

@inproceedings{PrioritizedGrammarEnumeration,
    author = {Worm, Tony and Chiu, Kenneth},
    title = {Prioritized Grammar Enumeration: Symbolic Regression by Dynamic Programming},
    year = {2013},
    isbn = {9781450319638},
    publisher = {Association for Computing Machinery},
    address = {New York, NY, USA},
note="Available at: \url{https://doi.org/10.1145/2463372.2463486}",
    doi = {10.1145/2463372.2463486},
    booktitle = {Proceedings of the 15th Annual Conference on Genetic and Evolutionary Computation},
    pages = {1021–1028},
    numpages = {8},
    location = {Amsterdam, The Netherlands},
    series = {GECCO '13}}

@inproceedings{ITWithCoeffsOptimization,
    author = {Aldeia, Guilherme Seidyo Imai and de{ }Fran{\c{c}}a, Fabr\'{\i}cio Olivetti},
    title = {Interaction-Transformation Evolutionary Algorithm with Coefficients Optimization},
    year = {2022},
    isbn = {9781450392686},
    publisher = {Association for Computing Machinery},
    address = {New York, NY, USA},
note="Available at: \url{https://doi.org/10.1145/3520304.3533987}",
    doi = {10.1145/3520304.3533987},
    booktitle = {Proceedings of the Genetic and Evolutionary Computation Conference Companion},
    pages = {2274–2281},
    numpages = {8},
    location = {Boston, Massachusetts},
    series = {GECCO '22}}

@ARTICLE{PTC2,
  author={Luke, S.},
  journal={IEEE Transactions on Evolutionary Computation}, 
  title={Two fast tree-creation algorithms for genetic programming}, 
  year={2000},
  volume={4},
  number={3},
  pages={274-283},
  doi={10.1109/4235.873237}}

@ARTICLE{NSGA-II,
  author={Deb, K. and Pratap, A. and Agarwal, S. and Meyarivan, T.},
  journal={IEEE Transactions on Evolutionary Computation}, 
  title={A fast and elitist multiobjective genetic algorithm: NSGA-II}, 
  year={2002},
  volume={6},
  number={2},
  pages={182-197},
  doi={10.1109/4235.996017}}

@article{SymbolicRegressionIsNPHard,
    title={Symbolic Regression is {NP}-hard},
    author={Marco Virgolin and Solon P Pissis},
    journal={Transactions on Machine Learning Research},
    issn={2835-8856},
    year={2022},
  volume={},
  number={},
    pages={1-11},
note="Available at: \url{https://openreview.net/forum?id=LTiaPxqe2e}"}

@article{ExhaustiveSR,
	title = {Exhaustive {Symbolic} {Regression}},
	issn = {1089-778X, 1089-778X, 1941-0026},
note="Available at: \url{https://ieeexplore.ieee.org/document/10136815/}",
	doi = {10.1109/TEVC.2023.3280250},
	urldate = {2023-12-12},
	journal = {IEEE Transactions on Evolutionary Computation},
	author = {Bartlett, Deaglan J. and Desmond, Harry and Ferreira, Pedro G.},
	year = {2023},
	pages = {1--1},}

@inproceedings{UnifiedFrameworkForDeepSR,
	title = {A {Unified} {Framework} for {Deep} {Symbolic} {Regression}},
	volume = {35},
note="Available at: \url{https://proceedings.neurips.cc/paper\_files/paper/2022/file/dbca58f35bddc6e4003b2dd80e42f838-Paper-Conference.pdf}",
	booktitle = {Advances in {Neural} {Information} {Processing} {Systems}},
	publisher = {Curran Associates, Inc.},
	author = {Landajuela, Mikel and Lee, Chak Shing and Yang, Jiachen and Glatt, Ruben and Santiago, Claudio P and Aravena, Ignacio and Mundhenk, Terrell and Mulcahy, Garrett and Petersen, Brenden K},
	editor = {Koyejo, S. and Mohamed, S. and Agarwal, A. and Belgrave, D. and Cho, K. and Oh, A.},
	year = {2022},
	pages = {33985--33998},}

@inproceedings{windturbinedynamics,
  title={Automatic Identification of Closed-Loop Wind Turbine Dynamics via Genetic Programming},
  author={La{ }Cava, William and Danai, Kourosh and Lackner, Matthew and Spector, Lee and Fleming, Paul and Wright, Alan},
  booktitle={Dynamic Systems and Control Conference},
  volume={57250},
  pages={V002T21A002},
  year={2015},
  organization={American Society of Mechanical Engineers}}

@article{Weng2020,
  doi = {10.1038/s41467-020-17263-9},
note="Available at: \url{https://doi.org/10.1038/s41467-020-17263-9}",
  year = {2020},
  month = jul,
  publisher = {Springer Science and Business Media {LLC}},
  volume = {11},
  number = {1},
  pages={1--8},
  author = {Baicheng Weng and Zhilong Song and Rilong Zhu and Qingyu Yan and Qingde Sun and Corey G. Grice and Yanfa Yan and Wan-Jian Yin},
  title = {Simple descriptor derived from symbolic regression accelerating the discovery of new perovskite catalysts},
  journal = {Nature Communications}}

@article{KUBALIK2021115210,
    title = {Multi-objective symbolic regression for physics-aware dynamic modeling},
    journal = {Expert Systems with Applications},
    volume = {182},
    pages = {115210},
    year = {2021},
    issn = {0957-4174},
    doi = {https://doi.org/10.1016/j.eswa.2021.115210},
note="Available at: \url{https://www.sciencedirect.com/science/article/pii/S0957417421006436}",
    author = {Jiří Kubalík and Erik Derner and Robert Babuška}}

@article{la_cava_flexible_2023,
	title = {A flexible symbolic regression method for constructing interpretable clinical prediction models},
	volume = {6},
	issn = {2398-6352},
note="Available at: \url{https://www.nature.com/articles/s41746-023-00833-8}",
	doi = {10.1038/s41746-023-00833-8},
	language = {en},
	number = {1},
	urldate = {2024-01-26},
	journal = {npj Digital Medicine},
	author = {La{ }Cava, William G. and Lee, Paul C. and Ajmal, Imran and Ding, Xiruo and Solanki, Priyanka and Cohen, Jordana B. and Moore, Jason H. and Herman, Daniel S.},
	month = jun,
	year = {2023},
	pages = {107},
}

@article{LACAVA2016892,
    title = {Automatic identification of wind turbine models using evolutionary multiobjective optimization},
    journal = {Renewable Energy},
    volume = {87},
    pages = {892-902},
    year = {2016},
    issn = {0960-1481},
    doi = {https://doi.org/10.1016/j.renene.2015.09.068},
    note="Available at: \url{https://www.sciencedirect.com/science/article/pii/S0960148115303475}",
    author = {William La{ }Cava and Kourosh Danai and Lee Spector and Paul Fleming and Alan Wright and Matthew Lackner}
    }

@article{Angelis2023,
  title = {Artificial Intelligence in Physical Sciences: Symbolic Regression Trends and Perspectives},
  volume = {30},
  ISSN = {1886-1784},
note="Available at: \url{http://dx.doi.org/10.1007/s11831-023-09922-z}",
  DOI = {10.1007/s11831-023-09922-z},
  number = {6},
  journal = {Archives of Computational Methods in Engineering},
  publisher = {Springer Science and Business Media LLC},
  author = {Angelis,  Dimitrios and Sofos,  Filippos and Karakasidis,  Theodoros E.},
  year = {2023},
  month = apr,
  pages = {3845–3865}
}

@article{WANG20042453,
    title = {Hybrid genetic algorithm for optimization problems with permutation property},
    journal = {Computers \& Operations Research},
    volume = {31},
    number = {14},
    pages = {2453-2471},
    year = {2004},
    issn = {0305-0548},
    doi = {https://doi.org/10.1016/S0305-0548(03)00198-9},
    author = {Hsiao-Fan Wang and Kuang-Yao Wu}}

@inproceedings{8393325,
  title={Elite bases regression: A real-time algorithm for symbolic regression},
  author={Chen, Chen and Luo, Changtong and Jiang, Zonglin},
  booktitle={2017 13th International conference on natural computation, fuzzy systems and knowledge discovery (ICNC-FSKD)},
  pages={529--535},
  year={2017},
  organization={IEEE}}

@article{Katoch2020,
  doi = {10.1007/s11042-020-10139-6},
note="Available at: \url{https://doi.org/10.1007/s11042-020-10139-6}",
  year = {2020},
  month = oct,
  publisher = {Springer Science and Business Media {LLC}},
  volume = {80},
  number = {5},
  pages = {8091--8126},
  author = {Sourabh Katoch and Sumit Singh Chauhan and Vijay Kumar},
  title = {A review on genetic algorithm: past,  present,  and future},
  journal = {Multimedia Tools and Applications}}

@article{10.1162/evco_a_00285,
    author = {de{ }Fran{\c{c}}a, F. O. and Aldeia, G. S. I.},
    title = "{Interaction–Transformation Evolutionary Algorithm for Symbolic Regression}",
    journal = {Evolutionary Computation},
    volume = {29},
    number = {3},
    pages = {367-390},
    year = {2021},
    month = {09},
    issn = {1063-6560},
    doi = {10.1162/evco\_a\_00285},
note="Available at: \url{https://doi.org/10.1162/evco\_a\_00285}",
    eprint = {https://direct.mit.edu/evco/article-pdf/29/3/367/1959462/evco\_a\_00285.pdf}}

@article{doi:10.1080/00031305.1989.10475644,
    author = { Simo   Puntanen  and  George P. H.   Styan },
    title = {The Equality of the Ordinary Least Squares Estimator and the Best Linear Unbiased Estimator},
    journal = {The American Statistician},
    volume = {43},
    number = {3},
    pages = {153-161},
    year  = {1989},
    publisher = {Taylor \& Francis},
    doi = {10.1080/00031305.1989.10475644},
note="Available at: \url{https://www.tandfonline.com/doi/abs/10.1080/00031305.1989.10475644}",
    eprint = {https://www.tandfonline.com/doi/pdf/10.1080/00031305.1989.10475644}}

@inproceedings{kamienny_end--end_nodate,
    title={End-to-end Symbolic Regression with Transformers},
    author={Pierre-Alexandre Kamienny and St{\'e}phane d'Ascoli and Guillaume Lample and Francois Charton},
    booktitle={Advances in Neural Information Processing Systems},
    editor={Alice H. Oh and Alekh Agarwal and Danielle Belgrave and Kyunghyun Cho},
    year={2022},
    pages = {1-13},
    note="Available at: \url{https://openreview.net/forum?id=GoOuIrDHG\_Y}"}

@inproceedings{shojaee_transformer-based_nodate,
    title={Transformer-based Planning for Symbolic Regression},
    author={Parshin Shojaee and Kazem Meidani and Amir Barati Farimani and Chandan K. Reddy},
    booktitle={Thirty-seventh Conference on Neural Information Processing Systems},
    year={2023},
    note="Available at: \url{https://openreview.net/forum?id=0rVXQEeFEL}"}

@inproceedings{udrescu_ai_2020,
	title = {{AI} {Feynman} 2.0: {Pareto}-optimal symbolic regression exploiting graph modularity},
	volume = {33},
note="Available at: \url{https://proceedings.neurips.cc/paper\_files/paper/2020/file/33a854e247155d590883b93bca53848a-Paper.pdf}",
	booktitle = {Advances in {Neural} {Information} {Processing} {Systems}},
	publisher = {Curran Associates, Inc.},
	author = {Udrescu, Silviu-Marian and Tan, Andrew and Feng, Jiahai and Neto, Orisvaldo and Wu, Tailin and Tegmark, Max},
	editor = {Larochelle, H. and Ranzato, M. and Hadsell, R. and Balcan, M. F. and Lin, H.},
	year = {2020},
	pages = {4860--4871},
}

@misc{defranca2023interpretable,
      title={Interpretable Symbolic Regression for Data Science: Analysis of the 2022 Competition}, 
      author={F. O. de{ }Fran{\c{c}}a and M. Virgolin and M. Kommenda and M. S. Majumder and M. Cranmer and G. Espada and L. Ingelse and A. Fonseca and M. Landajuela and B. Petersen and R. Glatt and N. Mundhenk and C. S. Lee and J. D. Hochhalter and D. L. Randall and P. Kamienny and H. Zhang and G. Dick and A. Simon and B. Burlacu and Jaan Kasak and Meera Machado and Casper Wilstrup and W. G. La{ }Cava},
      year={2023},
      eprint={2304.01117},
      archivePrefix={arXiv},
      pages = {1–13},
      primaryClass={cs.LG}
}

@article{Aldeia2022,
  title = {Interpretability in symbolic regression: a benchmark of explanatory methods using the Feynman data set},
  volume = {23},
  ISSN = {1573-7632},
note="Available at: \url{http://dx.doi.org/10.1007/s10710-022-09435-x}",
  DOI = {10.1007/s10710-022-09435-x},
  number = {3},
  journal = {Genetic Programming and Evolvable Machines},
  publisher = {Springer Science and Business Media LLC},
  author = {Aldeia,  Guilherme Seidyo Imai and de{ }Fran{\c{c}}a, Fabrício Olivetti},
  year = {2022},
  month = may,
  pages = {309–349}
}

@article{egklitz2020,
  title = {Benchmarking state-of-the-art symbolic regression algorithms},
  volume = {22},
  ISSN = {1573-7632},
  url = {http://dx.doi.org/10.1007/s10710-020-09387-0},
  DOI = {10.1007/s10710-020-09387-0},
  number = {1},
  journal = {Genetic Programming and Evolvable Machines},
  publisher = {Springer Science and Business Media LLC},
  author = { {\v{Z}}egklitz, Jan and Po{\v{s}}{\'i}k, Petr },
  year = {2020},
  month = mar,
  pages = {5–33}
}

@article{esg1352,
  title={Evaluation and Benchmarking of Performance of Machine Learning and Symbolic Regression: Datasets, Software Tools and Prediction Models},
  author={Abdalla, Jamal A and Naser, MZ and Alogla, Saleh M and Ghasemi, Alireza and Naser, Ahmad},
  journal={ES General},
  volume={7},
  pages={1352},
  year={2024},
  publisher={Engineered Science Publisher}
}

@article{matsubara2024rethinking,
title={Rethinking Symbolic Regression Datasets and Benchmarks for Scientific Discovery},
author={Yoshitomo Matsubara and Naoya Chiba and Ryo Igarashi and Yoshitaka Ushiku},
journal={Journal of Data-centric Machine Learning Research},
issn={XXXX-XXXX},
year={2024},
url={https://openreview.net/forum?id=qrUdrXsiXX},
note={}
}

@inproceedings{10.1145/3205455.3205539,
author = {Orzechowski, Patryk and La{ }Cava, William and Moore, Jason H.},
title = {Where are we now? a large benchmark study of recent symbolic regression methods},
year = {2018},
isbn = {9781450356183},
publisher = {Association for Computing Machinery},
address = {New York, NY, USA},
note="Available at: \url{https://doi.org/10.1145/3205455.3205539}",
doi = {10.1145/3205455.3205539},
booktitle = {Proceedings of the Genetic and Evolutionary Computation Conference},
pages = {1183–1190},
numpages = {8},
location = {Kyoto, Japan},
series = {GECCO '18}
}

@inproceedings{holt2023deep,
title={Deep Generative Symbolic Regression},
author={Samuel Holt and Zhaozhi Qian and Mihaela van der Schaar},
booktitle={The Eleventh International Conference on Learning Representations },
year={2023},
note="Available at: \url{https://openreview.net/forum?id=o7koEEMA1bR}"
}

@article{Wilstrup2022,
  title = {Combining symbolic regression with the Cox proportional hazards model improves prediction of heart failure deaths},
  volume = {22},
  ISSN = {1472-6947},
note="Available at: \url{http://dx.doi.org/10.1186/s12911-022-01943-1}",
  DOI = {10.1186/s12911-022-01943-1},
  number = {1},
  journal = {BMC Medical Informatics and Decision Making},
  publisher = {Springer Science and Business Media LLC},
  author = {Wilstrup,  Casper and Cave,  Chris},
  year = {2022},
pages = {1–7},
  month = jul 
}

@inproceedings{10.1145/3205455.3205604,
author = {Virgolin, Marco and Alderliesten, Tanja and Bel, Arjan and Witteveen, Cees and Bosman, Peter A. N.},
title = {Symbolic regression and feature construction with GP-GOMEA applied to radiotherapy dose reconstruction of childhood cancer survivors},
year = {2018},
isbn = {9781450356183},
publisher = {Association for Computing Machinery},
address = {New York, NY, USA},
note="Available at: \url{https://doi.org/10.1145/3205455.3205604}",
doi = {10.1145/3205455.3205604},
booktitle = {Proceedings of the Genetic and Evolutionary Computation Conference},
pages = {1395–1402},
numpages = {8},
location = {Kyoto, Japan},
series = {GECCO '18}
}

@article{Makke2024,
  title = {Interpretable scientific discovery with symbolic regression: a review},
  volume = {57},
  ISSN = {1573-7462},
note="Available at: \url{http://dx.doi.org/10.1007/s10462-023-10622-0}",
  DOI = {10.1007/s10462-023-10622-0},
  number = {1},
  journal = {Artificial Intelligence Review},
  publisher = {Springer Science and Business Media LLC},
  author = {Makke,  Nour and Chawla,  Sanjay},
  year = {2024},
pages = {1–32},
  month = jan 
}

@article{ZHANG2022101061,
title = {PS-Tree: A piecewise symbolic regression tree},
journal = {Swarm and Evolutionary Computation},
volume = {71},
pages = {101061},
year = {2022},
issn = {2210-6502},
doi = {https://doi.org/10.1016/j.swevo.2022.101061},
note="Available at: \url{https://www.sciencedirect.com/science/article/pii/S2210650222000335}",
author = {Hengzhe Zhang and Aimin Zhou and Hong Qian and Hu Zhang}
}

@article{10.1162/evco_a_00278,
    author = {Virgolin, M. and Alderliesten, T. and Witteveen, C. and Bosman, P. A. N.},
    title = "{Improving Model-Based Genetic Programming for Symbolic Regression of
                    Small Expressions}",
    journal = {Evolutionary Computation},
    volume = {29},
    number = {2},
    pages = {211-237},
    year = {2021},
    month = {06},
    issn = {1063-6560},
    doi = {10.1162/evco\_a\_00278},
note="Available at: \url{https://doi.org/10.1162/evco\_a\_00278}",
    eprint = {https://direct.mit.edu/evco/article-pdf/29/2/211/1921067/evco\_a\_00278.pdf},
}

@misc{petersen2021deep,
      title={Deep symbolic regression: Recovering mathematical expressions from data via risk-seeking policy gradients}, 
      author={Brenden K. Petersen and Mikel Landajuela and T. Nathan Mundhenk and Claudio P. Santiago and Soo K. Kim and Joanne T. Kim},
      year={2021},
      eprint={1912.04871},
      archivePrefix={arXiv},
      primaryClass={cs.LG}
}

@misc{wilstrup2021symbolic,
      title={Symbolic regression outperforms other models for small data sets}, 
      author={Casper Wilstrup and Jaan Kasak},
      year={2021},
      eprint={2103.15147},
      archivePrefix={arXiv},
      primaryClass={cs.LG}
}

@article{4a848dd1-54e3-3c3c-83c3-04977ded2e71,
 ISSN = {00905364},
note="Available at: \url{http://www.jstor.org/stable/2699986}",
 author = {Jerome H. Friedman},
 journal = {The Annals of Statistics},
 number = {5},
 pages = {1189--1232},
 publisher = {Institute of Mathematical Statistics},
 title = {Greedy Function Approximation: A Gradient Boosting Machine},
 urldate = {2024-01-30},
 volume = {29},
 year = {2001}
}

@book{feynman2006feynman,
  title={The Feynman Lectures on Physics},
  author={Feynman, R.P. and Leighton, R.B. and Sands, M.L.},
  number={vol. 2},
  isbn={9780805390476},
  lccn={20052802},
  series={The Feynman Lectures on Physics},
note="Available at: \url{https://books.google.com.br/books?id=AbruAAAAMAAJ}",
  year={2006},
  publisher={Pearson/Addison-Wesley}
}

@book{feynman2015feynman,
  title={The Feynman Lectures on Physics, Vol. I: The New Millennium Edition: Mainly Mechanics, Radiation, and Heat},
  author={Feynman, R.P. and Leighton, R.B. and Sands, M.},
  number={vol. 1},
  series={The Feynman Lectures on Physics},
  isbn={9780465040858},
note="Available at: \url{https://books.google.com.br/books?id=d76DBQAAQBAJ}",
  year={2015},
  publisher={Basic Books}
}

@inproceedings{gpgomea,
author = {Virgolin, Marco and Alderliesten, Tanja and Witteveen, Cees and Bosman, Peter A. N.},
title = {Scalable genetic programming by gene-pool optimal mixing and input-space entropy-based building-block learning},
year = {2017},
isbn = {9781450349208},
publisher = {Association for Computing Machinery},
address = {New York, NY, USA},
note="Available at: \url{https://doi.org/10.1145/3071178.3071287}",
doi = {10.1145/3071178.3071287},
booktitle = {Proceedings of the Genetic and Evolutionary Computation Conference},
pages = {1041–1048},
numpages = {8},
location = {Berlin, Germany},
series = {GECCO '17}
}

@inproceedings{operon,
author = {Burlacu, Bogdan and Kronberger, Gabriel and Kommenda, Michael},
title = {Operon C++: an efficient genetic programming framework for symbolic regression},
year = {2020},
isbn = {9781450371278},
publisher = {Association for Computing Machinery},
address = {New York, NY, USA},
note="Available at: \url{https://doi.org/10.1145/3377929.3398099}",
doi = {10.1145/3377929.3398099},
booktitle = {Proceedings of the 2020 Genetic and Evolutionary Computation Conference Companion},
pages = {1562–1570},
numpages = {9},
location = {Canc\'{u}n, Mexico},
series = {GECCO '20}
}

@misc{burlacu2025zobristhashbasedduplicatedetection,
      title={Zobrist Hash-based Duplicate Detection in Symbolic Regression}, 
      author={Bogdan Burlacu},
      year={2025},
      eprint={2508.13859},
      archivePrefix={arXiv},
      primaryClass={cs.NE},
      url={https://arxiv.org/abs/2508.13859}, 
}

@article{SymTree,
    title = "A greedy search tree heuristic for symbolic regression",
    journal = "Information Sciences",
    volume = "442-443",
    pages = "18 - 32",
    year = "2018",
    issn = "0020-0255",
    doi = "https://doi.org/10.1016/j.ins.2018.02.040",
note="Available at: \url{http://www.sciencedirect.com/science/article/pii/S0020025516308635}",
    author = "de{ }Fran{\c{c}}a, Fabr\'{\i}cio Olivetti",
}

@inproceedings{ExplainingSRPredictions,
    doi = {10.1109/cec48606.2020.9185683},
note="Available at: \url{https://doi.org/10.1109\%2Fcec48606.2020.9185683}",
    year = 2020,
    month = {jul},
    publisher = {{IEEE}},
    numpages = {8},
    address       = "New York",
    author = {Renato Miranda Filho and Anisio Lacerda and Gisele L. Pappa},
    title = {Explaining Symbolic Regression Predictions},
    booktitle = {2020 {IEEE} Congress on Evolutionary Computation ({CEC})}
}

@incollection{GainingDeeperInsightsInSymbolicRegression,
    author="Affenzeller, Michael
    and Winkler, Stephan M.
    and Kronberger, Gabriel
    and Kommenda, Michael
    and Burlacu, Bogdan
    and Wagner, Stefan",
    editor="Riolo, Rick
    and Moore, Jason H.
    and Kotanchek, Mark",
    title="Gaining Deeper Insights in Symbolic Regression",
    bookTitle="Genetic Programming Theory and Practice XI",
    year="2014",
    publisher="Springer New York",
    address="New York, NY",
    pages="175--190",
    isbn="978-1-4939-0375-7",
    doi="10.1007/978-1-4939-0375-7\_10",
    note="Available at: \url{https://doi.org/10.1007/978-1-4939-0375-7\_10}"
}

@inproceedings{ApplyingGeneticProgrammingToImproveInterpretability,
    doi = {10.1109/cec48606.2020.9185620},
note="Available at: \url{https://doi.org/10.1109\%2Fcec48606.2020.9185620}",
    year = 2020,
    month = {jul},
    publisher = {{IEEE}},
    address       = "New York",
    numpages = {8},
    author = {Leonardo Augusto Ferreira and Frederico Gadelha Guimaraes and Rodrigo Silva},
    title = {Applying Genetic Programming to Improve Interpretability in Machine Learning Models},
    booktitle = {2020 {IEEE} Congress on Evolutionary Computation ({CEC})}
}

@misc{Agarwal_Ceres_Solver_2022,
  author = {Agarwal, Sameer and Mierle, Keir and The Ceres Solver Team},
  title = {{Ceres Solver}},
  license = {Apache-2.0},
note="Available at: \url{https://github.com/ceres-solver/ceres-solver}",
  version = {2.2},
  year = {2023},
  month = {10}
}

@misc{song2024prove,
      title={Prove Symbolic Regression is NP-hard by Symbol Graph}, 
      author={Jinglu Song and Qiang Lu and Bozhou Tian and Jingwen Zhang and Jake Luo and Zhiguang Wang},
      year={2024},
      eprint={2404.13820},
      archivePrefix={arXiv},
      primaryClass={cs.CC}
}

@book{GPOnTheProgrammingOfComputersByMeansOfNaturalSelection,
    title={Genetic Programming: On the Programming of Computers by Means of Natural Selection},
    author={Koza, J.R.},
    isbn={9780262111706},
    lccn={92025785},
    series={A Bradford book},
note="Available at: \url{https://books.google.com.br/books?id=Bhtxo60BV0EC}",
    year={1992},
    publisher={Bradford},
    address={Bradford, PA}
}

@article{Reinbold2021,
  doi = {10.1038/s41467-021-23479-0},
note="Available at: \url{https://doi.org/10.1038/s41467-021-23479-0}",
  year = {2021},
  month = may,
  publisher = {Springer Science and Business Media {LLC}},
  volume = {12},
  number = {1},
  author = {Patrick A. K. Reinbold and Logan M. Kageorge and Michael F. Schatz and Roman O. Grigoriev},
  pages={1--8},
  title = {Robust learning from noisy,  incomplete,  high-dimensional experimental data via physically constrained symbolic regression},
  journal = {Nature Communications}
}

@ARTICLE{970573,
  author={Donghee Lee and Jongmoo Choi and Jong-Hun Kim and Noh, S.H. and Sang Lyul Min and Yookun Cho and Chong Sang Kim},
  journal={IEEE Transactions on Computers}, 
  title={LRFU: a spectrum of policies that subsumes the least recently used and least frequently used policies}, 
  year={2001},
  volume={50},
  number={12},
  pages={1352-1361},
  doi={10.1109/TC.2001.970573}}

@article{CHEN20181973,
title = {Block building programming for symbolic regression},
journal = {Neurocomputing},
volume = {275},
pages = {1973-1980},
year = {2018},
issn = {0925-2312},
doi = {https://doi.org/10.1016/j.neucom.2017.10.047},
note="Available at: \url{https://www.sciencedirect.com/science/article/pii/S0925231217316983}",
author = {Chen Chen and Changtong Luo and Zonglin Jiang}
}

@article{LUO20121182,
title = {Parse-matrix evolution for symbolic regression},
journal = {Engineering Applications of Artificial Intelligence},
volume = {25},
number = {6},
pages = {1182-1193},
year = {2012},
issn = {0952-1976},
doi = {https://doi.org/10.1016/j.engappai.2012.05.015},
note="Available at: \url{https://www.sciencedirect.com/science/article/pii/S0952197612001212}",
author = {Changtong Luo and Shao-Liang Zhang}
}

@misc{cranmer2023interpretablemachinelearningscience,
      title={Interpretable Machine Learning for Science with PySR and SymbolicRegression.jl}, 
      author={Miles Cranmer},
      year={2023},
      eprint={2305.01582},
      archivePrefix={arXiv},
      primaryClass={astro-ph.IM},
      url={https://arxiv.org/abs/2305.01582}, 
}

@incollection{McConaghy2011,
  doi = {10.1007/978-1-4614-1770-513},
note="Available at: \url{https://doi.org/10.1007/978-1-4614-1770-513}",
  year = {2011},
  publisher = {Springer New York},
  pages = {235--260},
  author = {Trent McConaghy},
  title = {{FFX}: Fast,  Scalable,  Deterministic Symbolic Regression Technology},
  booktitle = {Genetic and Evolutionary Computation}
}

@inproceedings{kantor2021simulated,
  title={Simulated annealing for symbolic regression},
  author={Kantor, Daniel and Von Zuben, Fernando J and de{ }Fran{\c{c}}a, Fabricio Olivetti},
  booktitle={Proceedings of the Genetic and Evolutionary Computation Conference},
  pages={592--599},
  year={2021}
}

@article{de2021interaction,
  title={Interaction-transformation symbolic regression with extreme learning machine},
  author={de{ }Fran{\c{c}}a, Fabricio Olivetti and de Lima, Maira Zabuscha},
  journal={Neurocomputing},
  volume={423},
  pages={609--619},
  year={2021},
  publisher={Elsevier}
}

@inproceedings{aldeia2021measuring,
  title={Measuring feature importance of symbolic regression models using partial effects},
  author={Aldeia, Guilherme Seidyo Imai and de{ }Fran{\c{c}}a, Fabr{\'\i}cio Olivetti},
  booktitle={Proceedings of the Genetic and Evolutionary Computation Conference},
  pages={750--758},
  year={2021}
}

@misc{jafari_survey_2021,
	title = {A {Survey} on {Locality} {Sensitive} {Hashing} {Algorithms} and their {Applications}},
note="Available at: \url{http://arxiv.org/abs/2102.08942}",
	language = {en},
	urldate = {2024-01-26},
	publisher = {arXiv},
	author = {Jafari, Omid and Maurya, Preeti and Nagarkar, Parth and Islam, Khandker Mushfiqul and Crushev, Chidambaram},
	month = feb,
	year = {2021},
	note = {arXiv:2102.08942 [cs]},
}

@misc{radwan_comparison_2024,
	title = {A {Comparison} of {Recent} {Algorithms} for {Symbolic} {Regression} to {Genetic} {Programming}},
	url = {http://arxiv.org/abs/2406.03585},
	doi = {10.48550/arXiv.2406.03585},
	language = {en},
	urldate = {2025-07-05},
	publisher = {arXiv},
	author = {Radwan, Yousef A. and Kronberger, Gabriel and Winkler, Stephan},
	month = jun,
	year = {2024},
	note = {arXiv:2406.03585 [cs]},
}

@article{LIU2024108986,
title = {A generalized grey model with symbolic regression algorithm and its application in predicting aircraft remaining useful life},
journal = {Engineering Applications of Artificial Intelligence},
volume = {136},
pages = {108986},
year = {2024},
issn = {0952-1976},
doi = {https://doi.org/10.1016/j.engappai.2024.108986},
url = {https://www.sciencedirect.com/science/article/pii/S0952197624011448},
author = {Lianyi Liu and Sifeng Liu and Yingjie Yang and Xiaojun Guo and Jinghe Sun}
}

@inproceedings{bomarito_bayesian_2022,
	address = {Boston Massachusetts},
	title = {Bayesian model selection for reducing bloat and overfitting in genetic programming for symbolic regression},
	isbn = {978-1-4503-9268-6},
note="Available at: \url{https://dl.acm.org/doi/10.1145/3520304.3528899}",
	doi = {10.1145/3520304.3528899},
	language = {en},
	urldate = {2024-01-26},
	booktitle = {Proceedings of the {Genetic} and {Evolutionary} {Computation} {Conference} {Companion}},
	publisher = {ACM},
	author = {Bomarito, G. F. and Leser, P. E. and Strauss, N. C. M. and Garbrecht, K. M. and Hochhalter, J. D.},
	month = jul,
	year = {2022},
	pages = {526--529},
}

@article{kulunchakov_creation_2017,
	title = {Creation of parametric rules to rewrite algebraic expressions in {Symbolic} {Regression}},
	volume = {3},
	issn = {22233792},
note="Available at: \url{http://jmlda.org/papers/doc/2017/no1/Kulunchakov2017RuleCreation.pdf}",
	doi = {10.21469/22233792.3.1.01},
	language = {en},
	number = {1},
	urldate = {2024-01-26},
	journal = {Machine Learning and Data Analysis},
  author={Kulunchakov, AS},
	year = {2017},
	pages = {6--19},
}

@article{gionis_similarity_nodate,
  title={Similarity search in high dimensions via hashing},
  author={Gionis, Aristides and Indyk, Piotr and Motwani, Rajeev and others},
  booktitle={Vldb},
  volume={99},
  number={6},
  pages={518--529},
  year={1999}
}

@inproceedings{helmuth_improving_2017,
	address = {Berlin Germany},
	title = {Improving generalization of evolved programs through automatic simplification},
	isbn = {978-1-4503-4920-8},
note="Available at: \url{https://dl.acm.org/doi/10.1145/3071178.3071330}",
	doi = {10.1145/3071178.3071330},
	language = {en},
	urldate = {2024-01-26},
	booktitle = {Proceedings of the {Genetic} and {Evolutionary} {Computation} {Conference}},
	publisher = {ACM},
	author = {Helmuth, Thomas and McPhee, Nicholas Freitag and Pantridge, Edward and Spector, Lee},
	month = jul,
	year = {2017},
	pages = {937--944},
}

@article{luke_comparison_2006,
	title = {A {Comparison} of {Bloat} {Control} {Methods} for {Genetic} {Programming}},
	volume = {14},
	issn = {1063-6560, 1530-9304},
note="Available at: \url{https://direct.mit.edu/evco/article/14/3/309-344/1242}",
	doi = {10.1162/evco.2006.14.3.309},
	language = {en},
	number = {3},
	urldate = {2024-01-26},
	journal = {Evolutionary Computation},
	author = {Luke, Sean and Panait, Liviu},
	month = sep,
	year = {2006},
	pages = {309--344},
}

@article{nguyen_semantic_2020,
	title = {Semantic approximation for reducing code bloat in {Genetic} {Programming}},
	volume = {58},
	issn = {22106502},
note="Available at: \url{https://linkinghub.elsevier.com/retrieve/pii/S2210650220303825}",
	doi = {10.1016/j.swevo.2020.100729},
	language = {en},
	urldate = {2024-01-26},
	journal = {Swarm and Evolutionary Computation},
	author = {Nguyen, Quang Uy and Chu, Thi Huong},
	month = nov,
	year = {2020},
	pages = {100729},
}

@inproceedings{10.1145/3583131.3590346,
author = {Fabricio Olivetti de{ }Fran{\c{c}}a and Gabriel Kronberger},
title = {Reducing Overparameterization of Symbolic Regression Models with Equality Saturation},
year = {2023},
isbn = {9798400701191},
publisher = {Association for Computing Machinery},
address = {New York, NY, USA},
note="Available at: \url{https://doi.org/10.1145/3583131.3590346}",
doi = {10.1145/3583131.3590346},
booktitle = {Proceedings of the Genetic and Evolutionary Computation Conference},
pages = {1064–1072},
numpages = {9},
location = {Lisbon, Portugal},
series = {GECCO '23}
}

@article{10.1145/502807.502808,
author = {Ch\'{a}vez, Edgar and Navarro, Gonzalo and Baeza-Yates, Ricardo and Marroqu\'{\i}n, Jos\'{e} Luis},
title = {Searching in metric spaces},
year = {2001},
issue_date = {September 2001},
publisher = {Association for Computing Machinery},
address = {New York, NY, USA},
volume = {33},
number = {3},
issn = {0360-0300},
note="Available at: \url{https://doi.org/10.1145/502807.502808}",
doi = {10.1145/502807.502808},
journal = {ACM Comput. Surv.},
month = {sep},
pages = {273–321},
numpages = {49}
}

@article{DEAP_JMLR2012, 
    author    = " F\'elix-Antoine Fortin and Fran{\c{c}}ois-Michel {De Rainville} and Marc-Andr\'e Gardner and Marc Parizeau and Christian Gagn\'e ",
    title     = { {DEAP}: Evolutionary Algorithms Made Easy },
    pages    = { 2171--2175 },
    volume    = { 13 },
    month     = { jul },
    year      = { 2012 },
    journal   = { Journal of Machine Learning Research }
}

@ARTICLE{2020SciPy-NMeth,
  author  = {Virtanen, Pauli and Gommers, Ralf and Oliphant, Travis E. and
            Haberland, Matt and Reddy, Tyler and Cournapeau, David and
            Burovski, Evgeni and Peterson, Pearu and Weckesser, Warren and
            Bright, Jonathan and {van der Walt}, St{\'e}fan J. and
            Brett, Matthew and Wilson, Joshua and Millman, K. Jarrod and
            Mayorov, Nikolay and Nelson, Andrew R. J. and Jones, Eric and
            Kern, Robert and Larson, Eric and Carey, C J and
            Polat, {\.I}lhan and Feng, Yu and Moore, Eric W. and
            {VanderPlas}, Jake and Laxalde, Denis and Perktold, Josef and
            Cimrman, Robert and Henriksen, Ian and Quintero, E. A. and
            Harris, Charles R. and Archibald, Anne M. and
            Ribeiro, Ant{\^o}nio H. and Pedregosa, Fabian and
            {van Mulbregt}, Paul and {SciPy 1.0 Contributors}},
  title   = {{{SciPy} 1.0: Fundamental Algorithms for Scientific
            Computing in Python}},
  journal = {Nature Methods},
  year    = {2020},
  volume  = {17},
  pages   = {261--272},
  adsurl  = {https://rdcu.be/b08Wh},
  doi     = {10.1038/s41592-019-0686-2},
}

@inproceedings{charikar2002similarity,
  title={Similarity estimation techniques from rounding algorithms},
  author={Charikar, Moses S},
  booktitle={Proceedings of the thiry-fourth annual ACM symposium on Theory of computing},
  pages={380--388},
  year={2002}
}

@article{moscato1989evolution,
  title={On evolution, search, optimization, genetic algorithms and martial arts: Towards memetic algorithms},
  author={Moscato, Pablo and others},
  journal={Caltech concurrent computation program, C3P Report},
  volume={826},
  number={1989},
  pages={37},
  year={1989}
}

@incollection{moscato2003gentle,
  title={A gentle introduction to memetic algorithms},
  author={Moscato, Pablo and Cotta, Carlos},
  booktitle={Handbook of metaheuristics},
  pages={105--144},
  year={2003},
  publisher={Springer}
}

@article{de2024srbenchplusplus,
  title={SRBench++: principled benchmarking of symbolic regression with domain-expert interpretation},
  author={de{ }Fran{\c{c}}a, FO and Virgolin, M and Kommenda, M and Majumder, MS and Cranmer, M and Espada, G and Ingelse, L and Fonseca, A and Landajuela, M and Petersen, B and others},
  journal={IEEE Transactions on Evolutionary Computation},
  year={2024},
  publisher={IEEE}
}

@article{lenat11984automated,
  author = {Lenat, Douglas B.},
  title = {Automated theory formation in mathematics},
  year = {1977},
  publisher = {Morgan Kaufmann Publishers Inc.},
  address = {San Francisco, CA, USA},
  booktitle = {Proceedings of the 5th International Joint Conference on Artificial Intelligence - Volume 2},
  pages = {833–842},
  numpages = {10},
  location = {Cambridge, USA},
  series = {IJCAI'77}
}

@inproceedings{langley1977bacon,
  title={BACON: A Production System That Discovers Empirical Laws.},
  author={Langley, Pat},
  booktitle={IJCAI},
  pages={344},
  year={1977}
}

@inproceedings{dvzeroski1993discovering,
  title={Discovering dynamics},
  author={D{\v{z}}eroski, Sa{\v{s}}o and Todorovski, Ljup{\v{c}}o},
  booktitle={Proc. tenth international conference on machine learning},
  pages={97--103},
  year={1993}
}

@Book{adaptationinnaturalartificial,
    title = { Adaptation in natural and artificial systems : an introductory analysis with applications to biology, control, and artificial intelligence },
    author = { Holland, John H.},
    publisher = { Ann Arbor : University of Michigan Press },
    year = { 1975 },
    type = { Book; Book/Illustrated },
    isbn = { 0472084607 },
    subjects = { Adaptation (Biology) -- Mathematical models },
    language = { English },
    note = { Includes index },
    catalogue-url = { https://trove.nla.gov.au/work/21158880 }
}

@article{holland_genetic_1992,
	title = {Genetic {Algorithms}},
	volume = {267},
	issn = {0036-8733},
	url = {https://www.jstor.org/stable/24939139},
	number = {1},
	urldate = {2025-08-28},
	journal = {Scientific American},
	author = {Holland, John H.},
	year = {1992},
	note = {Publisher: Scientific American, a division of Nature America, Inc.},
	pages = {66--73},
}

@article{cramer1985representation,
  title={A Representation for the Adaptive Generation of Simple Sequential Programs},
  author={CRAMER, NL},
  journal={ICGA-85},
  pages={183--187},
  year={1985}
}

@article{back1997handbook,
  title={Handbook of evolutionary computation},
  author={B{\"a}ck, Thomas and Fogel, David B and Michalewicz, Zbigniew},
  journal={Release},
  volume={97},
  number={1},
  pages={B1},
  year={1997}
}

@INPROCEEDINGS{selective_pressure,
  author={Back, T.},
  booktitle={Proceedings of the First IEEE Conference on Evolutionary Computation. IEEE World Congress on Computational Intelligence}, 
  title={Selective pressure in evolutionary algorithms: a characterization of selection mechanisms}, 
  year={1994},
  volume={},
  number={},
  pages={57-62 vol.1},
  doi={10.1109/ICEC.1994.350042}}

@inbook{Ye_2020,
   title={Benchmarking a $(\mu +\lambda )$ Genetic Algorithm with Configurable Crossover Probability},
   ISBN={9783030581152},
   ISSN={1611-3349},
   url={http://dx.doi.org/10.1007/978-3-030-58115-2\_49},
   DOI={10.1007/978-3-030-58115-2\_49},
   booktitle={Parallel Problem Solving from Nature - PPSN XVI},
   publisher={Springer International Publishing},
   author={Ye, Furong and Wang, Hao and Doerr, Carola and Bäck, Thomas},
   year={2020},
   pages={699-713} }

@article{Beyer2002,
  title = {Evolution strategies – A comprehensive introduction},
  volume = {1},
  ISSN = {1572-9796},
  url = {http://dx.doi.org/10.1023/A:1015059928466},
  DOI = {10.1023/a:1015059928466},
  number = {1},
  journal = {Natural Computing},
  publisher = {Springer Science and Business Media LLC},
  author = {Beyer,  Hans-Georg and Schwefel,  Hans-Paul},
  year = {2002},
  month = mar,
  pages = {3–52}
}

@article{simulatedannealing,
  author = {S. Kirkpatrick  and C. D. Gelatt  and M. P. Vecchi },
  title = {Optimization by Simulated Annealing},
  journal = {Science},
  volume = {220},
  number = {4598},
  pages = {671-680},
  year = {1983},
  doi = {10.1126/science.220.4598.671}
}

@article{dang_proof_2023,
	title = {A {Proof} {That} {Using} {Crossover} {Can} {Guarantee} {Exponential} {Speed}-{Ups} in {Evolutionary} {Multi}-{Objective} {Optimisation}},
	volume = {37},
	issn = {2374-3468, 2159-5399},
	url = {https://ojs.aaai.org/index.php/AAAI/article/view/26460},
	doi = {10.1609/aaai.v37i10.26460},
	language = {en},
	number = {10},
	urldate = {2025-07-05},
	journal = {Proceedings of the AAAI Conference on Artificial Intelligence},
	author = {Dang, Duc-Cuong and Opris, Andre and Salehi, Bahare and Sudholt, Dirk},
	month = jun,
	year = {2023},
	pages = {12390--12398},
}

@article{wolpert_no_1997,
	title = {No free lunch theorems for optimization},
	volume = {1},
	copyright = {https://ieeexplore.ieee.org/Xplorehelp/downloads/license-information/IEEE.html},
	issn = {1089778X},
	url = {http://ieeexplore.ieee.org/document/585893/},
	doi = {10.1109/4235.585893},
	language = {en},
	number = {1},
	urldate = {2025-08-28},
	journal = {IEEE Transactions on Evolutionary Computation},
	author = {Wolpert, D.H. and Macready, W.G.},
	month = apr,
	year = {1997},
	pages = {67--82},
}

@article{Keijzer2004,
  title = {Scaled Symbolic Regression},
  volume = {5},
  ISSN = {1573-7632},
  url = {http://dx.doi.org/10.1023/B:GENP.0000030195.77571.f9},
  DOI = {10.1023/b:genp.0000030195.77571.f9},
  number = {3},
  journal = {Genetic Programming and Evolvable Machines},
  publisher = {Springer Science and Business Media LLC},
  author = {Keijzer,  Maarten},
  year = {2004},
  month = sep,
  pages = {259–269}
}

@inproceedings{chen_relieving_2023,
	address = {Lisbon Portugal},
	title = {Relieving {Genetic} {Programming} from {Coefficient} {Learning} for {Symbolic} {Regression} via {Correlation} and {Linear} {Scaling}},
	isbn = {979-8-4007-0119-1},
	url = {https://dl.acm.org/doi/10.1145/3583131.3595918},
	doi = {10.1145/3583131.3595918},
	language = {en},
	urldate = {2025-09-13},
	booktitle = {Proceedings of the {Genetic} and {Evolutionary} {Computation} {Conference}},
	publisher = {ACM},
	author = {Chen, Qi and Xue, Bing and Banzhaf, Wolfgang and Zhang, Mengjie},
	month = jul,
	year = {2023},
	pages = {420--428},
}

@article{schmidt_distilling_2009,
	title = {Distilling {Free}-{Form} {Natural} {Laws} from {Experimental} {Data}},
	volume = {324},
	issn = {0036-8075, 1095-9203},
	url = {https://www.science.org/doi/10.1126/science.1165893},
	doi = {10.1126/science.1165893},
	language = {en},
	number = {5923},
	urldate = {2025-07-05},
	journal = {Science},
	author = {Schmidt, Michael and Lipson, Hod},
	month = apr,
	year = {2009},
	pages = {81--85},
}

@misc{molina_paradox_2025,
      title={The Paradox of Success in Evolutionary and Bioinspired Optimization: Revisiting Critical Issues, Key Studies, and Methodological Pathways}, 
      author={Daniel Molina and Javier Del Ser and Javier Poyatos and Francisco Herrera},
      year={2025},
      eprint={2501.07515},
      archivePrefix={arXiv},
      primaryClass={cs.NE},
      url={https://arxiv.org/abs/2501.07515}, 
}

@misc{martius_extrapolation_2016,
	title = {Extrapolation and learning equations},
	url = {http://arxiv.org/abs/1610.02995},
	doi = {10.48550/arXiv.1610.02995},
	urldate = {2025-08-28},
	publisher = {arXiv},
	author = {Martius, Georg and Lampert, Christoph H.},
	month = oct,
	year = {2016},
	note = {arXiv:1610.02995 [cs]},
}

@misc{wu_evolutionary_2024,
	title = {Evolutionary {Computation} in the {Era} of {Large} {Language} {Model}: {Survey} and {Roadmap}},
	shorttitle = {Evolutionary {Computation} in the {Era} of {Large} {Language} {Model}},
	url = {http://arxiv.org/abs/2401.10034},
	doi = {10.48550/arXiv.2401.10034},
	language = {en},
	urldate = {2025-07-05},
	publisher = {arXiv},
	author = {Wu, Xingyu and Wu, Sheng-hao and Wu, Jibin and Feng, Liang and Tan, Kay Chen},
	month = may,
	year = {2024},
	note = {arXiv:2401.10034 [cs]}
}

@misc{lehman_evolution_2022,
	title = {Evolution through {Large} {Models}},
	url = {http://arxiv.org/abs/2206.08896},
	doi = {10.48550/arXiv.2206.08896},
	language = {en},
	urldate = {2025-07-05},
	publisher = {arXiv},
	author = {Lehman, Joel and Gordon, Jonathan and Jain, Shawn and Ndousse, Kamal and Yeh, Cathy and Stanley, Kenneth O.},
	month = jun,
	year = {2022},
	note = {arXiv:2206.08896 [cs]},
}

@misc{voigt2025analyzinggeneralizationpretrainedsymbolic,
      title={Analyzing Generalization in Pre-Trained Symbolic Regression}, 
      author={Henrik Voigt and Paul Kahlmeyer and Kai Lawonn and Michael Habeck and Joachim Giesen},
      year={2025},
      eprint={2509.19849},
      archivePrefix={arXiv},
      primaryClass={cs.LG},
      url={https://arxiv.org/abs/2509.19849}, 
}

@misc{constantin_statistical_2024,
	title = {Statistical {Patterns} in the {Equations} of {Physics} and the {Emergence} of a {Meta}-{Law} of {Nature}},
	url = {http://arxiv.org/abs/2408.11065},
	doi = {10.48550/arXiv.2408.11065},
	urldate = {2025-07-05},
	publisher = {arXiv},
	author = {Constantin, Andrei and Bartlett, Deaglan and Desmond, Harry and Ferreira, Pedro G.},
	month = aug,
	year = {2024},
	note = {arXiv:2408.11065 [physics]},
}

@article{eplex,
    author = {La{ }Cava, William and Helmuth, Thomas and Spector, Lee and Moore, Jason H.},
    title = {A Probabilistic and Multi-Objective Analysis of Lexicase Selection and epsilon-Lexicase Selection},
    journal = {Evolutionary Computation},
    volume = {27},
    number = {3},
    pages = {377-402},
    year = {2019},
    month = {09},
    issn = {1063-6560},
    doi = {10.1162/evco\_a\_00224},
    url = {https://doi.org/10.1162/evco\_a\_00224},
    eprint = {https://direct.mit.edu/evco/article-pdf/27/3/377/1858632/evco\_a\_00224.pdf},
}

@inproceedings{espada2022data,
  author={Guilherme Espada and Leon Ingelse and Paulo Canelas and Pedro Barbosa and Alcides Fonseca},
  editor    = {Bernhard Scholz and Yukiyoshi Kameyama},
  title={Datatypes as a More Ergonomic Frontend for Grammar-Guided Genetic Programming},
  booktitle = {{GPCE} '22: Concepts and Experiences, Auckland, NZ, December 6 - 7, 2022},
  pages     = {1},
  publisher = {{ACM}},
  year      = {2022},
}

@inproceedings{10.1145/3377930.3390237,
  author = {Dick, Grant and Owen, Caitlin A. and Whigham, Peter A.},
  title = {Feature standardisation and coefficient optimisation for effective symbolic regression},
  year = {2020},
  isbn = {9781450371285},
  publisher = {Association for Computing Machinery},
  address = {New York, NY, USA},
  url = {https://doi.org/10.1145/3377930.3390237},
  doi = {10.1145/3377930.3390237},
  booktitle = {Proceedings of the 2020 Genetic and Evolutionary Computation Conference},
  pages = {306–314},
  numpages = {9},
  location = {Canc\'{u}n, Mexico},
  series = {GECCO '20}
}

@inproceedings{tir,
author={de{ }Fran{\c{c}}a, Fabr{\'\i}cio Olivetti},
title = {Transformation-interaction-rational representation for symbolic regression},
year = {2022},
isbn = {9781450392372},
publisher = {Association for Computing Machinery},
address = {New York, NY, USA},
url = {https://doi.org/10.1145/3512290.3528695},
doi = {10.1145/3512290.3528695},
booktitle = {Proceedings of the Genetic and Evolutionary Computation Conference},
pages = {920-928},
numpages = {9},
location = {Boston, Massachusetts},
series = {GECCO '22}
}

@misc{cava2018learning,
	title = {Learning concise representations for regression by evolving networks of trees},
note="Available at: \url{http://arxiv.org/abs/1807.00981}",
	language = {en},
	urldate = {2024-01-26},
	publisher = {arXiv},
	author = {La{ }Cava, William and Singh, Tilak Raj and Taggart, James and Suri, Srinivas and Moore, Jason H.},
	month = mar,
	year = {2019},
	note = {arXiv:1807.00981 [cs]},
}

@InProceedings{pmlrv80sahoo18a,
  title = 	 {Learning Equations for Extrapolation and Control},
  author =       {Sahoo, Subham and Lampert, Christoph and Martius, Georg},
  booktitle = 	 {Proceedings of the 35th International Conference on Machine Learning},
  pages = 	 {4442--4450},
  year = 	 {2018},
  editor = 	 {Dy, Jennifer and Krause, Andreas},
  volume = 	 {80},
  series = 	 {Proceedings of Machine Learning Research},
  month = 	 {10--15 Jul},
  publisher =    {PMLR},
  url = 	 {https://proceedings.mlr.press/v80/sahoo18a.html}
}

@article{Kartelj2023,
  title = {RILS-ROLS: robust symbolic regression via iterated local search and ordinary least squares},
  volume = {10},
  ISSN = {2196-1115},
  url = {http://dx.doi.org/10.1186/s40537-023-00743-2},
  DOI = {10.1186/s40537-023-00743-2},
  number = {1},
  journal = {Journal of Big Data},
  publisher = {Springer Science and Business Media LLC},
  author = {Kartelj,  Aleksandar and Djukanović,  Marko},
  year = {2023},
  month = may 
}

@misc{jin2020bayesiansymbolicregression,
      title={Bayesian Symbolic Regression}, 
      author={Ying Jin and Weilin Fu and Jian Kang and Jiadong Guo and Jian Guo},
      year={2020},
      eprint={1910.08892},
      archivePrefix={arXiv},
      primaryClass={stat.ME},
      url={https://arxiv.org/abs/1910.08892}, 
}

@article{StopExplainingBlackBoxMachineLearningModels,
    archivePrefix = {arXiv},
    arxivId = {1811.10154},
    author = {Rudin, Cynthia},
    doi = {10.1038/s42256-019-0048-x},
    eprint = {1811.10154},
    issn = {25225839},
    journal = {Nature Machine Intelligence},
    number = {5},
    pages = {206--215},
    title = {{Stop explaining black box machine learning models for high stakes decisions and use interpretable models instead}},
    volume = {1},
    year = {2019}
}

@article{DoWeNeedHundredsOfClassifiers,
  author  = {Manuel Fern{\'{a}}ndez-Delgado and Eva Cernadas and Sen{{\'e}}n Barro and Dinani Amorim},
  title   = {Do we Need Hundreds of Classifiers to Solve Real World Classification Problems?},
  journal = {Journal of Machine Learning Research},
  year    = {2014},
  volume  = {15},
  number  = {90},
  pages   = {3133-3181},
  url     = {http://jmlr.org/papers/v15/delgado14a.html}
}

@article{Kronberger2022,
  title = {Shape-Constrained Symbolic Regression—Improving Extrapolation with Prior Knowledge},
  volume = {30},
  ISSN = {1530-9304},
  url = {http://dx.doi.org/10.1162/evco\_a\_00294},
  DOI = {10.1162/evco\_a\_00294},
  number = {1},
  journal = {Evolutionary Computation},
  publisher = {MIT Press},
  author = {Kronberger,  G. and de{ }Franca,  F. O. and Burlacu,  B. and Haider,  C. and Kommenda,  M.},
  year = {2022},
  pages = {75-98}
}

@article{martinek44shape,
  title={Shape Constraints in Symbolic Regression using Penalized Least Squares},
  author={Martinek, Viktor and Reuter, Julia and Frotscher, Ophelia and Mostaghim, Sanaz and Richter, Markus and Herzog, Roland},
  journal={arXiv preprint arXiv:2405.20800},
  year={2024}
}

@inproceedings{randall2022bingo,
  title={Bingo: a customizable framework for symbolic regression with genetic programming},
  author={Randall, David L and Townsend, Tyler S and Hochhalter, Jacob D and Bomarito, Geoffrey F},
  booktitle={Proceedings of the Genetic and Evolutionary Computation Conference Companion},
  pages={2282--2288},
  year={2022}
}

@misc{haut_correlation_2022,
	title = {Correlation versus {RMSE} {Loss} {Functions} in {Symbolic} {Regression} {Tasks}},
	url = {http://arxiv.org/abs/2205.15990},
	doi = {10.48550/arXiv.2205.15990},
	urldate = {2025-07-05},
	publisher = {arXiv},
	author = {Haut, Nathan and Banzhaf, Wolfgang and Punch, Bill},
	month = jul,
	year = {2022},
	note = {arXiv:2205.15990 [cs]},
}

@article{10.1145/3236009,
author = {Guidotti, Riccardo and Monreale, Anna and Ruggieri, Salvatore and Turini, Franco and Giannotti, Fosca and Pedreschi, Dino},
title = {A Survey of Methods for Explaining Black Box Models},
year = {2018},
issue_date = {September 2019},
publisher = {Association for Computing Machinery},
address = {New York, NY, USA},
volume = {51},
number = {5},
issn = {0360-0300},
url = {https://doi.org/10.1145/3236009},
doi = {10.1145/3236009},
journal = {ACM Comput. Surv.},
month = aug,
articleno = {93},
numpages = {42}
}

@article{vladislavleva_order_2009,
	title = {Order of {Nonlinearity} as a {Complexity} {Measure} for {Models} {Generated} by {Symbolic} {Regression} via {Pareto} {Genetic} {Programming}},
	volume = {13},
	issn = {1941-0026},
	url = {https://ieeexplore.ieee.org/document/4632147/},
	doi = {10.1109/TEVC.2008.926486},
	number = {2},
	urldate = {2025-08-29},
	journal = {IEEE Transactions on Evolutionary Computation},
	author = {Vladislavleva, Ekaterina J. and Smits, Guido F. and den Hertog, Dick},
	month = apr,
	year = {2009},
	pages = {333--349},
}

@inbook{Schmidt2010,
  title = {Age-Fitness Pareto Optimization},
  ISBN = {9781441977472},
  ISSN = {1932-0167},
  url = {http://dx.doi.org/10.1007/978-1-4419-7747-2\_8},
  DOI = {10.1007/978-1-4419-7747-2\_8},
  booktitle = {Genetic Programming Theory and Practice VIII},
  publisher = {Springer New York},
  author = {Schmidt,  Michael and Lipson,  Hod},
  year = {2010},
  month = oct,
  pages = {129–146}
}

@book{ehrgott_multicriteria_2005,
	address = {Berlin Heidelberg New York},
	edition = {Second edition},
	title = {Multicriteria optimization: with 88 figures and 12 tables},
	isbn = {978-3-540-21398-7 978-3-540-27659-3},
	shorttitle = {Multicriteria optimization},
	language = {en},
	publisher = {Springer},
	author = {Ehrgott, Matthias},
	year = {2005},
}

@article{emmerich_tutorial_2018,
	title = {A tutorial on multiobjective optimization: fundamentals and evolutionary methods},
	volume = {17},
	issn = {1572-9796},
	shorttitle = {A tutorial on multiobjective optimization},
	url = {https://doi.org/10.1007/s11047-018-9685-y},
	doi = {10.1007/s11047-018-9685-y},
	language = {en},
	number = {3},
	urldate = {2025-08-27},
	journal = {Natural Computing},
	author = {Emmerich, Michael T. M. and Deutz, André H.},
	month = sep,
	year = {2018},
	pages = {585--609},
}

@book{mcdm,
author = {Deb, Kalyanmoy and Kalyanmoy, Deb},
title = {Multi-Objective Optimization Using Evolutionary Algorithms},
year = {2001},
isbn = {047187339X},
publisher = {John Wiley \& Sons, Inc.},
address = {USA}
}

@article{goldberg1989messy,
  title={Messy genetic algorithms: Motivation, analysis, and first results},
  author={Goldberg, David E and Korb, Bradley and Deb, Kalyanmoy},
  journal={Complex systems},
  volume={3},
  number={5},
  pages={493--530},
  year={1989},
  publisher={Complex Systems Publications, Champaign, IL, USA}
}

@inproceedings{keijzer2003improving,
  title={Improving symbolic regression with interval arithmetic and linear scaling},
  author={Keijzer, Maarten},
  booktitle={European Conference on Genetic Programming},
  pages={70--82},
  year={2003},
  organization={Springer}
}

@inproceedings{arnaldo2014multiple,
  title={Multiple regression genetic programming},
  author={Arnaldo, Ignacio and Krawiec, Krzysztof and O'Reilly, Una-May},
  booktitle={Proceedings of the 2014 annual conference on genetic and evolutionary computation},
  pages={879--886},
  year={2014}
}

@inbook{neutrality,
author = {Hu, Ting and Banzhaf, Wolfgang and Ochoa, Gabriela},
title = {How Neutrality Shapes Evolution: Simplicity Bias and Search},
year = {2025},
isbn = {9798400714658},
publisher = {Association for Computing Machinery},
address = {New York, NY, USA},
url = {https://doi.org/10.1145/3712256.3726454},
booktitle = {Proceedings of the Genetic and Evolutionary Computation Conference},
pages = {1008–1016},
numpages = {9}
}

@incollection{winkler_how_2024,
	address = {Singapore},
	title = {How the {Combinatorics} of {Neutral} {Spaces} {Leads} {Genetic} {Programming} to {Discover} {Simple} {Solutions}},
	isbn = {978-981-99-8412-1 978-981-99-8413-8},
	url = {https://link.springer.com/10.1007/978-981-99-8413-8\_4},
	language = {en},
	urldate = {2025-09-02},
	booktitle = {Genetic {Programming} {Theory} and {Practice} {XX}},
	publisher = {Springer Nature Singapore},
	author = {Banzhaf, Wolfgang and Hu, Ting and Ochoa, Gabriela},
	editor = {Winkler, Stephan and Trujillo, Leonardo and Ofria, Charles and Hu, Ting},
	year = {2024},
	doi = {10.1007/978-981-99-8413-8\_4},
	note = {Series Title: Genetic and Evolutionary Computation},
	pages = {65--86},
}

@article{Olson2017PMLB,
    author="Olson, Randal S. and La{ }Cava, William and Orzechowski, Patryk and Urbanowicz, Ryan J. and Moore, Jason H.",
    title="PMLB: a large benchmark suite for machine learning evaluation and comparison",
    journal="BioData Mining",
    year="2017",
    month="Dec",
    day="11",
    volume="10",
    number="1",
    pages="36",
    issn="1756-0381",
    doi="10.1186/s13040-017-0154-4",
    url="https://doi.org/10.1186/s13040-017-0154-4"
}

@article{thing2025cp3,
  title={cp3-bench: a tool for benchmarking symbolic regression algorithms demonstrated with cosmology},
  author={Thing, ME and Koksbang, SM},
  journal={Journal of Cosmology and Astroparticle Physics},
  volume={2025},
  number={01},
  pages={040},
  year={2025},
  publisher={IOP Publishing}
}

@inproceedings{dick2022genetic,
  title={Genetic programming, standardisation, and stochastic gradient descent revisited: Initial findings on srbench},
  author={Dick, Grant},
  booktitle={Proceedings of the Genetic and Evolutionary Computation Conference Companion},
  pages={2265--2273},
  year={2022}
}

@article{whitley1999island,
  title={The island model genetic algorithm: On separability, population size and convergence},
  author={Whitley, Darrell and Rana, Soraya and Heckendorn, Robert B},
  journal={Journal of computing and information technology},
  volume={7},
  number={1},
  pages={33--47},
  year={1999},
  publisher={Sveu{\v{c}}ili{\v{s}}te u Zagrebu Sveu{\v{c}}ili{\v{s}}ni ra{\v{c}}unski centar}
}

@misc{reis_benchmarking_2024,
	title = {Benchmarking symbolic regression constant optimization schemes},
	url = {http://arxiv.org/abs/2412.02126},
	doi = {10.48550/arXiv.2412.02126},
	urldate = {2025-09-13},
	publisher = {arXiv},
	author = {Reis, L. G. A. dos and Caminha, V. L. P. S. and Penna, T. J. P.},
	month = dec,
	year = {2024},
	note = {arXiv:2412.02126 [cs]},
}

@misc{franca_improving_2025,
	title = {Improving {Genetic} {Programming} for {Symbolic} {Regression} with {Equality} {Graphs}},
	url = {http://arxiv.org/abs/2501.17848},
	doi = {10.1145/3712256.3726383},
	language = {en},
	urldate = {2025-07-05},
	author = {de{ }Fran{\c{c}}a, Fabricio Olivetti and Kronberger, Gabriel},
	month = apr,
	year = {2025},
	note = {arXiv:2501.17848 [cs]},
}

@article{gujarati2003multicollinearity,
  title={Multicollinearity: What happens if the regressors are correlated},
  author={Gujarati, Damodar and Porter, Dawn},
  journal={Basic econometrics},
  volume={363},
  year={2003},
  publisher={McGraw-Hill New York}
}

@article{kronberger_effects_2025,
	title = {Effects of reducing redundant parameters in parameter optimization for symbolic regression using genetic programming},
	volume = {129},
	issn = {07477171},
	url = {https://linkinghub.elsevier.com/retrieve/pii/S0747717124001172},
	doi = {10.1016/j.jsc.2024.102413},
	language = {en},
	urldate = {2025-09-13},
	journal = {Journal of Symbolic Computation},
	author = {Kronberger, Gabriel and Olivetti de{ }Fran{\c{c}}a, Fabrício},
	month = jul,
	year = {2025},
	pages = {102413}
}

@techreport{brooks1989airfoil,
  title={Airfoil self-noise and prediction},
  author={Brooks, Thomas F and Pope, D Stuart and Marcolini, Michael A},
  year={1989}
}

@misc{geiger_was_2025,
	title = {Was {Tournament} {Selection} {All} {We} {Ever} {Needed}? {A} {Critical} {Reflection} on {Lexicase} {Selection}},
	shorttitle = {Was {Tournament} {Selection} {All} {We} {Ever} {Needed}?},
	url = {http://arxiv.org/abs/2502.18093},
	doi = {10.48550/arXiv.2502.18093},
	urldate = {2025-09-16},
	publisher = {arXiv},
	author = {Geiger, Alina and Briesch, Martin and Sobania, Dominik and Rothlauf, Franz},
	month = feb,
	year = {2025},
	note = {arXiv:2502.18093 [cs]},
}

@article{geiger_performance_2025,
	title = {A {Performance} {Analysis} of {Lexicase}-{Based} and {Traditional} {Selection} {Methods} in {GP} for {Symbolic} {Regression}},
	issn = {2688-3007},
	url = {https://dl.acm.org/doi/10.1145/3725860},
	doi = {10.1145/3725860},
	language = {en},
	urldate = {2025-09-16},
	journal = {ACM Transactions on Evolutionary Learning and Optimization},
	author = {Geiger, Alina and Sobania, Dominik and Rothlauf, Franz},
	month = mar,
	year = {2025},
	pages = {3725860},
}

@inproceedings{helmuth_benchmarking_2020,
	address = {Cancún Mexico},
	title = {Benchmarking parent selection for program synthesis by genetic programming},
	isbn = {978-1-4503-7127-8},
	url = {https://dl.acm.org/doi/10.1145/3377929.3389987},
	doi = {10.1145/3377929.3389987},
	language = {en},
	urldate = {2025-07-05},
	booktitle = {Proceedings of the 2020 {Genetic} and {Evolutionary} {Computation} {Conference} {Companion}},
	publisher = {ACM},
	author = {Helmuth, Thomas and Abdelhady, Amr},
	month = jul,
	year = {2020},
	pages = {237--238},
}

@inproceedings{dolson_calculating_2023,
	address = {Lisbon Portugal},
	title = {Calculating lexicase selection probabilities is {NP}-{Hard}},
	isbn = {979-8-4007-0119-1},
	url = {https://dl.acm.org/doi/10.1145/3583131.3590356},
	doi = {10.1145/3583131.3590356},
	language = {en},
	urldate = {2025-09-16},
	booktitle = {Proceedings of the {Genetic} and {Evolutionary} {Computation} {Conference}},
	publisher = {ACM},
	author = {Dolson, Emily},
	month = jul,
	year = {2023},
	pages = {1575--1583},
}

@article{demvsar2006statistical,
  title={Statistical comparisons of classifiers over multiple data sets},
  author={Dem{\v{s}}ar, Janez},
  journal={Journal of Machine learning research},
  volume={7},
  number={Jan},
  pages={1--30},
  year={2006}
}

@inbook{Ingelse2023,
  title = {Domain-Aware Feature Learning with Grammar-Guided Genetic Programming},
  ISBN = {9783031295737},
  ISSN = {1611-3349},
  url = {http://dx.doi.org/10.1007/978-3-031-29573-7\_15},
  DOI = {10.1007/978-3-031-29573-7\_15},
  booktitle = {Genetic Programming},
  publisher = {Springer Nature Switzerland},
  author = {Ingelse,  Leon and Fonseca,  Alcides},
  year = {2023},
  pages = {227–243}
}

@book{bonnet1764contemplation,
  title={Contemplation de la nature},
  author={Bonnet, C.},
  number={v. 2},
  series={Contemplation de la nature},
  url={https://books.google.com/books?id=Sm8GAAAAQAAJ},
  year={1764},
  publisher={M.M. Rey}
}

@article{Hubble1929,
  title = {A relation between distance and radial velocity among extra-galactic nebulae},
  volume = {15},
  ISSN = {1091-6490},
  url = {http://dx.doi.org/10.1073/pnas.15.3.168},
  DOI = {10.1073/pnas.15.3.168},
  number = {3},
  journal = {Proceedings of the National Academy of Sciences},
  publisher = {Proceedings of the National Academy of Sciences},
  author = {Hubble,  Edwin},
  year = {1929},
  month = mar,
  pages = {168–173}
}

@book{keplerHarmonicesMundi1619,
  title = {Harmonices {{Mundi}}},
  author = {Kepler, Johannes},
  year = {1619},
  publisher = {{Lincii Austriæ}}
}

@article{leavittPeriods25Variable1912,
  title = {Periods of 25 {{Variable Stars}} in the {{Small Magellanic Cloud}}.},
  author = {Leavitt, Henrietta S. and Pickering, Edward C.},
  year = {1912},
  month = mar,
  journal = {Harvard College Observatory Circular},
  volume = {173},
  pages = {1--3},
  urldate = {2022-01-26}
}

@article{tullyNewMethodDetermining1977,
  title = {A New Method of Determining Distance to Galaxies.},
  author = {Tully, R. B. and Fisher, J. R.},
  year = {1977},
  month = feb,
  journal = {Astronomy and Astrophysics},
  volume = {500},
  pages = {105--117},
  issn = {0004-6361},
  urldate = {2022-01-26},
  langid = {english}
}

@article{dolan2002benchmarking,
  title={Benchmarking optimization software with performance profiles},
  author={Dolan, Elizabeth D and Mor{\'e}, Jorge J},
  journal={Mathematical programming},
  volume={91},
  pages={201--213},
  year={2002},
  publisher={Springer}
}

@article{10.7717/peerj-cs.103,
     title = {SymPy: symbolic computing in Python},
     author = {Meurer, Aaron and Smith, Christopher P. and Paprocki, Mateusz and {\v{c}}ert\'{i}k, Ond{\v{r}}ej and Kirpichev, Sergey B. and Rocklin, Matthew and Kumar, AMiT and Ivanov, Sergiu and Moore, Jason K. and Singh, Sartaj and Rathnayake, Thilina and Vig, Sean and Granger, Brian E. and Muller, Richard P. and Bonazzi, Francesco and Gupta, Harsh and Vats, Shivam and Johansson, Fredrik and Pedregosa, Fabian and Curry, Matthew J. and Terrel, Andy R. and Rou{\v{c}}ka, {\v{S}}t{\v{e}}p\'{a}n and Saboo, Ashutosh and Fernando, Isuru and Kulal, Sumith and Cimrman, Robert and Scopatz, Anthony},
     year = 2017,
     month = jan,
     volume = 3,
     pages = {e103},
     journal = {PeerJ Computer Science},
     issn = {2376-5992},
     doi = {10.7717/peerj-cs.103}
    }

@misc{hooker2020hardwarelottery,
      title={The Hardware Lottery}, 
      author={Sara Hooker},
      year={2020},
      eprint={2009.06489},
      archivePrefix={arXiv},
      primaryClass={cs.CY},
      url={https://arxiv.org/abs/2009.06489}, 
}

@article{Budennyy2022,
  title = {eco2AI: Carbon Emissions Tracking of Machine Learning Models as the First Step Towards Sustainable AI},
  volume = {106},
  ISSN = {1531-8362},
  url = {http://dx.doi.org/10.1134/S1064562422060230},
  DOI = {10.1134/s1064562422060230},
  number = {S1},
  journal = {Doklady Mathematics},
  publisher = {Pleiades Publishing Ltd},
  author = {Budennyy,  S. A. and Lazarev,  V. D. and Zakharenko,  N. N. and Korovin,  A. N. and Plosskaya,  O. A. and Dimitrov,  D. V. and Akhripkin,  V. S. and Pavlov,  I. V. and Oseledets,  I. V. and Barsola,  I. S. and Egorov,  I. V. and Kosterina,  A. A. and Zhukov,  L. E.},
  year = {2022},
  month = dec,
  pages = {S118–S128}
}

@article{la2016inference,
  title={Inference of compact nonlinear dynamic models by epigenetic local search},
  author={La{ }Cava, William and Danai, Kourosh and Spector, Lee},
  journal={Engineering Applications of Artificial Intelligence},
  volume={55},
  pages={292--306},
  year={2016},
  publisher={Elsevier}
}

@misc{brush_multi_armed_bandits,
  author       = {CavaLab},
  title        = {Brush: An Interpretable Machine Learning Library},
  year         = {2025},
  url          = {https://github.com/cavalab/brush/},
  note         = {Accessed: 2025-03-29},
}

@misc{Stephens2025trevorstephens,
	author = {Stephens, Trevor and van Kemenade, Hugo and Rai, Anshul and Price, Ben and Watson, Colin and Bell, Ian and McDermott, James and de Vos, Nico and Niculae, Stefan and Di Caprio, Ulderico and {sun ao}},
	year = {2022},
	month = {may 3},
	title = {trevorstephens/gplearn},
	url = {https://github.com/trevorstephens/gplearn},
	howpublished = {https://github.com/trevorstephens/gplearn},
}

@InProceedings{pmlr-v139-biggio21a,
  title = 	 {Neural Symbolic Regression that scales},
  author =       {Biggio, Luca and Bendinelli, Tommaso and Neitz, Alexander and Lucchi, Aurelien and Parascandolo, Giambattista},
  booktitle = 	 {Proceedings of the 38th International Conference on Machine Learning},
  pages = 	 {936--945},
  year = 	 {2021},
  editor = 	 {Meila, Marina and Zhang, Tong},
  volume = 	 {139},
  series = 	 {Proceedings of Machine Learning Research},
  month = 	 {18--24 Jul},
  publisher =    {PMLR},
  url = 	 {https://proceedings.mlr.press/v139/biggio21a.html}
}

@misc{broløs2021approachsymbolicregressionusing,
      title={An Approach to Symbolic Regression Using Feyn}, 
      author={Kevin René Broløs and Meera Vieira Machado and Chris Cave and Jaan Kasak and Valdemar Stentoft-Hansen and Victor Galindo Batanero and Tom Jelen and Casper Wilstrup},
      year={2021},
      eprint={2104.05417},
      archivePrefix={arXiv},
      primaryClass={cs.LG},
      url={https://arxiv.org/abs/2104.05417}
}

@inproceedings{dacosta_adaptive_2008,
	address = {Atlanta GA USA},
	title = {Adaptive operator selection with dynamic multi-armed bandits},
	isbn = {978-1-60558-130-9},
	url = {https://dl.acm.org/doi/10.1145/1389095.1389272},
	doi = {10.1145/1389095.1389272},
	language = {en},
	urldate = {2025-07-05},
	booktitle = {Proceedings of the 10th annual conference on {Genetic} and evolutionary computation},
	publisher = {ACM},
	author = {DaCosta, Luis and Fialho, Alvaro and Schoenauer, Marc and Sebag, Michèle},
	month = jul,
	year = {2008},
	pages = {913--920},
}

@inproceedings{thanh_multi-armed_2021,
	address = {Kraków, Poland},
	title = {Multi-{Armed} {Bandits} for {Many}-{Task} {Evolutionary} {Optimization}},
	copyright = {https://ieeexplore.ieee.org/Xplorehelp/downloads/license-information/IEEE.html},
	isbn = {978-1-7281-8393-0},
	url = {https://ieeexplore.ieee.org/document/9504691/},
	doi = {10.1109/CEC45853.2021.9504691},
	language = {en},
	urldate = {2025-07-05},
	booktitle = {2021 {IEEE} {Congress} on {Evolutionary} {Computation} ({CEC})},
	publisher = {IEEE},
	author = {Thanh, Le Tien and Van Cuong, Le and Thang, Ta Bao and Thi Thanh Binh, Huynh},
	month = jun,
	year = {2021},
	pages = {1664--1671},
}

@inproceedings{belluz_operator_2015,
	address = {Madrid Spain},
	title = {Operator {Selection} using {Improved} {Dynamic} {Multi}-{Armed} {Bandit}},
	isbn = {978-1-4503-3472-3},
	url = {https://dl.acm.org/doi/10.1145/2739480.2754712},
	doi = {10.1145/2739480.2754712},
	language = {en},
	urldate = {2025-07-05},
	booktitle = {Proceedings of the 2015 {Annual} {Conference} on {Genetic} and {Evolutionary} {Computation}},
	publisher = {ACM},
	author = {Belluz, Jany and Gaudesi, Marco and Squillero, Giovanni and Tonda, Alberto},
	month = jul,
	year = {2015},
	pages = {1311--1317},
}

@inproceedings{burlacu2019hash,
  title={Hash-based tree similarity and simplification in genetic programming for symbolic regression},
  author={Burlacu, Bogdan and Kammerer, Lukas and Affenzeller, Michael and Kronberger, Gabriel},
  booktitle={International conference on computer aided systems theory},
  pages={361--369},
  year={2019},
  organization={Springer}
}

@article{10.1145/3572895,
author = {Smith-Miles, Kate and Mu\~{n}oz, Mario Andr\'{e}s},
title = {Instance Space Analysis for Algorithm Testing: Methodology and Software Tools},
year = {2023},
issue_date = {December 2023},
publisher = {Association for Computing Machinery},
address = {New York, NY, USA},
volume = {55},
number = {12},
issn = {0360-0300},
url = {https://doi.org/10.1145/3572895},
doi = {10.1145/3572895},
journal = {ACM Comput. Surv.},
month = mar,
articleno = {255},
numpages = {31},
}

@inproceedings{10.1145/3292500.3330701,
author = {Akiba, Takuya and Sano, Shotaro and Yanase, Toshihiko and Ohta, Takeru and Koyama, Masanori},
title = {Optuna: A Next-generation Hyperparameter Optimization Framework},
year = {2019},
isbn = {9781450362016},
publisher = {Association for Computing Machinery},
address = {New York, NY, USA},
url = {https://doi.org/10.1145/3292500.3330701},
doi = {10.1145/3292500.3330701},
booktitle = {Proceedings of the 25th ACM SIGKDD International Conference on Knowledge Discovery \& Data Mining},
pages = {2623-2631},
numpages = {9},
location = {Anchorage, AK, USA},
series = {KDD '19}
}

@software{codecarbon2024,
  author       = {Benoit Courty and
                  Victor Schmidt and
                  Sasha Luccioni and
                  Kamal Goyal and
                  Marion Coutarel and
                  Boris Feld and
                  {J{\'e}r{\'e}my Lecourt} and
                  Liam Connell and
                  Amine Saboni and
                  Inimaz and
                  Supatomic and
                  {Mathilde L{\'e}val} and
                  Luis Blanche and
                  Alexis Cruveiller and
                  Oumina Sara and
                  Franklin Zhao and
                  Aditya Joshi and
                  Alexis Bogroff and
                  Hugues de Lavoreille and
                  Niko Laskaris and
                  Edoardo Abati and
                  Douglas Blank and
                  Ziyao Wang and
                  Armin Catovic and
                  Marc Alen{\c{c}}on and
                  {Micha{\l} St{\k{e}}ch{\l}y} and
                  Christian Bauer and
                  {Lucas Ot{\'a}vio N. de Ara{\'u}jo} and
                  JPW and
                  MinervaBooks},
  title        = {mlco2/codecarbon: v2.4.1},
  month        = may,
  year         = {2024},
  publisher    = {Zenodo},
  version      = {v2.4.1},
  doi          = {10.5281/zenodo.11171501},
  url          = {https://doi.org/10.5281/zenodo.11171501}
}

@inbook{10.1145/3712256.3726385,
author = {de{ }Fran{\c{c}}a, Fabricio Olivetti and Kronberger, Gabriel},
title = {rEGGression: an Interactive and Agnostic Tool for the Exploration of Symbolic Regression Models},
year = {2025},
isbn = {9798400714658},
publisher = {Association for Computing Machinery},
address = {New York, NY, USA},
url = {https://doi.org/10.1145/3712256.3726385},
booktitle = {Proceedings of the Genetic and Evolutionary Computation Conference},
pages = {4–12},
numpages = {9}
}

@INPROCEEDINGS{8477951,
  author={Imai Aldeia, Guilherme Seidyo and de Franca, Fabrício Olivetti},
  booktitle={2018 IEEE Congress on Evolutionary Computation (CEC)}, 
  title={Lightweight Symbolic Regression with the Interaction - Transformation Representation}, 
  year={2018},
  volume={},
  number={},
  pages={1-8},
  doi={10.1109/CEC.2018.8477951}}

@inproceedings{10.1007/978-3-031-14721-0_34,
author = {Helmuth, Thomas and Lengler, Johannes and La Cava, William},
title = {Population Diversity Leads to Short Running Times of Lexicase Selection},
year = {2022},
isbn = {978-3-031-14720-3},
publisher = {Springer-Verlag},
address = {Berlin, Heidelberg},
url = {https://doi.org/10.1007/978-3-031-14721-0\_34},
doi = {10.1007/978-3-031-14721-0\_34},
booktitle = {Parallel Problem Solving from Nature - PPSN XVII: 17th International Conference, PPSN 2022, Dortmund, Germany, September 10-14, 2022, Proceedings, Part II},
pages = {485-498},
numpages = {14},
location = {Dortmund, Germany}
}
